\documentclass[a4paper]{cas-sc}

\usepackage[numbers]{natbib}
\usepackage{multirow}
\usepackage{hyperref}
\usepackage{bbm}
\usepackage{float}

\usepackage{amsmath,amssymb,amsfonts}
\usepackage{algorithm}
\usepackage{algorithmic}
\usepackage{graphicx}
\usepackage{textcomp}
\usepackage{xcolor}
\usepackage{subfigure}
\usepackage{float}
\usepackage{multirow}
\usepackage{ulem}

\def\tsc#1{\csdef{#1}{\textsc{\lowercase{#1}}\xspace}}
\tsc{WGM}
\tsc{QE}
\tsc{EP}
\tsc{PMS}
\tsc{BEC}
\tsc{DE}
\begin{document}
\let\WriteBookmarks\relax
\def\floatpagepagefraction{1}
\def\textpagefraction{.001}
% \linespread{2.0}

\shorttitle{Modeling Global Distribution for Federated Learning with Label Distribution Skew}

% \begin{highlights}
% \item We propose a novel federated learning method, FedMGD, which achieves the global modeling of decentralized data distribution by Realistic Score, and effectively eliminates the performance degradation caused by label distribution skew without privacy leakage;
% \item The experimental results demonstrate that the proposed FedMGD significantly outperforms the state-of-the-art on several public benchmarks.
% \item A series of ablation experiments demonstrate that FedMGD can correctly model the global data distribution in the scenario of label distribution skew and provide a novel solution to the data source for downstream tasks.
% \end{highlights}

\shortauthors{Tao Sheng et~al.}

\title [mode = title]{Modeling Global Distribution for Federated Learning with Label Distribution Skew}

\author[1]{Tao Sheng}[type=editor, style=chinese]

% \cormark[1]



% % URL of the first author
% \ead[url]{www.cvr.cc, cvr@sayahna.org}

% \credit{Conceptualization of this study, Methodology, Software}

%\affiliation[1]{organization={School of Computer Science and Engineering, Central South University},
%    % addressline={South Lu Shan Road 932}, 
%    city={Changsha},
%    % citysep={}, % Uncomment if no comma needed between city and postcode
%    % postcode={1043 NX}, 
%    % state={},
%    country={China}}
%\affiliation[2]{organization={Department of Electrical Engineering, University of South Florida},
%    % addressline={South Lu Shan Road 932}, 
%    city={Tampa},
%    % citysep={}, % Uncomment if no comma needed between city and postcode
%    % postcode={1043 NX}, 
%    % state={},
%    country={USA}}

\address[1]{School of Computer Science and Engineering, Central South University, Changsha, China} 
\address[2]{Department of Electrical Engineering, University of South Florida, Tampa, USA} 

\author[1]{Chengchao Shen}[style=chinese, orcid=0000-0003-0249-764X]
\cormark[1]
\ead{scc.cs@csu.edu.cn}
\tnotetext[1]{Corresponding author.}
% \affiliation[2]{organization={School of Computer Science, Central South University},
%     addressline={South Lu Shan Road 932}, 
%     city={Changsha},
%     % citysep={}, % Uncomment if no comma needed between city and postcode
%     % postcode={1043 NX}, 
%     % state={},
%     country={China}}

\author[1]{Yuan Liu}[style=chinese]
% \affiliation[3]{organization={School of Computer Science, Central South University},
%     addressline={South Lu Shan Road 932}, 
%     city={Changsha},
%     % citysep={}, % Uncomment if no comma needed between city and postcode
%     % postcode={1043 NX}, 
%     % state={},
%     country={China}}

% \credit{Data curation, Writing - Original draft preparation}

\author[1]{Yeyu Ou}[style=chinese]
% \affiliation[4]{organization={School of Computer Science, Central South University},
%     addressline={South Lu Shan Road 932}, 
%     city={Changsha},
%     % citysep={}, % Uncomment if no comma needed between city and postcode
%     % postcode={1043 NX}, 
%     % state={},
%     country={China}}
\author[2]{Zhe Qu}[style=chinese]

\author[1]{Jianxin Wang}[style=chinese]

% \affiliation[5]{organization={School of Computer Science, Central South University},
%     addressline={South Lu Shan Road 932}, 
%     city={Changsha},
%     % citysep={}, % Uncomment if no comma needed between city and postcode
%     % postcode={1043 NX}, 
%     % state={},
%     country={China}}

% % Corresponding author text
% \cortext[cor1]{Corresponding author}
% \cortext[cor2]{Principal corresponding author}

% % Footnote text
% \fntext[fn1]{This is the first author footnote. but is common to third
% \fntext[fn2]{Another author footnote, this is a very long footnote and
%   it should be a really long footnote. But this footnote is not yet
%   sufficiently long enough to make two lines of footnote text.}

% % For a title note without a number/mark
% \nonumnote{This note has no numbers. In this work we demonstrate $a_b$
%   the formation Y\_1 of a new type of polariton on the interface
%   between a cuprous oxide slab and a polystyrene micro-sphere placed

\begin{abstract}
Federated learning achieves joint training of deep models by connecting decentralized data sources, which can significantly mitigate the risk of privacy leakage. However, in a more general case, the distributions of labels among clients are different, called ``label distribution skew''. Directly applying conventional federated learning without consideration of label distribution skew issue significantly hurts the performance of the global model. To this end, we propose a novel federated learning method, named FedMGD, to alleviate the performance degradation caused by the label distribution skew issue. It introduces a global Generative Adversarial Network to model the global data distribution without access to local datasets, so the global model can be trained using the global information of data distribution without privacy leakage. The experimental results demonstrate that our proposed method significantly outperforms the state-of-the-art on several public benchmarks. Code is available at \url{https://github.com/Sheng-T/FedMGD}.
\end{abstract}

% \begin{highlights}
% \item We propose a novel federated learning algorithm through global distribution modeling.
% \item Our method achieves global distribution modeling without collecting local privacy data.
% \item Our method solves the label distribution skew problem in federated learning.
% \item It provides additional high-quality data source for federated downstream tasks.
% \item Experiments demonstrate that our method outperforms state-of-the-art methods.
% \end{highlights}

% Keywords

\maketitle

% \begin{itemize} \item document style \item baselineskip \item front
% matter \item keywords and MSC codes \item theorems, definitions and
% proofs \item lables of enumerations \item citation style and labeling.
% \end{itemize}

\section{Introduction}
With the process of deep learning technology, massive training data play a vital role.
However, due to some security or privacy issues in some scenarios, access to data is under strict restriction. 
The data cannot flow out of the corresponding institution, and the lack of data results in the ineffective training of machine learning models. 
The above data dilemma significantly limits the performance of the trained model and even leads to serious consequences. 
For example, Watson, a famous artificial intelligence in the medical field, may cause death once it mistakenly gives a drug. 
To alleviate the above dilemma, a novel framework for machine learning, federated learning~\cite{2} is proposed, which provides more data sources for model training through multi-party collaboration without sharing local private data.
However, there is generally a large difference in the distribution among clients, which leads to performance degradation of conventional federated learning methods, named Non-Independent and Identically Distributed (Non-IID). Non-IID is one of the important causes of training bias in federated learning, and it has different expressions in different scenarios. Non-IID scenarios in federated learning are carefully delineated in~\cite{3}, where data heterogeneity caused by differences in the distribution of labels is called label distribution skew. This is the most common scenario in distributed environments that manifests itself in a different distribution of labels among clients, for example, when the client’s data is limited by factors such as geography, the distribution of labels appears significantly different --- pandas are only found in China or zoos.

To this end, several methods~\cite{4,13,21,22} are proposed to restrict the differences between local and server models in the parameter space, which have shown success in some applications.
However, when the data or labels among clients are highly heterogeneous, the above approaches can not take full use of the information in dispersed data, and degrade the performance of the aggregated global model. 
Therefore, a potential way to address the label distribution skew issue in federated learning is to model the global data distribution among clients. Most existing global data distribution modeling federated learning methods require some auxiliary information such as proxy dataset~\cite{5, 23} or pre-known knowledge~\cite{6, 7, 8}. 
Unfortunately, the requirement of these methods is to directly collect private data or data information from clients, which violates the privacy preserving policy and is not applicable to federated learning.

In this paper, we propose a novel federated learning method, named FedMGD, which allows the global model to obtain global information about data distribution across clients without incurring the privacy leakage issue from clients and thus improves its performance. 
Specifically, we use the global distribution information obtained from modeling to refine the aggregation model, by which we can reduce the differences in distribution among clients and thereby mitigate the hazards caused by label distribution skew. In addition, to ensure that the learned global information generated is more similar to the real data, we constrain by both image authenticity and semantic authenticity, which makes FedMGD still maintain high performance in some noisy datasets.
To achieve our design goal, our federated learning scheme follows a two-stage procedure. 
The advantage of this design is to avoid the inability of the model to accurately describe the local label distribution information of the client after federated learning aggregation.
In the first stage, we adopt a federated Generative Adversarial Network (GAN) framework, where the generator in the server generates samples as the ones from clients and the discriminator in the corresponding client distinguishes generated samples from the local ones. After the training of federated GAN accomplishes, the generator can effectively capture global information about data distribution across clients. 
% However, the unsupervised approach used by federated GAN ignores the importance of labeling information. Therefore, we introduce a \emph{Realistic Score} to enhance the local data distribution.
However, directly aggregating the generator parameters used by the federated GAN is also affected by the label distribution skew, which is not conducive to modeling the global distribution. Therefore, we adapted the architecture of the federated GAN and introduced a Realistic Score to more accurately describe the distribution information of the local data.
In the second stage, we first aggregate the local trained models from clients and then refine the aggregated model using the samples synthesized by the above generator.
In this way, the aggregated global model can significantly benefit from the global information captured by the generator.

In summary, our main contributions are listed as follows:
\begin{itemize}
	\item We propose a novel federated learning method, FedMGD, which achieves the global modeling of decentralized data distribution by Realistic Score, and effectively eliminates the performance degradation caused by label distribution skew without privacy leakage.
	\item The experimental results demonstrate that the proposed FedMGD significantly outperforms the state-of-the-art on several public benchmarks.
    \item A series of ablation experiments demonstrate that FedMGD can correctly model the global data distribution in the scenario of label distribution skew, and provide an  additional high-quality data source for downstream tasks.
\end{itemize}

\section{Related Work}
\subsection{Federated Learning}
Federated Learning (FL) is a distributed machine learning scheme that aims to address data collaboration and protect privacy preserving. It allows multiple participants to train machine learning models collaboratively without exposing local data. Federated learning was first proposed by Google~\cite{2} to address the problem of updating models locally on Android phones for users. Since federated learning provides an effective solution to the current "Isolated Data Island" problem~\cite{12}, it has been gradually used in finance~\cite{33}, security~\cite{34}, healthcare~\cite{35,36,37,38}, recommendation systems~\cite{39,40,41}, and other fields.

However, the collaborative multi-party approach in federated learning introduces new issues, mainly in terms of data privacy security and Non-IID datasets. For data privacy security, it has been shown in~\cite{25} that federated learning does not fully guarantee data security, and even only gradient exchange among the participants may cause data privacy leakage. The current popular solutions mainly combine Secure Multi-Party Computation (MPC) and Differential Privacy (DP) to ensure privacy security issues on the federated learning process and results~\cite{42,43,44}. On the other hand, since most of the real world data are distributed in a Non-IID way, this distribution makes the training of models in federated learning very difficult.~\cite{3} provides a detailed division of Non-IID scenarios in federated learning, in which the Non-IID caused by the behaviors and habits of different participants is called Feature distribution skew~\cite{45,46}; and the difference of label distribution among participants due to the limitation of geographical location and other factors, this Non-IID is called Label distribution skew, which is also the problem focused on in this paper.

\subsection{Label Distribution Skew in Federated Learning}
The label distribution skew in federation learning refers to the differences in the distribution of data labels among the clients participating in federated learning.
FedAvg~\cite{2} is a classical algorithm for solving federated learning problems, which has attracted wide attention because of its simplicity and low communication cost. It learns knowledge from the decentralized data through the transfer and aggregation of model parameters. Unfortunately, in the scenario of label distribution skew, the performance of FedAvg will drop significantly~\cite{5}.

Recently, many studies have developed different solutions to solve the problem caused by label distribution skew. Some studies restrict the differences between local and server models in the parameter space.
For example, FedProx~\cite{4} introduced dynamic regularization to ensure the stability of the model in a highly heterogeneous environment by penalizing updates away from the server model. SCAFFOLD \cite{13} corrects deviation in local updates by introducing additional control variables. However, these approaches focus on reducing the differences between the local and global models without taking full advantage of the information contained in the dispersed data.

Other methods are based on knowledge distillation~\cite{6,7,8}. By using a small portion of the data collected from the clients to guide global model training, the global model can integrate knowledge from the local and mitigate the negative effects of aggregating different data distributions. 
In addition to the above methods, there are many studies that also strive to reduce the hazards due to label distribution skew by changing the global target or data distribution. Among them, several studies~\cite{21,22} force the data to obey uniform distribution by modeling the target distribution. 
Other studies~\cite{5,23,24} suggest that collecting a small portion of data from clients to build a globally shared dataset can reduce the differences in data distribution across clients and improve the performance of the model on label distribution skew data. 
However, the prerequisites for collecting data from clients may make this approach infeasible in many scenarios with strict privacy requirements. Therefore, we need a method to mitigate the label distribution skew problem in federated learning by obtaining the same data as the client label distribution without violating user privacy.

To solve the privacy issues associated with data collection in federated distillation methods, Zhu et al.~\cite{27} proposed FedGen, a data-free knowledge distillation method. It integrates knowledge from the prediction results of the local model by learning a lightweight generator at the server and then distributes the learned knowledge via broadcast and as an inductive bias to regulate the local training. Although FedGen does not collect data from the client directly, the client needs to count and upload the label information of this participating training during each round of communication with the server. This way of obtaining the label distribution also causes the leakage of local distribution information.
% Unlike FedGen, we introduce the idea of distributed generative adversarial in FedMGD, and can directly model the global distribution by the proposed \emph{Realistic Score}, avoiding exposing the label distribution information of individual clients. In addition, unlike the features of FedGen generated data, we consider that the data samples themselves contain more information, and hence FedMGD refines the global model through the generated raw sample data. 

\subsection{GAN in Federated Learning}
In order to better learn global information from the decentralized data distribution, we introduce a distributed GAN to model the global data distribution, thus alleviating the performance degradation caused by label distribution skew in federated learning.

In the traditional GAN~\cite{14}, the generator and discriminator learn the data distribution through mutual confrontation. 
Distributed GAN is another application of GAN in a distributed environment.
In federated learning, distributed GAN is mainly active in two aspects: one is how to use GAN to attack the client in order to obtain its private data.
In the studies of Hitaj et al.~\cite{25} and Wang et al.~\cite{26}, they used GAN to perform recovery attacks on the client's private data from other clients and the server side, respectively.
The other focuses on how to generate higher quality data within the constraints of federated learning environments. 
For example, McMahan et al.~\cite{15} propose to train a generator with different privacy protection levels in federated learning, allowing the model to examine data without direct access to real data. FedGAN~\cite{16} was proposed to train GAN by parameter exchange in a federated scenario. Hardy et al.~\cite{9} proposed a new GAN structure for a distributed environment. On this basis, Chang et al. \cite{11} supplemented the missing modes of medical images through generation.

However, these methods default to data belonging to IID, which is not suitable for data modeling in label offset scenarios. Recently, Yonetani et al.~\cite{10} proposed a strategy to solve the problem of label distribution skew in different clients, called Forgiver-First Update (F2U). Although F2U can be used in learning the distribution of some rare classes, the unsupervised approach used ignores the importance of labeling (semantic) information. In the unsupervised label distribution skew scenario, the generated data is likely to be concentrated in only a few classes, which is not conducive to modeling the global data distribution, and thus the use of labels to constrain the type of generated data is necessary. In addition, unlabeled data bring new challenges for the subsequent training of downstream models.

To address the above issues we propose an update of Realistic Score and apply it to FedMGD. Realistic Score adds a semantic truth restriction on the data under different distributions and can better model the data distribution under label distribution skew.

\section{Background and Motivation}

\subsection{Federated Learning Objective}
A Federated learning framework typically include multiple participants who collaboratively train a global consensus model by learning the model locally and periodically communicating with a central server. Suppose that $K$ clients are in the federated learning system, and the overall optimization objective is defined as:

\begin{equation}
    \underset{\omega}{\min}F(\omega)=\sum_{i=1}^{K}\frac{N_i}{N}F_i(\omega),
    \label{global_objective}
\end{equation}
where $\omega$ is the parameter of the global model.
For client $i \in [K]$, we define the dataset as $\xi_i = \{(x_m^{(i)}, y_m^{(i)})\}_{m=1}^{N_i}$, where $N_i$ is the number of data samples. As such, the global dataset across clients can be denoted as $\xi = \cup_{i=1}^K \xi_{i}$, and the total number of participating training data $N=\sum_{i}N_i$. In each communication round, client $i$ receives model parameters $\omega$ from the central server and optimizes the local objective function using local data samples, i.e., $F_i(\omega)=\frac{1}{N_i}\sum_{m=1}^{N_i}\mathcal{L}(\omega; x_{m}^{(i)}, y_m^{(i)})$, where $\mathcal{L}(\cdot)$ is the local loss function in each client, e.g., cross-entropy. In this paper, we consider that $F_i(\omega^{(t)})$ is non-convex and use a local optimizer such as stochastic gradient descent to process the local training update, which is widely used in existing federated learning works \cite{2,4}. In the communication round $t$, client $i$ samples $B$ mini-batches from its dataset $\xi_i$ and performs $E$ local epochs to minimize the objective function $F_i(\omega)$ as:

\begin{equation}
    \omega_{i}^{(t+1)}=\omega_{i}^{(t)}-\sum_{e=1}^{E}\eta_i\triangledown F_i(\omega_{i,e}^{(t)};b),
    \label{local_objective}
\end{equation}
where $b \in B$ is one batch in the local epoch $e$, $\eta_i$ is the learning rate of local training, and $\triangledown F_i(\cdot)$ is the local model gradient.

\subsection{Design Motivation}

\begin{figure}[t]
	\centering
	\includegraphics[width=5.8in]{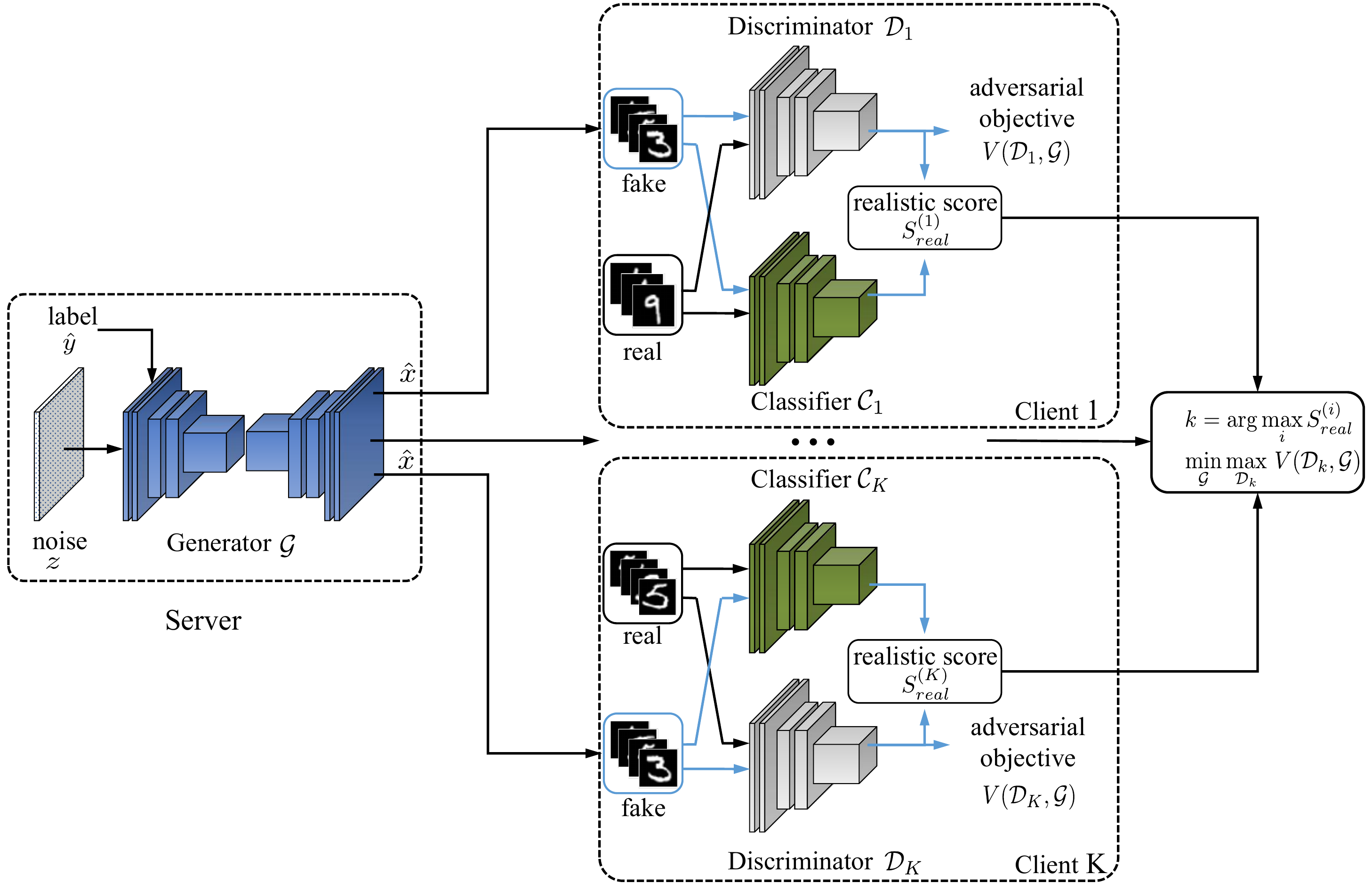}
    \caption{The process of communication between the server and clients in the Generative Adversarial Stage of FedMGD.}
	\label{thesis_fig1}
\end{figure}

In the traditional distributed IID assumption, the data samples $\xi_i$ are uniformly and randomly distributed among different clients, when $\mathbb{E}_{\xi_i}[F_i(\omega)]=F(\omega)$. However, Non-IID data distribution is a more practical issue caused by different user habits, geographic locations, and other factors, the client optimizes toward different local optimal target $F_i$. As a result, it leads to a significant degradation or even failure to converge. For two different clients $i$ and $j$, the input data $x$ and its corresponding label $y$, $(x,y)\sim p_i(x,y)$ denotes a sample data drawn randomly from the local data distribution in the client $i$. In formal, for the different two clients $i$ and $j$, the Non-IID represents $p_i(x,y)\neq p_j(x,y)$. $p(x,y)$ can be written in the form of $p(x|y)p(y)$, and we $p_i(x|y)= p_j(x|y)$ refer to the case of but $p_i(y)\neq p_j(y)$ as label distribution skew.

Some existing works have developed improved algorithms to address the label distribution skew problem, which can be divided into two categories: (1) Designing new aggregation rule \cite{4,13,6}. They have been shown the less efficiency and also incur large performance degradation. (2) Modeling the global distribution~\cite{5,23,24}. They require a gloally shared dataset to reduce the client-side label distribution differences. However, this approach relies on the way data is collected from local clients, which makes this approach unusable in scenarios due to the strict privacy policy in federated learning.

\begin{figure}[t]
	\centering
	\includegraphics[width=5.5in]{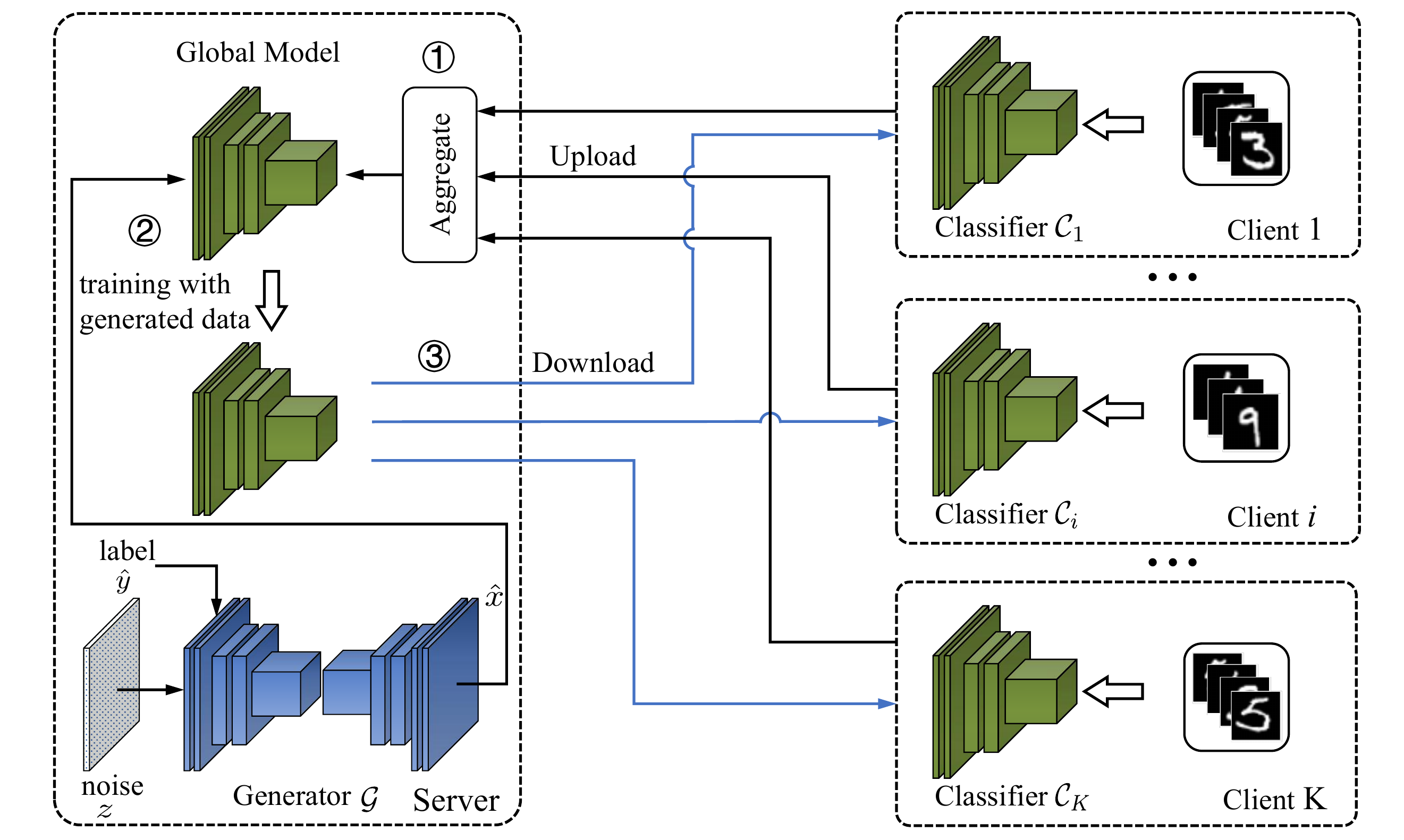}
	\caption{The process of communication between the server and clients in the Federated Enhancement Stage of FedMGD.}
	\label{thesis_fig2}
\end{figure}

% Hence, we introduce generator $\mathcal{G}$, which is trained to model the global data distribution across the given clients, to refine the aggregated model and improve the compatibility of models among clients. 

According to the above statements, how to model the global distribution without sharing any private data from local clients should be seriously considered. To address this problem, we propose a new federated learning method, named FedMGD, to model the global distribution by introducing a global generative adversarial network to obtain information about the distribution of the global data, which is not required any private data in clients and only manipulate the distribution. We build the generator $\mathcal{G}$ on the server side, which is trained to model the global data distribution across a given client to refine the aggregation model and improve the compatibility of the model between clients. 
% In the next section, we will present a two-step training procedure and explain why our proposed FedMGD can successfully avoid these privacy issues.

In particular, in order to avoid the federated learning aggregated model returning wrong local label distribution information and to obtain a well-initialized local model before the federated learning training starts, we divide FedMGD into two stages: the adversarial generation stage and the federated enhancement stage.
As shown in Figure~\ref{thesis_fig1}, in the generative adversarial stage, we adopt the Generative Adversarial Network framework, where generator $\mathcal{G}$ is set on the server side and the discriminators $\{\mathcal{D}_i\}_{i=1}^K$ are set on the corresponding clients, respectively. 
Additionally, we introduce local classifiers $\{\mathcal{C}_i\}_{i=1}^K$, one for each client.
These classifiers are respectively trained using local data and are aimed to discriminate the semantic consistency between the predictions of the classifier and the preset labels.
In the federated enhancement stage, the local classifiers are initialized by the ones trained in the generative adversarial stage and then aggregated into a global model in the server. 
As shown in Figure~\ref{thesis_fig2}, the global model is refined with the samples synthesized by generator $\mathcal{G}$ and used to update the local model in the clients, which significantly reduces weight divergence during the federated learning. The detailed training procedure of FedMGD will be introduced in the next section.

\section{Method}

In this section, we elaborate on our proposed method. 
We proposed the method which consists of two stages, generative adversarial stage and 
federated enhancement stage.
Then we describe the above stages in detail in Section~\ref{3.2} and Section~\ref{3.3}, respectively.

\subsection{Generative Adversarial Stage}\label{3.2}

\begin{algorithm}[t]
	\caption{FedMGD: Generative Adversarial Stage.}
	\label{alg1}
	\begin{algorithmic}[1] %这个1 表示每一行都显示数字
		\REQUIRE  %算法的输入参数：Input
		The number of clients $K$, local datasets $\xi_{i}$.
		\ENSURE  %算法的输出：Output
		Generator $\mathcal{G}$, classifier $\{\mathcal{C}_i\}_{i=1}^K$ and discriminator $\{\mathcal{D}_i\}_{i=1}^K$.
		\STATE Initialize the parameters of generator $\mathcal{G}^{(0)}$, classifier $\{\mathcal{C}_i^{(0)}\}_{i=1}^K$ and discriminator $\{\mathcal{D}_i^{(0)}\}_{i=1}^K$;
		\FOR {$t = 1,..., T$}
			\STATE Uniformly sample a set of clients $\mathcal{A}^{(t)} \in [K]$;
            \STATE {Set the preset labels by server sampling and synthesize dataset $\xi_{syn}^{(t)}$ by the generator $\mathcal{G}^{(t-1)}$.}
            \STATE Send the synthesized dataset $\xi_{syn}^{(t)}$ to a set of clients $\mathcal{A}^{(t)}$;
			\FOR {each client $i \in \mathcal{A}^{(t)}$ in parallel}	
				\STATE Train discriminator $\mathcal{D}_i^{(t)}$ in client $i$ using $\xi_i$ and $\xi_{syn}^{(t)}$ by Eq.~\eqref{eq5};
				\STATE Train classifier $\mathcal{C}_i^{(t)}$ in client $i$ using $\xi_i$ by cross entropy loss;
				\STATE Compute Realistic Score $S_{real}^{(i,t)}$ for client $i$ on $\xi_{syn}^{(t)}$  by Eq.~\eqref{eq3};
		\ENDFOR
		\STATE Select client $k$ according to $\{S_{real}^{(i,t)}\}_{i \in \mathcal{A}}$ by Eq.~\eqref{eq:select};
		\STATE Update generator $\mathcal{G}^{(t)}$ using the loss computed by client $k$ as Eq.~\eqref{eq5}.
		\ENDFOR
	\end{algorithmic}
\end{algorithm}

To prevent the leakage of real data from clients, we split the conventional GAN into two parts: the generator $\mathcal{G}$ on the server side and discriminators $\{\mathcal{D}_i\}_{i=1}^K$ on the client side. 
In this way, the server can access synthesized samples from global data distribution by generator $\mathcal{G}$ but without the leakage of real data from clients. 
Let $p_i$ denote the data distribution of the $i$-th client, where $i \in [K]$, 
and $p_g$ denote the distribution learned by generator $\mathcal{G}$. 
The goal of generator $\mathcal{G}$ is to learn the global data distribution $p_{data}$. 
In the conventional distributed GAN~\cite{9}, the distribution of different clients is assumed to follow the same distribution, that is, $p_i = p_j$ for every pair of different clients, where $i \neq j$. 
Hence, generator $\mathcal{G}$ is trained to approximate the distribution of real data $p_{data}$ by fitting $\frac{1}{K}\sum\nolimits_{i=1}^K p_i(x)$.
However, due to the effect of label distribution skew, the distribution of local clients is significantly different from each other, that is, $p_i \neq p_j$.
Therefore, conventional distributed GAN fails to model global data distribution in the federated learning with label distribution skew. 

To solve this problem, an approximate solution was proposed in F2U~\cite{10}, which points out that $p_{max}(x)$ as Eq.~\eqref{pmax} can be regarded as the global optimum for $p_{data}$ in distributed GAN. 

\begin{equation}
    \begin{split}
    & p_{max}(x) = \frac{1}{Z} \underset{i}{\max}\:p_{i}(x), \\
    & Z=\int_{x}\underset{i}{\max}\:p_{i}(x)dx,
    \end{split}
    \label{pmax}
\end{equation}
where $Z$ is the normalization constant, $p_{max}(x)$ contains all classes across clients, including rare classes that are only available on a few clients. 
Since it is infeasible to obtain the true distribution of each client $p_i (x)$, we can leverage an alternative way to update the generator $\mathcal{G}$.
That is, for the synthesized sample $\hat{x}$, the generator $\mathcal{G}$ selects the discriminator with the largest discriminant probability by Eq.~\eqref{largest_score} to be updated, which is given as follows:
\begin{equation}
\mathcal{D}_{max}(\hat{x})=\underset{i}{max}\mathcal{D}_{i}(\hat{x})
\label{largest_score}
\end{equation}

This training approach allows learning the rare classes and avoids the negative effect of poorly trained model on the clients.
However, the unsupervised approach used by F2U ignores the importance of labeling information. In the unsupervised label distribution skew scenario, the generated data is likely to be concentrated in only a few classes, which is not conducive to modeling the global data distribution, and thus the use of labels to constrain the type of generated data is necessary. In addition, unlabeled data bring new challenges for the subsequent training of downstream models.

To address the issues mentioned above, we propose a modeling method that can describe the global distribution more comprehensively without compromising client privacy.
In the proposed method to better guide the generator $\mathcal{G}$ for global modeling, we set the discriminator $\mathcal{D}_i$ and the classifier $\mathcal{C}_i$ on the client side to constrain the generated data in terms of both semantic truth and image truth, respectively.
For the synthetic samples whose preset labels are consistent with the local labels, $\mathcal{D}_i$ and $\mathcal{C}_i$ give scores in terms of both realism and semantics, which in turn guide the generator $\mathcal{G}$ for the next update.
The communication process between the server and the client in the FedMGD generative adversarial stage is shown in Figure~\ref{thesis_fig1}.
Specifically, in the communication round $t$, the generator $\mathcal{G}^{(t)}$ synthesizes dataset $\xi_{syn}^{(t)}$ and feeds them to the randomly selected clients. The synthetic dataset $\xi_{syn}^{(t)}$ is a batch of samples $\hat{x}$ generated according to the specified label $\hat{y}$.
Discriminator $\mathcal{D}_{i}^{(t)}$ and classifier $\mathcal{C}_{i}^{(t)}$ in the selected client $i$ output discriminant probability $\mathcal{D}_{i}^{(t)}(\hat{x})$ and classification probability $\mathcal{C}_{i}^{(t)}(\hat{x})$, respectively. 
Then, we further measure the difference between the given label $\hat{y}$ and classification probability $\mathcal{C}_{i}^{(t)}(\hat{x})$ using cross entropy criterion.
Combining discriminant score and classification score, we introduce a novel Realistic Score $S_{real}^{(i,t)}$ for client $i$ at training round $t$ as follows:
\begin{equation}
	S_{real}^{(i,t)} =  \mathcal{D}_{i}^{(t)} \left (\hat{x} \right)-  \mathcal{L}_{xen} \left (\mathcal{C}_{i}^{(t)} \left (\hat{x} \right ),\hat{y} \right ),
	\label{eq3}
\end{equation}
where $\mathcal{L}_{xen}(\cdot)$ denotes cross entropy metric.
The Realistic Score $S_{real}^{(i,t)}$ is used to measure both the realism and semantics of generated images, which provides more comprehensive information for generator $\mathcal{G}$. 
We verified the validity of Realistic Score in Section~\ref{4.3.1}.

To model the global distribution of data across clients with label distribution skew, we follow and improve the solution of F2U. 
Specifically, we select the discriminator according to Realistic Score $S_{real}^{(i,t)}$, instead of simple discriminant probability.
The selected client $k$, which is trained with generator $\mathcal{G}^{(t)}$, can be denoted as:
\begin{equation}
	k = \arg \max\limits_{i \in \mathcal{A}^{(t)}} \: S_{real}^{(i,t)},
	\label{eq:select}
\end{equation}
% where $\mathcal{A}$ denotes the set of the select clients.
where $\mathcal{A}^{(t)}$ denotes the set of clients selected by the server in the current training round.

Finally, the objective function of the proposed GAN can be presented as:

\begin{equation}
	\begin{split}
	\min\limits_{\mathcal{G}^{(t)}} \max\limits_{\mathcal{D}_k^{(t)}}  V(\mathcal{D}_k^{(t)}, \mathcal{G}^{(t)})  = & \mathbb{E}_{x\sim p_{data}(x)}[\log \mathcal{D}_k^{(t)}(x)] \: +\\
	&  \mathbb{E}_{z\sim p_{g}(z)}[\log (1-\mathcal{D}_k^{(t)}(\mathcal{G}^{(t)}(z|\hat{y})))+ \mathcal{L}_{xen}(\hat{y}, \mathcal{C}_k^{(t)}(\mathcal{G}^{(t)}(z|\hat{y})))],
	\end{split}
	\label{eq5}
	\end{equation}
where $z$ is random Gaussian noise.
The overall algorithm for generative adversarial stage is shown in Algorithm~\ref{alg1}. 

In this way, we can obtain additional global data without violating client privacy for federated learning, which can further improves the performance of the global model in the scenario of label distribution skew. 
In addition, we adopt a conditional generator in our GAN, which synthesizes samples by the given labels in a controllable manner. 
Specifically, the training process ensures that the data for each class is fully trained by preset labels for the generator $\mathcal{G}$. In this case, the preset labels are sampled using Server Sampling, i.e., uniform sampling among all classes at the server side. We will discuss the proposed sampling method in Section~\ref{4.3.3}.
% And the classifier of the generative adversarial stage can also be directly used in the subsequent federated enhancement stage. 
Finally, the classifier from the generative adversarial stage can also be directly used in the subsequent federated enhancement stage.

\subsection{Federated Enhancement Stage}\label{3.3}

In federated enhancement stage, we exploit the global generator $\mathcal{G}$ trained in generative adversarial stage to alleviate the performance degradation caused by label distribution skew issue.
By modeling the global distribution, the samples synthesized by generator $\mathcal{G}$ can effectively reduce the differences in label distribution among clients.
Specifically, the training procedure of this stage can be introduced as follows:
\begin{enumerate}
    \item We initialize the classifiers $\{\mathcal{C}_i\}_{i=1}^K$ of clients with the ones trained in generative adversarial stage. 
    \item In the communication round $t$, we uniformly sample a subset of clients $\mathcal{A}^{(t)} \subseteq [K]$ and train their classifiers using local data.
    \item We upload the parameters of classifiers $\{\mathcal{C}_i^{(t)}\}_{i \in \mathcal{A}^{(t)}}$ to server and further aggregate the uploaded parameters into the global model $\tilde{\mathcal{C}}^{(t)}$.
    \item Generator $\mathcal{G}$ synthesizes samples, which follow the global distribution of data across clients, to refine the aggregated global model $\tilde{\mathcal{C}}^{(t)}$. Specifically, in this step, we only select generated samples, whose labels are consistent with the predictions of global model $\tilde{\mathcal{C}}^{(t)}$, to train global model $\tilde{\mathcal{C}}^{(t)}$.
    \item The trained global model $\tilde{\mathcal{C}}^{(t)}$ is downloaded to the corresponding clients in $\mathcal{A}^{(t)}$ and the training in the next round continues.
\end{enumerate}
The overall algorithm is summarized into Algorithm~\ref{alg2}.
In this stage, we use the knowledge obtained from global modeling to refine the aggregation model and mitigate the hazards caused by the label distribution skew problem. In this way, we re-correct the problem that the aggregation model is inconsistent with the global optimization direction due to local distribution differences. And because the generator provides additional globally distributed data to the aggregation model, it further improves the performance and generalization of the aggregation model. We prove the validity of the proposed method in Section~\ref{4.2}.

\begin{algorithm}[h]
	\caption{ FedMGD: Federated Enhancement Stage.}
	\label{alg2}
	\begin{algorithmic}[1]
		\REQUIRE
			The number of clients $K$, generator $\mathcal{G}$, local datasets $\cup_{i=1}^K \xi_{i}$, initial classifier parameters $\{\hat{\omega}_{i}\}_{i=1}^K$.
		\ENSURE
			The global classifier $\tilde{\mathcal{C}}^{(t)}$.
		\FOR {$i$ = 1 to $K$}
			\STATE Initialize classifier parameters by $\omega_{i}^{(0)} \leftarrow \hat{\omega}_{i}$;
		\ENDFOR
		\FOR {$t$ = 1 to $T$}
			\STATE Uniformly sample a subset of clients $\mathcal{A}^{(t)} \subseteq [K]$;
			\FOR {each client $i \in \mathcal{A}^{(t)}$ in parallel}	
				\STATE Train classifier $\mathcal{C}_{i}^{(t)}$ in client $i$ for $E$ epochs;
				\STATE Send the parameters $\omega_{i}^{(t)}$ of classifier $\mathcal{C}_{i}^{(t)}$ to the server;
			\ENDFOR
			\STATE Aggregate client model weights $\omega_{i}^{(t)}$ by:
			$$\omega^{(t)} = \sum_{i=1}^{K}\frac{N_{i}}{N}\omega^{(t)}_{i}$$
			where $N$ is the number of all data across clients, $N_i$ is the number of data for client $i$, and $K$ is the number of clients;
			% \STATE Synthesize samples $\xi_{syn}^{(t)}$ by generator $\mathcal{G}$;
            \STATE {Set the preset labels by server sampling and synthesize dataset $\xi_{syn}^{(t)}$ by the generator $\mathcal{G}$.}
			\STATE Select the consistent synthesize samples $\xi_{syn}^{(t)}$ to refine the global classifier $\tilde{\mathcal{C}}^{(t)}$;
			\STATE Download the parameters $\omega^{(t)}$ to update the ones in clients $[K]$.
		\ENDFOR	
	\end{algorithmic}
\end{algorithm}

\section{Experiments}\label{4}
In this section, we experimentally verify the effectiveness of FedMGD, and summarize the implementation details in Section ~\ref{4.1}. 
We compare FedMGD with several baseline algorithms in Section ~\ref{4.2} and further analyze FedMGD by ablation experiments in Section ~\ref{4.3}.

\subsection{Experimental Setup}\label{4.1}

\subsubsection{Dataset} \label{4.1.1}
We conduct experiments on the following image datasets:  EMNIST~\cite{18}, FashionMNIST~\cite{19}, SVHN~\cite{28} and CIFAR10~\cite{32}. Among them, EMNIST includes 26 handwritten letters with different kinds of labels and FashionMNIST contains 10 different kinds of clothing images. SVHN is a dataset consisting of different house numbers in street view images. CIFAR10 is a dataset including 10 classes of color images from the real world, making the task more difficult because it contains more noise. We divide 10\% of the data as public test set and distribute the rest data over the clients for locally training and testing. The whole process is controlled by random seeds. To ensure the consistency of image resolution, we resize all images to $32 \times 32$.

\begin{figure}[t]
	\centering
	\subfigure[The EMNIST dataset is divided under 5 clients according to the Dirichlet distribution.]{
		\begin{minipage}[t]{0.33\linewidth}
			\centering
			\includegraphics[width=1.9in]{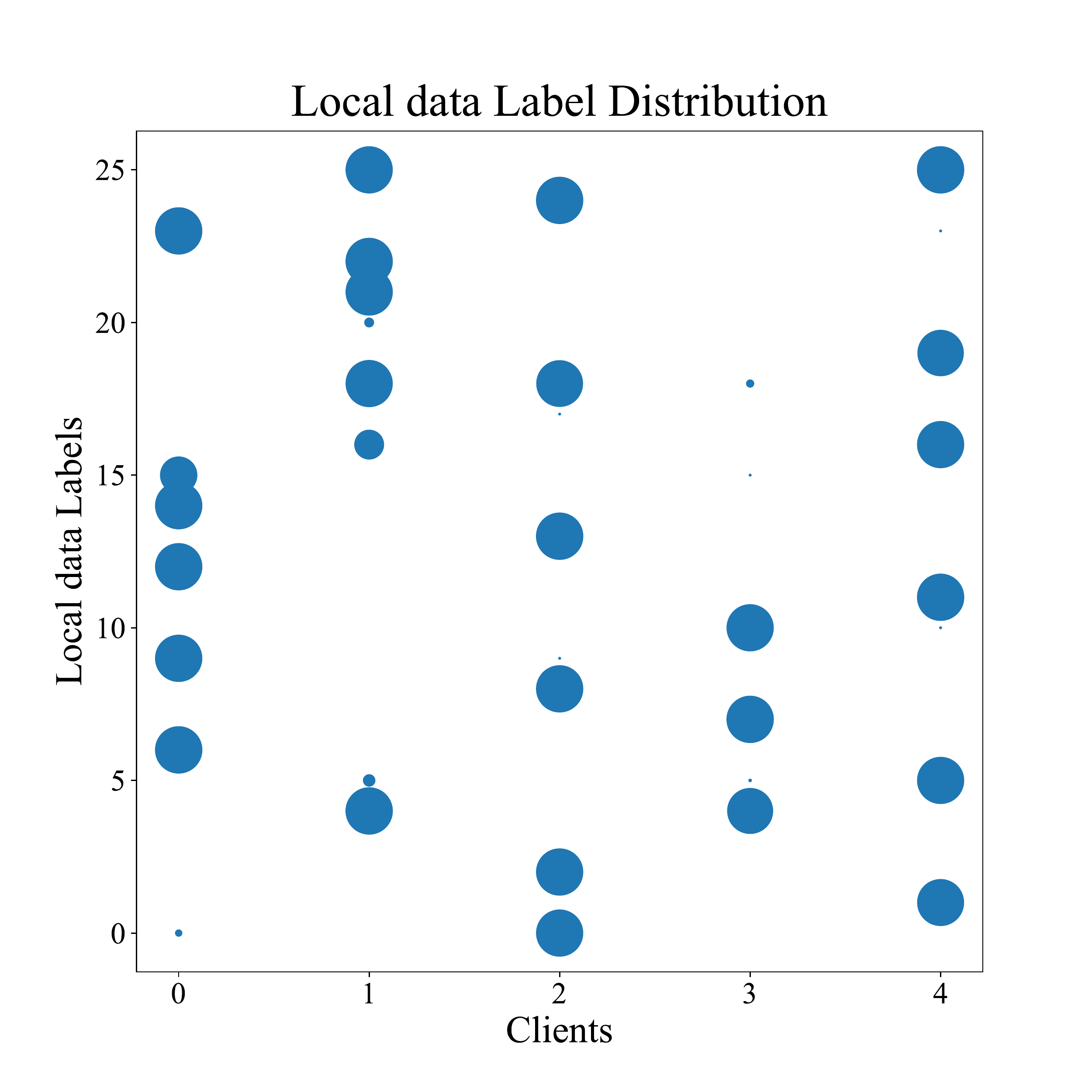}
			%\caption{fig1}
		\end{minipage}%
		\begin{minipage}[t]{0.33\linewidth}
			\centering
			\includegraphics[width=1.9in]{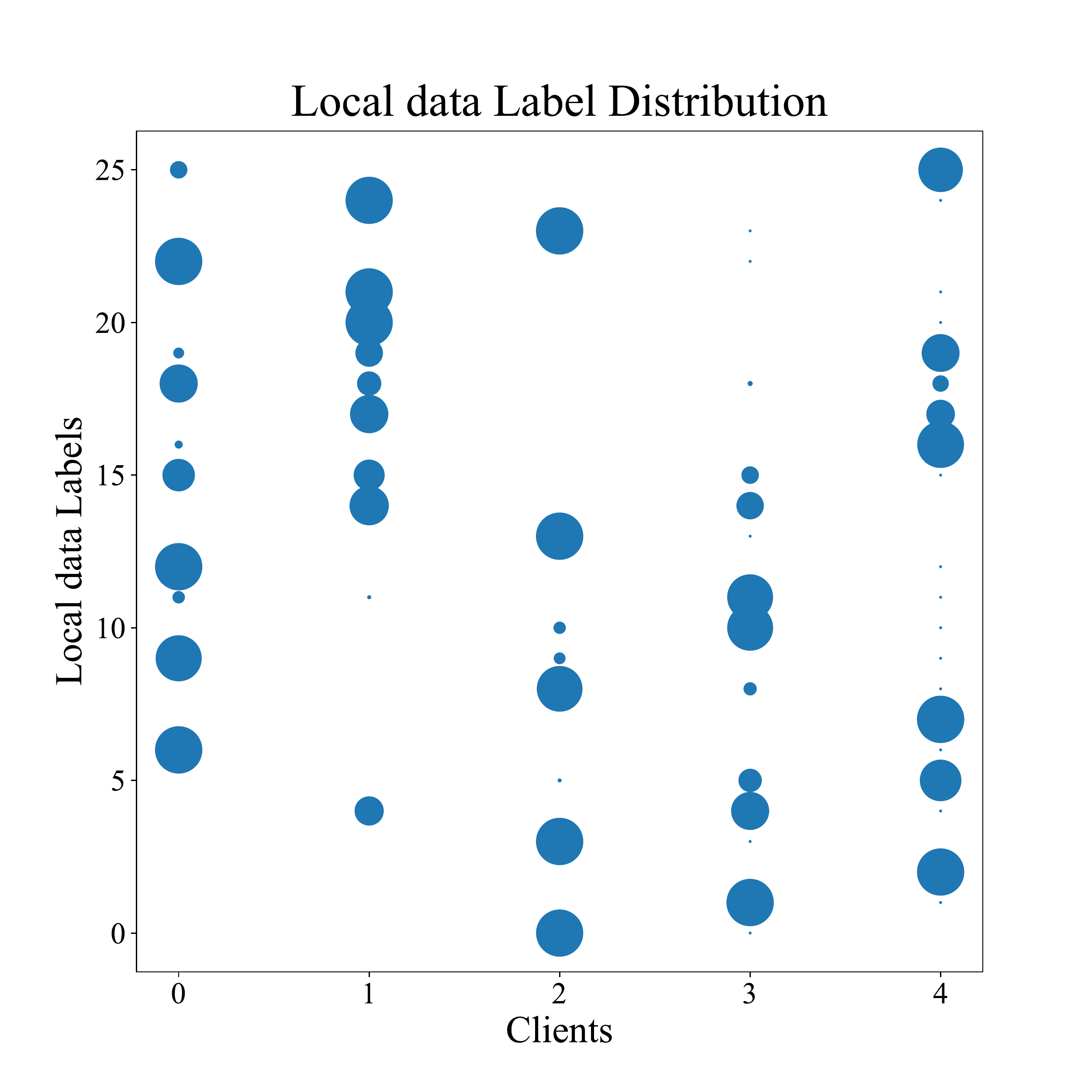}
			%\caption{fig1}
		\end{minipage}%
		\begin{minipage}[t]{0.33\linewidth}
			\centering
			\includegraphics[width=1.9in]{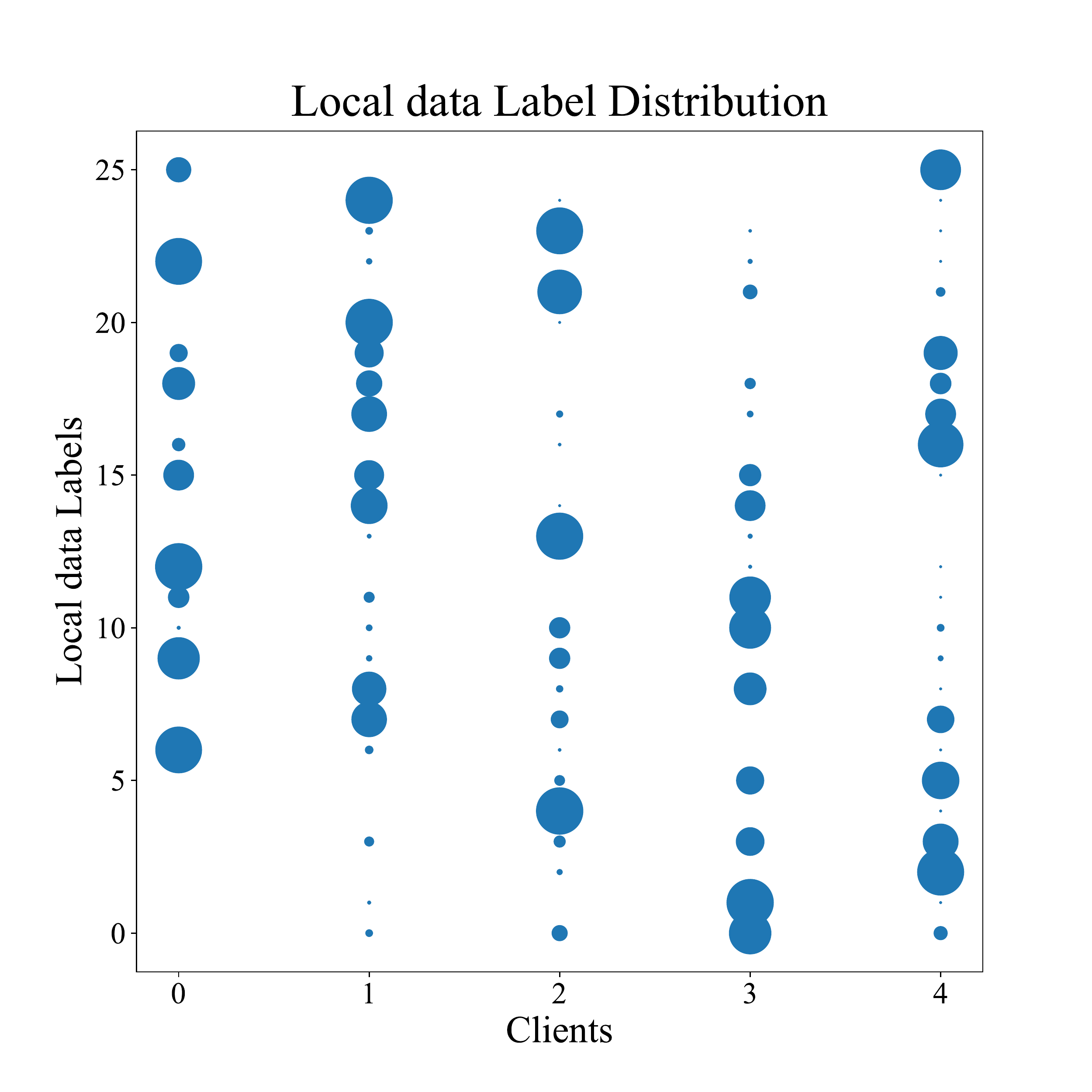}
			%\caption{fig1}
		\end{minipage}%
	}
	\subfigure[The FashionMNIST dataset is divided under 5 clients according to the Dirichlet distribution.]{
		\begin{minipage}[t]{0.33\linewidth}
			\centering
			\includegraphics[width=1.9in]{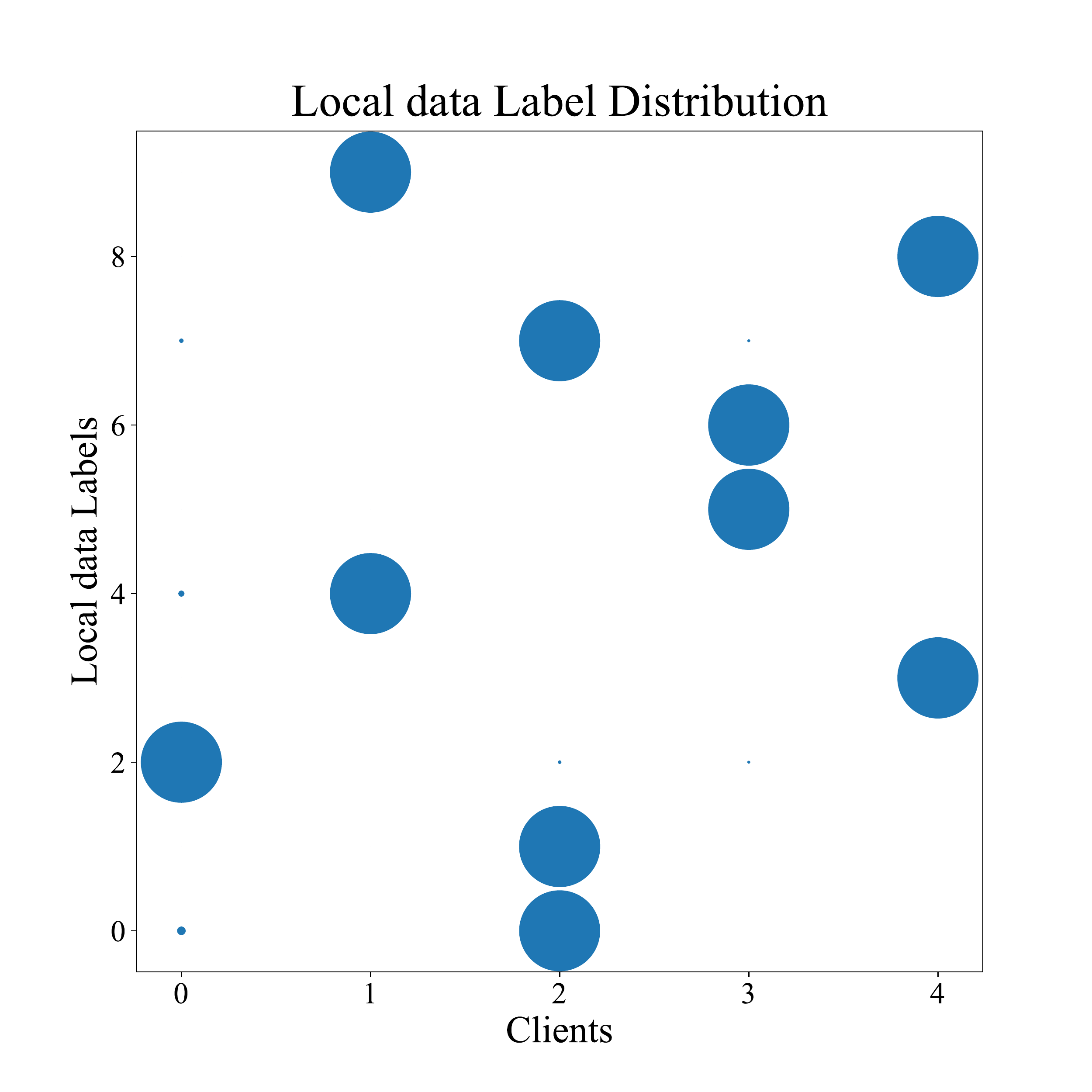}
			%\caption{fig1}
		\end{minipage}%
		\begin{minipage}[t]{0.33\linewidth}
			\centering
			\includegraphics[width=1.9in]{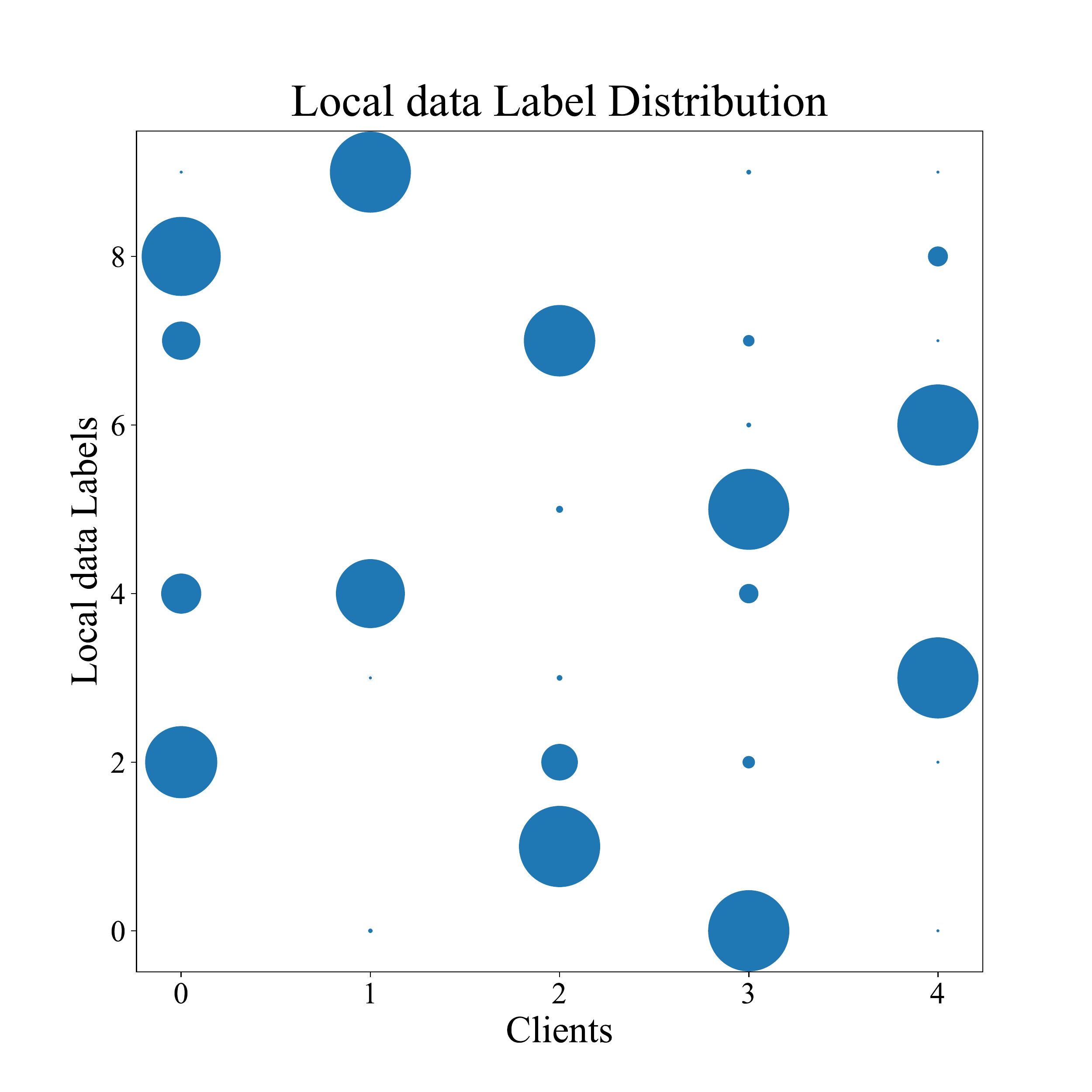}
			%\caption{fig1}
		\end{minipage}%
		\begin{minipage}[t]{0.33\linewidth}
			\centering
			\includegraphics[width=1.9in]{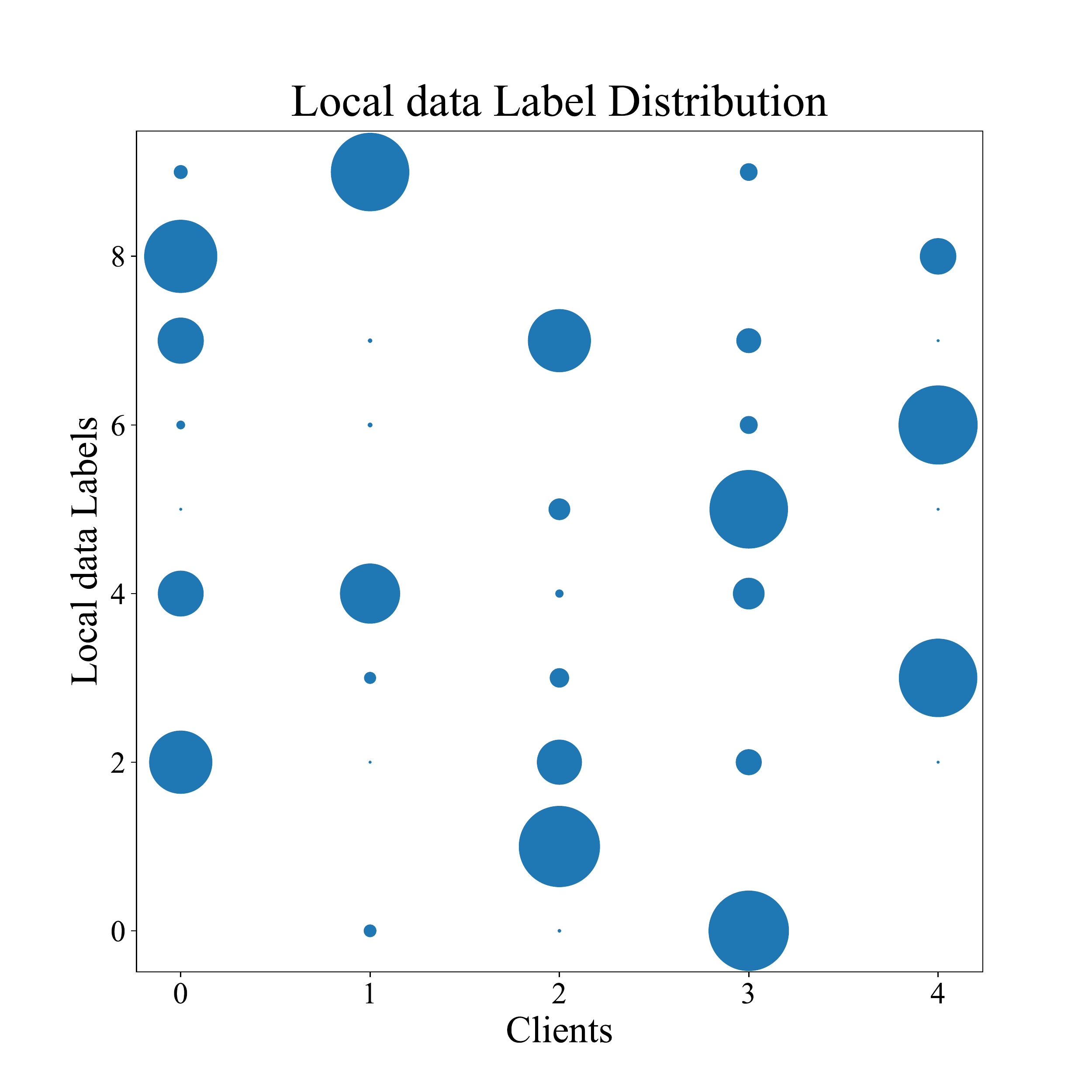}
			%\caption{fig1}
		\end{minipage}%
	}
    \subfigure[The SVHN dataset is divided under 5 clients according to the Dirichlet distribution.]{
		\begin{minipage}[t]{0.33\linewidth}
			\centering
			\includegraphics[width=1.9in]{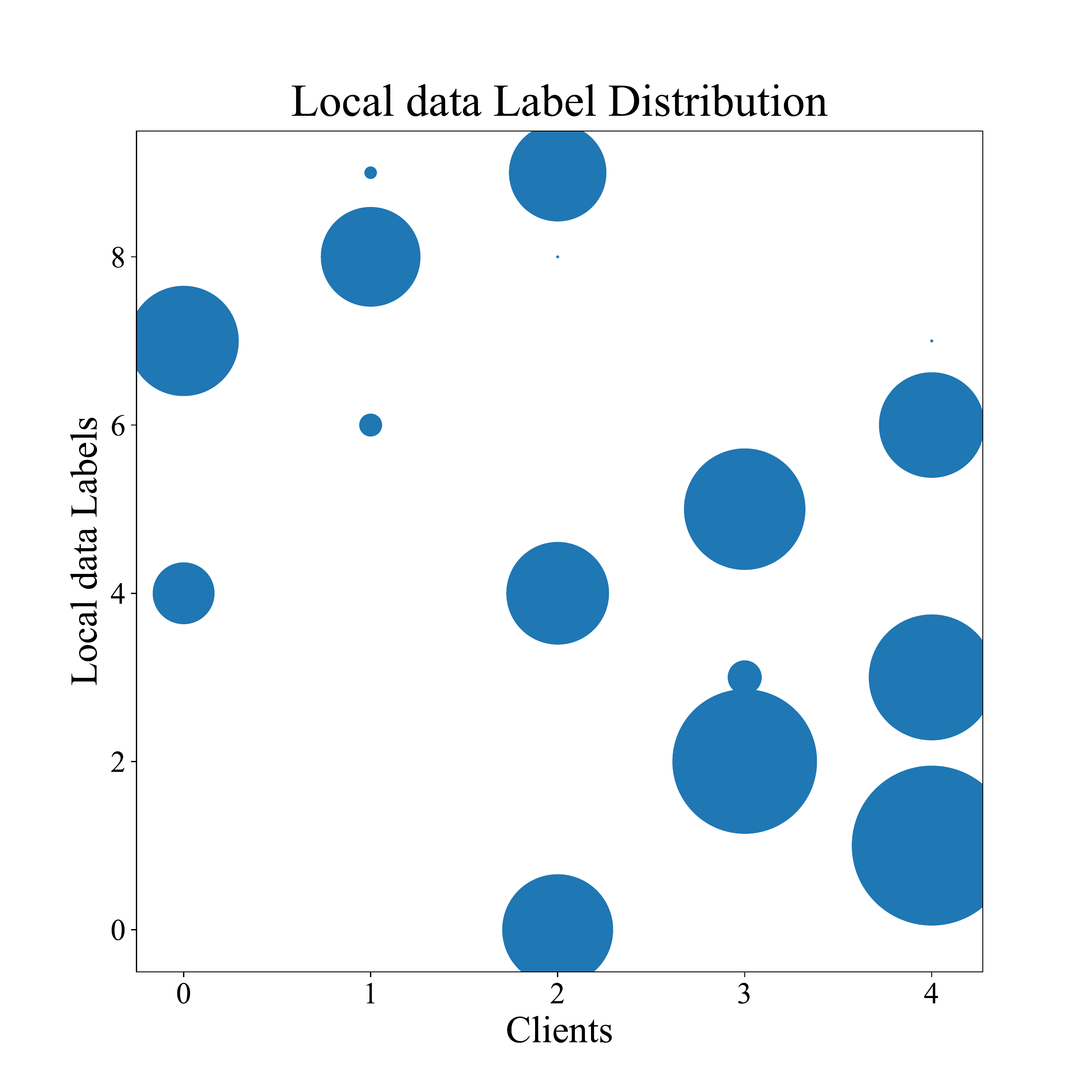}
			%\caption{fig1}
		\end{minipage}%
		\begin{minipage}[t]{0.33\linewidth}
			\centering
			\includegraphics[width=1.9in]{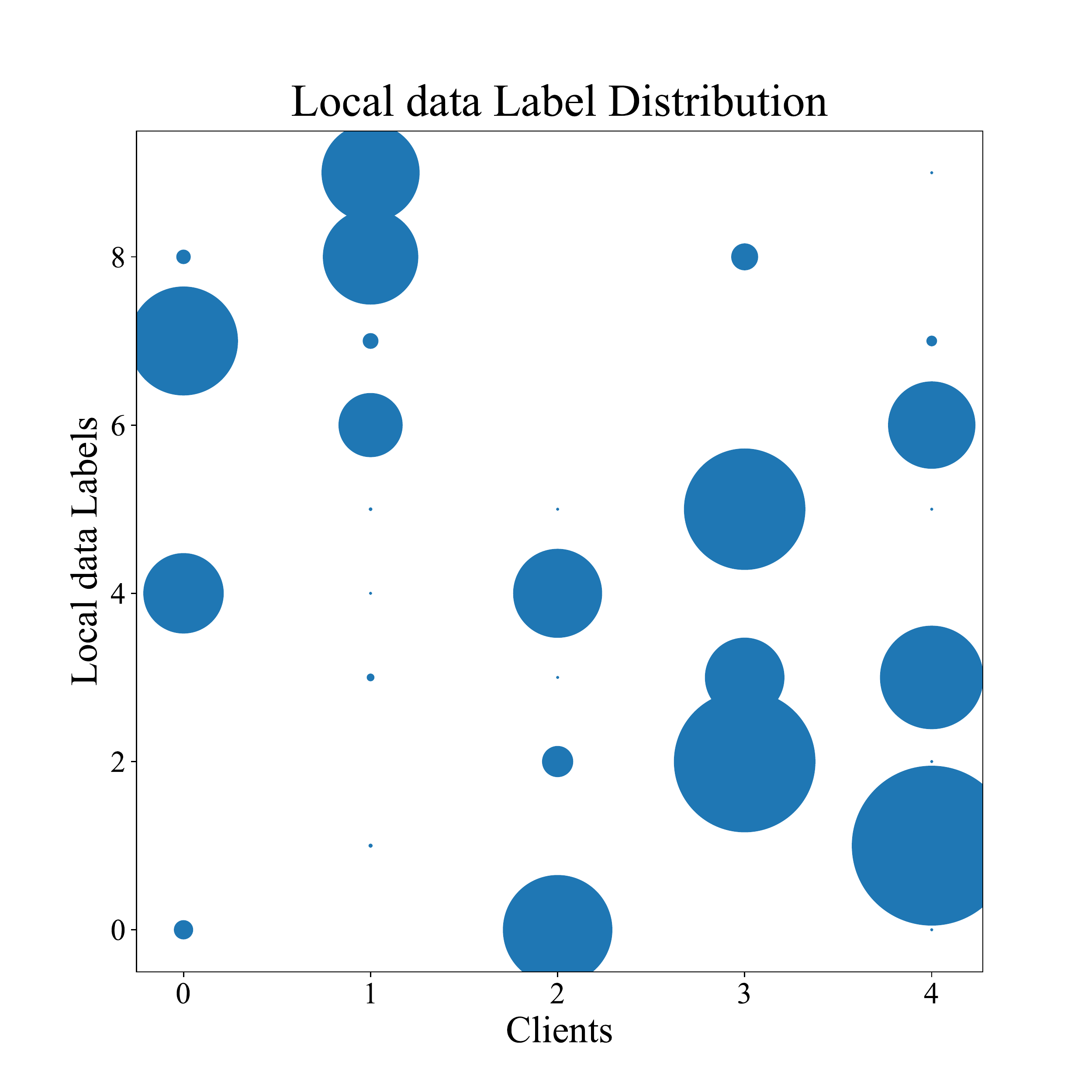}
			%\caption{fig1}
		\end{minipage}%
		\begin{minipage}[t]{0.33\linewidth}
			\centering
			\includegraphics[width=1.9in]{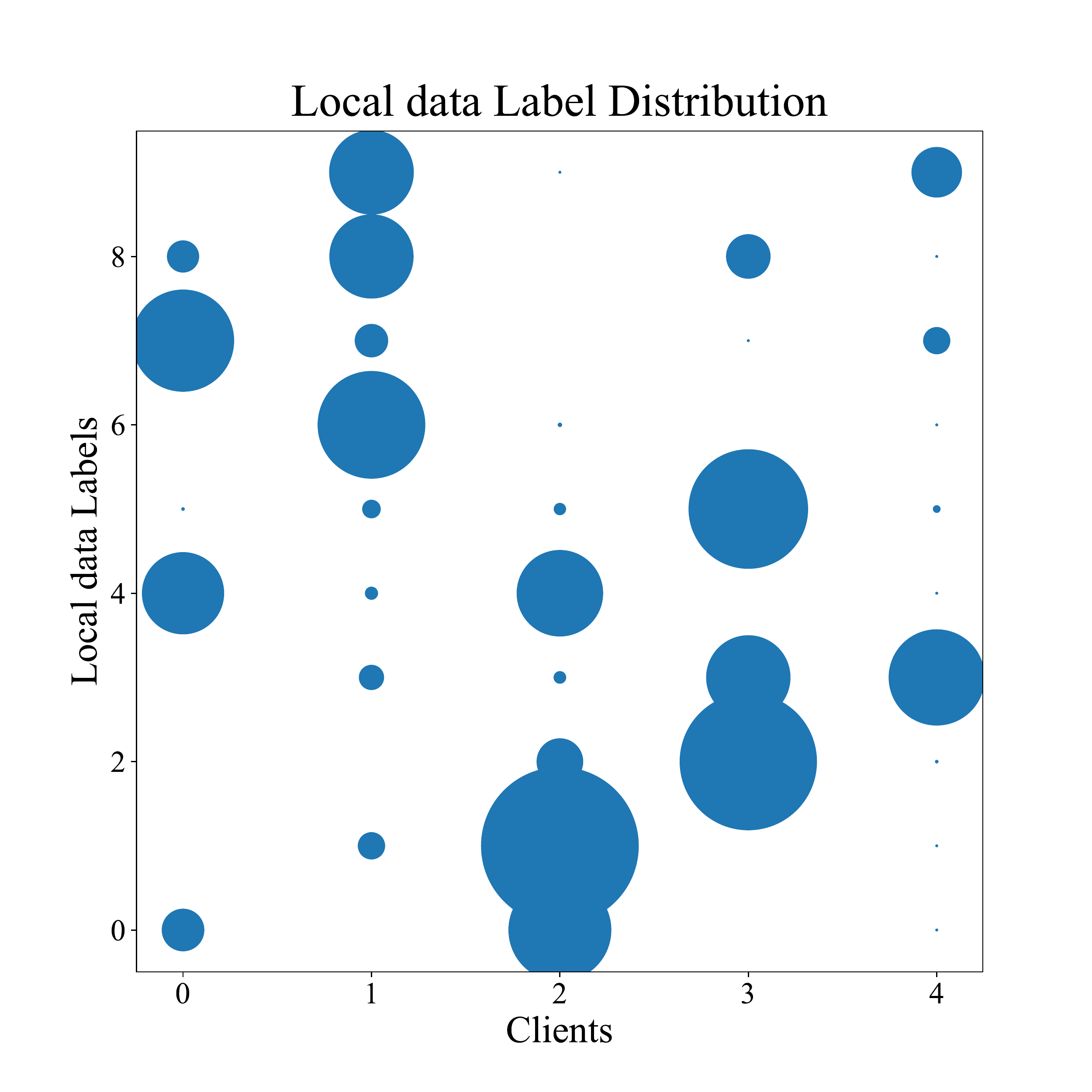}
			%\caption{fig1}
		\end{minipage}%
	}
	\centering
	\caption{The specific distribution of client labels in different datasets with 5 clients and label distribution skewness ($\alpha$). Where, the circle represents that the client contains that class of data, and the circle size represents the proportion of that class of labels in the total data of all clients. From left to right $\alpha$ is 0.01, 0.05, and 0.1.}
	\label{fig4}
\end{figure}

\begin{figure}[h]
	\centering
	\subfigure[The EMNIST dataset is divided under 10 clients according to the Dirichlet distribution.]{
		\begin{minipage}[t]{0.33\linewidth}
			\centering
			\includegraphics[width=1.9in]{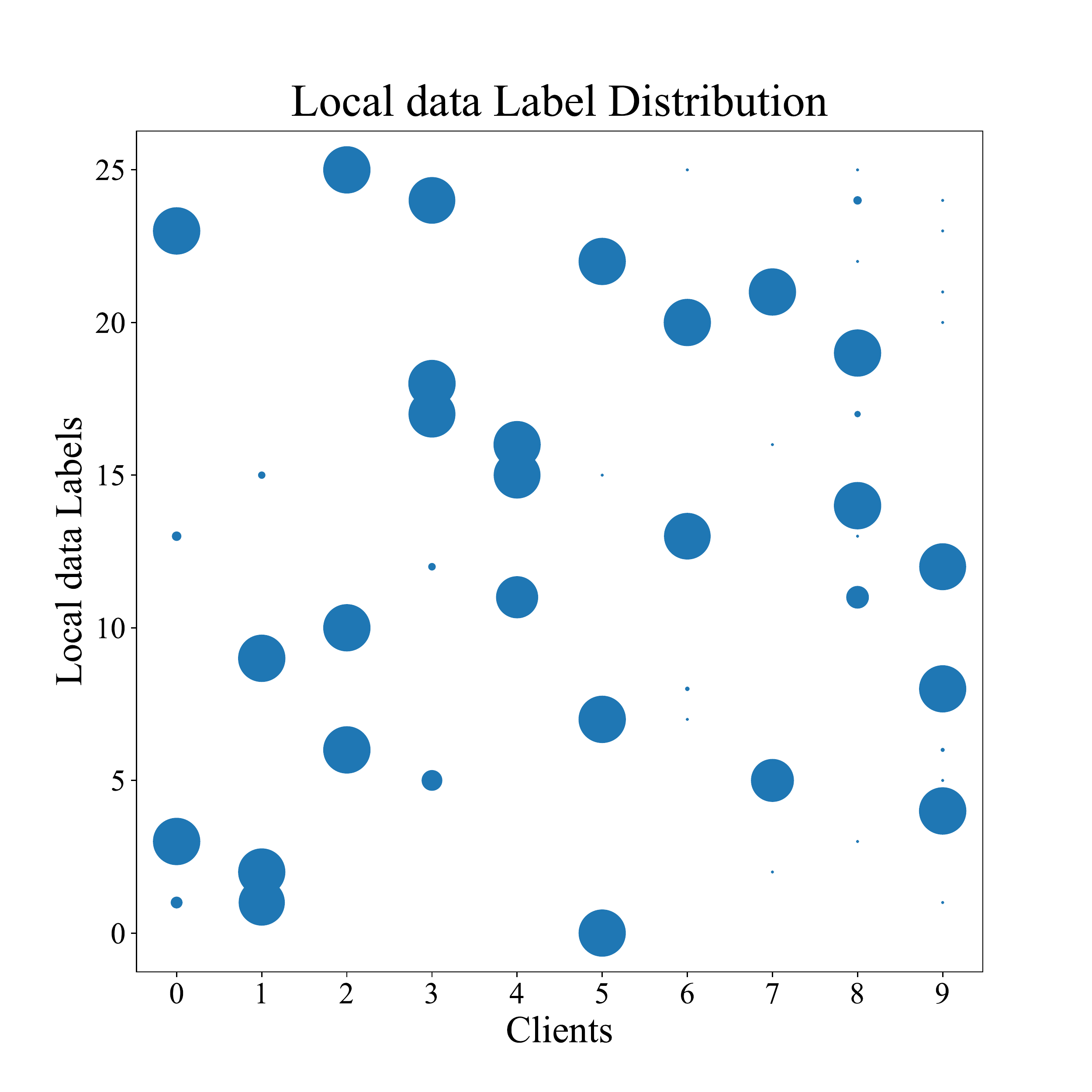}
			%\caption{fig1}
		\end{minipage}%
		\begin{minipage}[t]{0.33\linewidth}
			\centering
			\includegraphics[width=1.9in]{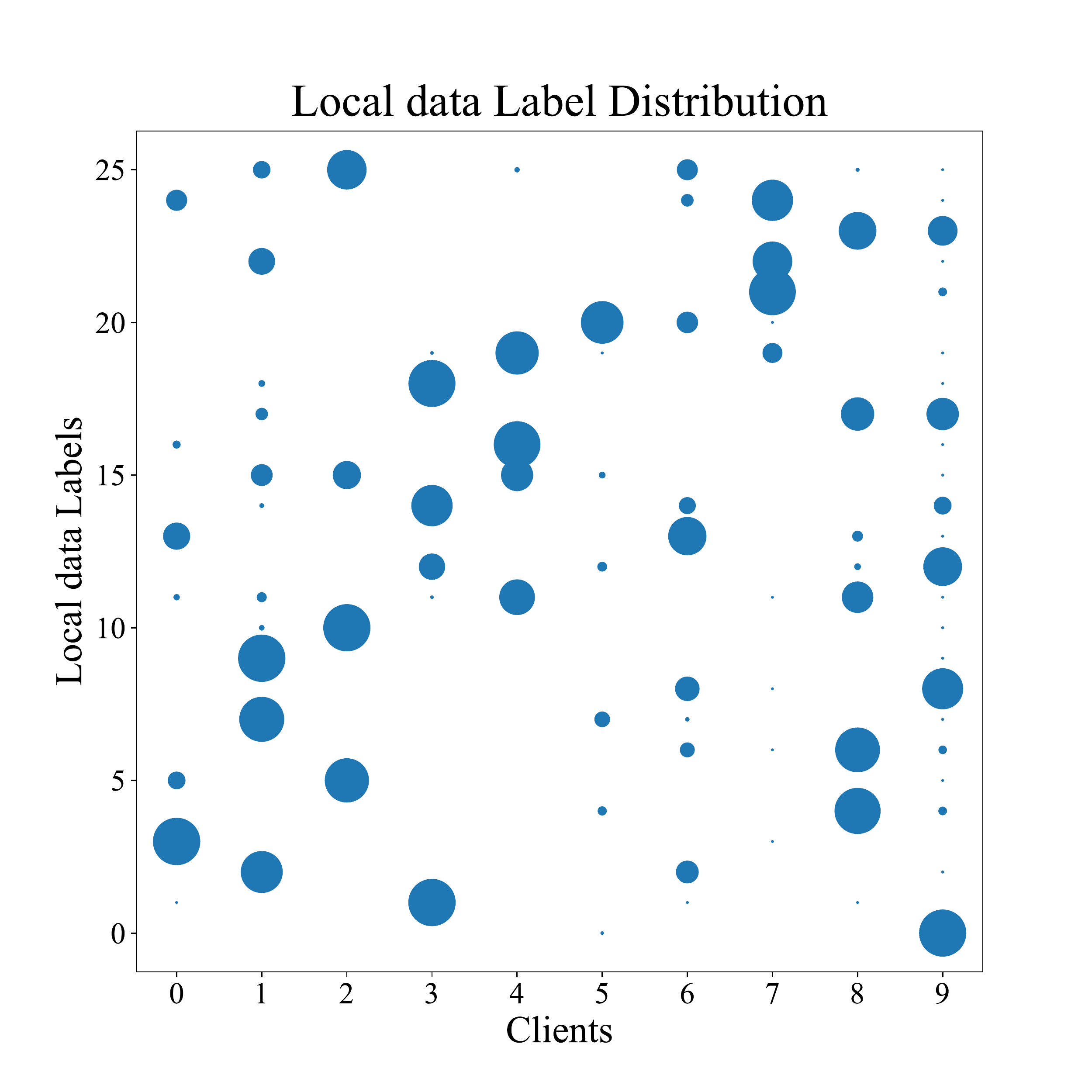}
			%\caption{fig1}
		\end{minipage}%
		\begin{minipage}[t]{0.33\linewidth}
			\centering
			\includegraphics[width=1.9in]{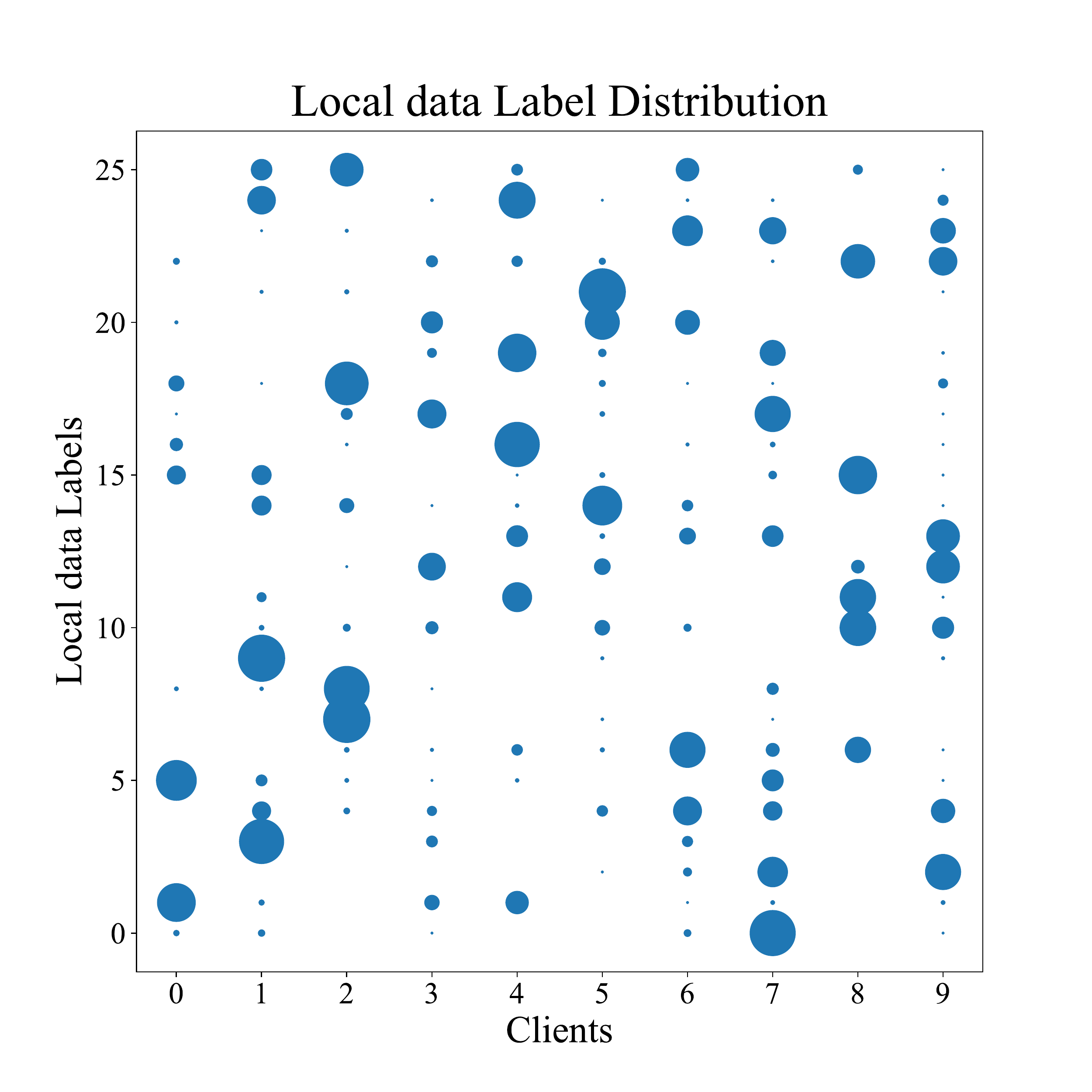}
			%\caption{fig1}
		\end{minipage}%
	}
	\subfigure[The FashionMNIST dataset is divided under 10 clients according to the Dirichlet distribution.]{
		\begin{minipage}[t]{0.33\linewidth}
			\centering
			\includegraphics[width=1.9in]{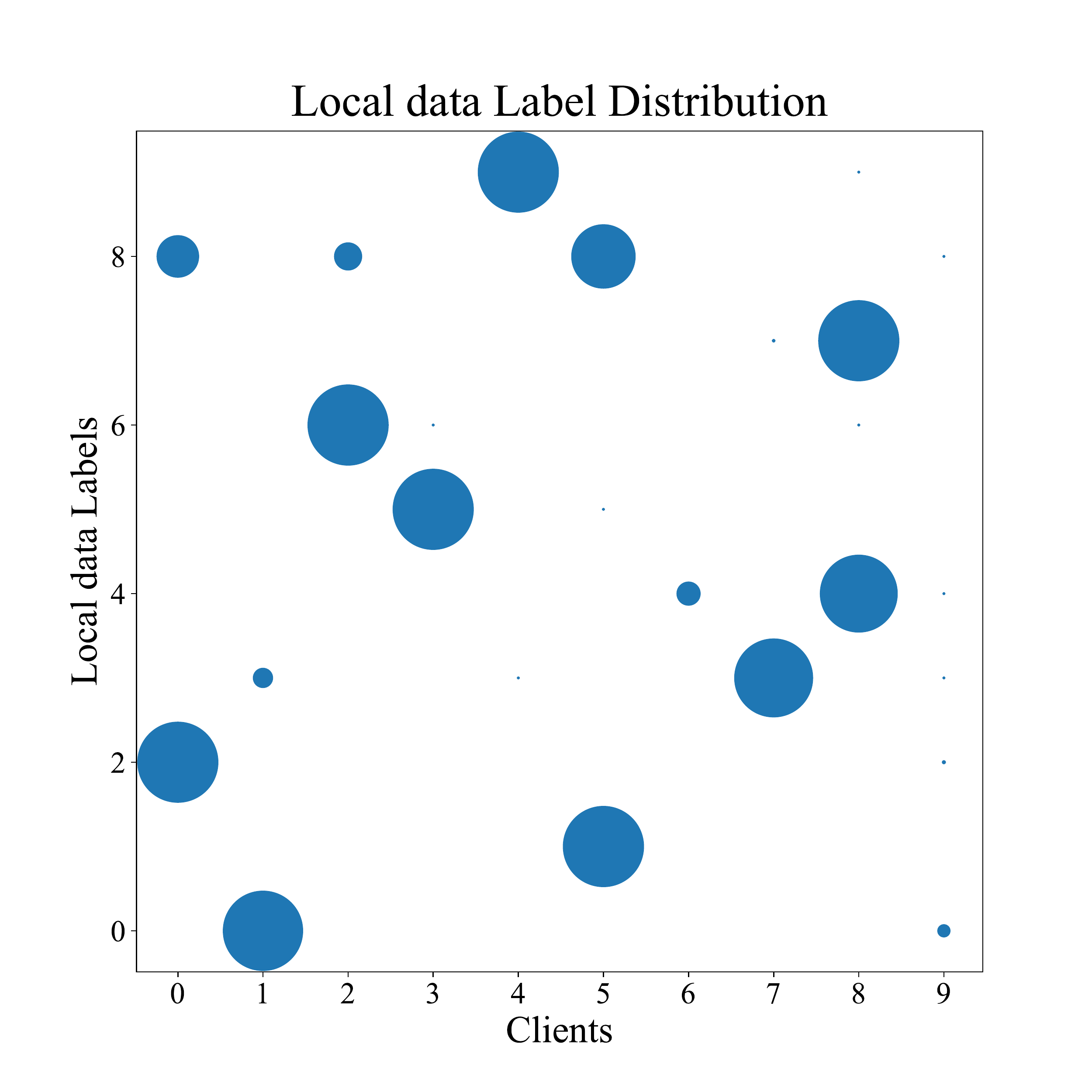}
			%\caption{fig1}
		\end{minipage}%
		\begin{minipage}[t]{0.33\linewidth}
			\centering
			\includegraphics[width=1.9in]{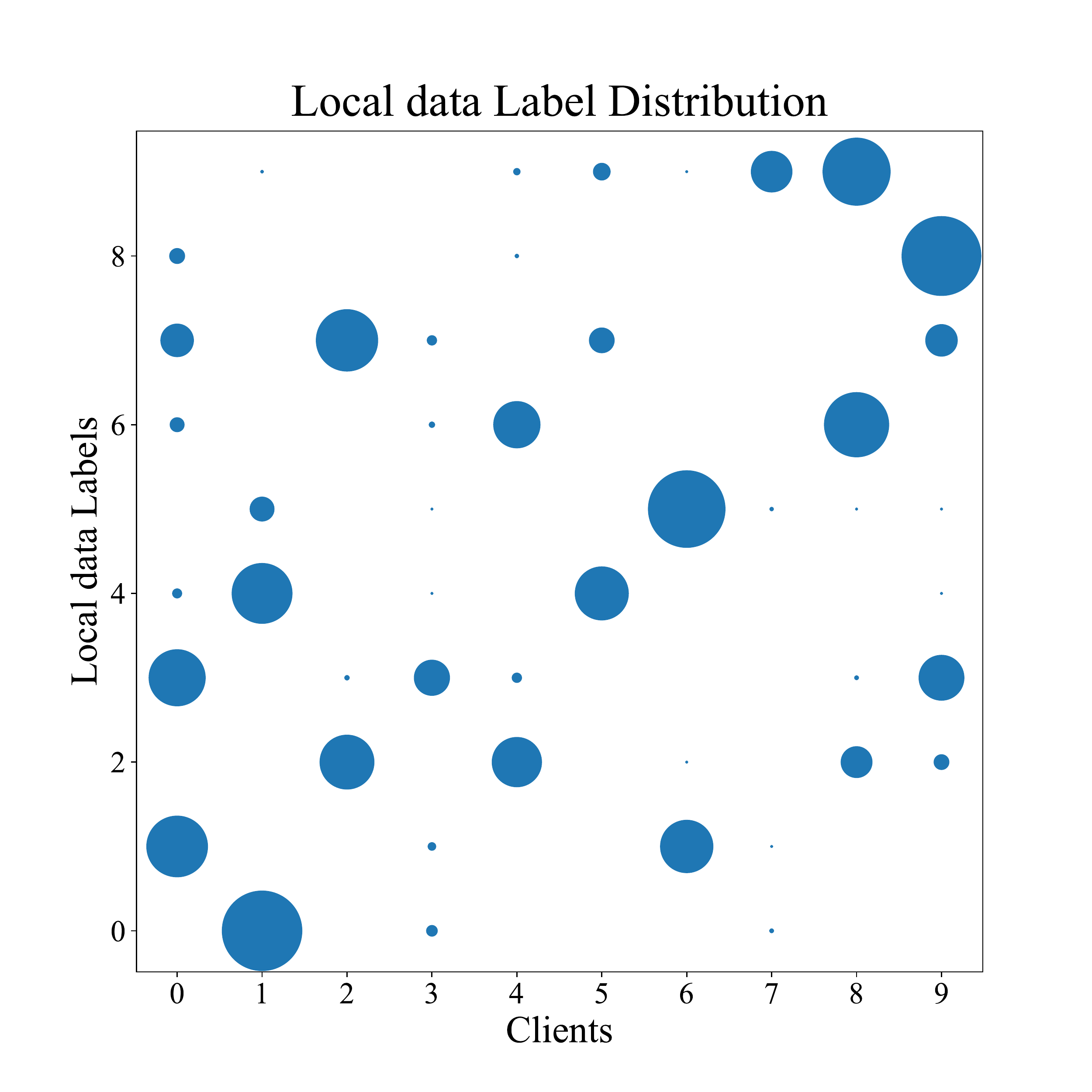}
			%\caption{fig1}
		\end{minipage}%
		\begin{minipage}[t]{0.33\linewidth}
			\centering
			\includegraphics[width=1.9in]{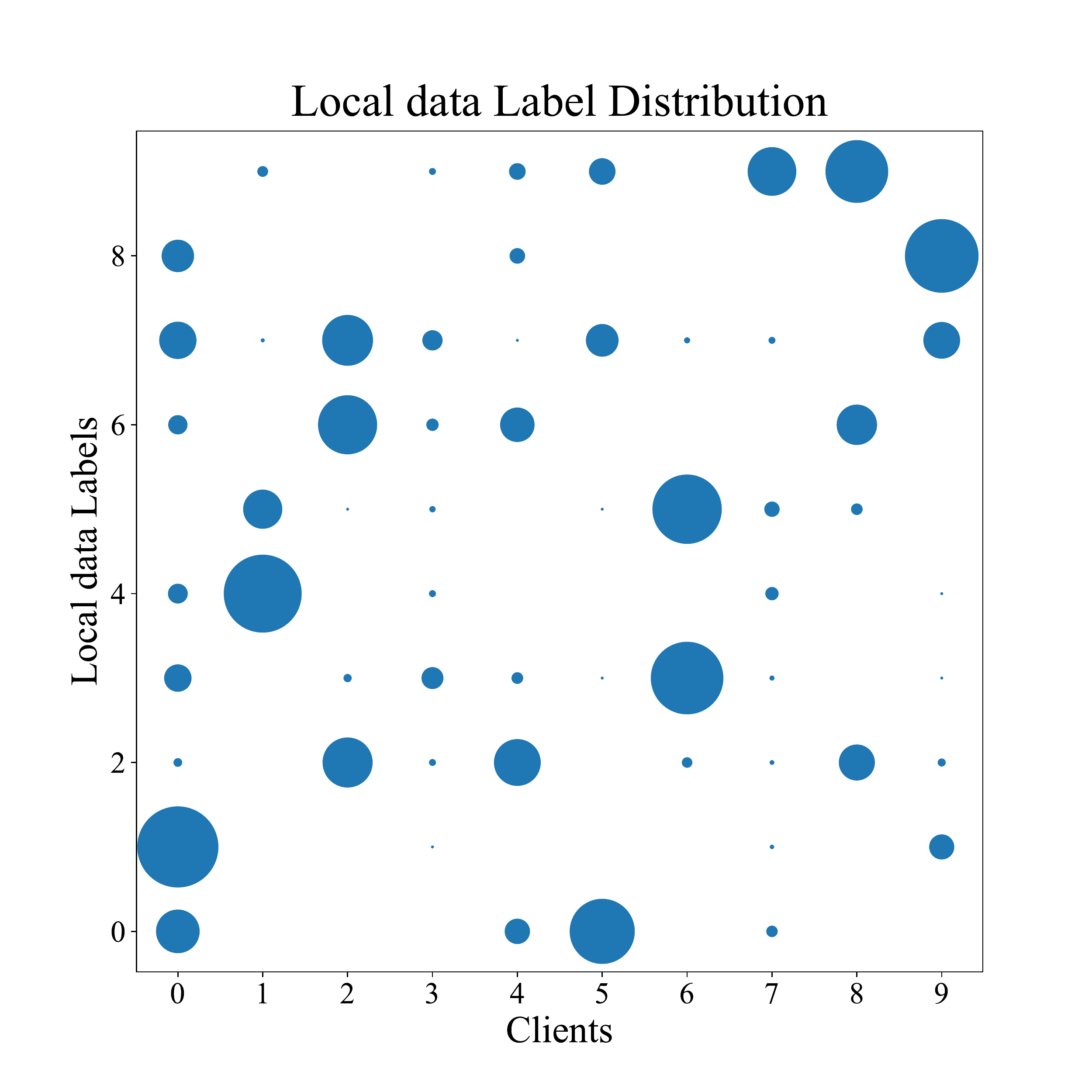}
			%\caption{fig1}
		\end{minipage}%
	}
    \subfigure[The SVHN dataset is divided under 10 clients according to the Dirichlet distribution.]{
		\begin{minipage}[t]{0.33\linewidth}
			\centering
			\includegraphics[width=1.9in]{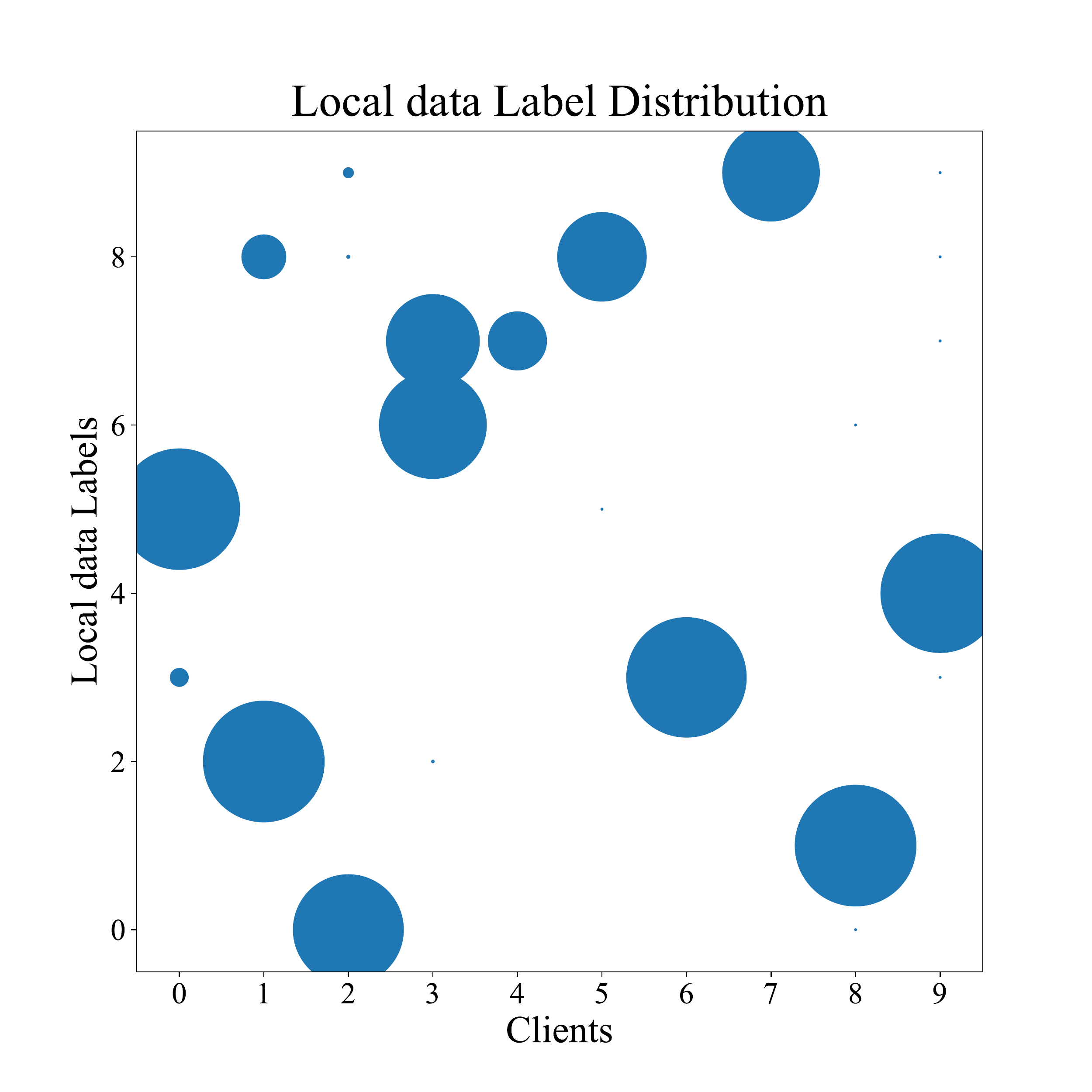}
			%\caption{fig1}
		\end{minipage}%
		\begin{minipage}[t]{0.33\linewidth}
			\centering
			\includegraphics[width=1.9in]{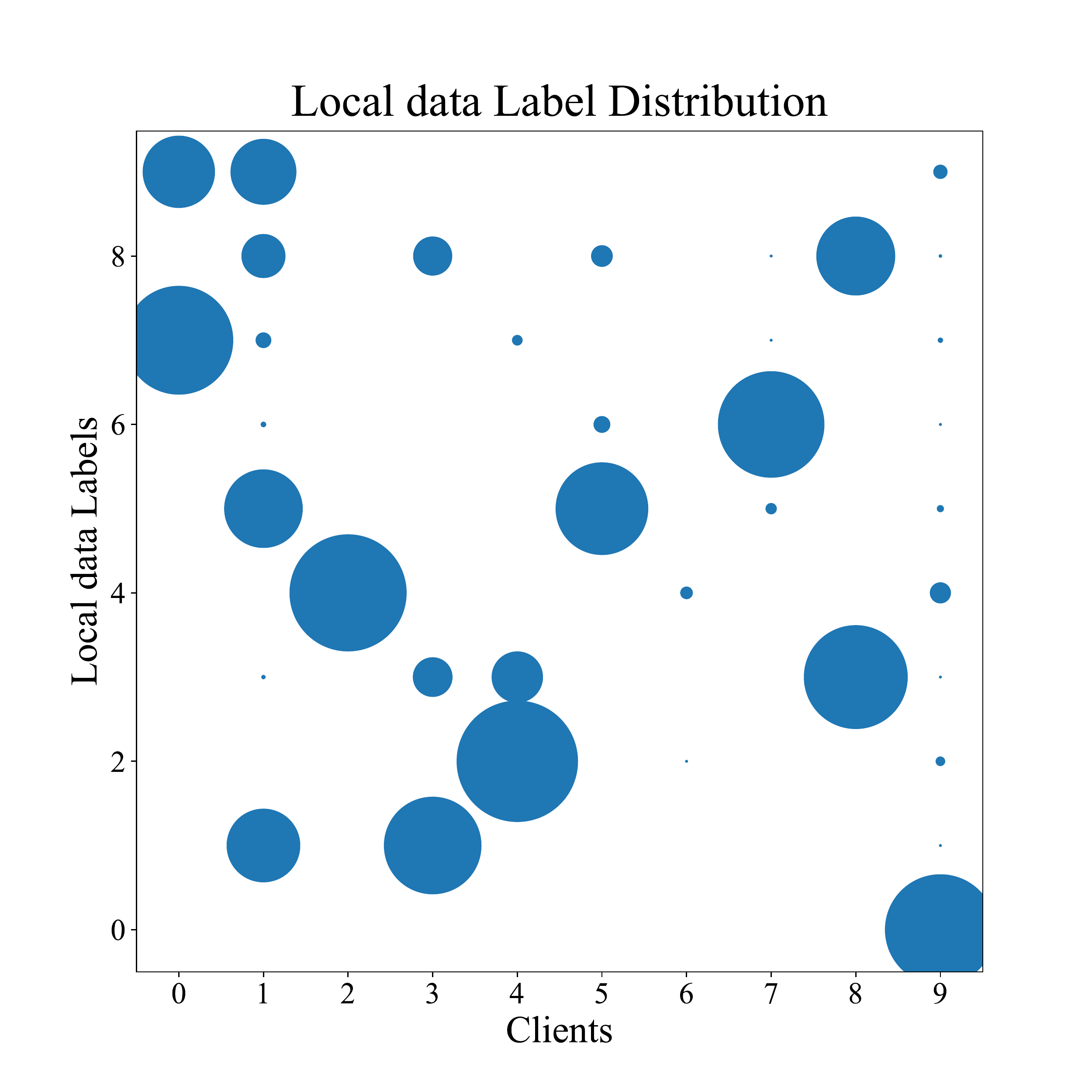}
			%\caption{fig1}
		\end{minipage}%
		\begin{minipage}[t]{0.33\linewidth}
			\centering
			\includegraphics[width=1.9in]{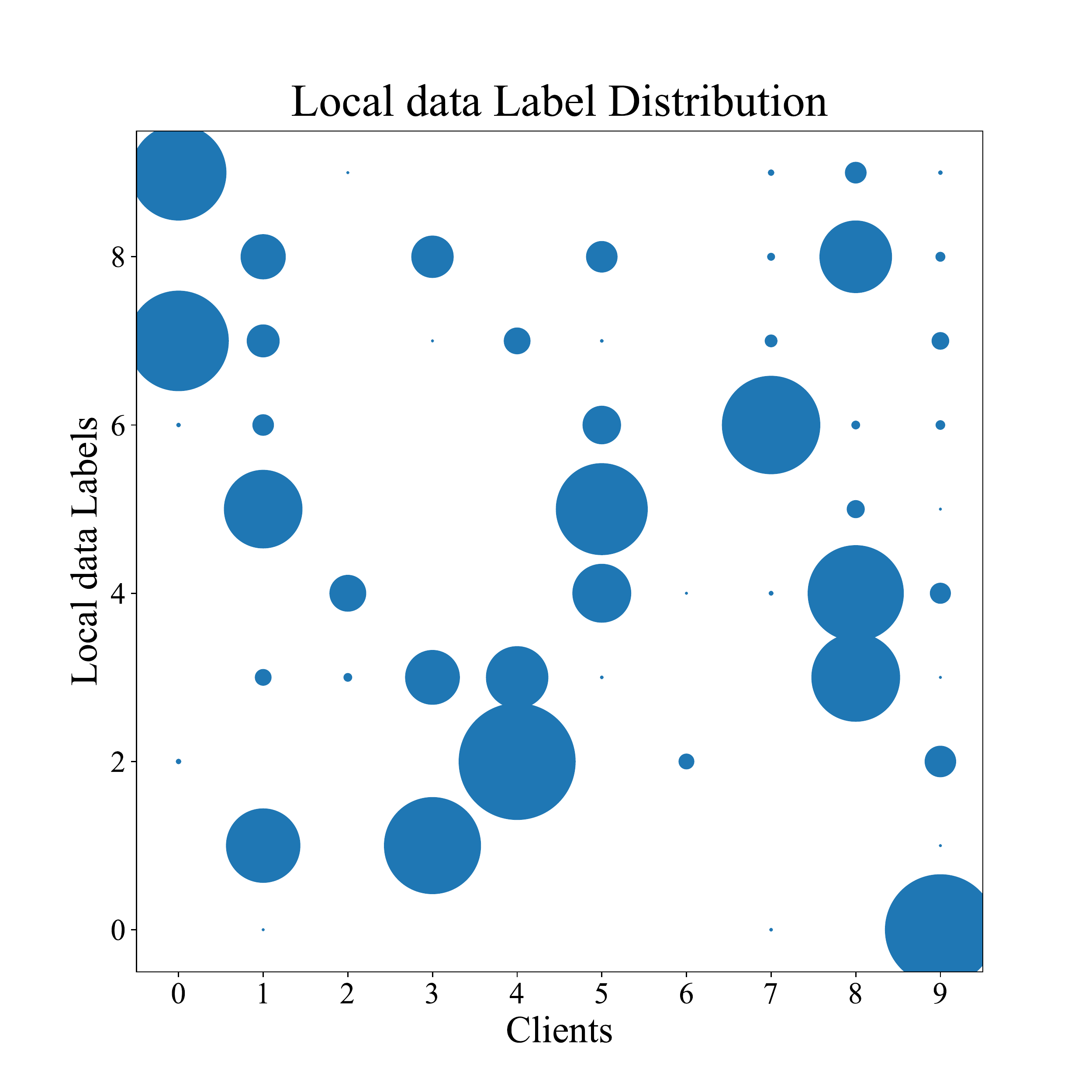}
			%\caption{fig1}
		\end{minipage}%
	}
	\centering
	\caption{The specific distribution of client labels in different datasets with 10 clients and label distribution skewness ($\alpha$). Where, the circle represents that the client contains that class of data, and the circle size represents the proportion of that class of labels in the total data of all clients. From left to right $\alpha$ is 0.01, 0.05, and 0.1.}
	\label{fig4.4}
\end{figure}

\subsubsection{Distribution of Labels among Clients} \label{4.1.2}
Following the existing works \cite{29,6,27}, we use Dirichlet distribution $\textbf{Dir}(\alpha)$ to simulate the data distribution among clients in the scenario of label distribution skew, where the value of $\alpha$ controls the degree of label distribution skew. The larger the value of $\alpha$ means the smaller the difference in label distribution among clients.
Specifically, we set $\alpha$ to 0.1, 0.05, and 0.01 to compare the effect of $\alpha$ on algorithms. 

In Figure~\ref{fig4}, we show the specific distribution of client labels under 5 clients and label distribution skewness ($\alpha$). The circle represents that the client contains that category of data, and the circle size represents the proportion of that class of labels in the total data of all clients. In addition, we present in Figure~\ref{fig4.4} the division of the different datasets according to the Dirichlet distribution for 10 clients.
% Here we only list the division of the EMNIST dataset in clients, and the division of the rest of the dataset can be found in the Appendix~\ref{Appendix}.

\subsubsection{Baselines}
We compare FedMGD with the following baselines to evaluate the effectiveness of FedMGD from different perspectives. First, we compare FedMGD with several state-of-the-art federated learning benchmarks: FedAvg~\cite{2}, FedProx~\cite{4}, SCAFFOLD~\cite{13}, FedDF~\cite{6}, and FedGen~\cite{27}. In addition, to clearly demonstrate the performance of our proposed method for global distribution modeling, we also compare a series of existing distributed generative adversarial network methods: F2U~\cite{10}, MD-GAN~\cite{9}, and FedGAN~\cite{16}.

% First, to verify the performance of FedMGD in the scenario of label distribution skew, we compare the accuracy of FedMGD with FedAvg~\cite{2}, FedProx~\cite{4}, SCAFFOLD~\cite{13}, FedDF~\cite{6}, and FedGen~\cite{27} on classification tasks.
% And, we compare the variability of the performance of different models across clients, which we call local fairness.
% Then, in subsequent ablation experiments, we selected FedDF, which uses real data collected from the client as a distillation data source, and a series of distributed generation methods: F2U, MD-GAN, and FedGAN, respectively, as baseline algorithms for comparison in a scenario with label distribution skew.
% Then, in subsequent ablation experiments, we selected FedDF~\cite{6}, which uses real data collected from the client as a distillation data source, and a series of distributed generation methods: F2U~\cite{10}, MD-GAN~\cite{9}, and FedGAN~\cite{16}, respectively, as baseline algorithms for comparison in a scenario with label distribution skew.

\begin{table}[t]
	\begin{center}
		\caption{\textbf{Accuracy comparison on the global test set.} FedMGD and state-of-the-art methods train the same model on scenarios with different data sets and data distributions and test the accuracy of the model using a global test set (\%).}
		\renewcommand{\arraystretch}{1.2}
		\resizebox{1.\columnwidth}{!}{
			\begin{tabular}{c|c|c|c|c|c|c|c|c|c}
\hline
\textbf{Dataset}                                                                  & \textbf{\begin{tabular}[c]{@{}c@{}}Client\\ Num\end{tabular}} & $\boldsymbol{\alpha}$ & \textbf{Local} & \textbf{FedAvg}     & \textbf{FedProx} & \textbf{FedDF}      & \textbf{FedGen} & \textbf{SCAFFOLD}   & \textbf{FedMGD}     \\ \hline
\multirow{6}{*}{\textbf{EMNIST}}                                                  & \multirow{3}{*}{5}                                  & 0.01     & 24.36±0.23     & 86.56±0.95          & 85.43±0.61       & 88.06±0.37          & 82.41±2.34      & 85.30±0.37          & \textbf{89.00±0.93} \textcolor{green!70!black}{($\uparrow$2.44)} \\
                                                                                  &                                                               & 0.05     & 33.20±0.29     & 89.33±0.16          & 87.97±0.40       & 89.27±0.27          & 86.86±0.89      & 89.22±0.21          & \textbf{91.15±0.35} \textcolor{green!70!black}{($\uparrow$1.82)} \\
                                                                                  &                                                               & 0.1      & 36.86±0.26     & 90.85±0.31          & 89.36±0.55       & 90.32±0.26          & 90.12±0.63      & \textbf{91.65±0.33} & 91.52±0.17 \textcolor{red!70!black}{($\downarrow$0.13)}          \\ \cline{2-10} 
                                                                                  & \multirow{3}{*}{10}                                  & 0.01     & 13.38±0.08     & 65.98±3.95          & 77.09±1.49       & 65.72±1.33          & 66.74±8.45      & 69.23±1.47          & \textbf{85.04±0.81} \textcolor{green!70!black}{($\uparrow$15.82)} \\
                                                                                  &                                                               & 0.05     & 19.03±0.03     & 82.32±0.35          & 83.23±0.71       & 83.19±1.27          & 81.05±1.69      & 84.06±1.24          & \textbf{84.87±0.29} \textcolor{green!70!black}{($\uparrow$0.81)} \\
                                                                                  &                                                               & 0.1      & 32.22±0.02     & 88.69±0.47          & 87.68±0.47       & \textbf{88.97±0.32} & 88.28±0.49      & 87.88±0.81          & 87.83±0.33 \textcolor{red!70!black}{($\downarrow$1.14)}          \\ \hline
\multirow{6}{*}{\textbf{\begin{tabular}[c]{@{}c@{}}Fashion\\ MNIST\end{tabular}}} & \multirow{3}{*}{5}                                   & 0.01     & 25.71±0.85     & 68.38±6.69          & 69.64±2.14       & 71.02±0.72          & 55.67±7.13      & 55.58±1.90          & \textbf{84.04±0.58} \textcolor{green!70!black}{($\uparrow$13.02)} \\
                                                                                  &                                                               & 0.05     & 39.54±1.04     & 83.37±1.84          & 81.25±1.13       & 83.67±0.80          & 79.61±2.59      & 77.27±0.68          & \textbf{87.57±0.40} \textcolor{green!70!black}{($\uparrow$3.90)} \\
                                                                                  &                                                               & 0.1      & 49.15±0.19     & 88.28±0.89 & 86.68±0.89       & 87.53±0.95          & 84.15±2.21      & 86.10±1.20          & \textbf{89.29±1.33} \textcolor{green!70!black}{($\uparrow$1.01)} \\ \cline{2-10} 
                                                                                  & \multirow{3}{*}{10}                                  & 0.01     & 20.65±0.85     & 53.24±3.20          & 73.03±2.24       & 57.48±1.91          & 56.74±3.61      & 45.42±8.53          & \textbf{74.37±4.44} \textcolor{green!70!black}{($\uparrow$1.34)} \\
                                                                                  &                                                               & 0.05     & 31.23±0.21     & 80.32±3.25          & 84.30±1.12       & 79.65±4.39          & 80.95±1.23      & 81.70±0.49          & \textbf{85.71±0.45} \textcolor{green!70!black}{($\uparrow$3.90)} \\
                                                                                  &                                                               & 0.1      & 41.61±0.73     & 85.94±1.51          & 86.76±0.17       & 83.97±3.62          & 85.23±2.44      & 86.50±0.45          & \textbf{86.79±0.90} \textcolor{green!70!black}{($\uparrow$0.29)} \\ \hline
\multirow{6}{*}{\textbf{SVHN}}                                                    & \multirow{3}{*}{5}                                   & 0.01     & 18.11±0.09     & 73.07±0.25          & 76.70±0.92       & 72.67±0.79          & 57.57±1.47      & 83.05±0.87          & \textbf{84.14±0.91} \textcolor{green!70!black}{($\uparrow$1.09)} \\
                                                                                  &                                                               & 0.05     & 26.80±0.38     & 84.98±0.92          & 85.60±0.68       & 84.99±0.69          & 67.34±0.84      & 87.66±0.69          & \textbf{88.62±0.39} \textcolor{green!70!black}{($\uparrow$0.96)} \\
                                                                                  &                                                               & 0.1      & 29.23±0.44     & 88.43±0.91          & 87.91±0.18       & 88.65±0.22          & 68.45±4.17      & 89.83±0.47          & \textbf{90.47±0.37} \textcolor{green!70!black}{($\uparrow$0.64)} \\ \cline{2-10} 
                                                                                  & \multirow{3}{*}{10}                                  & 0.01     & 14.26±0.18     & 46.39±1.90          & 61.71±1.58       & 47.47±2.56          & 24.41±1.23      & 53.39±0.99          & \textbf{76.00±0.79} \textcolor{green!70!black}{($\uparrow$14.29)} \\
                                                                                  &                                                               & 0.05     & 13.17±0.06     & 70.62±0.93          & 75.09±0.52       & 72.63±2.20          & 52.02±2.94      & 76.04±1.27          & \textbf{76.87±1.26} \textcolor{green!70!black}{($\uparrow$0.83)} \\
                                                                                  &                                                               & 0.1      & 14.12±0.95     & 80.09±0.51          & 78.11±0.17       & \textbf{80.17±0.52} & 55.55±5.75      & 79.77±0.44          & 79.85±0.10  \textcolor{red!70!black}{($\downarrow$0.32)}         \\ \hline
\multicolumn{1}{l|}{\multirow{6}{*}{\textbf{CIFAR10}}}                            & \multirow{3}{*}{5}                                   & 0.01     & 16.69±0.39     & 46.73±0.90          & 51.71±1.21       & 47.25±1.40          & 26.73±1.19      & 54.46±0.99          & \textbf{62.61±1.54} \textcolor{green!70!black}{($\uparrow$8.15)} \\
\multicolumn{1}{l|}{}                                                             &                                                               & 0.05     & 28.72±0.39     & 61.61±1.36          & 60.08±3.19       & 60.27±0.39          & 41.86±0.47      & 64.28±1.43          & \textbf{66.52±0.49} \textcolor{green!70!black}{($\uparrow$2.24)} \\
\multicolumn{1}{l|}{}                                                             &                                                               & 0.1      & 33.00±1.16     & 65.77±1.77          & 65.07±0.40       & 64.58±0.95          & 46.61±2.88      & 67.37±1.02          & \textbf{69.31±0.33} \textcolor{green!70!black}{($\uparrow$1.94)} \\ \cline{2-10} 
\multicolumn{1}{l|}{}                                                             & \multirow{3}{*}{10}                                  & 0.01     & 15.76±0.04     & 38.79±4.97          & 45.98±0.58       & 37.06±1.26          & 26.67±1.10      & 46.09±2.50          & \textbf{55.13±0.72} \textcolor{green!70!black}{($\uparrow$9.04)}           \\
\multicolumn{1}{l|}{}                                                             &                                                               & 0.05     & 24.95±0.87     & 52.96±0.24          & 51.68±0.32       & 52.07±1.97          & 27.51±1.76      & 53.01±0.74          & \textbf{58.26±1.31} \textcolor{green!70!black}{($\uparrow$5.25)} \\
\multicolumn{1}{l|}{}                                                             &                                                               & 0.1      & 35.04±1.54     & 58.15±0.94          & 56.36±0.26       & 57.89±1.00          & 43.08±0.55      & 60.04±1.08          & \textbf{60.60±0.62} \textcolor{green!70!black}{($\uparrow$0.56)} \\ \hline
\end{tabular}
		}
% 		}
		\label {tab1}
	\end{center}
\end{table}

\subsubsection{Implementation Details}
We implement all the code of FedMGD in PyTorch, where the structure of the generator and multiple discriminators is based on PatchGAN~\cite{31} and implement with the 9-blocks of ResNet~\cite{20}.
For FedGen's generator, we use the structure in the original paper~\cite{27} to output the feature representation of samples after the input has passed through a hidden layer.
Furthermore, the classifier in all experiments is a standard Convolutional Neural Network (CNN), which consists of two convolutional layers and one fully connected (FC) layer.

For all methods, we set the number of local training epochs $E = 10$, the size of training batch $= 32$, and are optimized by Adam optimizer with an initial learning rate of 0.0002. 
For FedMGD, in the generative adversarial stage the server sends generated data of size 64 to the clients for GAN training at each round of communication. In the federated enhancement stage, the generator synthesizes samples of size 2048 per round to refine the global model.
% In the classification task we conducted a number of experiments and selected the accuracy of the last round after averaging as the final result.
In the classification task, we conduct several experiments and choose the accuracy after the last round of averaging as the final result.
In addition, some of the curves are smoothed for better presentation of the results.
Finally, to ensure the effectiveness of our method on different number of clients, we have verified it on 5 and 10 clients respectively.

\subsection{Comparison with State-of-the-art Methods} \label{4.2}
In this section, we compare the performance of FedMGD and state-of-the-art methods from two main perspectives: global and local. First, from a global perspective, we use accuracy as a metric to verify the performance of FedMGD and SOAT algorithms in the global test set. Then, from a local perspective, we compare the difference in performance performance between FedMGD and state-of-the-art algorithms in terms of the fairness of the model among clients.

\subsubsection{Accuracy on The Global Test Set.}
\begin{figure}[t!]
	\centering
	\subfigure[$\alpha$=0.01]{
    	\begin{minipage}[t]{0.32\linewidth}
    		\centering
    		\includegraphics[width=1.9in]{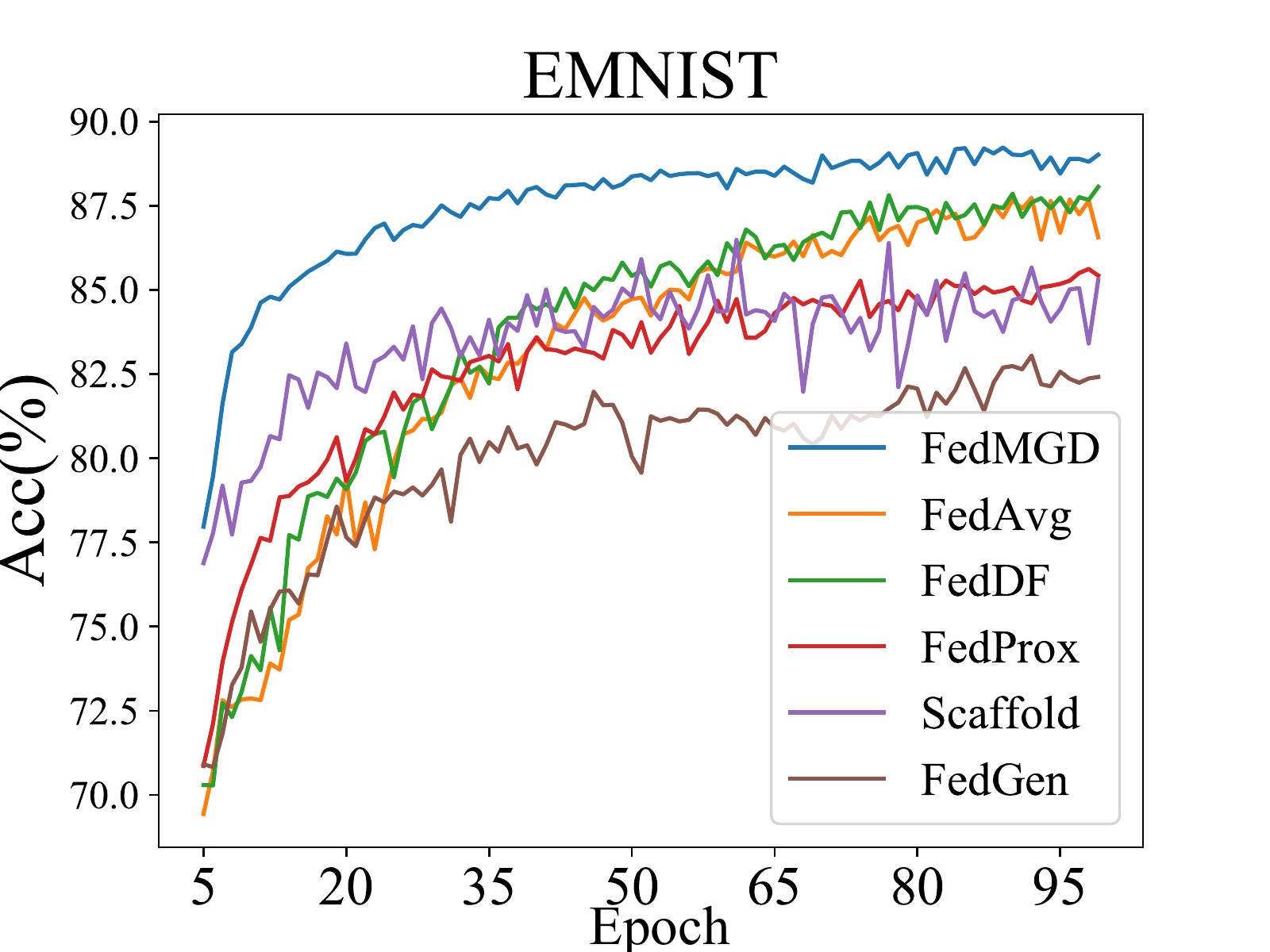}
    		\includegraphics[width=1.9in]{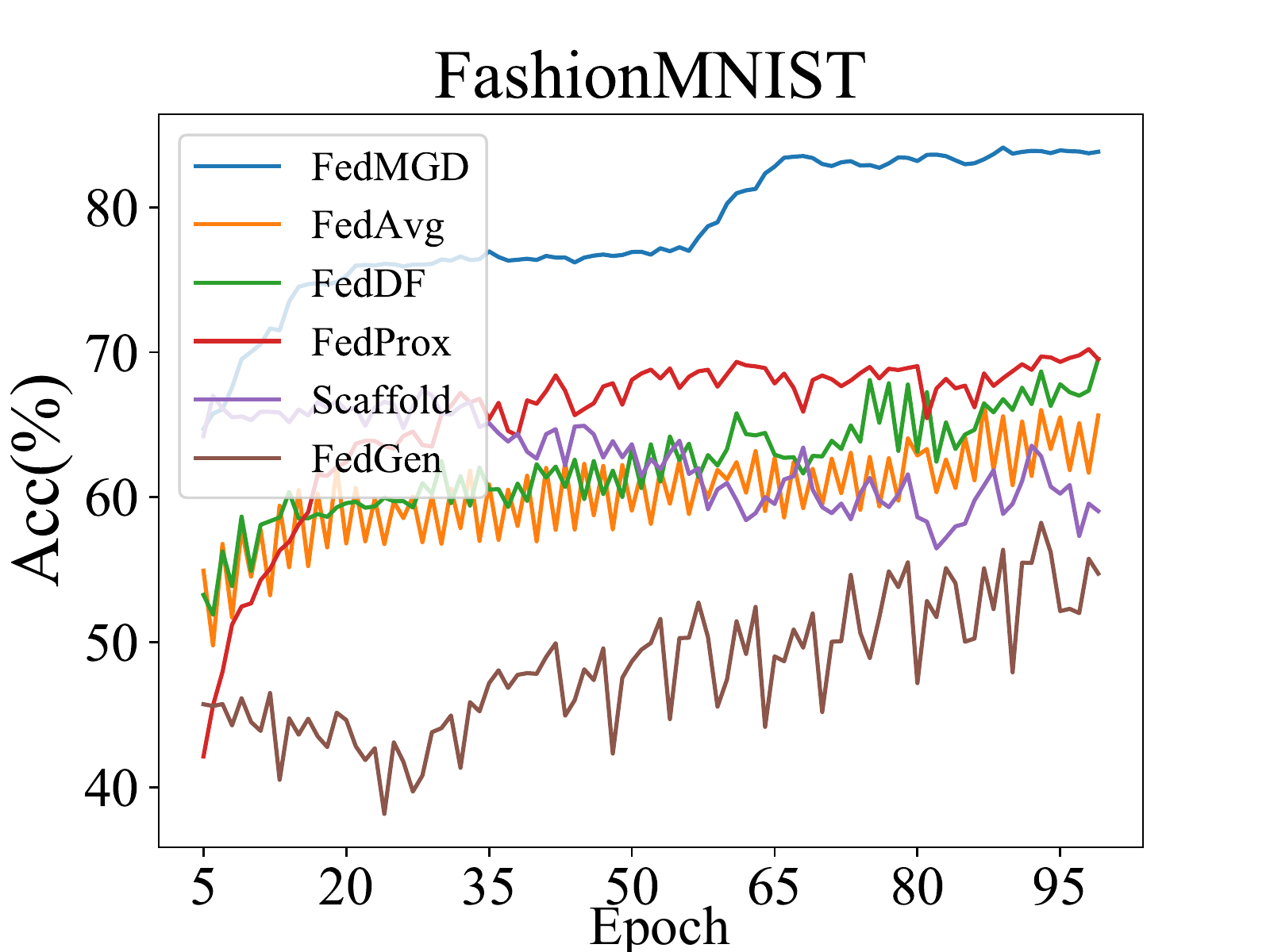}
    		\includegraphics[width=1.9in]{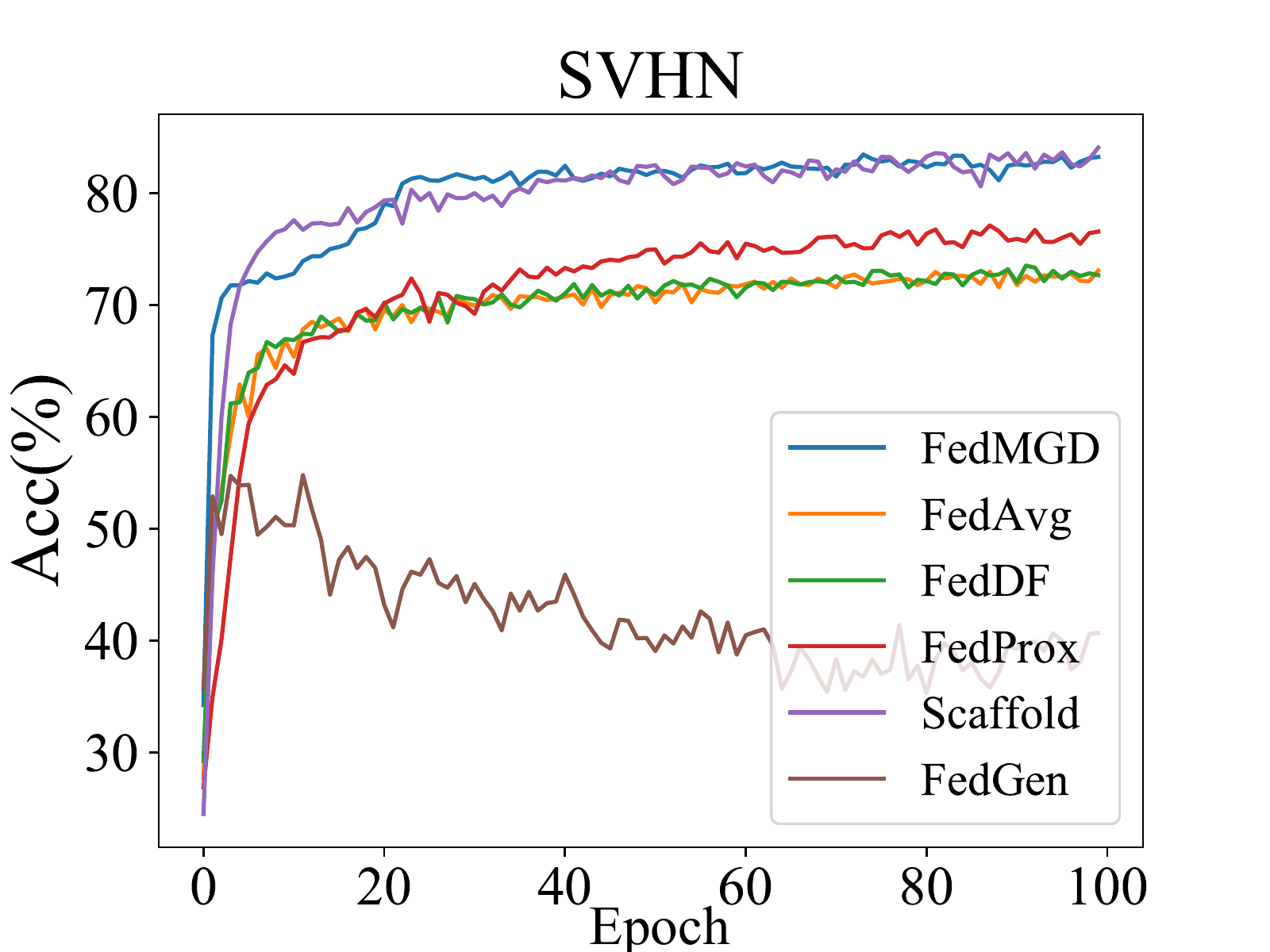}
    		\includegraphics[width=1.9in]{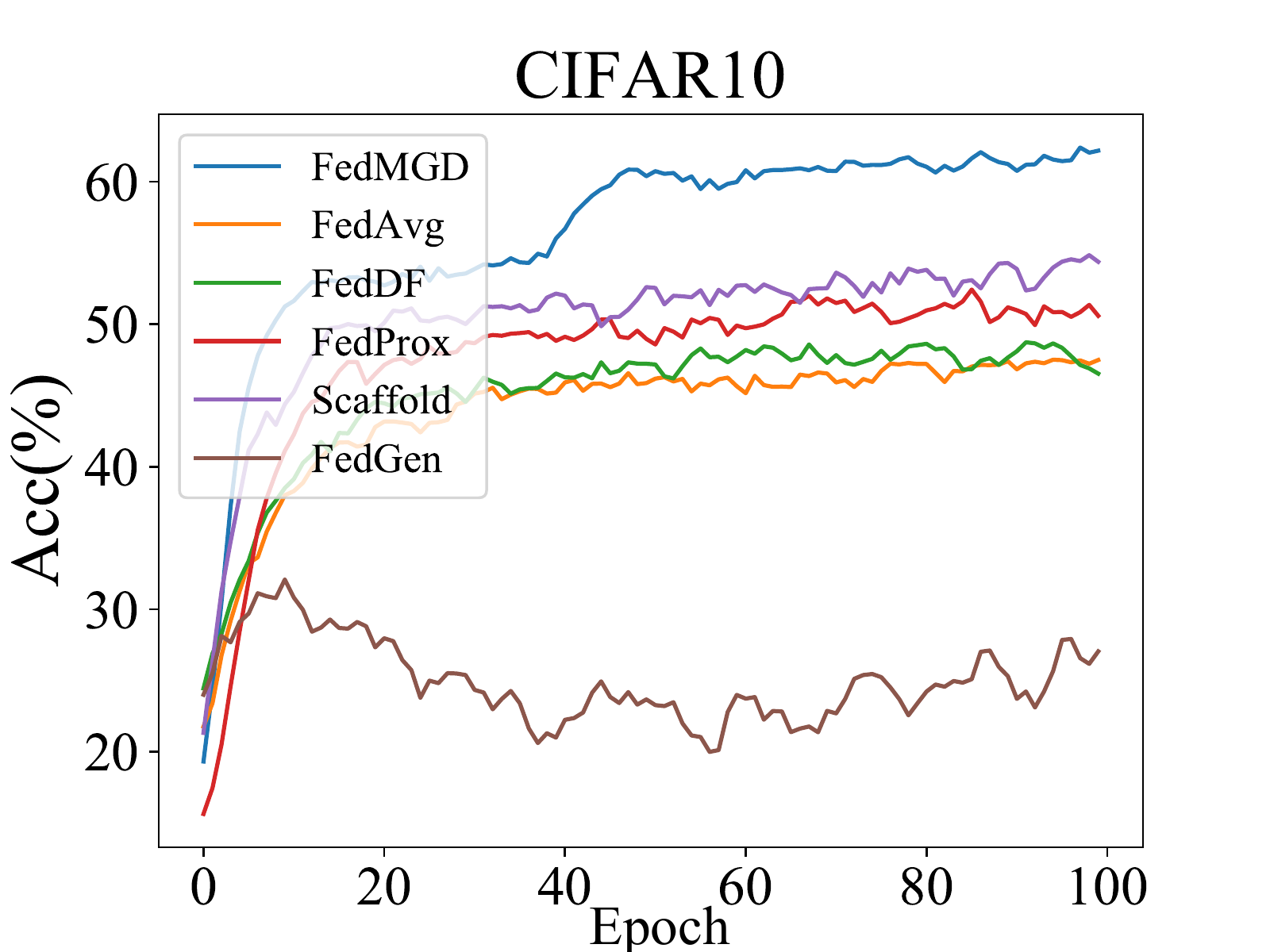}
    		%\caption{fig1}
    	\end{minipage}%
	}
	\subfigure[$\alpha$=0.05]{
    	\begin{minipage}[t]{0.32\linewidth}
    		\centering
    		\includegraphics[width=1.9in]{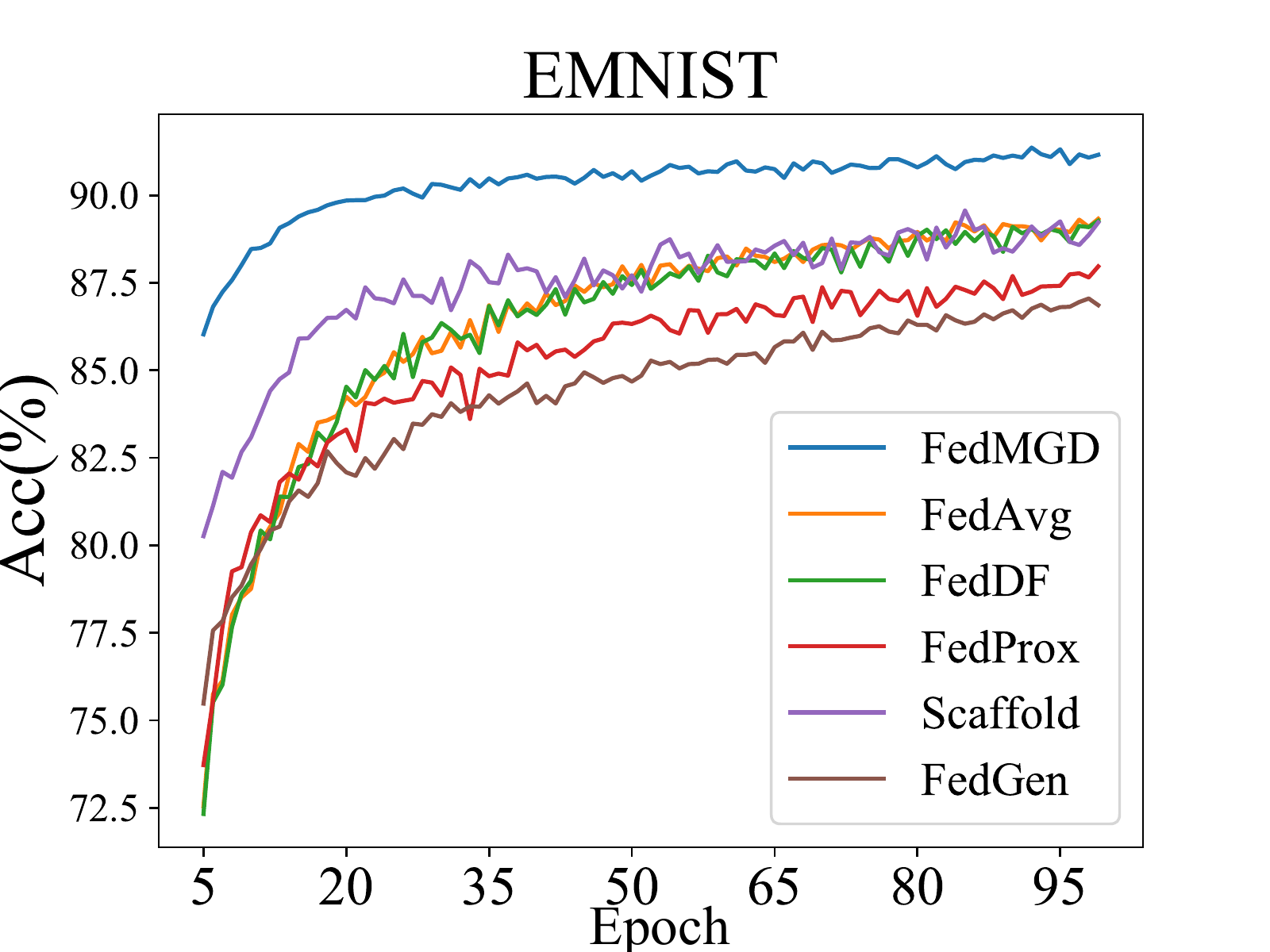}
    		\includegraphics[width=1.9in]{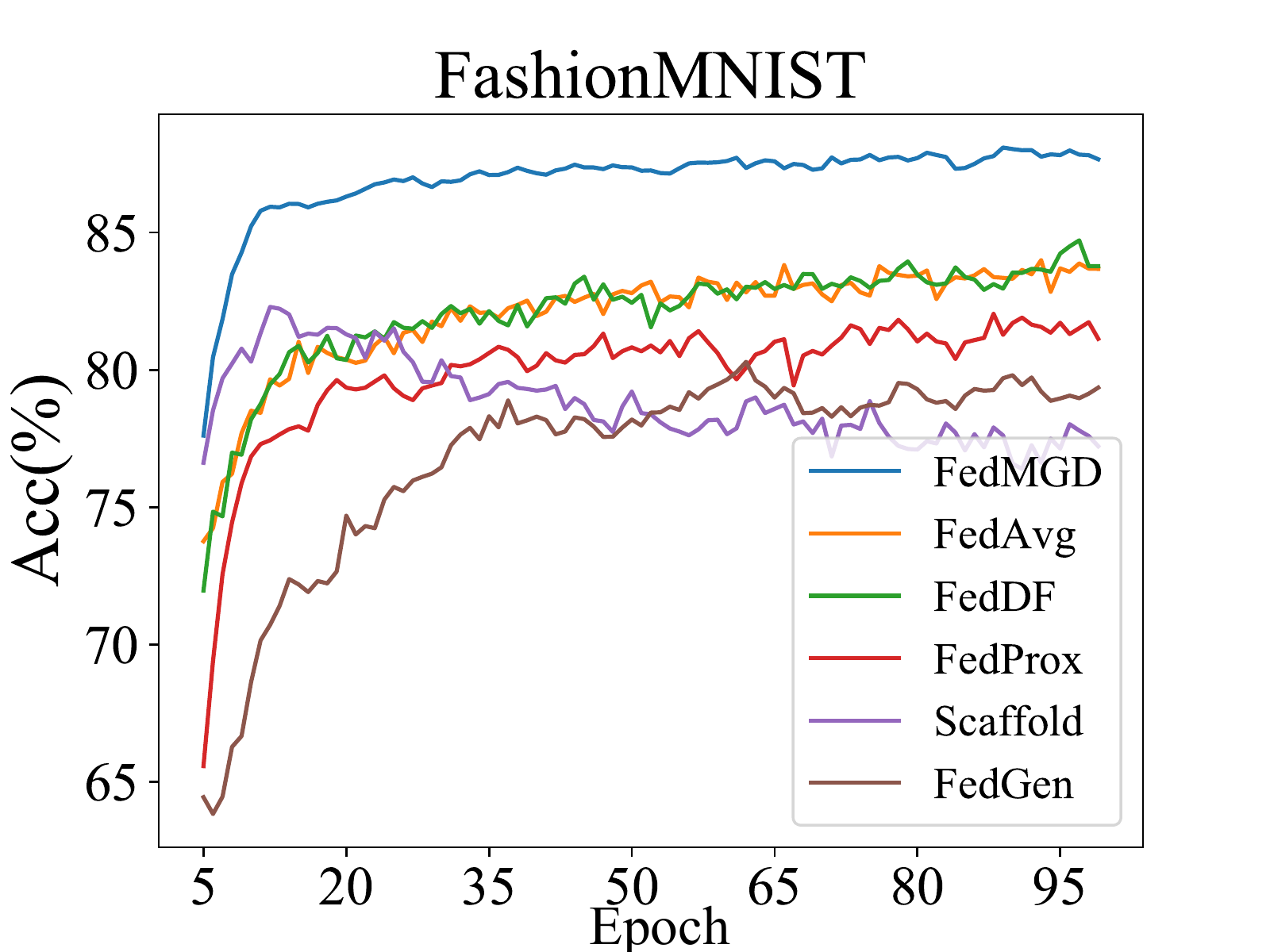}
    		\includegraphics[width=1.9in]{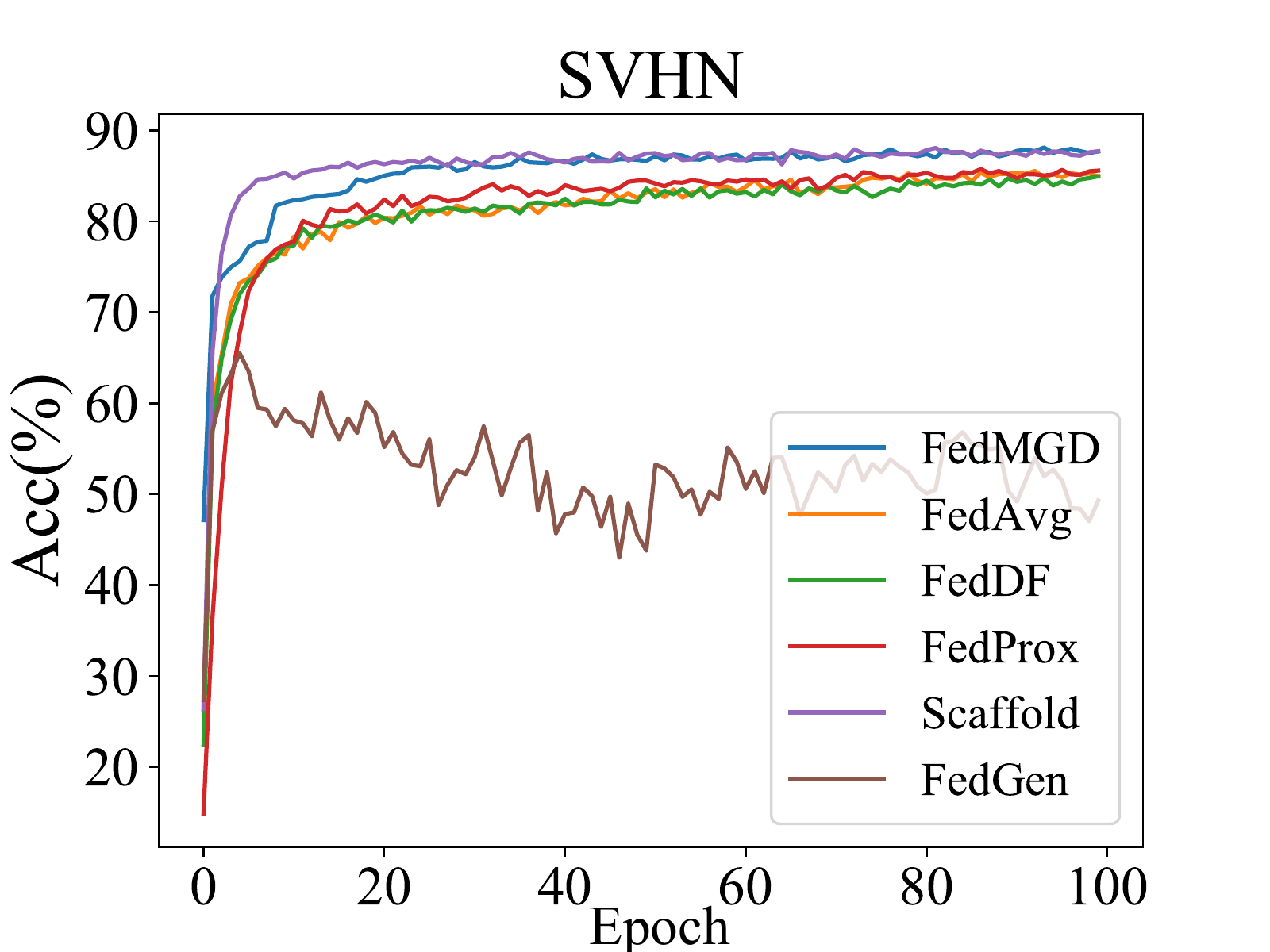}
    		\includegraphics[width=1.9in]{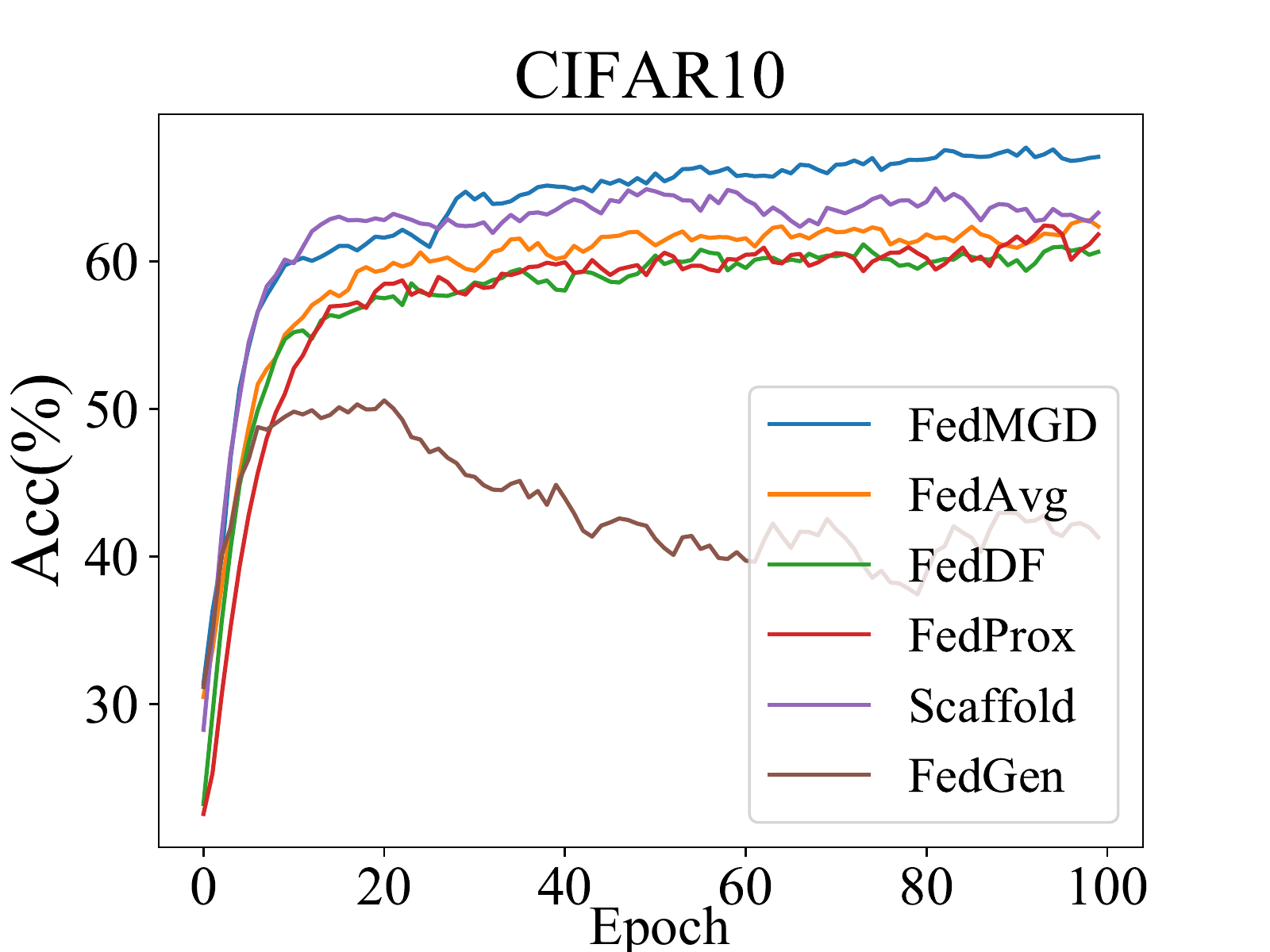}
    		%\caption{fig1}
    	\end{minipage}%
	}
	\subfigure[$\alpha$=0.1]{
    	\begin{minipage}[t]{0.32\linewidth}
    		\centering
    		\includegraphics[width=1.9in]{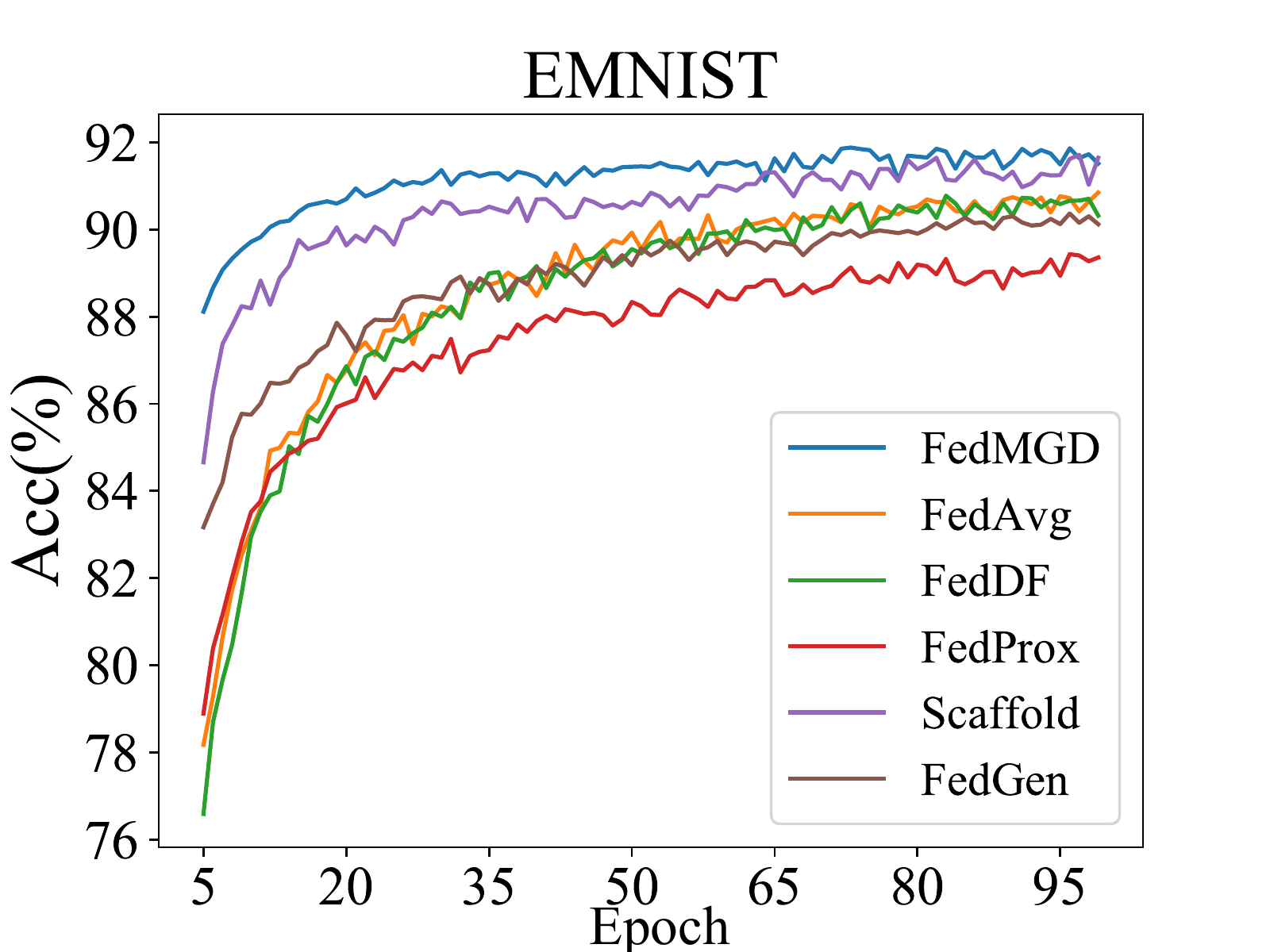}
    		\includegraphics[width=1.9in]{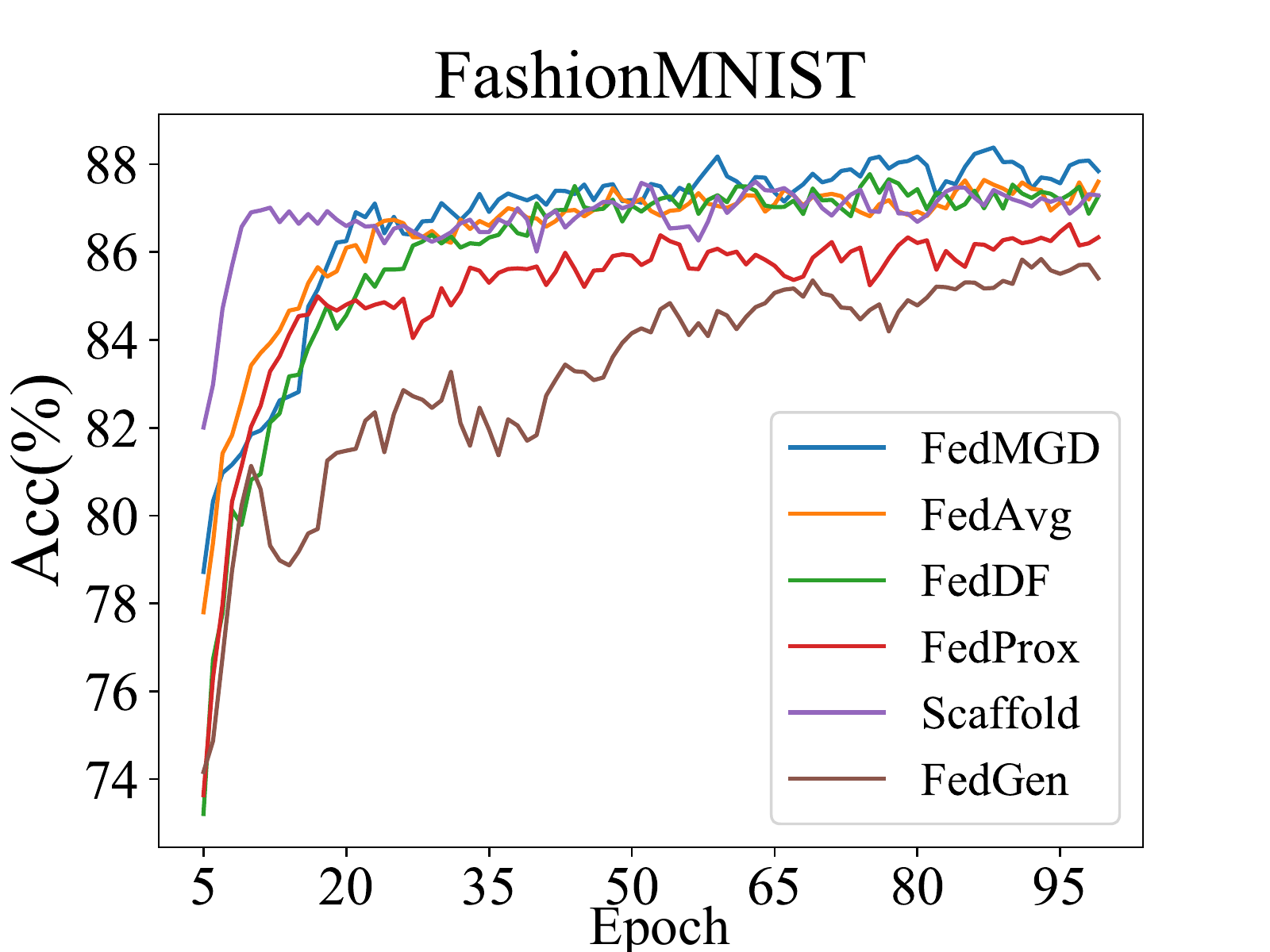}
    		\includegraphics[width=1.9in]{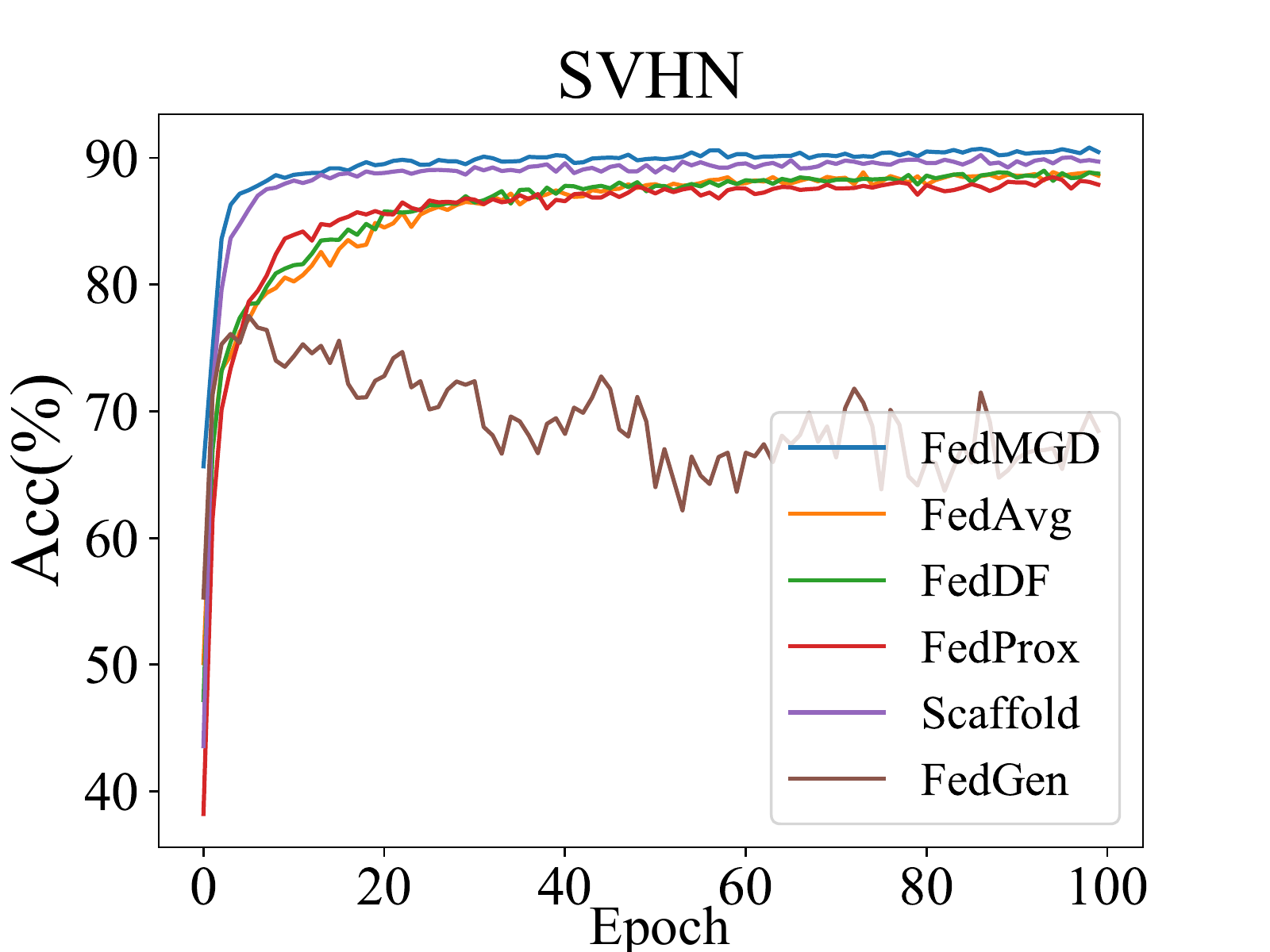}
    		\includegraphics[width=1.9in]{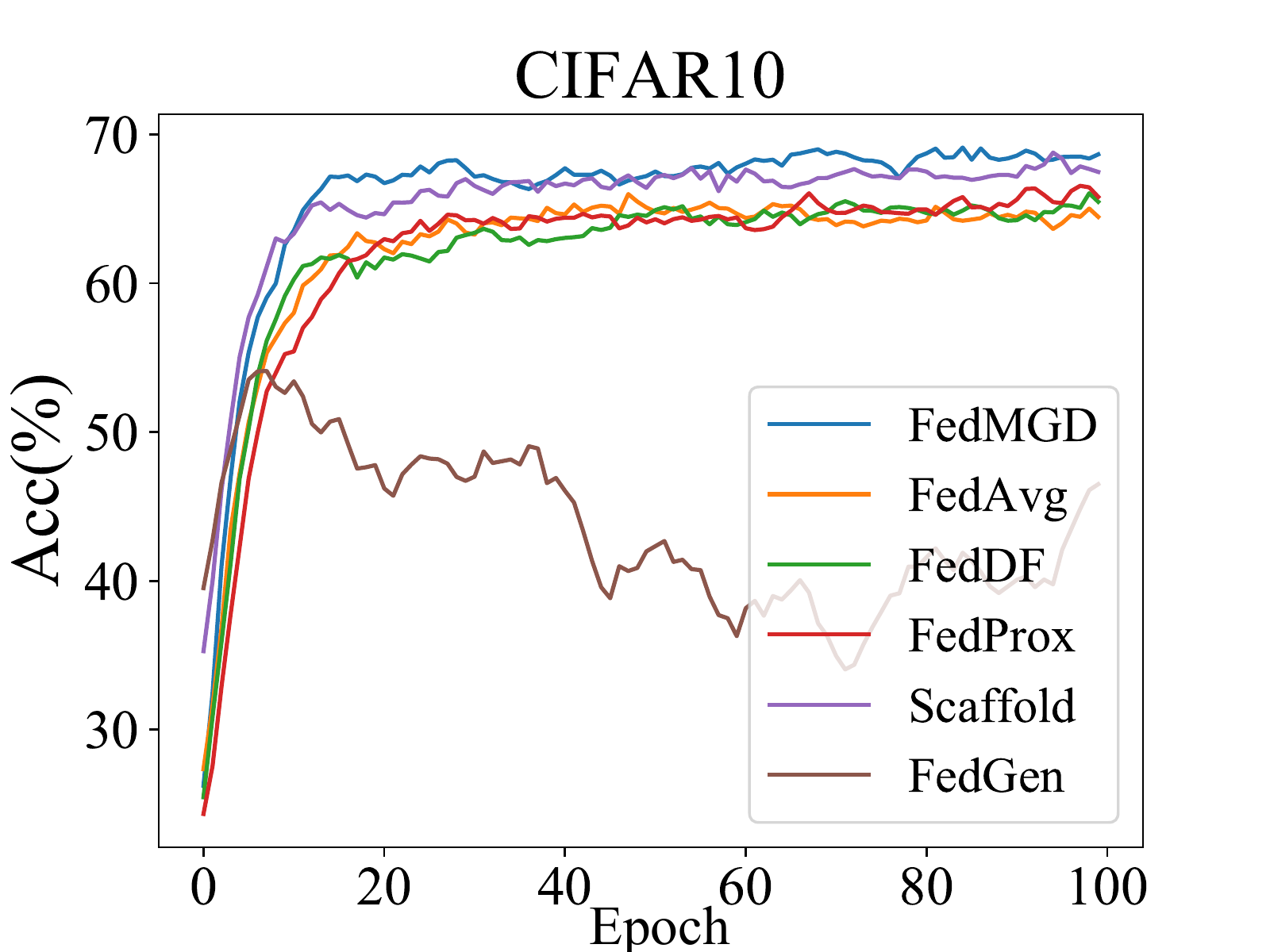}
    		%\caption{fig1}
    	\end{minipage}%
	}
	\caption{Visualization of algorithm performance at different degrees of label distribution skew. The case where the number of clients is 5 is shown here.}
	\label{fig5}
\end{figure}

We compare the accuracy of all algorithms in the global test set under different degrees of label distribution skew, and the results are shown in Table~\ref{tab1}.
% We compare the accuracy of the global test set for all algorithms with different degrees of label distribution skew, and the results are shown in Table~\ref{tab1}.
We observe that FedMGD outperforms other state-of-the-art methods in most cases, especially in highly heterogeneous scenario.
% In addition, we explored the performance of FedMGD on different number of clients. As shown in Table~\ref{tab1}, the performance of FedMGD does not change significantly over the number of clients.
Figure~\ref{fig5} visualizes the performance of all algorithms under different degrees of label distribution skew.
The figure shows that FedMGD exhibits more stable performance when the heterogeneity of the data changes. 
Compared with other methods, FedGen's accuracy decreases when there is more noise in training data. This may be because the lightweight generator in FedGen is vulnerable to noise in the sample, resulting in less information about the features captured from the sample.
In contrast, FedMGD uses a more sophisticated approach to data generation, thus ensuring the validity of the generated data.

\begin{figure}[h]
	\centering
		\centering
        \subfigure[$\alpha$=0.01]{
		\begin{minipage}[t]{0.32\linewidth}	
		\centering
			\includegraphics[width=1.9in]{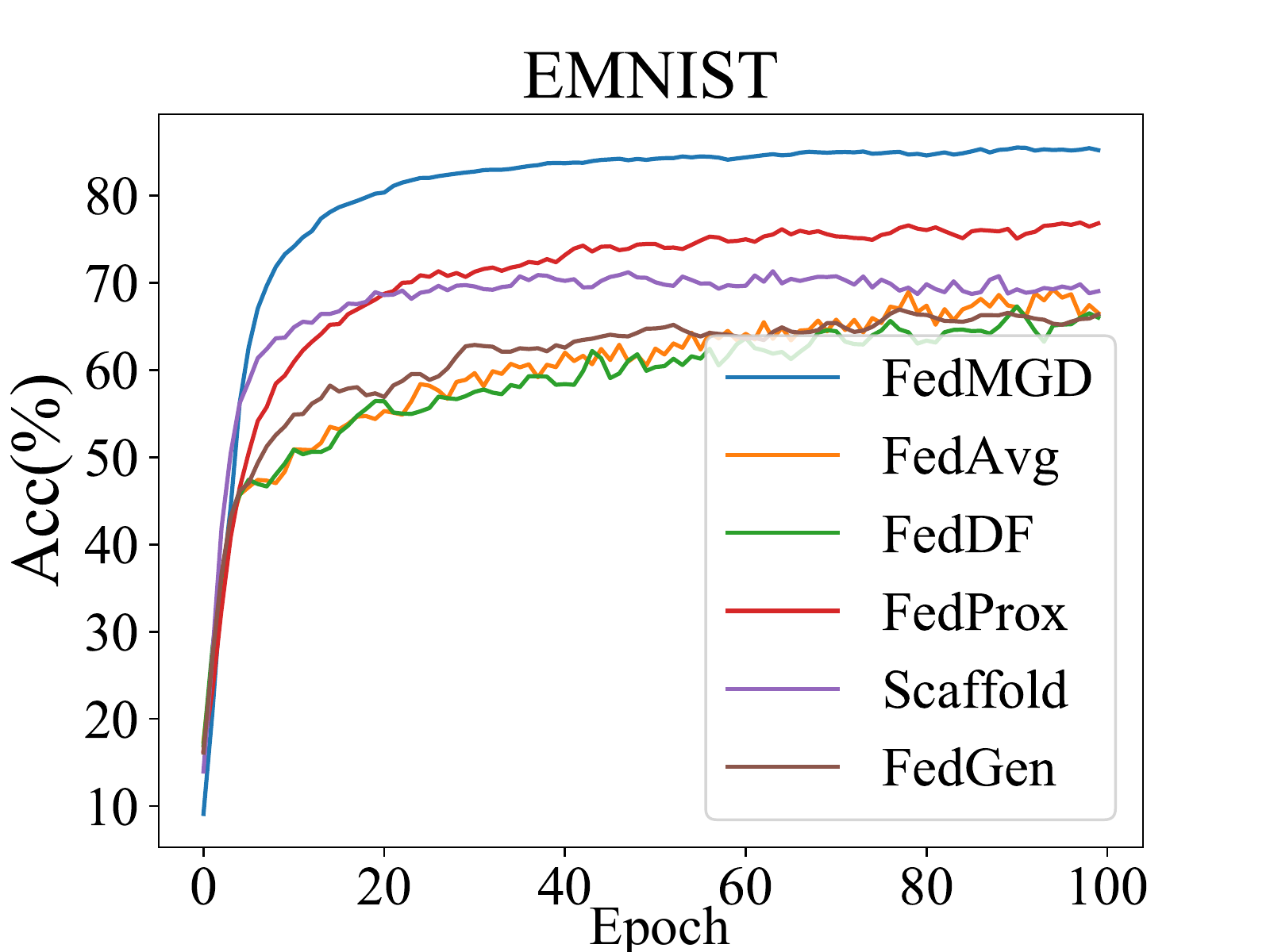}
			\includegraphics[width=1.9in]{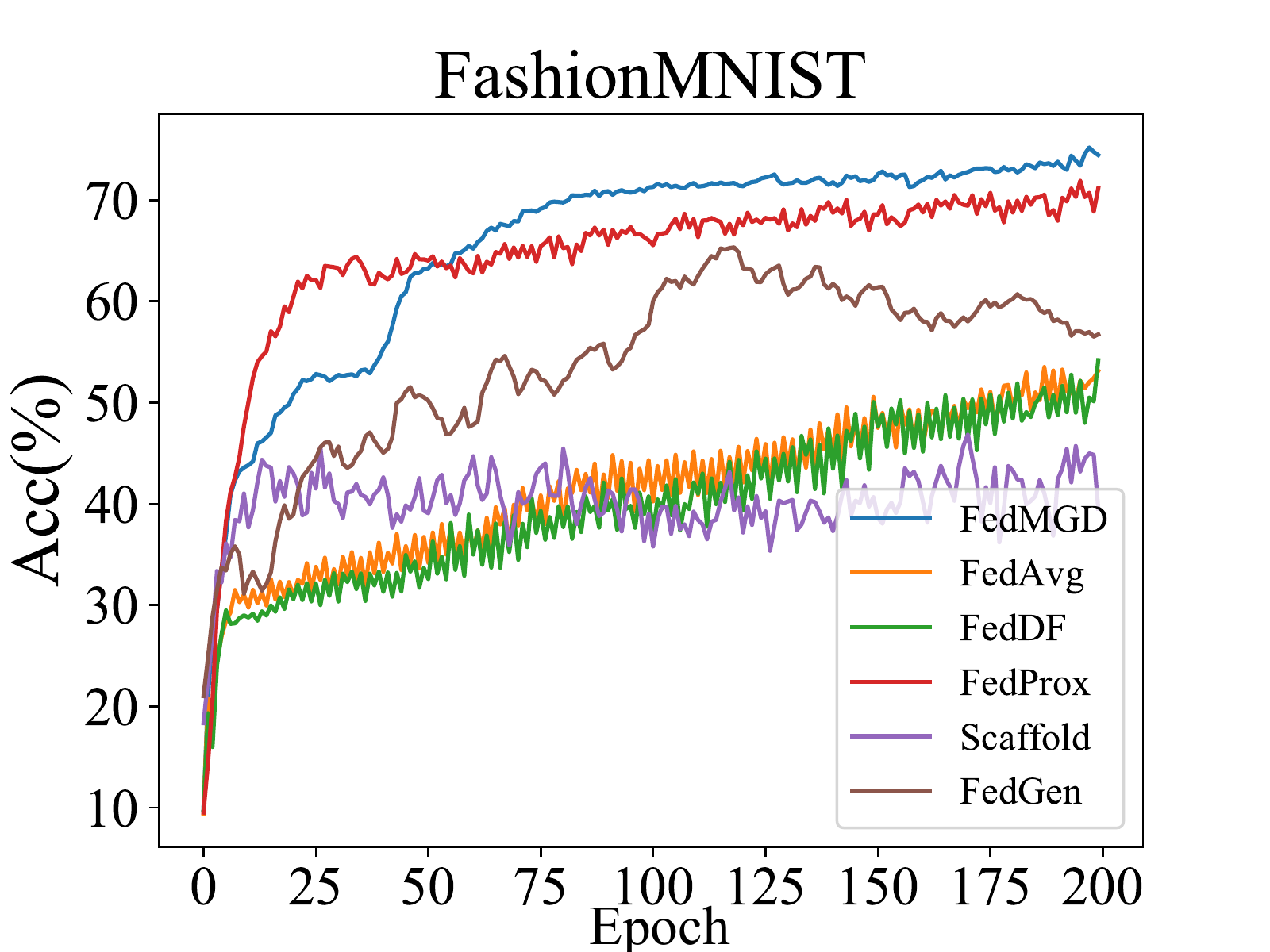}	
			\includegraphics[width=1.9in]{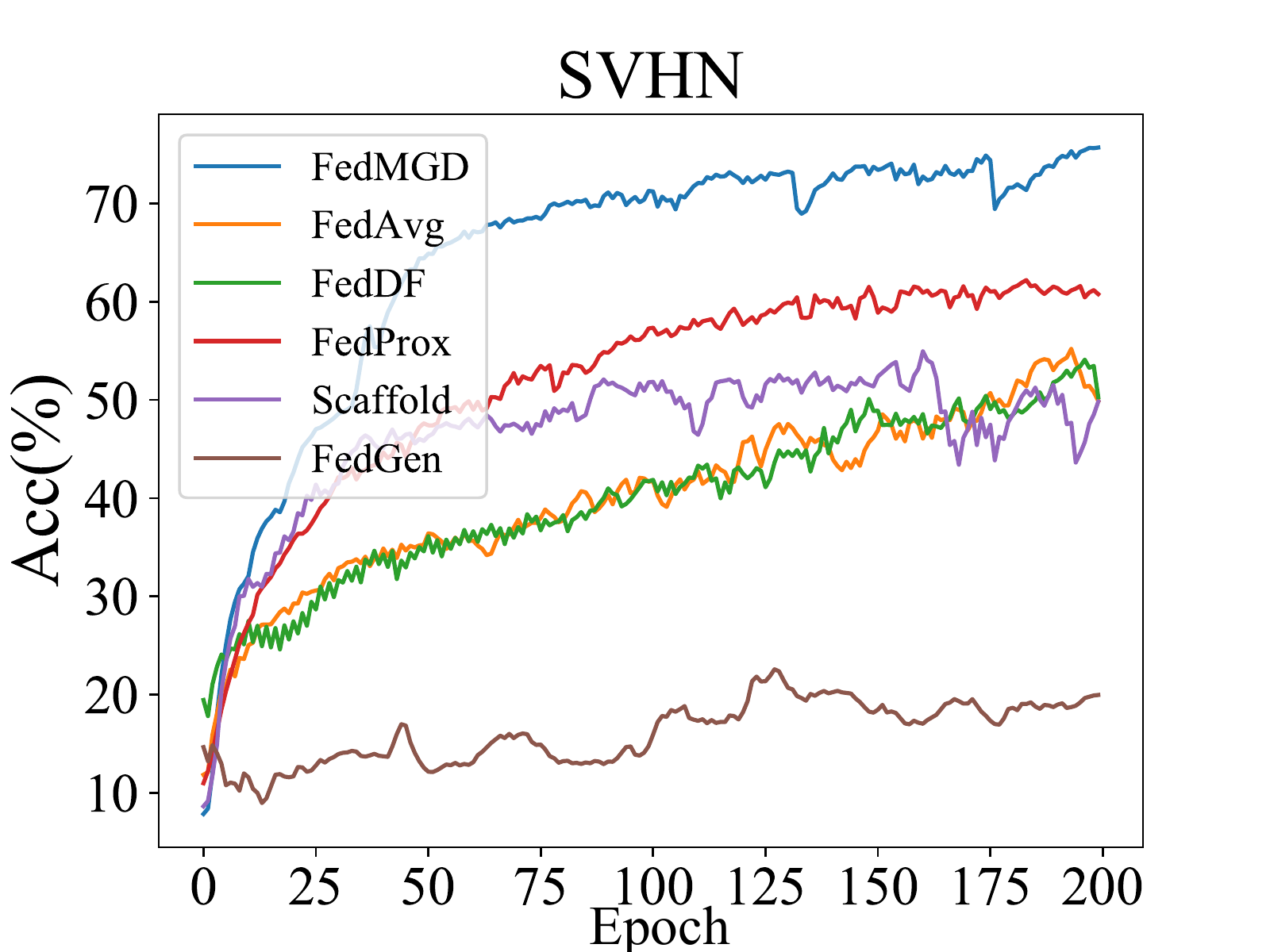}
			\includegraphics[width=1.9in]{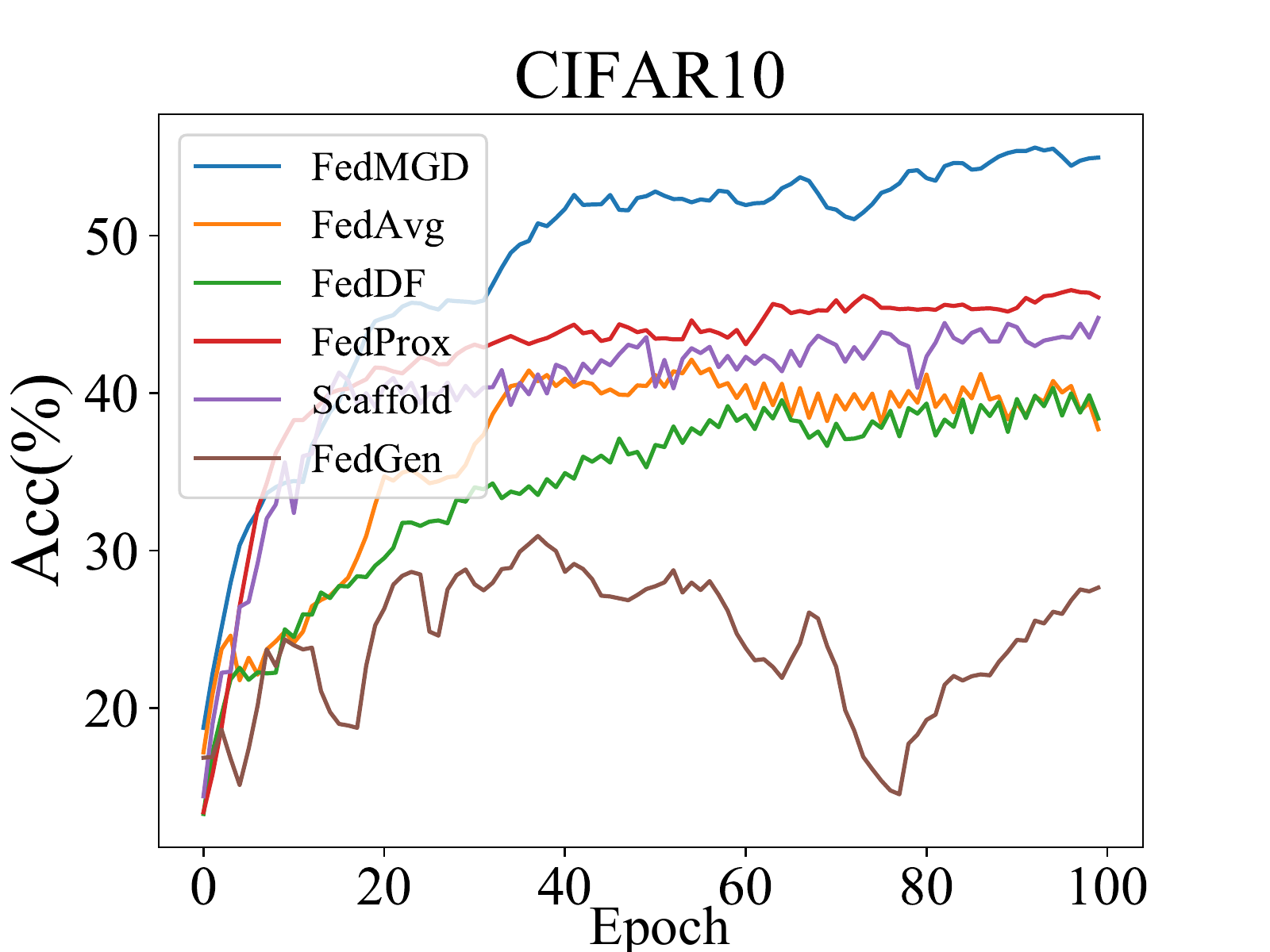}
		\end{minipage}%
        }
        \subfigure[$\alpha$=0.05]{
		\begin{minipage}[t]{0.32\linewidth}
		\centering
			\includegraphics[width=1.9in]{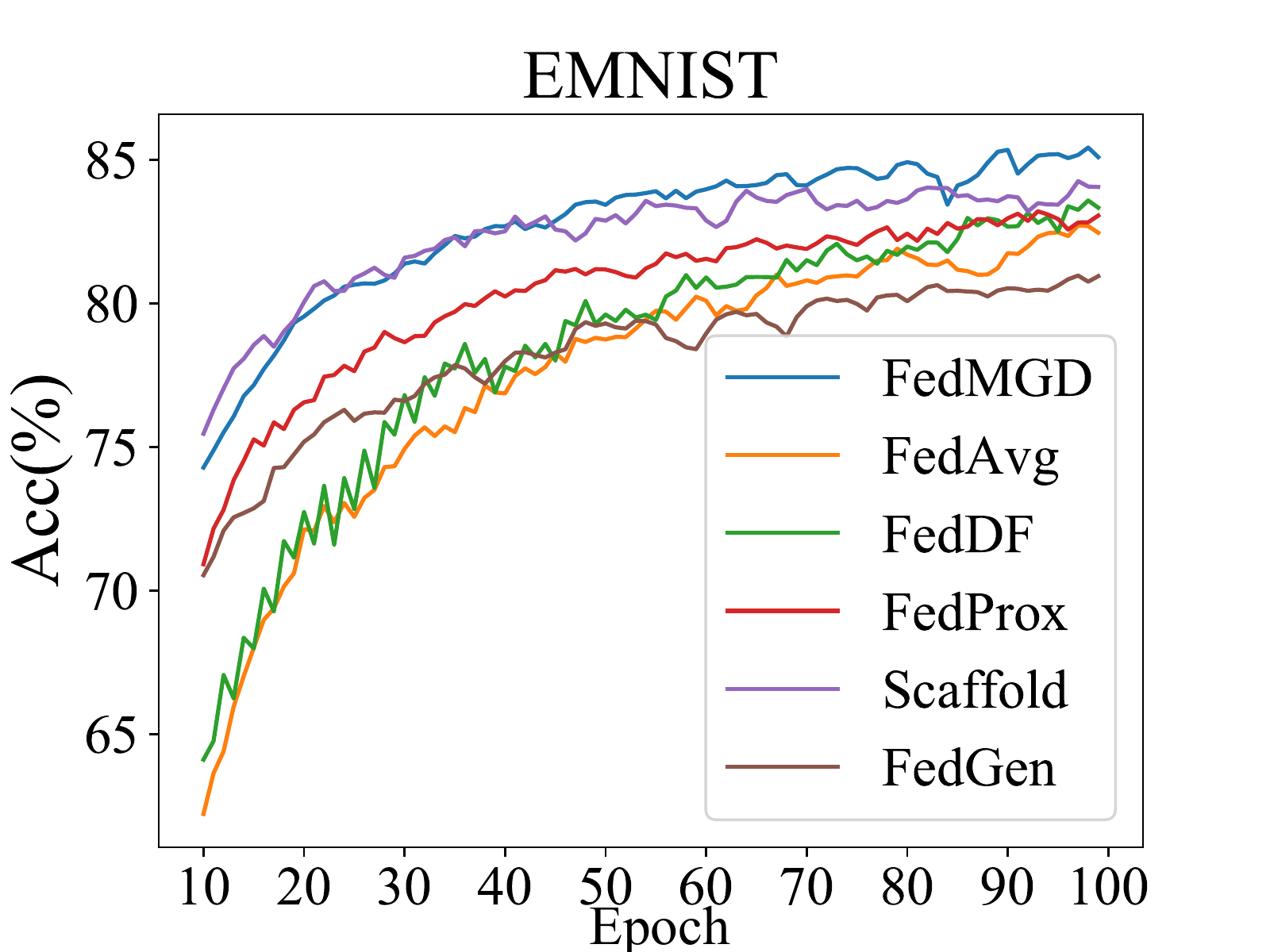} 
			%\caption{fig1}
			\includegraphics[width=1.9in]{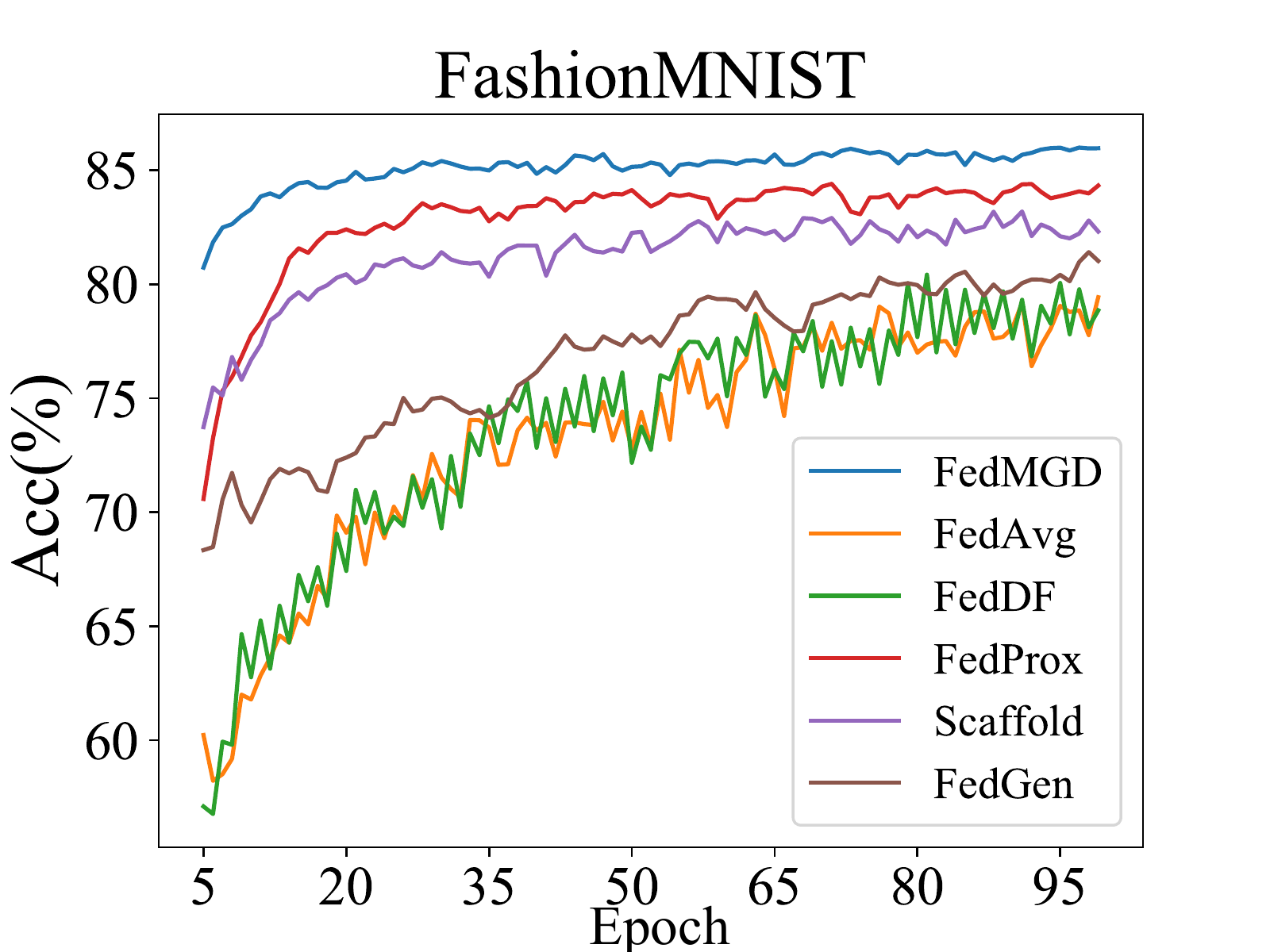}
			%\caption{fig1}
			\includegraphics[width=1.9in]{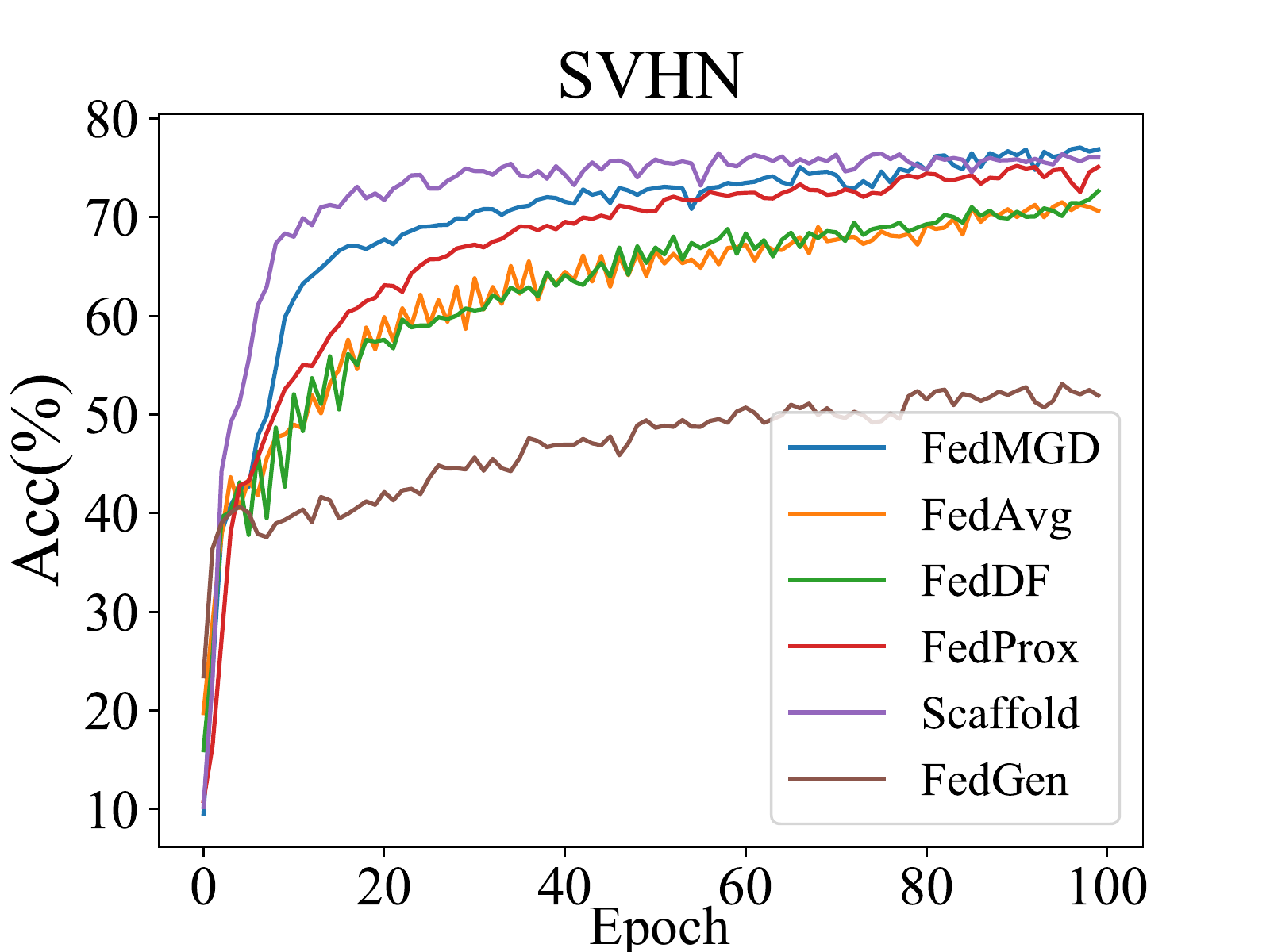}
			\includegraphics[width=1.9in]{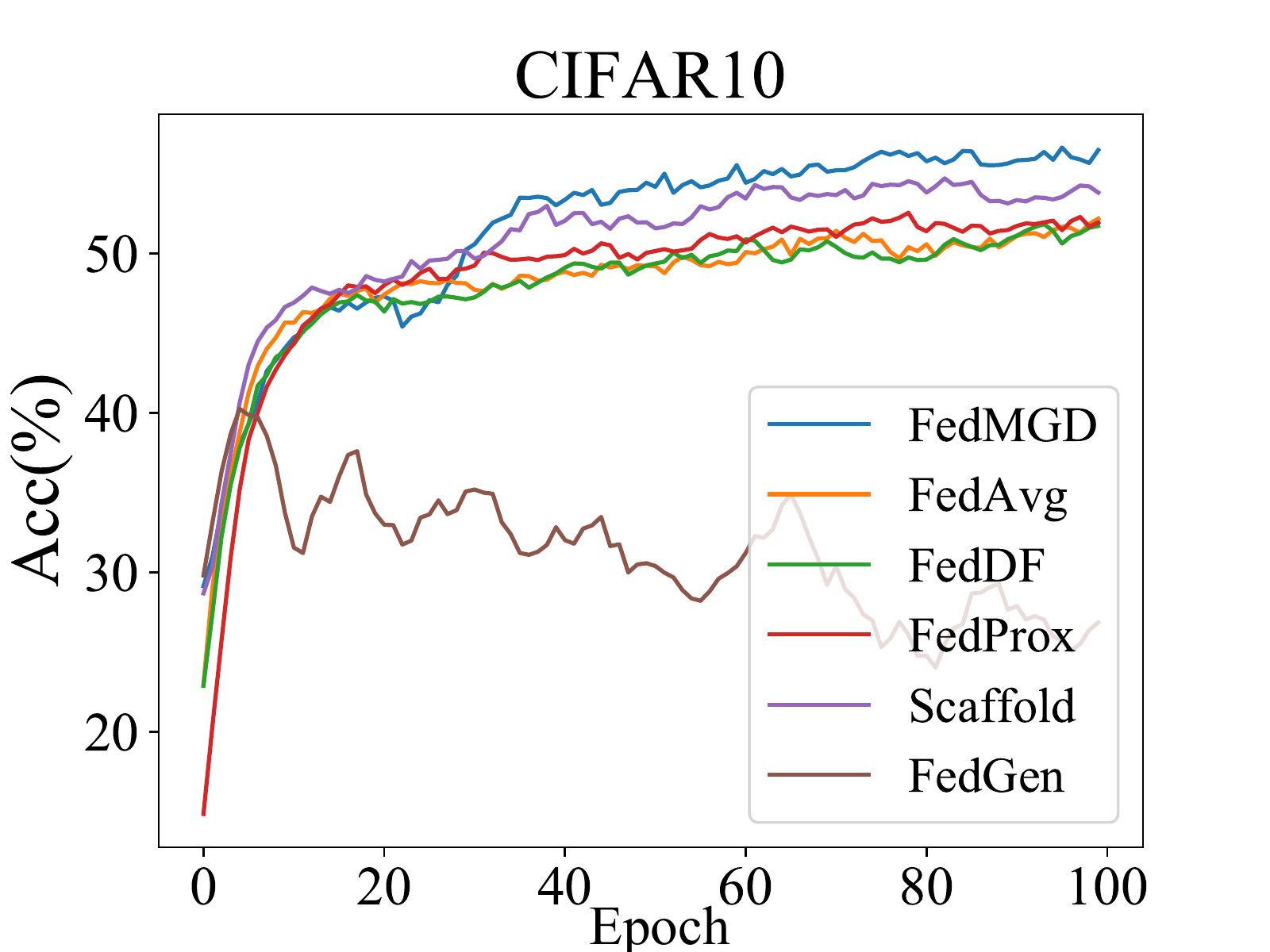}
		\end{minipage}%
        }
        \subfigure[$\alpha$=0.1]{
		\begin{minipage}[t]{0.32\linewidth}
		\centering
			\includegraphics[width=1.9in]{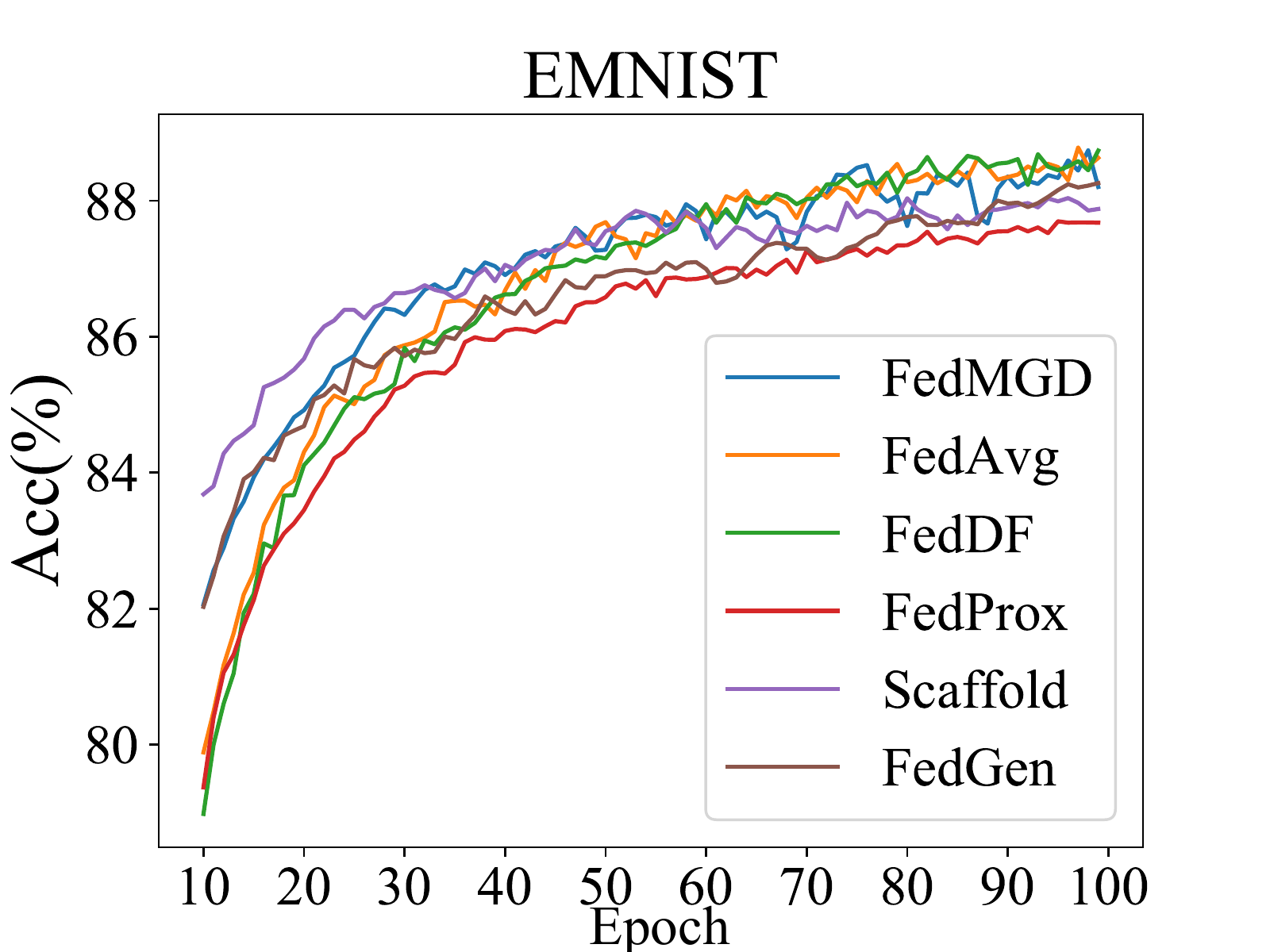} 
			%\caption{fig1}
			\includegraphics[width=1.9in]{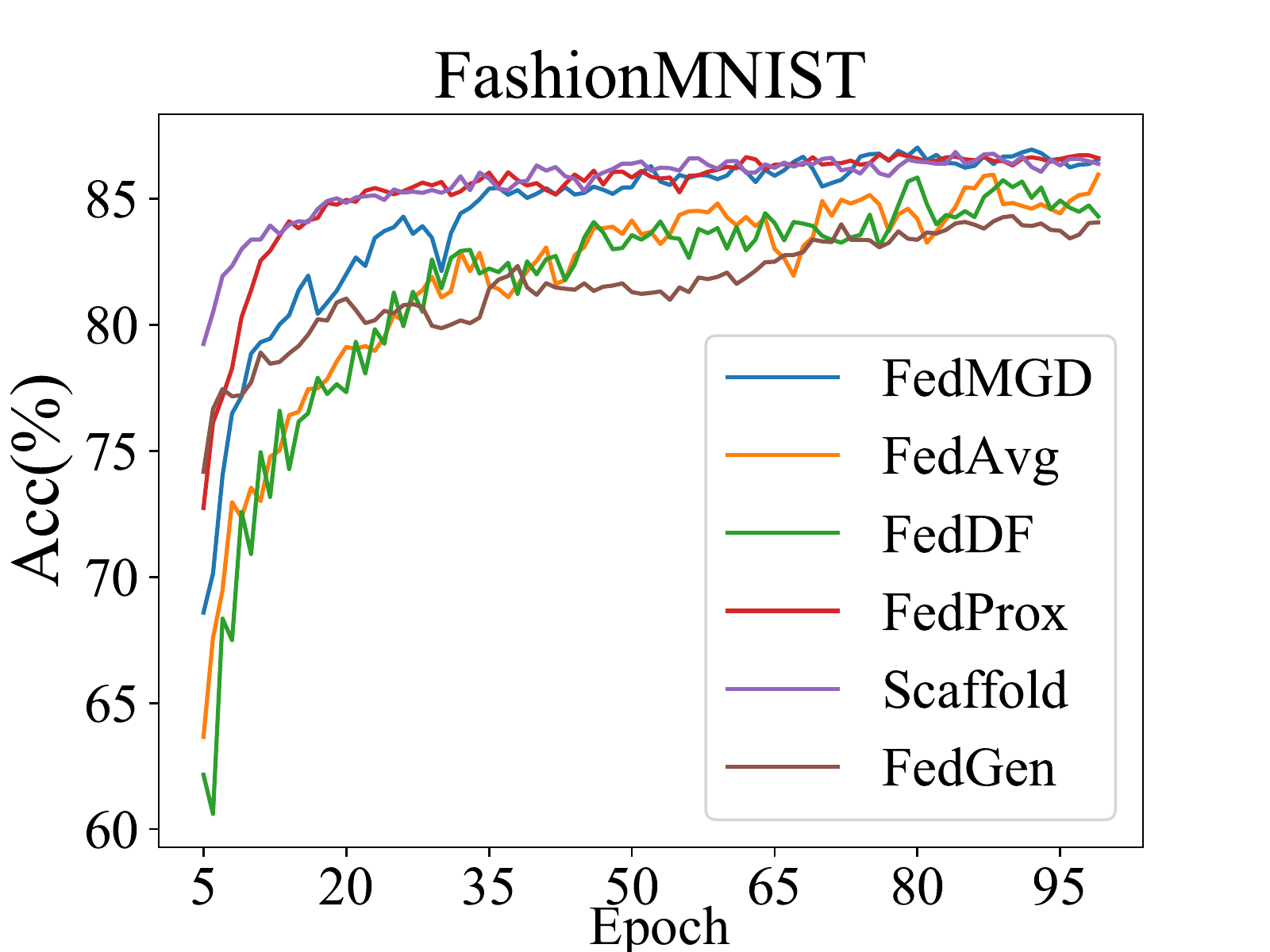}
			%\caption{fig1}
			\includegraphics[width=1.9in]{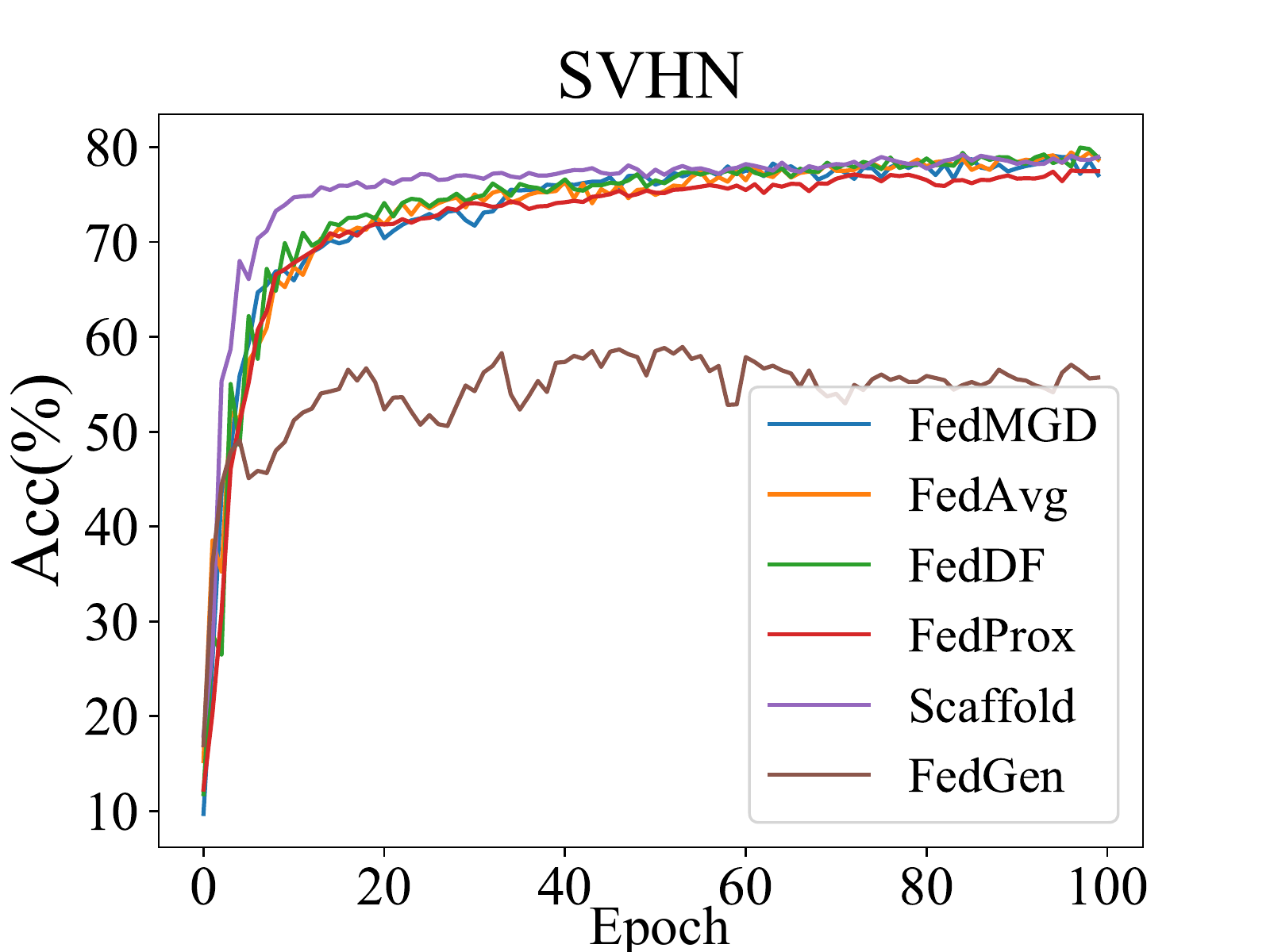}
			\includegraphics[width=1.9in]{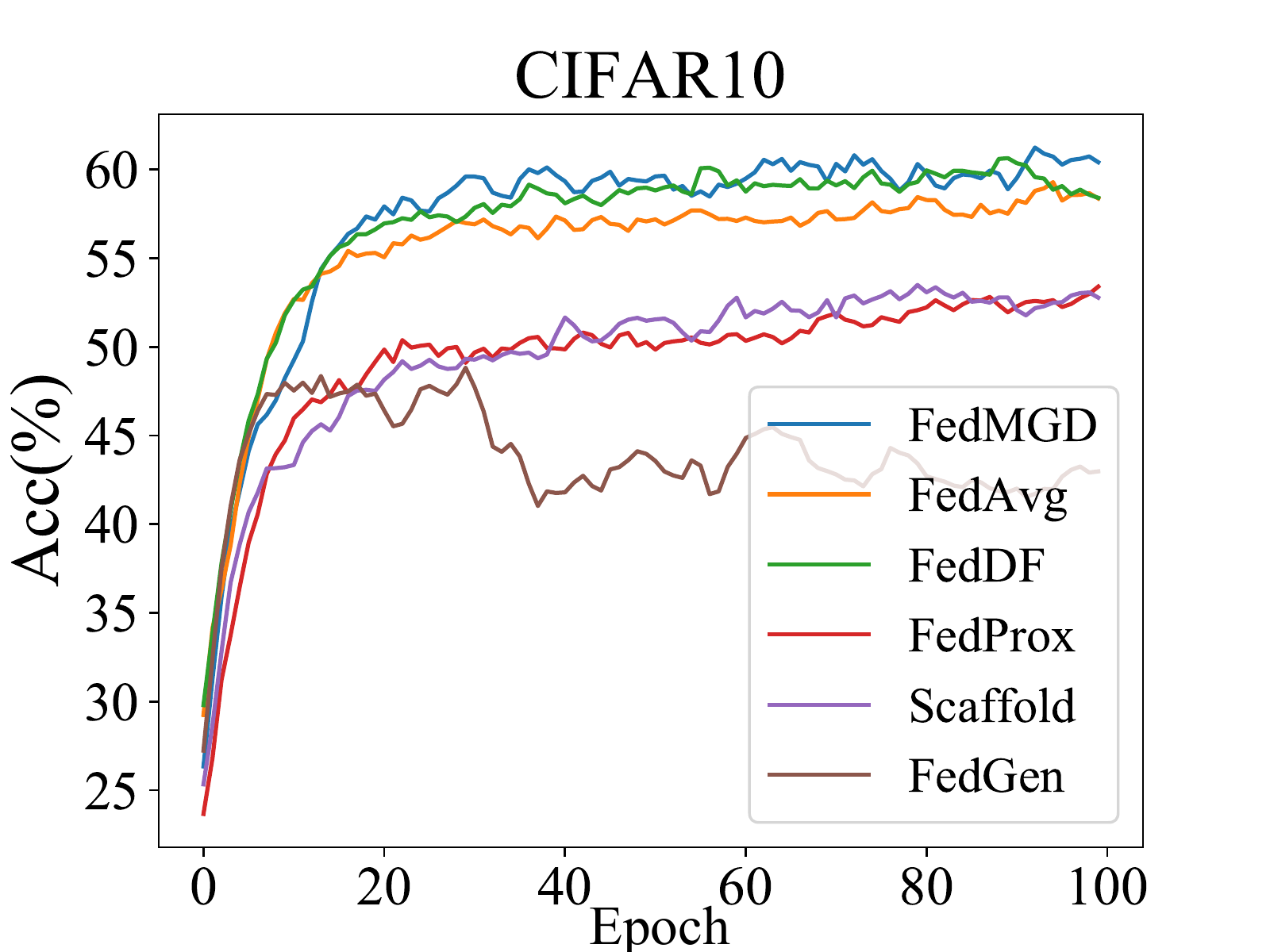}
		\end{minipage}%
        }
% 		}
	\centering
	\caption{Visualization of algorithm performance at different degrees of label distribution skew. The case where the number of clients is 10 is shown here.}
    \label{fig10}
\end{figure}

To increase the degree of label distribution skew among the clients, we distribute the labels among 10 clients. Each client has fewer label classes compared to the case with 5 clients.
The results are shown in Table~\ref{tab1}, where FedMGD can still maintain higher performance compared to other algorithms in the 10 clients scenario.
Figure~\ref{fig10} shows how the performance of the different algorithms changes when the data is divided into scenarios with 10 clients.
Compared to the scenario with 5 clients, FedMGD has a greater performance improvement compared to the baseline algorithm in the case of a more extreme label distribution.
In other words, in scenarios with a more skewed label distribution, the harm caused by the skewed label distribution to the model is reduced because FedMGD uses a global modeling approach to correct the model.

\subsubsection{Local Fairness of Different Models.}
\begin{table}[h]
	\begin{center}
		\caption{\textbf{Fairness comparison of FedMGD and state-of-the-art methods on local test sets.} This is measured by the standard deviation of the performance of the different algorithms on the local test set.}
		\renewcommand{\arraystretch}{1.05}
		% \resizebox{0.8\columnwidth}{!}{
		\setlength{\tabcolsep}{3.8mm}{
		\begin{tabular}{c|c|c|c|c|c|c|c}
			\hline
			\textbf{Dataset}                                                                                       & $\boldsymbol{\alpha}$ & \textbf{FedAvg} & \textbf{FedProx} & \textbf{FedDF} & \textbf{FedGen} & \textbf{SCAFFOLD} & \textbf{FedMGD} \\ \hline
			\multicolumn{1}{c|}{\multirow{3}{*}{\textbf{EMNIST}}}                                                  & 0.01     & 5.26            & 7.15             & 2.96           & 13.16           & 7.30              & \textbf{2.78} \textcolor{green!70!black}{($\downarrow$0.81)}   \\
			\multicolumn{1}{c|}{}                                                                                  & 0.05     & 3.81            & 4.53             & 4.11           & 5.53            & 6.50              & \textbf{2.20} \textcolor{green!70!black}{($\downarrow$1.61)}   \\
			\multicolumn{1}{c|}{}                                                                                  & 0.1      & 1.69            & 2.39             & 1.94           & 2.05            & \textbf{1.56}     & 2.61 \textcolor{red!70!black}{($\uparrow$1.05)}           \\ \hline
			\multicolumn{1}{c|}{\multirow{3}{*}{\textbf{\begin{tabular}[c]{@{}c@{}}Fashion\\ MNIST\end{tabular}}}} & 0.01     & 35.36           & 23.91            & 35.87          & 32.93           & 39.19             & \textbf{11.35} \textcolor{green!70!black}{($\downarrow$12.56)}  \\
			\multicolumn{1}{c|}{}                                                                                  & 0.05     & 19.04           & 18.82            & 18.11          & 24.87           & 28.09             & \textbf{6.18} \textcolor{green!70!black}{($\downarrow$11.93)}   \\
			\multicolumn{1}{c|}{}                                                                                  & 0.1      & 9.01            & 10.75            & 9.00           & 10.46           & 12.00             & \textbf{5.98} \textcolor{green!70!black}{($\downarrow$3.02)}   \\ \hline
			\multicolumn{1}{c|}{\multirow{3}{*}{\textbf{SVHN}}}                                                    & 0.01     & 39.31           & 21.90            & 39.12          & 33.21           & 13.83             & \textbf{10.01} \textcolor{green!70!black}{($\downarrow$3.82)}  \\
			\multicolumn{1}{c|}{}                                                                                  & 0.05     & 8.79            & 3.21             & 9.54           & 8.31            & \textbf{2.41}     & 3.93 \textcolor{red!70!black}{($\uparrow$1.52)}           \\
			\multicolumn{1}{c|}{}                                                                                  & 0.1      & 4.55            & 3.38             & 3.98           & 17.43           & 1.81              & \textbf{1.56} \textcolor{green!70!black}{($\downarrow$0.25)}   \\ \hline
			\multicolumn{1}{l|}{\multirow{3}{*}{\textbf{CIFAR10}}}                                                 & 0.01     & 27.77           & 23.31            & 27.48          & 18.92           & 18.57             & \textbf{7.54} \textcolor{green!70!black}{($\downarrow$11.03)}   \\
			\multicolumn{1}{l|}{}                                                                                  & 0.05     & 22.74           & 20.35            & 21.72          & 20.62           & 10.95             & \textbf{10.20} \textcolor{green!70!black}{($\downarrow$0.75)}  \\
			\multicolumn{1}{l|}{}                                                                                  & 0.1      & 13.71           & 14.58            & 14.26          & 22.31           & 15.22             & \textbf{7.21} \textcolor{green!70!black}{($\downarrow$6.50)}   \\ \hline
		\end{tabular}
	}
		% }
		\label{tab2}
	\end{center}
\end{table}

In the label distribution skew scenario, the same model may have different performance in different clients, and we call this difference in performance between clients the local fairness of the model.
We compare the performance of models trained by FedMGD with state-of-the-art Methods on the client test set and demonstrate that the fairness of the models on the client is effectively improved by learning the knowledge of the global distribution.
% the difference in model accuracy performance across clients is effectively mitigated by learning the knowledge of the global distribution.
% We evaluated the performance differences of each algorithm locally on the clients using standard deviation
We use the standard deviation of the accuracy of the algorithms on the local test sets of different clients as a measure, and the results are shown in Table~\ref{tab2}.
We have observed that FedMGD maintains a small difference in performance across clients in all scenarios.
This is due to the fact that FedMGD improves the compatibility of the global model under heterogeneous data distribution by using global information to refine the aggregated model, thus allowing the model to perform more fairly (with less performance difference) across clients.
% FedMGD learns about the global distribution, which effectively mitigates the difference in model performance across clients.

\begin{figure}[h]
	\centering
	\subfigure[$\alpha$=0.01]{
	\begin{minipage}[t]{0.32\linewidth}
		\centering
		\includegraphics[width=1.9in]{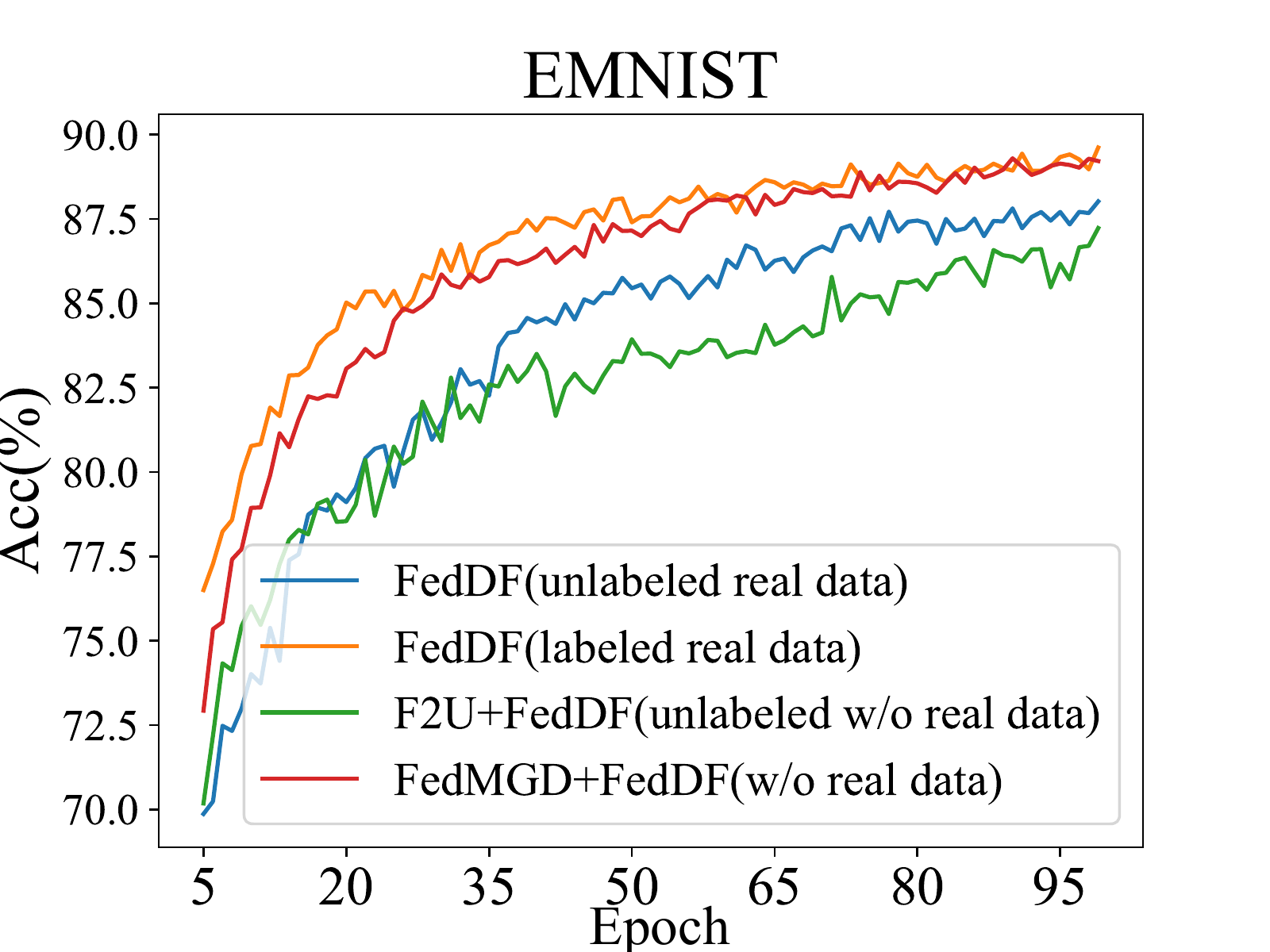}
		\includegraphics[width=1.9in]{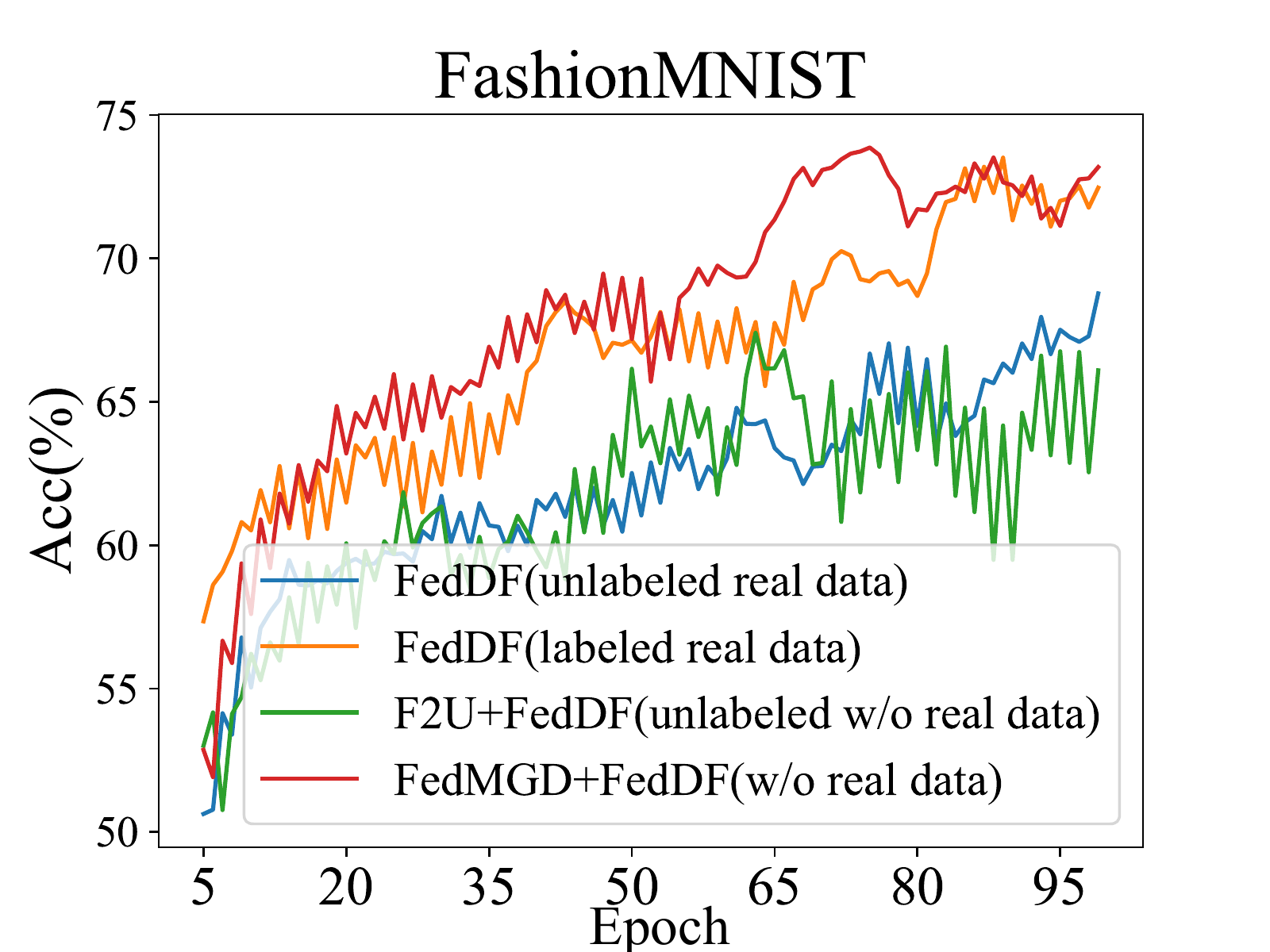}
		\includegraphics[width=1.9in]{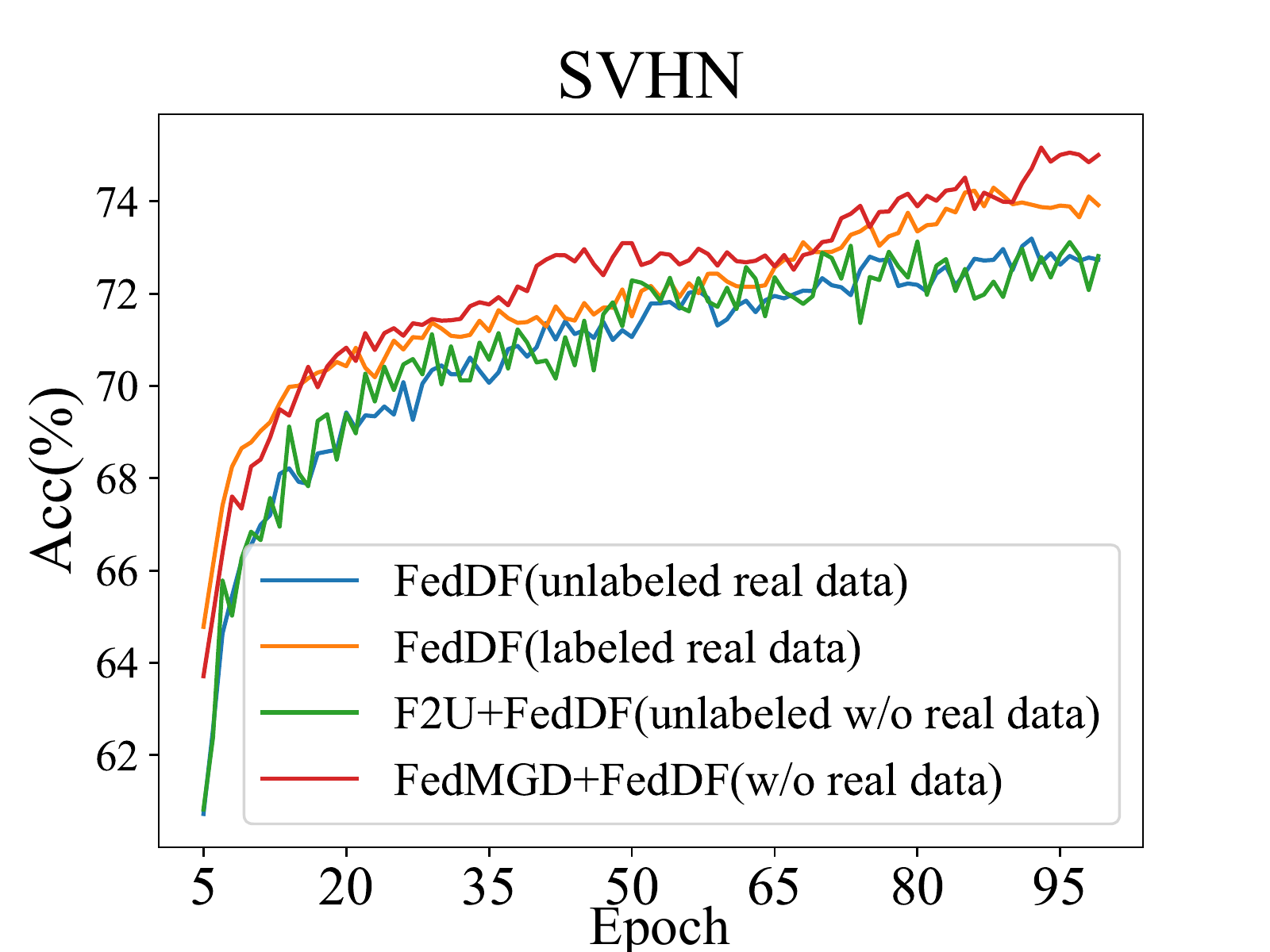}
 		\includegraphics[width=1.9in]{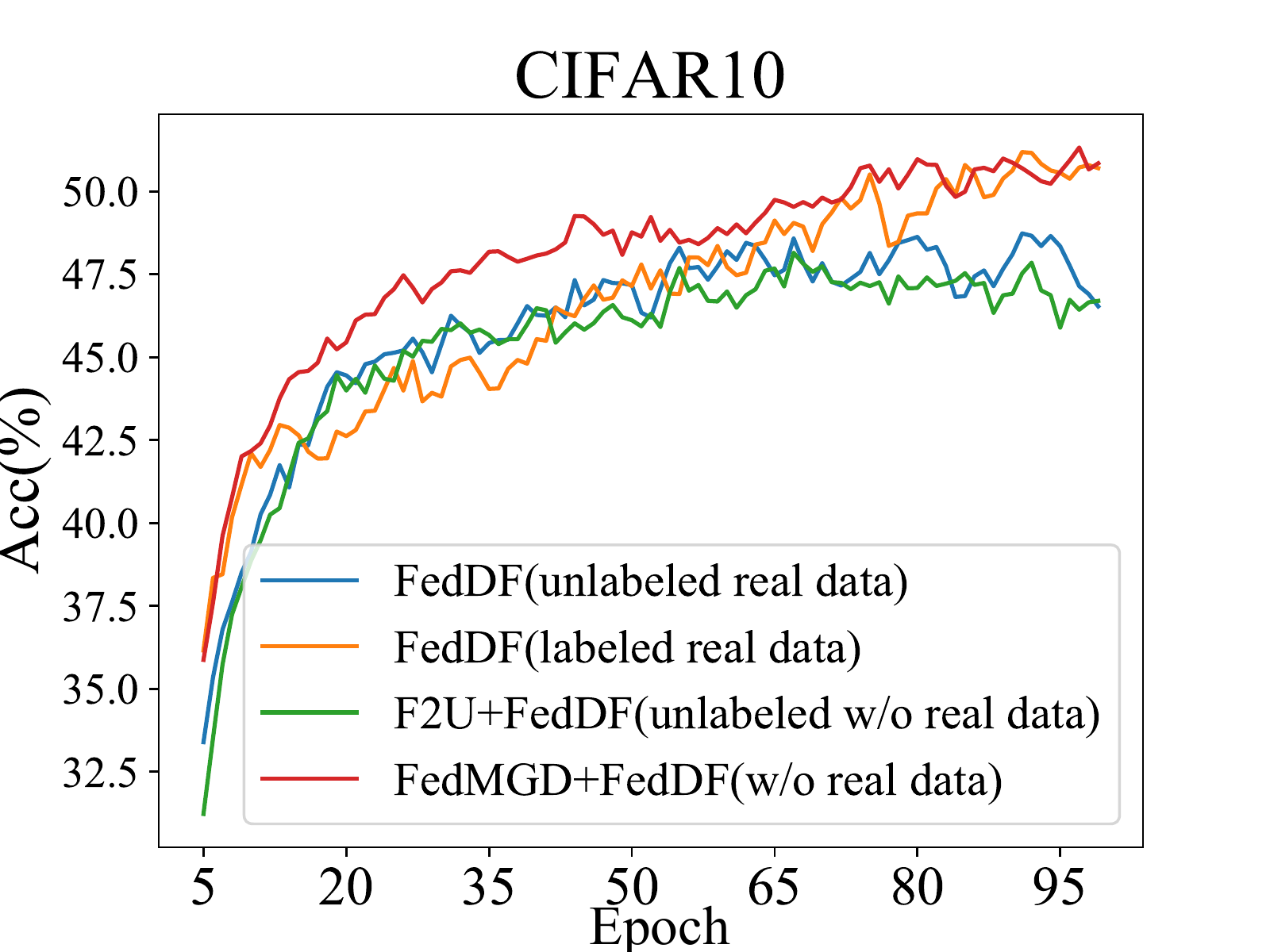}
		%\caption{fig1}
	\end{minipage}%
	}
	\subfigure[$\alpha$=0.05]{
	\begin{minipage}[t]{0.32\linewidth}
		\centering
		\includegraphics[width=1.9in]{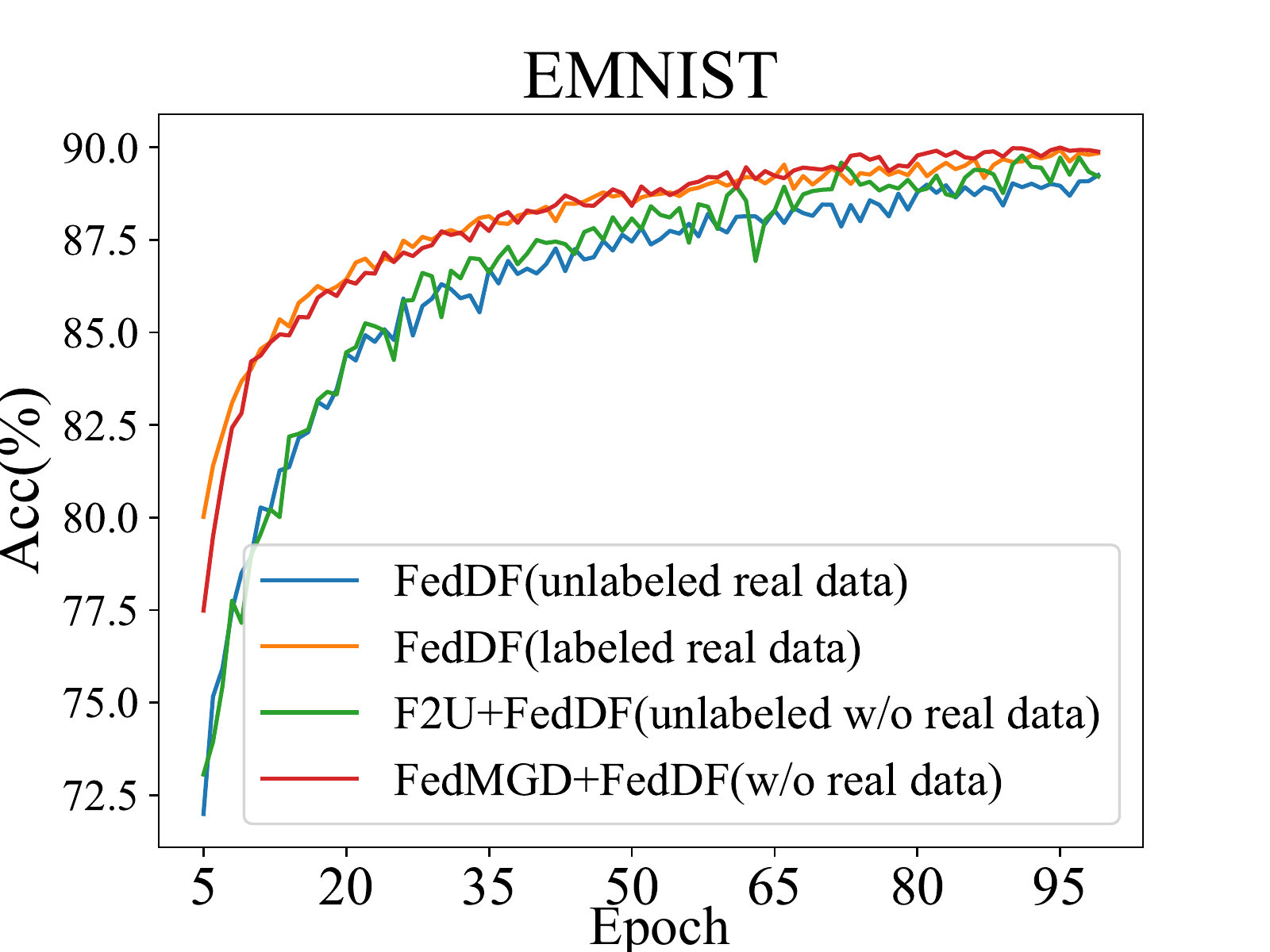}
		\includegraphics[width=1.9in]{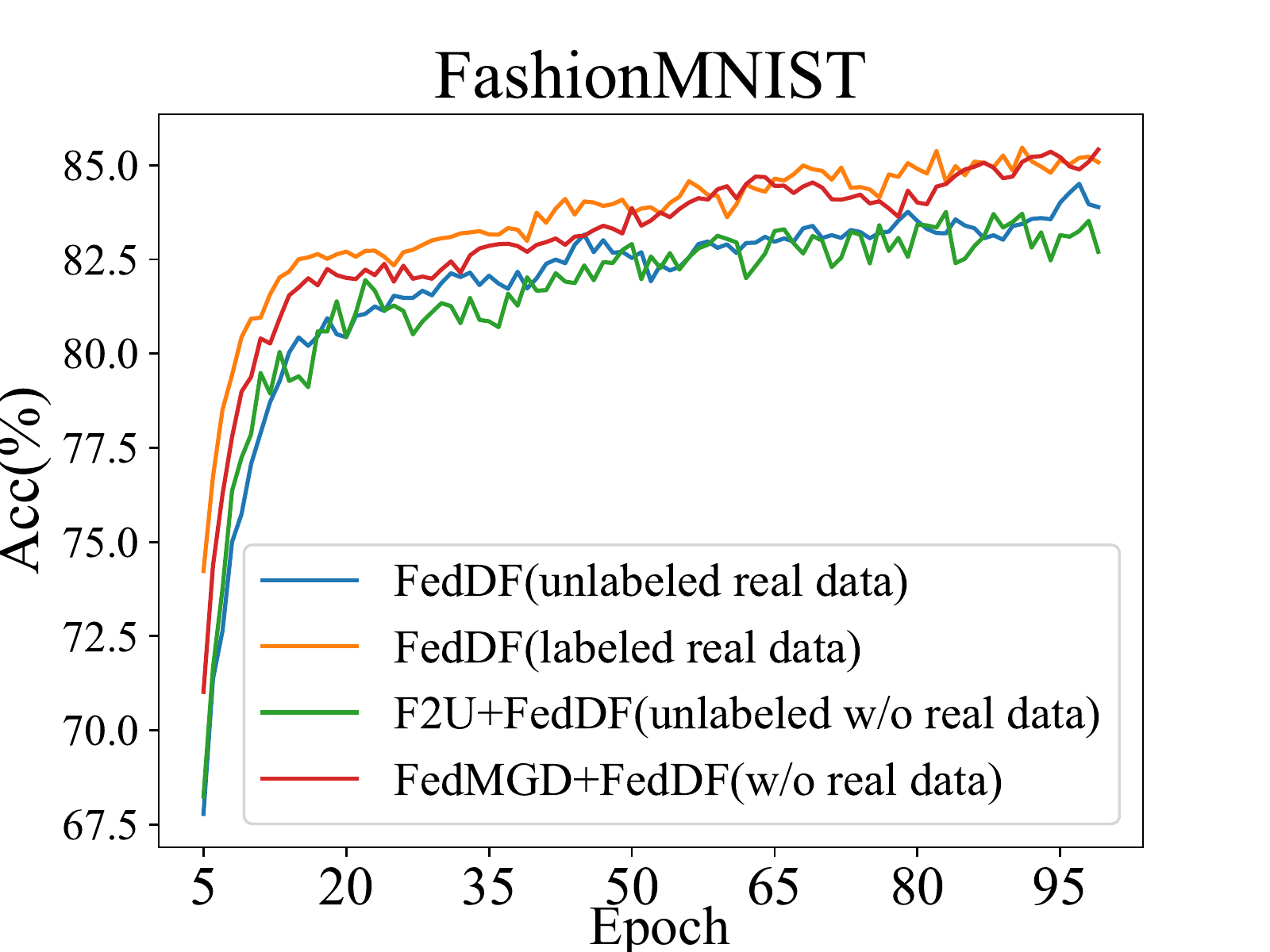}
		\includegraphics[width=1.9in]{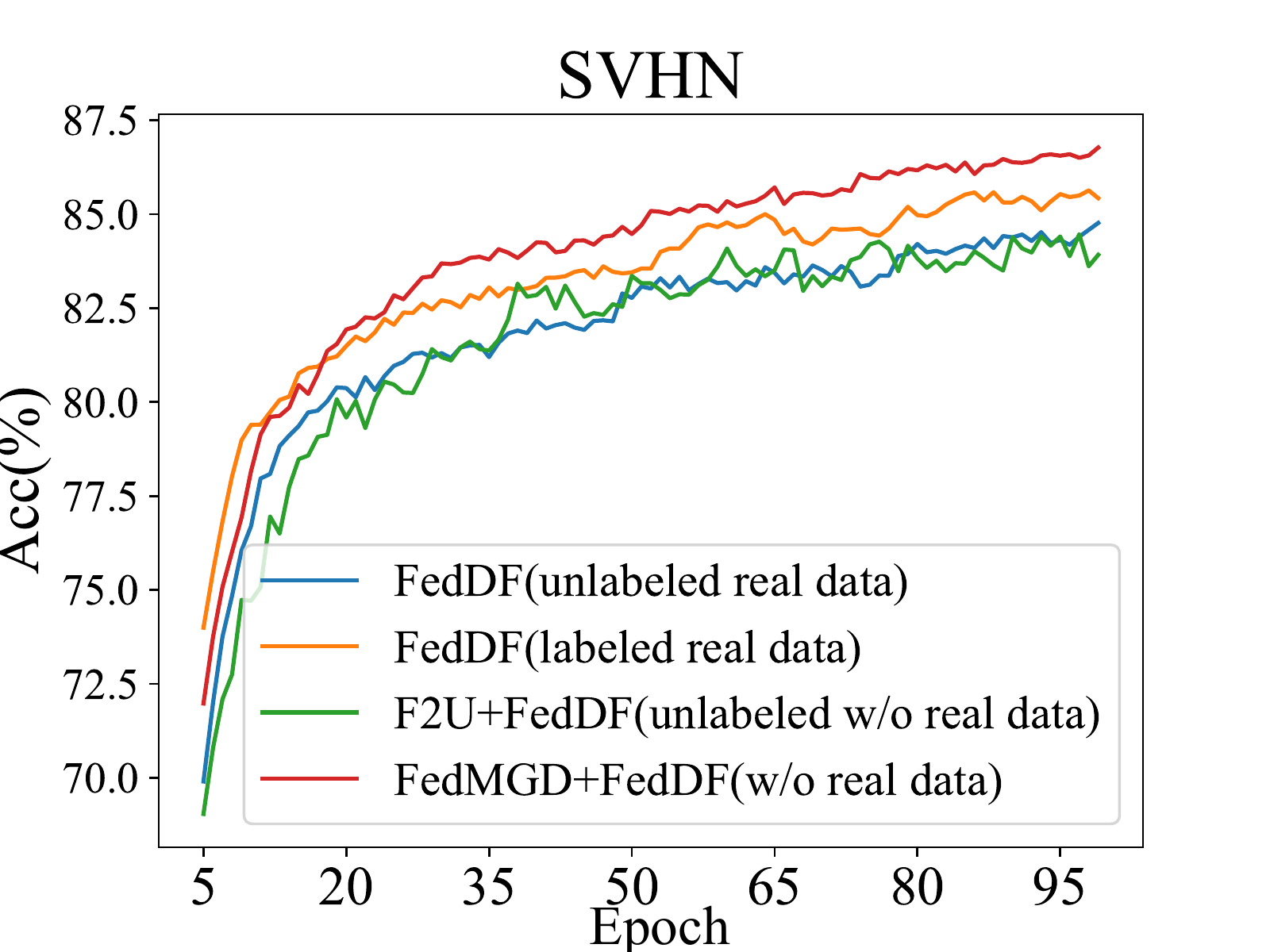}
		\includegraphics[width=1.9in]{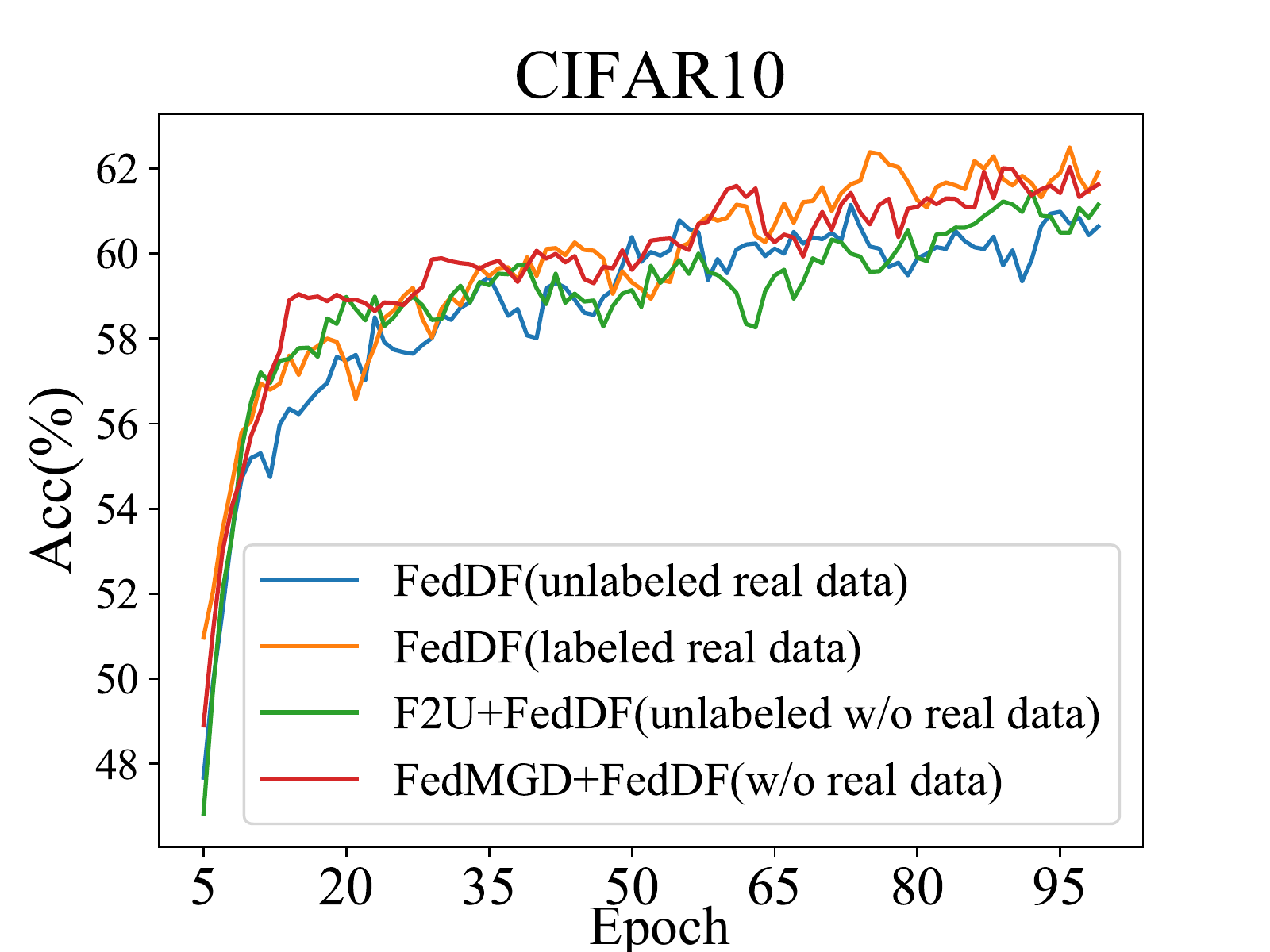}
		%\caption{fig1}
	\end{minipage}%
	}
	\subfigure[$\alpha$=0.1]{
	\begin{minipage}[t]{0.32\linewidth}
		\centering
		\includegraphics[width=1.9in]{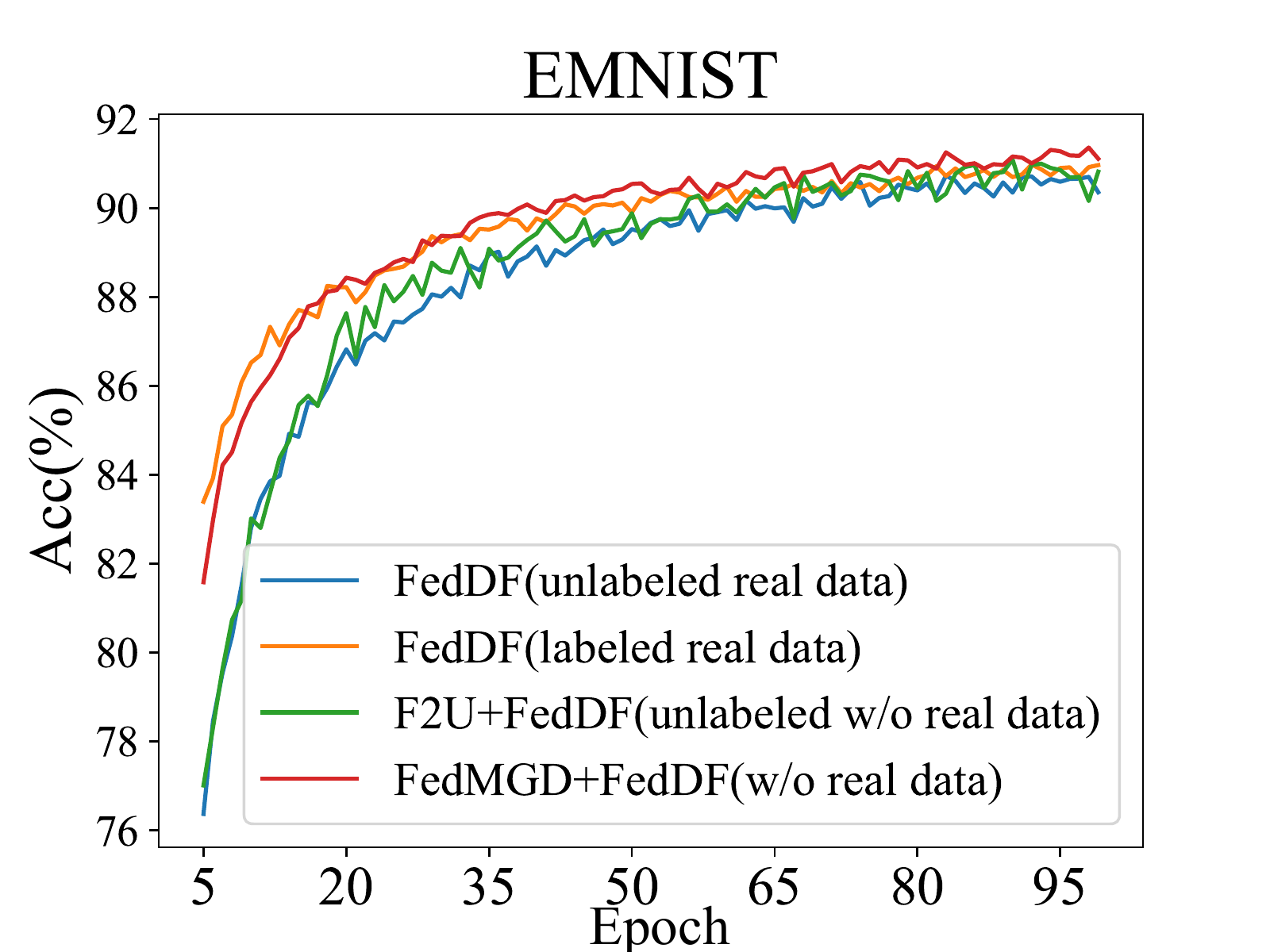}
		\includegraphics[width=1.9in]{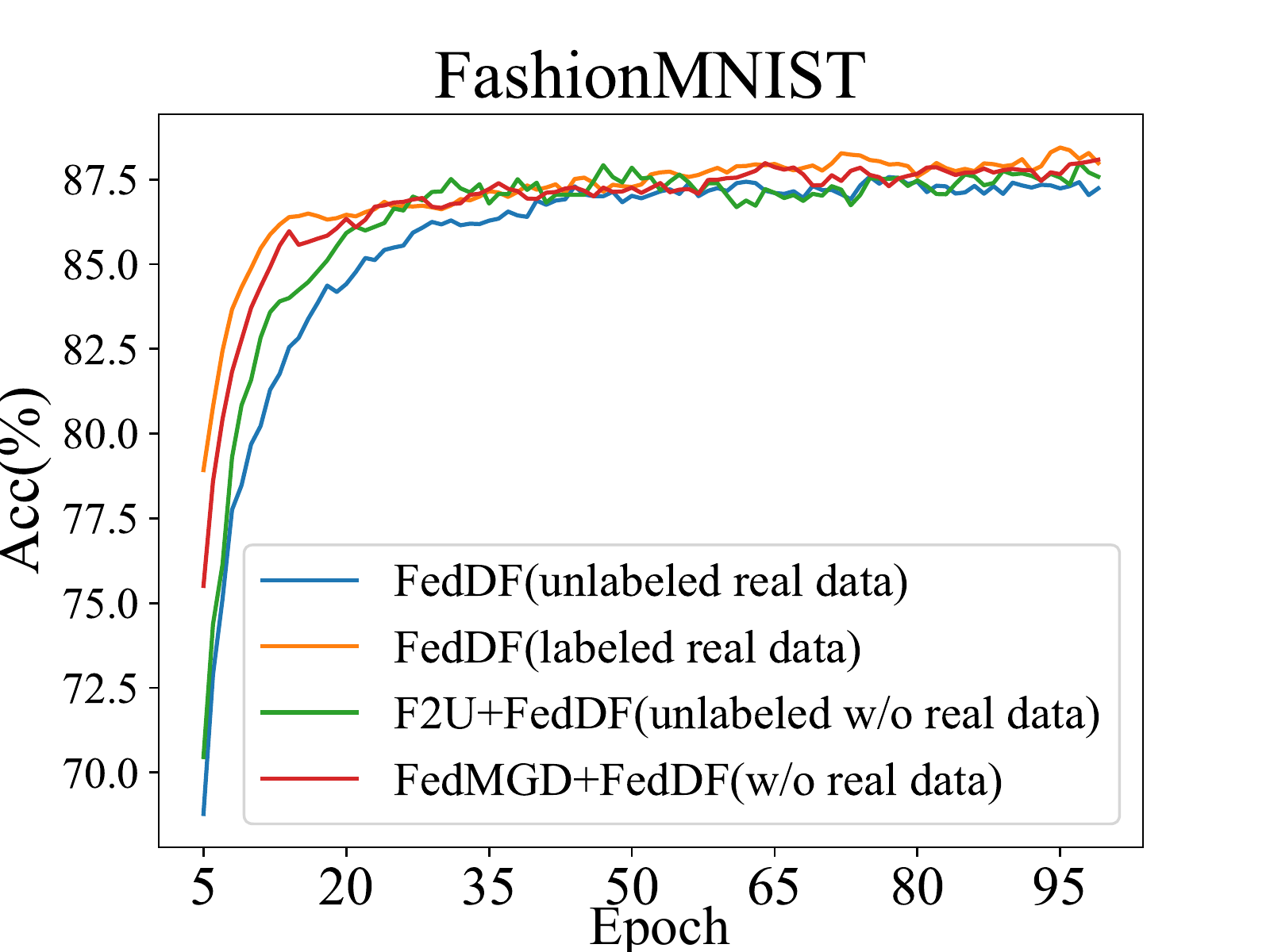}
		\includegraphics[width=1.9in]{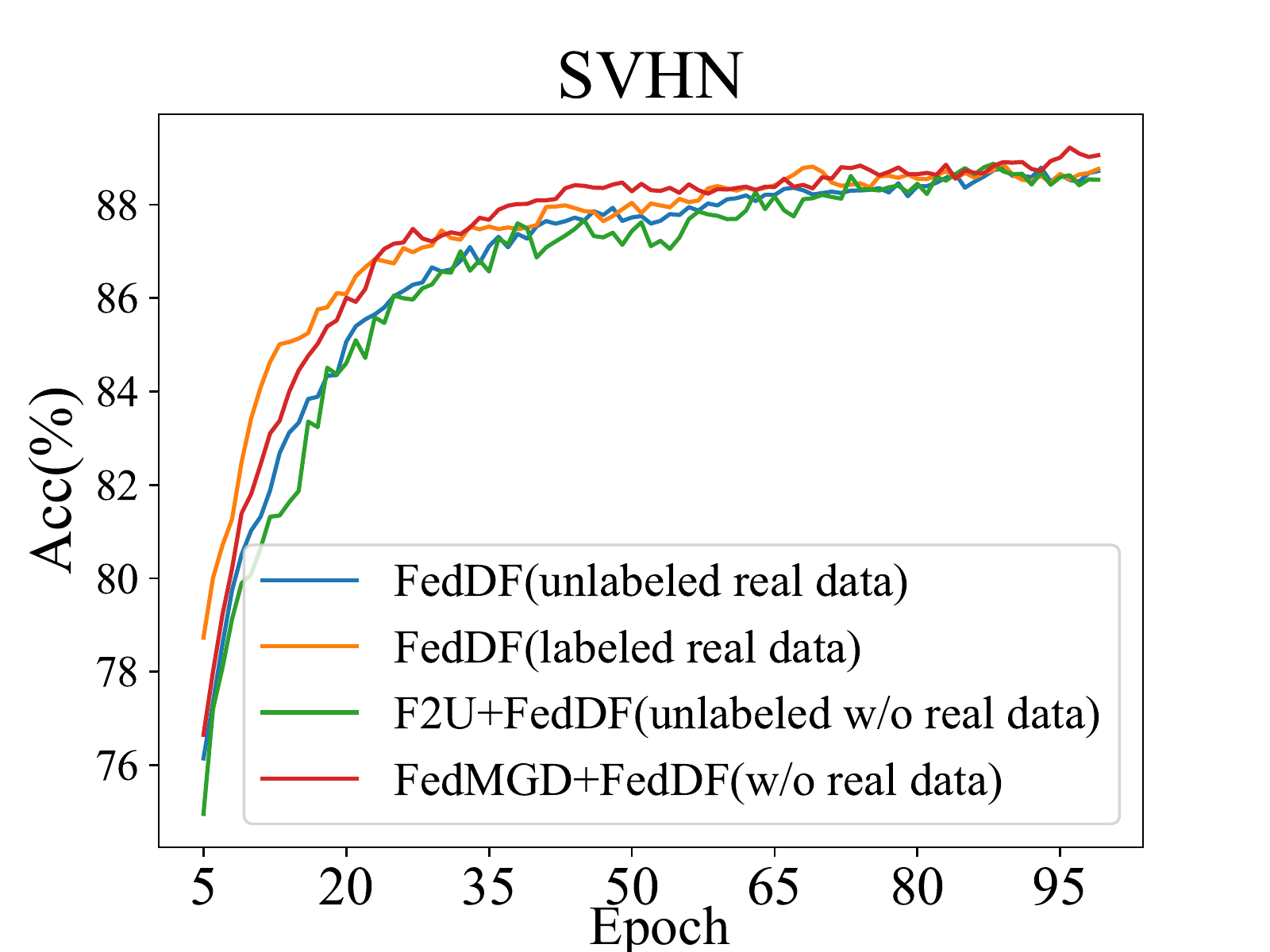}
		\includegraphics[width=1.9in]{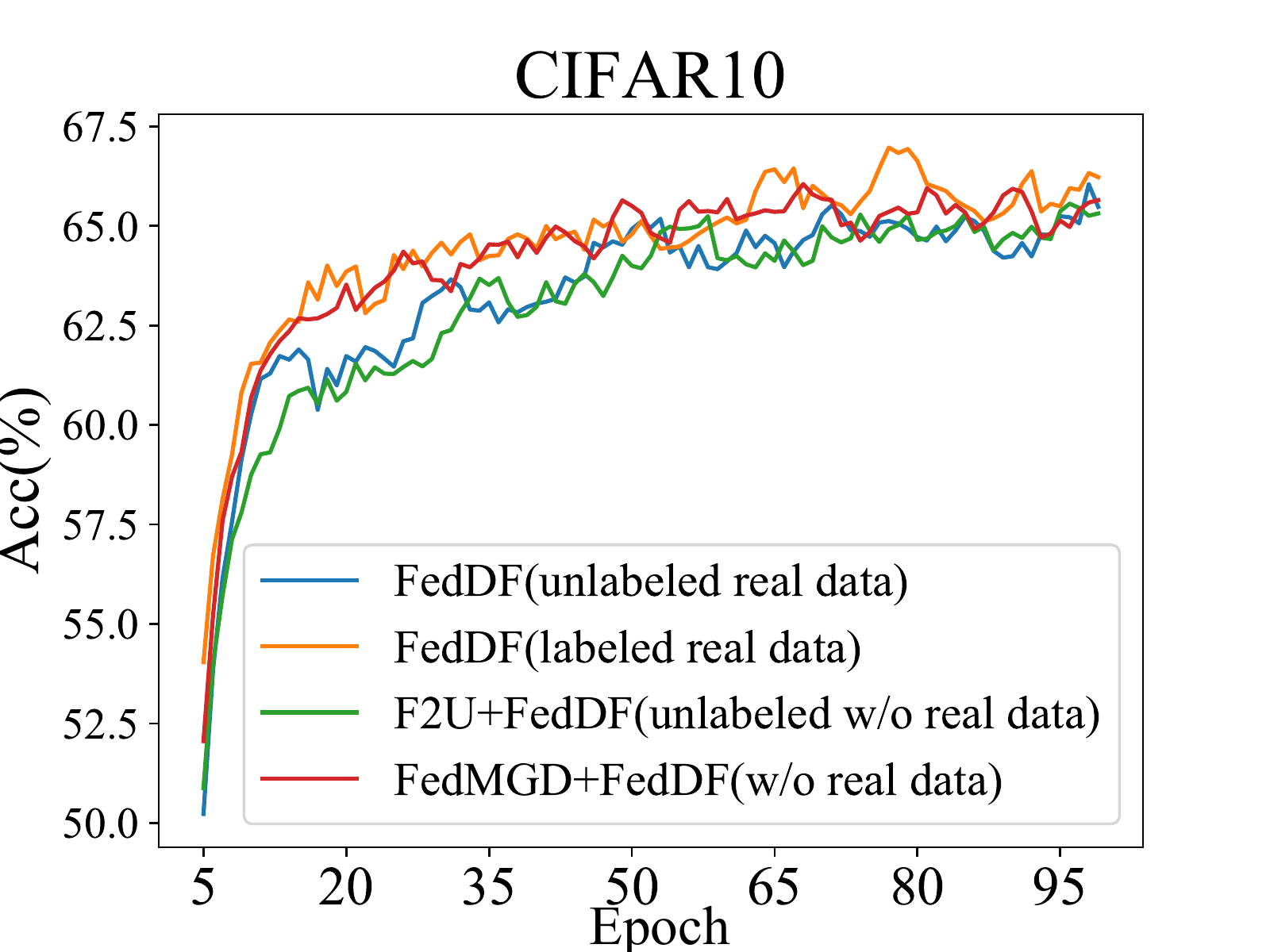}
		%\caption{fig1}
	\end{minipage}%
	}
	\centering
	\caption{Visualization of the federated distillation performance of different data sources in the label distribution skew scenarios.}
	\label{fig6}
\end{figure}

\subsection{Ablation Experiment} \label{4.3}

In this section, we mainly demonstrate the rationality of the proposed FedMGD through several ablation experiments. First, we verify that the proposed Realistic Score is more suitable for the case of label distribution skewing by comparing it with the collected data and F2U. Then, to demonstrate the quality of the data generated by FedMGD in the case of skewed label distribution, we compare it with several distributed generation models. Finally, we validate the rationality of FedMGD's preset label sampling approach.

\subsubsection{Realistic Score in FedMGD} \label{4.3.1}

In this subsection, we demonstrate from different perspectives that the proposed Realistic Score (Eq.~\eqref{eq3}) is beneficial for modeling the global data distribution. Before that, we need a downstream task that can use different data sources and use the model accuracy of the downstream task as a measure. We choose to experiment on a federated distillation algorithm (FedDF [18]) that requires an additional sources, and we can choose different data sources for the additional distillation dataset.

\begin{table}[h]
	\begin{center}
		\caption{\textbf{Verify the importance of label information.} We used 5\% of the real data collected from the client as an additional data source and compared it under different label distribution skew scenarios using both unlabeled and labeled methods.}
		\renewcommand{\arraystretch}{1.1}
		% \resizebox{1.\columnwidth}{!}{
     	\setlength{\tabcolsep}{2.6mm}{
                \begin{tabular}{c|c|c|c|c}
                \hline
                \textbf{Dataset}                                                                         & \textbf{Method}                     & \textbf{$\alpha$=0.01}      & \textbf{$\alpha$=0.05}      & \textbf{$\alpha$=0.1}       \\ \hline
                \multirow{2}{*}{\textbf{EMNIST}}                                                  & \textbf{FedDF(unlabeled real data)} & 87.63±0.30                  & 89.27±0.27                  & 90.32±0.26                  \\ \cline{2-5} 
                                                                                                  & \textbf{FedDF(labeled real data)}   & \textbf{89.69±0.49 \textcolor{green!70!black}{(↑2.06)}} & \textbf{90.91±0.17 \textcolor{green!70!black}{(↑1.64)}} & \textbf{90.97±0.30 \textcolor{green!70!black}{(↑0.65)}} \\ \hline
                \multirow{2}{*}{\textbf{\begin{tabular}[c]{@{}c@{}}Fashion\\ MNIST\end{tabular}}} & \textbf{FedDF(unlabeled real data)} & 71.02±0.72                  & 83.67±0.80                  & 87.53±0.95                  \\ \cline{2-5} 
                                                                                                  & \textbf{FedDF(labeled real data)}   & \textbf{74.03±1.14 \textcolor{green!70!black}{(↑3.01)}} & \textbf{85.19±1.78 \textcolor{green!70!black}{(↑1.52)}} & \textbf{88.61±0.44 \textcolor{green!70!black}{(↑1.08)}} \\ \hline
                \multirow{2}{*}{\textbf{SVHN}}                                                    & \textbf{FedDF(unlabeled real data)} & 72.67±0.79                  & 84.99±0.69                  & 88.65±0.22                  \\ \cline{2-5} 
                                                                                                  & \textbf{FedDF(labeled real data)}   & \textbf{73.74±0.84 \textcolor{green!70!black}{(↑1.07)}} & \textbf{85.23±0.51 \textcolor{green!70!black}{(↑0.24)}} & \textbf{88.88±0.19 \textcolor{green!70!black}{(↑0.23)}} \\ \hline
                \multirow{2}{*}{\textbf{CIFAR10}}                                                 & \textbf{FedDF(unlabeled real data)} & 47.25±1.40                  & 60.27±0.39                  & 64.58±0.95                  \\ \cline{2-5} 
                                                                                                  & \textbf{FedDF(labeled real data)}   & \textbf{52.86±1.98 \textcolor{green!70!black}{(↑5.61)}} & \textbf{62.50±0.19 \textcolor{green!70!black}{(↑2.23)}} & \textbf{66.78±0.23 \textcolor{green!70!black}{(↑2.20)}} \\ \hline
                \end{tabular}
            }
        % }
    \label{feddf_real_data}
	\end{center}
\end{table}

\textbf{(1) The effect of Realistic Score.} We verify the impact of label information of data on model performance when using real data sources as additional data sources. 
% We used 5\% of the real data collected from the client as an additional data source and compared it under different label offset scenarios using both unlabeled and labeled methods.
We use 5\% real data collected from customers as an additional data source and compare it under different label bias scenarios using unlabeled and labeled methods.
As shown in Table~\ref{feddf_real_data}, the model has better performance when the labeled real data is used as an additional data source, but using the way the data is collected may leak more users' privacy.
As shown in Figure~\ref{fig6}, FedDF(labeled real data) and FedDF(unlabeled real data) are the accuracies of training models using 5\% of real data with and without labels, respectively. The figure demonstrates that the accuracy of the trained model using real labeled data is significantly higher than that of the trained model using unlabeled real data. This proves that labels play an important role in model training and the necessity of introducing semantic truth in the Realistic Score.

\begin{table}[h]
	\begin{center}
		\caption{\textbf{Verify the effect of the generated data.} The generator(FedMGD) trained by Realistic Score  is used as the data source for FedDF and compared with the collected real data with labels (gray areas).}
		\renewcommand{\arraystretch}{1.1}
		% \resizebox{1.\columnwidth}{!}{
     		% \setlength{\tabcolsep}{1.5mm}{
    %  		\rowcolors{2}{red!20}{blue!20}
                
                \begin{tabular}{c|c|c|c|c}
                
                \hline
                \textbf{Dataset}                                                                         & \textbf{Method}                      & \textbf{$\alpha$=0.01}      & \textbf{$\alpha$=0.05}      & \textbf{$\alpha$=0.1}       \\ \hline
                
                \multirow{2}{*}{\textbf{EMNIST}}                                               & \cellcolor{gray!20} \textbf{FedDF(labeled real data)}    & \cellcolor{gray!20} 89.69±0.49                  & \cellcolor{gray!20} 90.91±0.17                  & \cellcolor{gray!20} 90.97±0.30                  \\ \cline{2-5} 
                                                                                                  & \textbf{FedMGD+FedDF(w/o real data)} & 89.19±0.10                  & 89.89±0.41                  & \textbf{91.09±0.25 \textcolor{green!70!black}{(↑0.12)}} \\  \hline
                \multirow{2}{*}{\textbf{\begin{tabular}[c]{@{}c@{}}Fashion\\ MNIST\end{tabular}}} & \cellcolor{gray!20} \textbf{FedDF(labeled real data)}    & \cellcolor{gray!20} 74.03±1.14                  & \cellcolor{gray!20} 85.19±1.78                  & \cellcolor{gray!20} 88.61±0.44                  \\ \cline{2-5} 
                                                                                                  & \textbf{FedMGD+FedDF(w/o real data)} & 73.22±1.21                  & \textbf{85.90±0.49 \textcolor{green!70!black}{(↑0.71)}} & 88.17±0.39                  \\ \hline
                \multirow{2}{*}{\textbf{SVHN}}                                                    & \cellcolor{gray!20} \textbf{FedDF(labeled real data)}    & \cellcolor{gray!20} 73.74±0.84                  & \cellcolor{gray!20} 85.23±0.51                  & \cellcolor{gray!20} 88.88±0.19                  \\ \cline{2-5} 
                                                                                                  & \textbf{FedMGD+FedDF(w/o real data)} & \textbf{75.16±1.28 \textcolor{green!70!black}{(↑1.42)}} & \textbf{86.98±0.80 \textcolor{green!70!black}{(↑1.72)}} & \textbf{89.09±0.34 \textcolor{green!70!black}{(↑0.21)}} \\ \hline
                \multirow{2}{*}{\textbf{CIFAR10}}                                                 & \cellcolor{gray!20} \textbf{FedDF(labeled real data)}    & \cellcolor{gray!20} 52.86±1.98                  & \cellcolor{gray!20} 62.50±0.19                  & \cellcolor{gray!20} 66.78±0.23                  \\ \cline{2-5} 
                                                                                                  & \textbf{FedMGD+FedDF(w/o real data)} & 51.12±1.01                  & 61.83±0.98                  & 65.74±0.52                  \\ \hline
                \end{tabular}
            % }
        % }
    \label{feddf_fedmgd_real}
	\end{center}
\end{table}

\textbf{(2) The effect of generated data.} In this experiment, we use the generator of FedMGD trained by Realistic Score as the data source of FedDF and compare it with the collected real data with labels. The experimental results are shown in Table~\ref{feddf_fedmgd_f2u}. The generated data using the generator trained by Realistic Score as an additional data source can approximate the effect of real data. We show the accuracy variation of the two approaches in Figure~\ref{fig6}, where FedDF(labeled real data) is the accuracy of training the model using 5\% of the real data with labels, and FedMGD+FedDF(w/o real data) is the accuracy of training the model using the generated data after modeling with FedMGD. The figure shows that FedMGD can approximate an accuracy similar to that of using real labeled data by global modeling.

\textbf{(3) Comparison with F2U.} In this experiment, we compare the proposed Realistic Score with the generator trained by the Largest Score proposed in F2U. As shown in Table~\ref{feddf_fedmgd_real}, the generator of FedMGD trained by Realistic Score can reach better results as the data source of FedDF. Because only the data realness is concerned in F2U and ignore the label realness information, the Realistic Score approach is more suitable for global modeling of data under label distribution skew.
As shown in Figure~\ref{fig6}, F2U+FedDF(unlabeled w/o real data) and FedMGD+FedDF(w/o real data) are the accuracy curves of the trained models using F2U and FedMGD as the data source of FedDF, respectively. Since FedMGD learns more information about the label distribution using Realistic Score, the accuracy of the training model is higher than that of the F2U method.

\begin{table}[t]
	\begin{center}
		\caption{\textbf{Comparison of FedMGD with F2U.} The generator obtained after FedMGD and F2U training is used as an additional data source for FedDF, to compare the performance(\%) achieved by the two modeling methods on downstream tasks, respectively.}
		\renewcommand{\arraystretch}{1.1}
		% \resizebox{1.\columnwidth}{!}{
     		\setlength{\tabcolsep}{1.5mm}{
                \begin{tabular}{c|c|c|c|c}
                \hline
                \textbf{Dataset}                                                                         & \textbf{Method}                             & \textbf{$\alpha$=0.01}      & \textbf{$\alpha$=0.05}      & \textbf{$\alpha$=0.1}       \\ \hline
                \multirow{2}{*}{\textbf{EMNIST}}                                                  & \textbf{F2U+FedDF(unlabeled w/o real data)} & 87.88±1.04                  & 89.45±0.34                  & 90.71±0.21                  \\ \cline{2-5} 
                                                                                                  & \textbf{FedMGD+FedDF(w/o real data)}        & \textbf{89.19±0.10 \textcolor{green!70!black}{(↑1.22)}} & \textbf{89.89±0.41 \textcolor{green!70!black}{(↑0.44)}} & \textbf{91.09±0.25 \textcolor{green!70!black}{(↑0.38)}} \\ \hline
                \multirow{2}{*}{\textbf{\begin{tabular}[c]{@{}c@{}}Fashion\\ MNIST\end{tabular}}} & \textbf{F2U+FedDF(unlabeled w/o real data)} & 71.76±0.53                  & 81.47±0.82                  & 87.61±0.37                  \\ \cline{2-5} 
                                                                                                  & \textbf{FedMGD+FedDF(w/o real data)}        & \textbf{73.22±1.21 \textcolor{green!70!black}{(↑1.46)}} & \textbf{85.90±0.49 \textcolor{green!70!black}{(↑4.43)}} & \textbf{88.17±0.39 \textcolor{green!70!black}{(↑0.56)}} \\ \hline
                \multirow{2}{*}{\textbf{SVHN}}                                                    & \textbf{F2U+FedDF(unlabeled w/o real data)} & 73.40±0.72                  & 84.30±0.53                  & 88.63±0.25                  \\ \cline{2-5} 
                                                                                                  & \textbf{FedMGD+FedDF(w/o real data)}        & \textbf{75.16±1.28 \textcolor{green!70!black}{(↑1.76)}} & \textbf{86.98±0.80 \textcolor{green!70!black}{(↑2.68)}} & \textbf{89.09±0.34 \textcolor{green!70!black}{(↑0.46)}} \\ \hline
                \multirow{2}{*}{\textbf{CIFAR10}}                                                 & \textbf{F2U+FedDF(unlabeled w/o real data)} & 45.99±0.32                  & 61.61±0.40                  & 65.74±0.47                  \\ \cline{2-5} 
                                                                                                  & \textbf{FedMGD+FedDF(w/o real data)}        & \textbf{51.12±1.01 \textcolor{green!70!black}{(↑5.13)}} & \textbf{61.83±0.98 \textcolor{green!70!black}{(↑0.22)}} & \textbf{65.74±0.52 \textcolor{green!70!black}{(↑0.05)}} \\ \hline
                \end{tabular}
            }
        % }
    \label{feddf_fedmgd_f2u}
	\end{center}
\end{table}

\subsubsection{Compare with Other Distributed GANs.} \label{4.3.2}
\begin{table}[t]
	\begin{center}
	    \small	
		\caption{\textbf{Quality evaluation of images generated by FedMGD.} Comparison of the FID of the images generated by FedMGD and other distributed GANs under different data distributions.}
		\renewcommand{\arraystretch}{1.05}
		% \resizebox{0.85\columnwidth}{!}{
 		\setlength{\tabcolsep}{7.3mm}{

			\begin{tabular}{c|c|c|c|c|c}

				\hline
				\textbf{Dataset}                                                                  & $\boldsymbol{\alpha}$ & \textbf{FedGAN} & \textbf{MD-GAN} & \textbf{F2U} & \textbf{FedMGD} \\ \hline
				\multirow{3}{*}{\textbf{EMNIST}}                                                  & 0.01     & 198.37          & 129.59          & 39.79        & \textbf{22.86} \textcolor{green!70!black}{($\downarrow$16.93)}  \\
				& 0.05     & 121.08          & 128.59          & 33.63        & \textbf{19.99} \textcolor{green!70!black}{($\downarrow$13.64)}  \\
				& 0.1      & 136.83          & 95.27           & 33.35        & \textbf{18.69} \textcolor{green!70!black}{($\downarrow$14.66)}  \\ \hline
				\multirow{3}{*}{\textbf{\begin{tabular}[c]{@{}c@{}}Fashion\\ MNIST\end{tabular}}} & 0.01     & 230.96          & 152.29          & 41.53        & \textbf{39.71} \textcolor{green!70!black}{($\downarrow$1.82)}  \\
				& 0.05     & 196.19          & 158.26          & 47.35        & \textbf{36.49} \textcolor{green!70!black}{($\downarrow$10.86)}  \\
				& 0.1      & 215.98 & 141.51          & 46.18        & \textbf{40.21} \textcolor{green!70!black}{($\downarrow$5.97)}  \\ \hline
				\multirow{3}{*}{\textbf{SVHN}}                                                    & 0.01     & 245.91          & 209.26          & 140.53       & \textbf{126.04} \textcolor{green!70!black}{($\downarrow$14.94)} \\
				& 0.05     & 199.41          & 206.34          & 145.69       & \textbf{139.47} \textcolor{green!70!black}{($\downarrow$6.22)} \\
				& 0.1      & 226.36          & 208.28          & 145.37       & \textbf{121.01} \textcolor{green!70!black}{($\downarrow$24.36)} \\ \hline
				\multicolumn{1}{l|}{\multirow{3}{*}{\textbf{CIFAR10}}}                            & 0.01     & 264.69          & 306.03          & 218.26       & \textbf{206.17} \textcolor{green!70!black}{($\downarrow$12.09)} \\
				\multicolumn{1}{l|}{}                                                             & 0.05     & 255.31          & 281.67          & 206.02       & \textbf{201.67} \textcolor{green!70!black}{($\downarrow$4.35)} \\
				\multicolumn{1}{l|}{}                                                             & 0.1      & 296.09          & 315.36          & 206.09       & \textbf{202.17} \textcolor{green!70!black}{($\downarrow$3.92)} \\ \hline
			\end{tabular}
		}
	% }
		\label{tab4}
	\end{center}
\end{table}

In this subsection, we validate the image quality generated by FedMGD using the Realistic Score approach in scenarios with label distribution skew and the training effect in downstream tasks by comparing with other distributed GANs.

\textbf{(1) Quality evaluation of images generated by FedMGD.} We use F2U~\cite{10}, MD-GAN~\cite{9}, and FedGAN~\cite{16} as baseline algorithms, and use the distance (FID) between the generated data and the real data distribution as a measure of image quality. In Table~\ref{tab4}, we compare the FID of images generated by the FedMGD method and the other three methods. It can be found that FedMGD has obvious advantages in the label distribution skew scenario.

\begin{figure}[t]
	\centering
    \subfigure[FedMGD]{
		\begin{minipage}[t]{0.32\linewidth}
			\centering
			\includegraphics[width=1.9in]{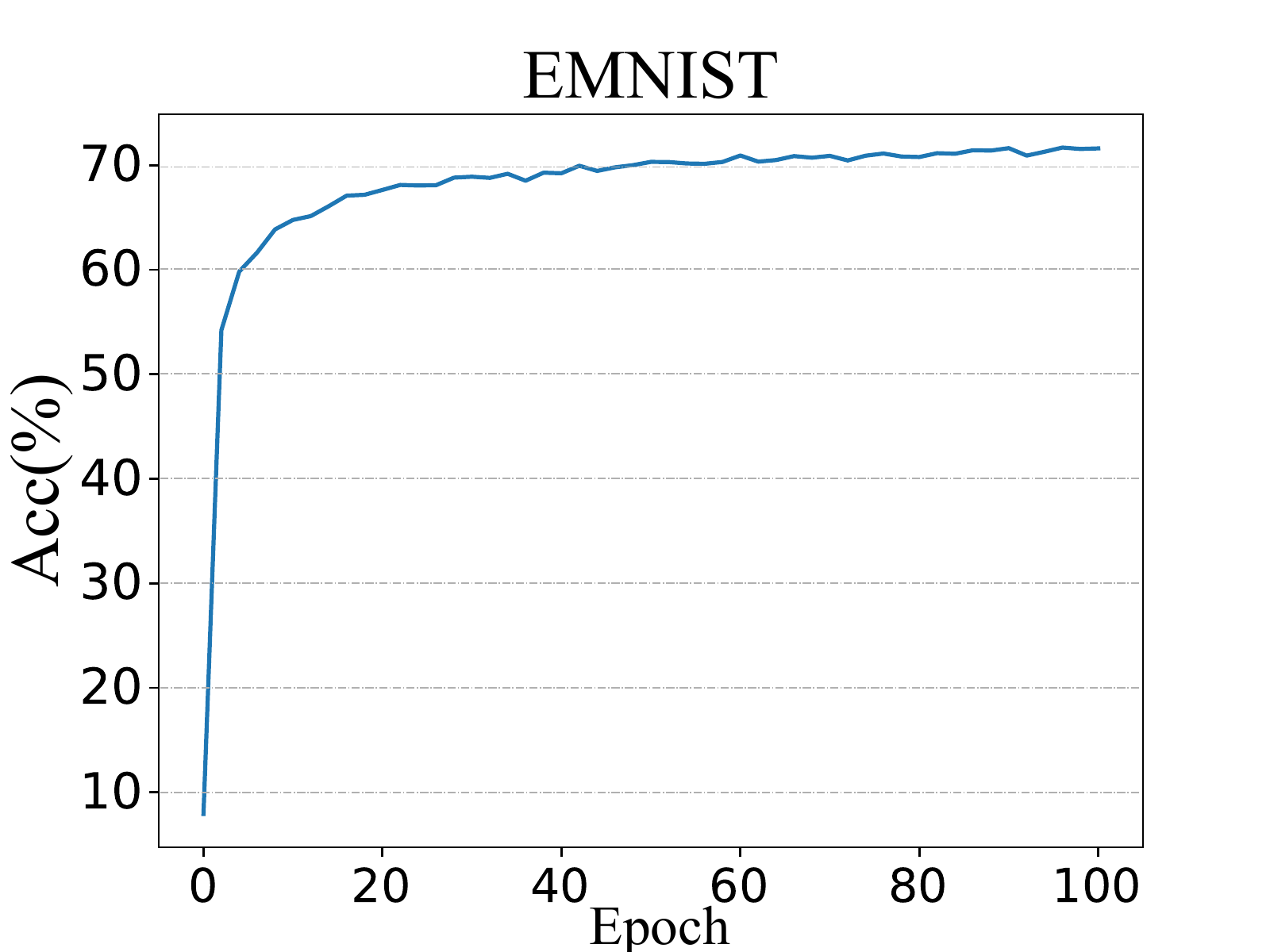}
            \includegraphics[width=1.9in]{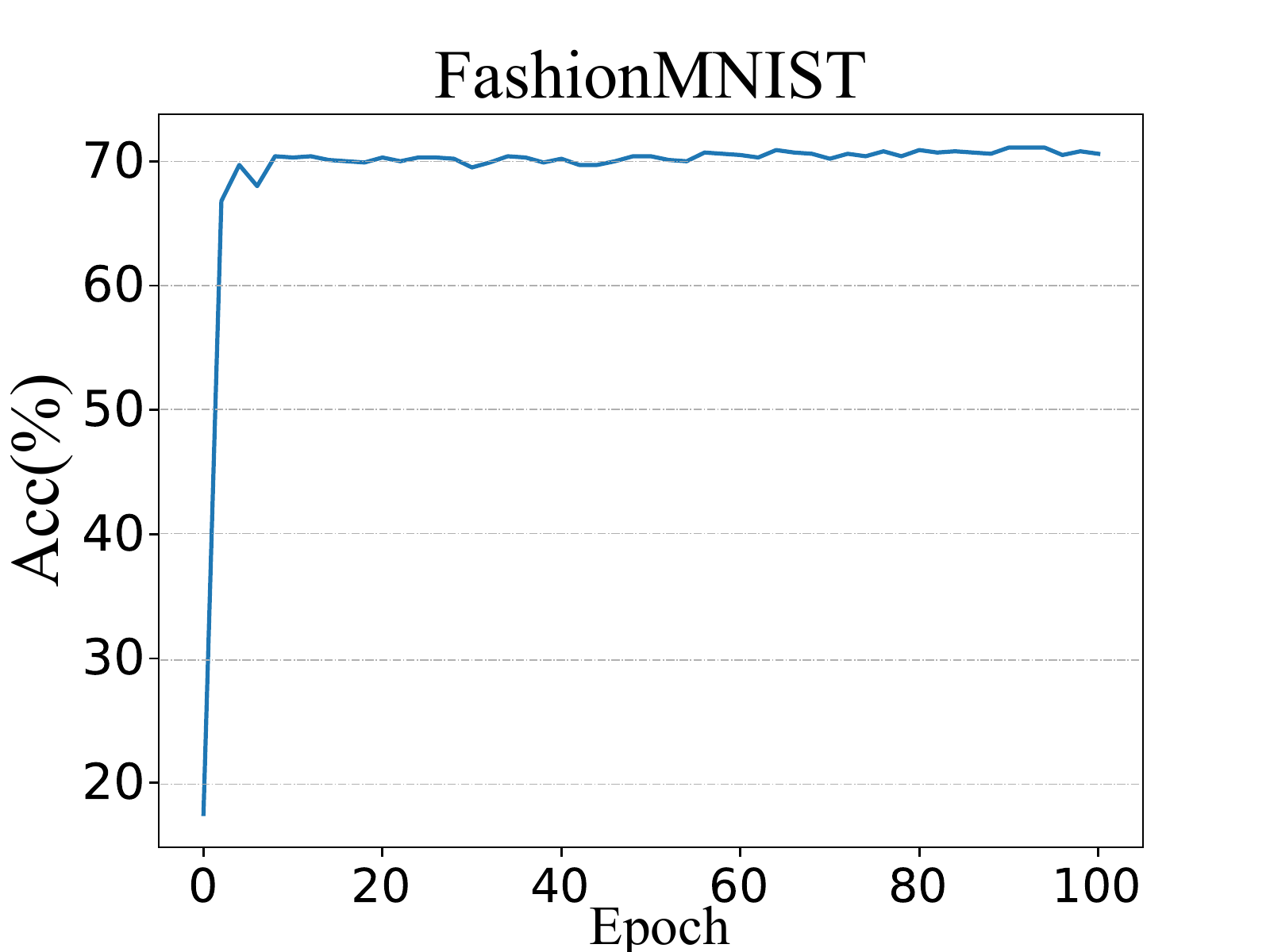}
            \includegraphics[width=1.9in]{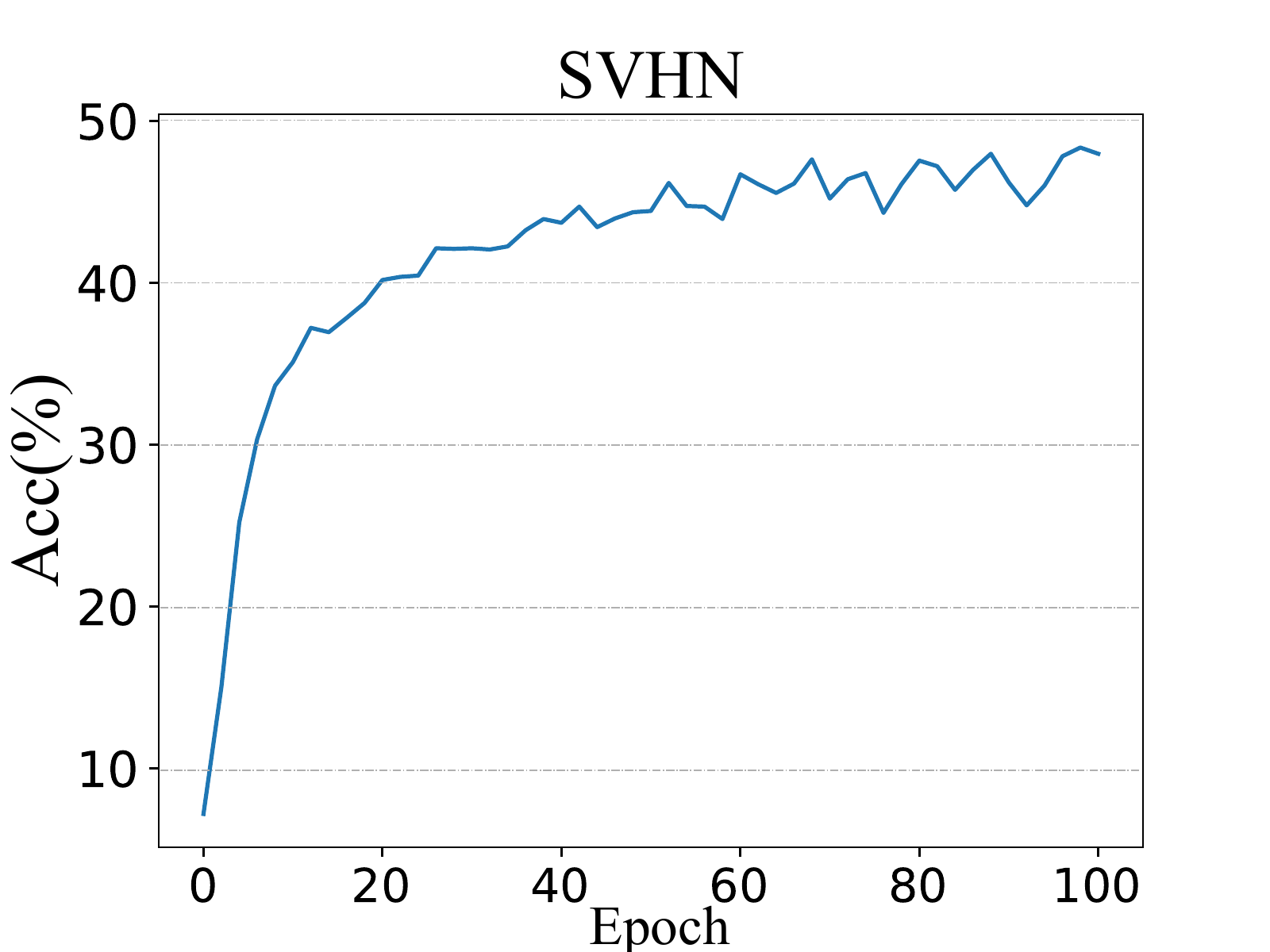}
			%\caption{fig1}
		\end{minipage}%
	}%
	\subfigure[MD-GAN]{
		\begin{minipage}[t]{0.32\linewidth}
			\centering
            \includegraphics[width=1.9in]{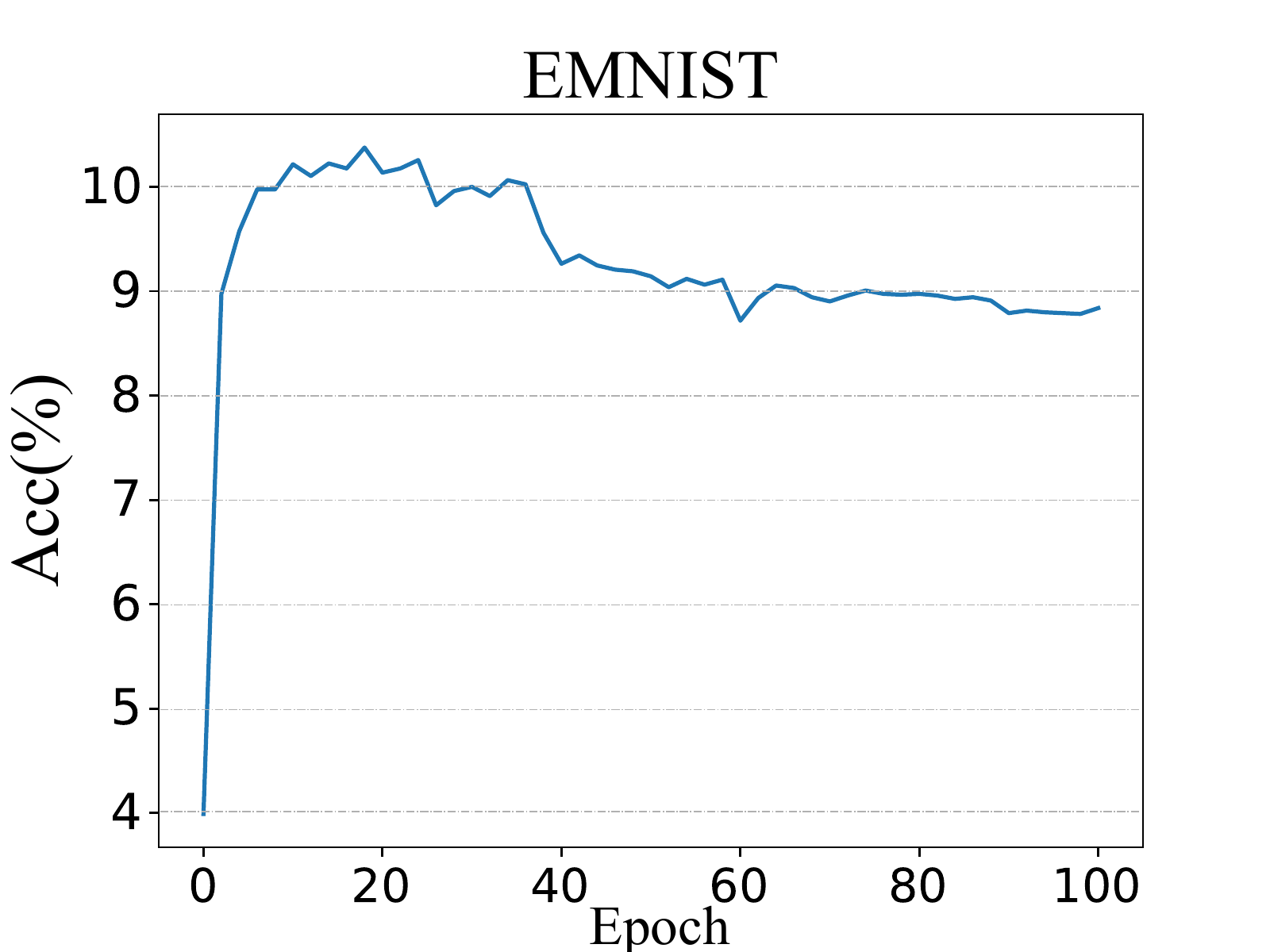}
			\includegraphics[width=1.9in]{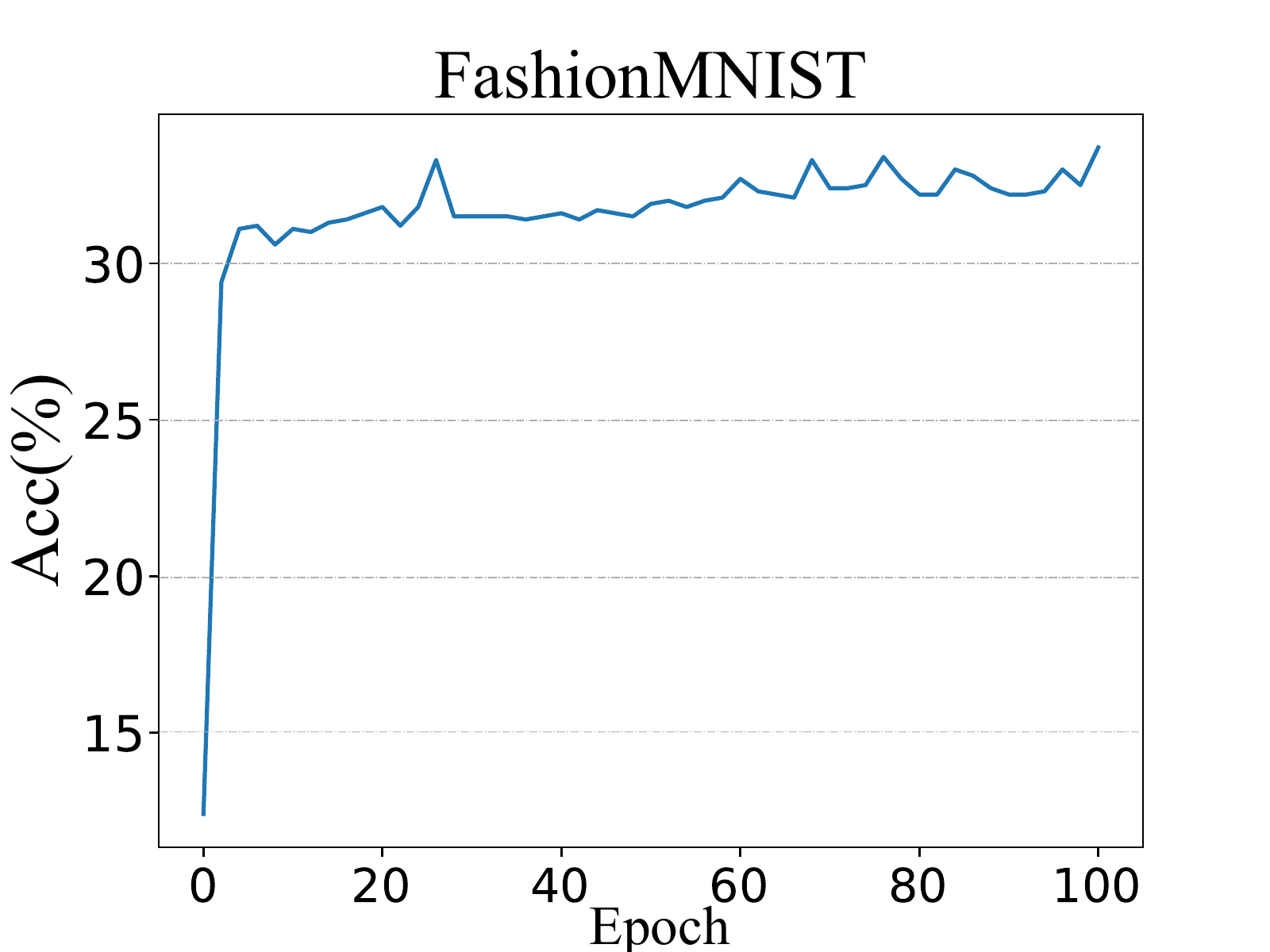}
            \includegraphics[width=1.9in]{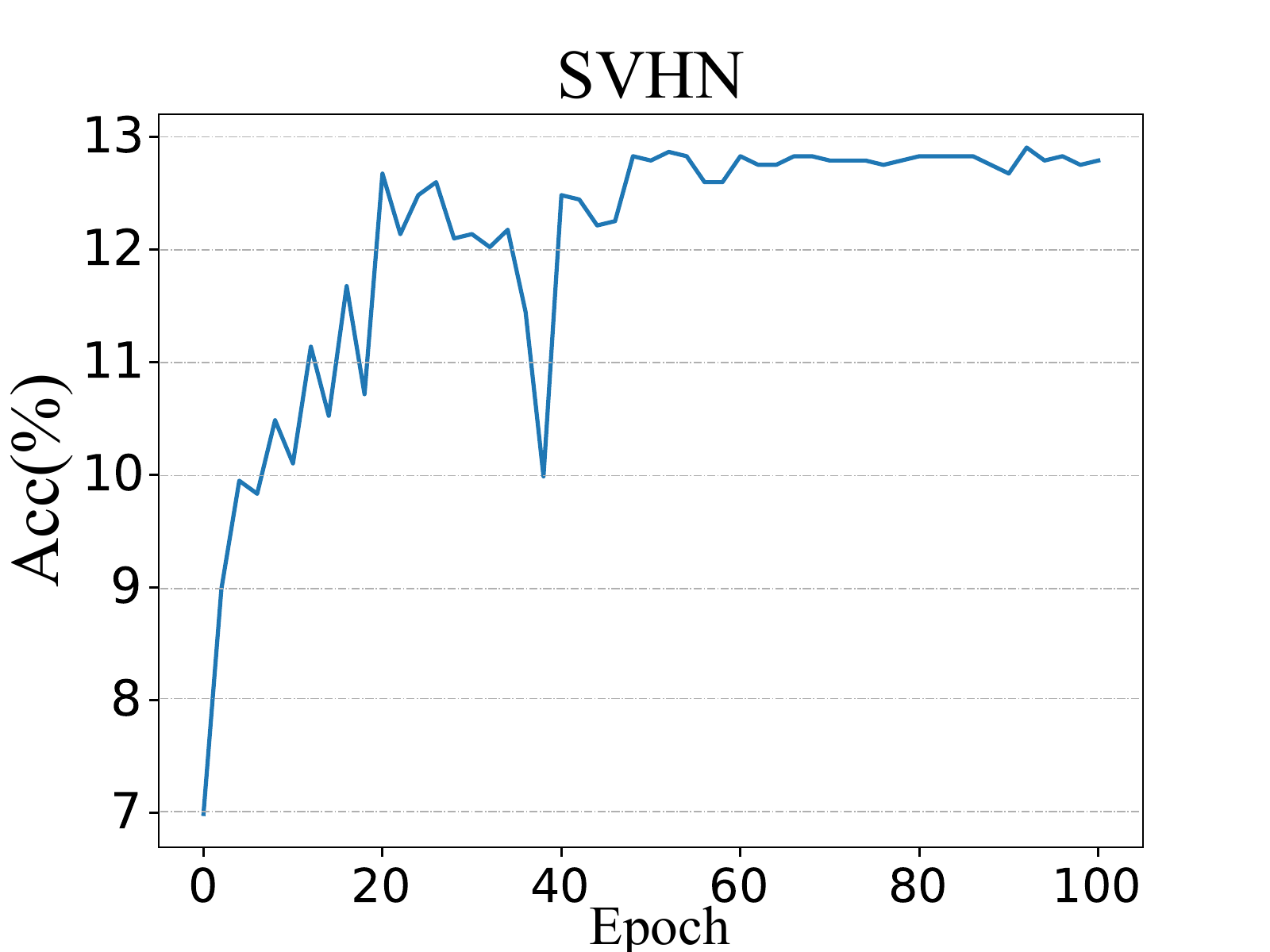}
		\end{minipage}%
	}
	\subfigure[FedGAN]{
		\begin{minipage}[t]{0.32\linewidth}
			\centering
            \includegraphics[width=1.9in]{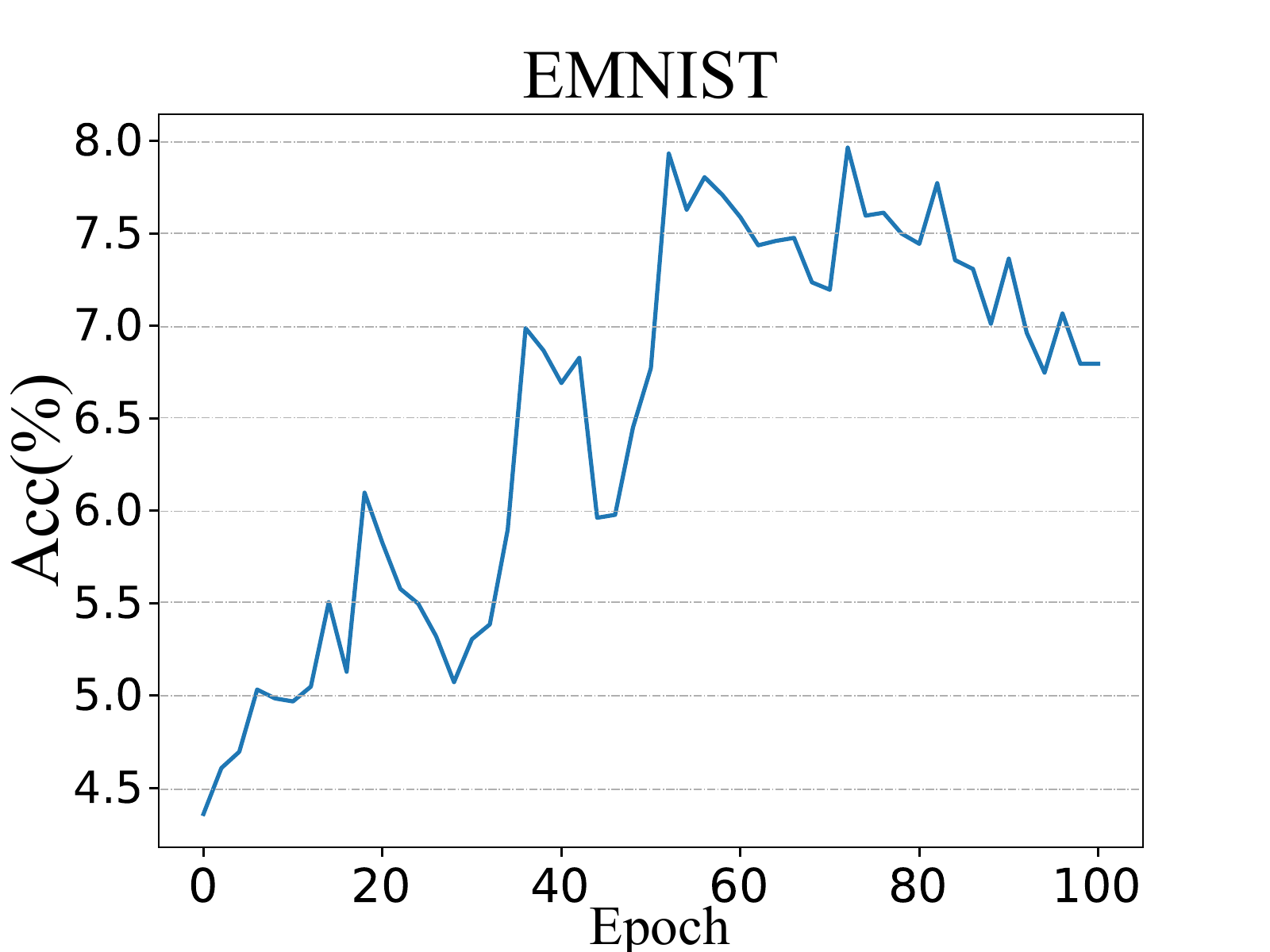}	
            \includegraphics[width=1.9in]{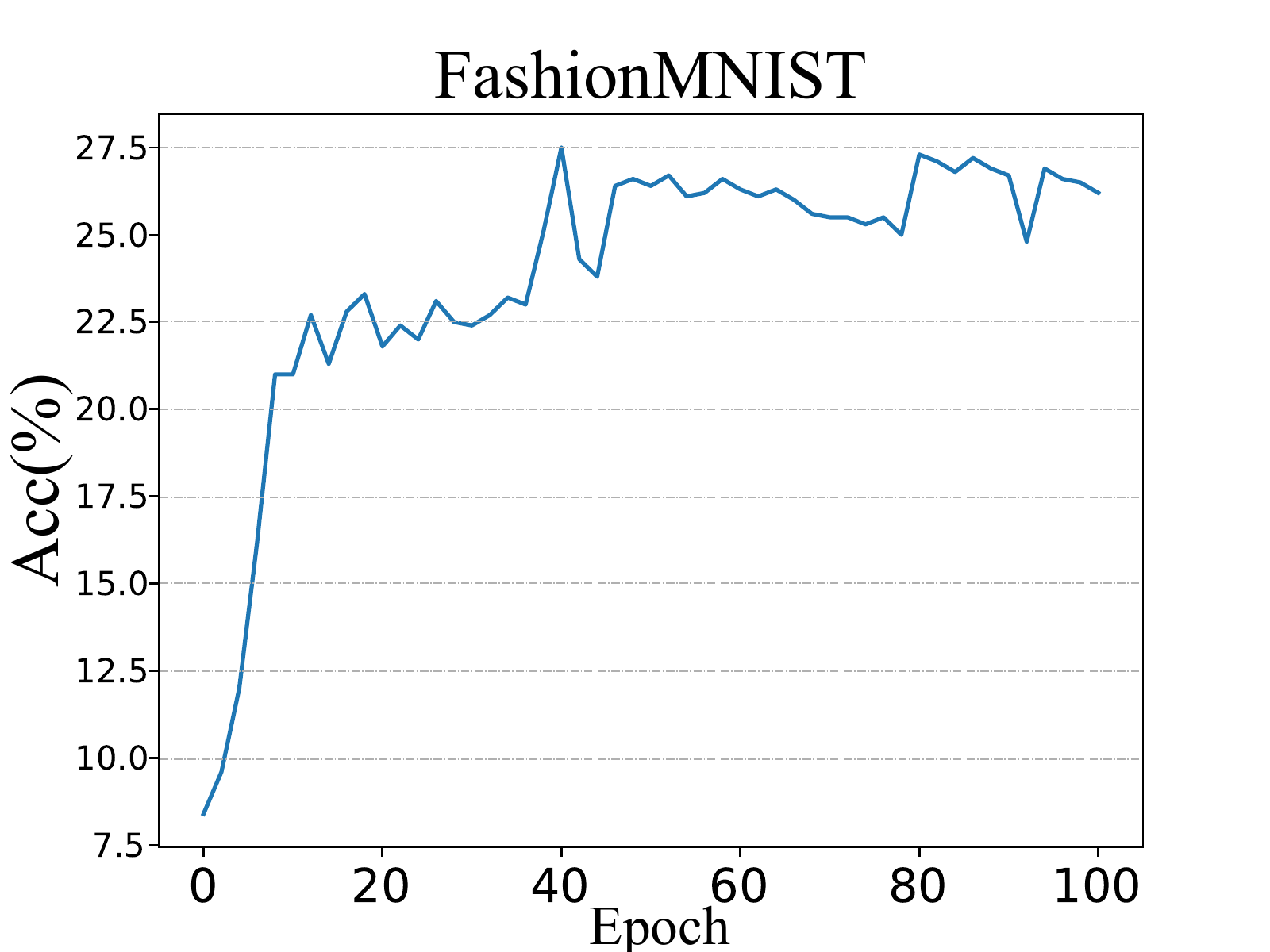}
            \includegraphics[width=1.9in]{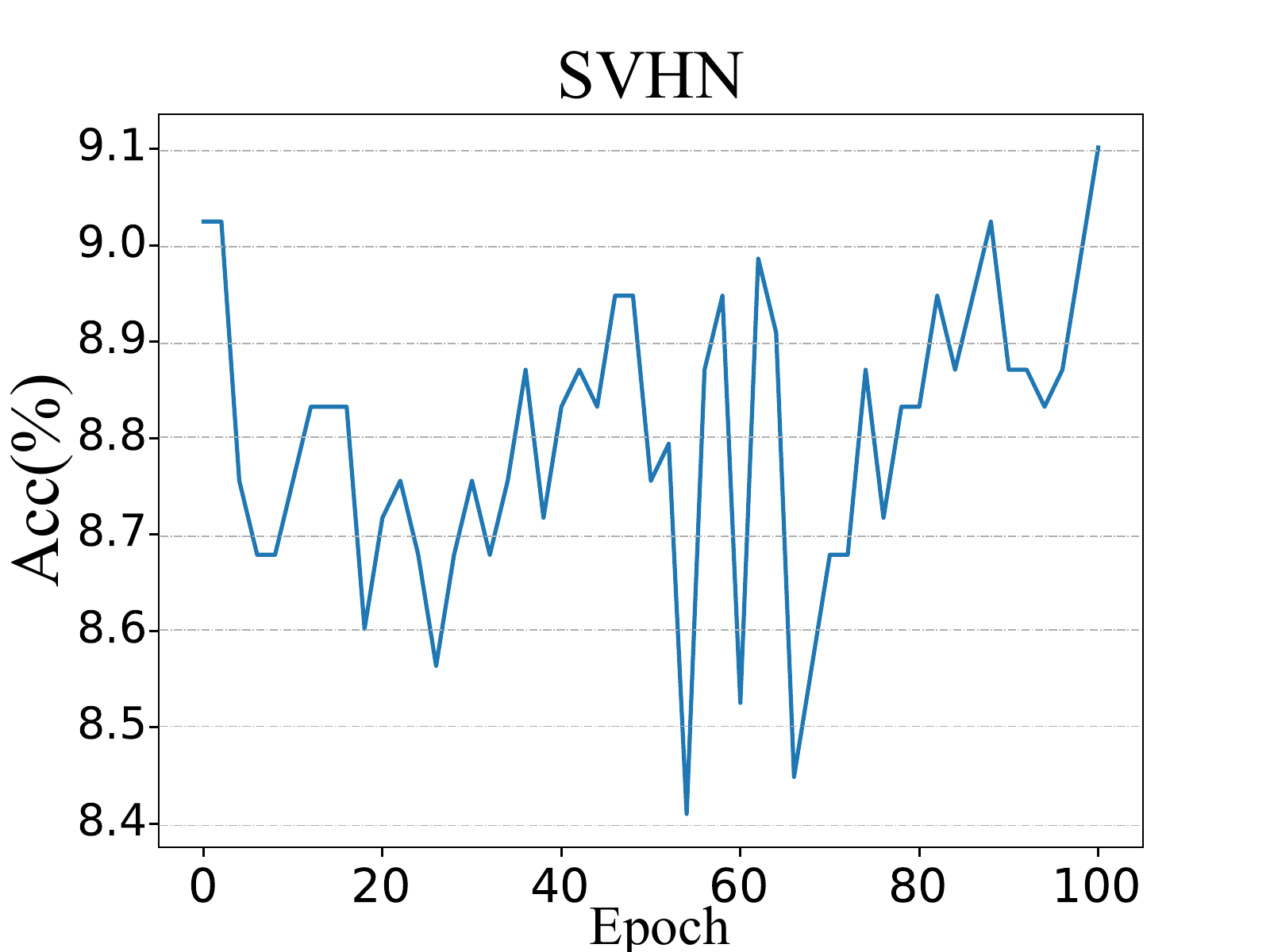}
		\end{minipage}%
    }	
	\centering
	\caption{Performance visualization of a separate classifier trained using the generator as data source. At this point $\alpha$ is 0.01.}
	\label{fig7}
\end{figure}

\textbf{(2) Verify the effectiveness of the generator for downstream tasks.} 
% To further verify whether the trained generator can be used alone for downstream tasks, we trained a separate classifier using the trained generator as the data source (since F2U can only generate unlabeled data, it did not participate in this Compare). 
To further verify that the trained generator can be used alone for downstream tasks, we train a separate classifier using the trained generator as the data source (F2U is not involved in this comparison since it can only generate unlabeled data).
As shown in Figure~\ref{fig7}, compared with other methods, classification model training using the data generated by the FedMGD generator can achieve good performance, which proves that the generator trained using FedMGD can be used for downstream tasks.

\subsubsection{Two-Stage vs. One-Stage FedMGD}
In this subsection, we experimentally validate the rationality of using two-stage of learning for FedMGD. In order to set up comparison experiments, we propose FedMGD using one-stage for training by fusing the federated enhancement stage and the generative adversarial stage of the two-stage. Specifically, we perform the aggregation of the local classifier $\{\mathcal{C}_i\}_{i=1}^K$ at the same time as $\mathcal{G}$ is updated by Realistic Score in Algorithm~\ref{alg1}. Then, we refine the aggregated model $\tilde{\mathcal{C}}$ by the global information currently learned by the generator. 

We conducted experiments on two datasets, FashionMNIST and SVHN, and the results are shown in Table~\ref{stage}. We can observe that the accuracy of FedMGD using two-stage for updating is significantly higher on different datasets than the approach that takes one stage for updating. This is due to the fact that in the original generative adversarial stage, the local classifier only uses the client-side local dataset for learning and describes the local label distribution information to the server through Realistic Score. However, in the process of using a stage update, as the classification model is aggregated in the global modeling process, the local model using aggregation cannot accurately describe the local label distribution information of the client. 
The generator $\mathcal{G}$ biases the modeling of the global label distribution by the Realistic Score returned locally, which eventually exhibits a degradation of the model performance.
% finally shows a degradation of the model performance. 

\begin{figure}[t]
	\centering
	\subfigure[$\alpha$=0.01]{
    	\begin{minipage}[t]{0.32\linewidth}
    		\centering
    		\includegraphics[width=1.9in]{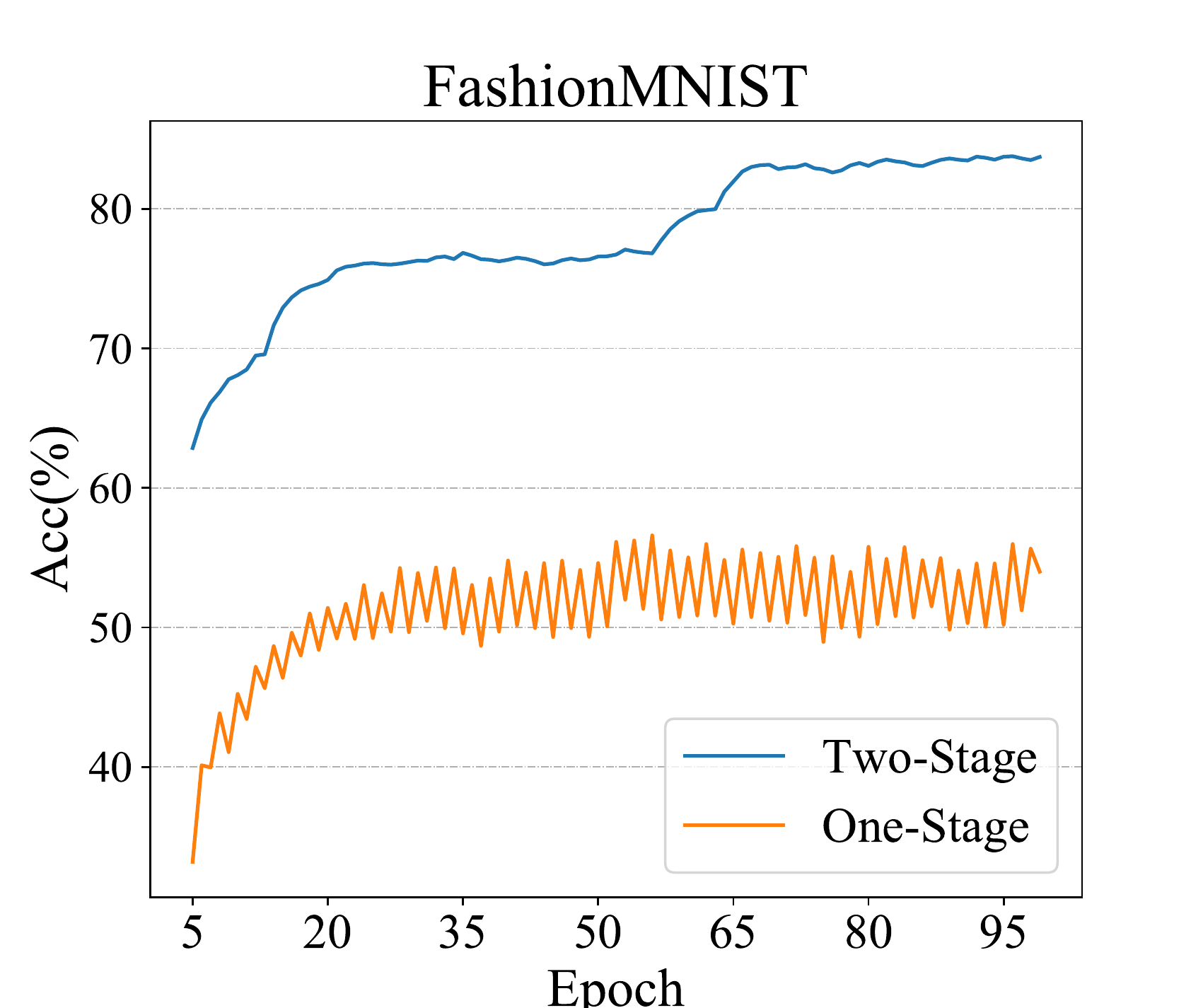}
      \includegraphics[width=1.9in]{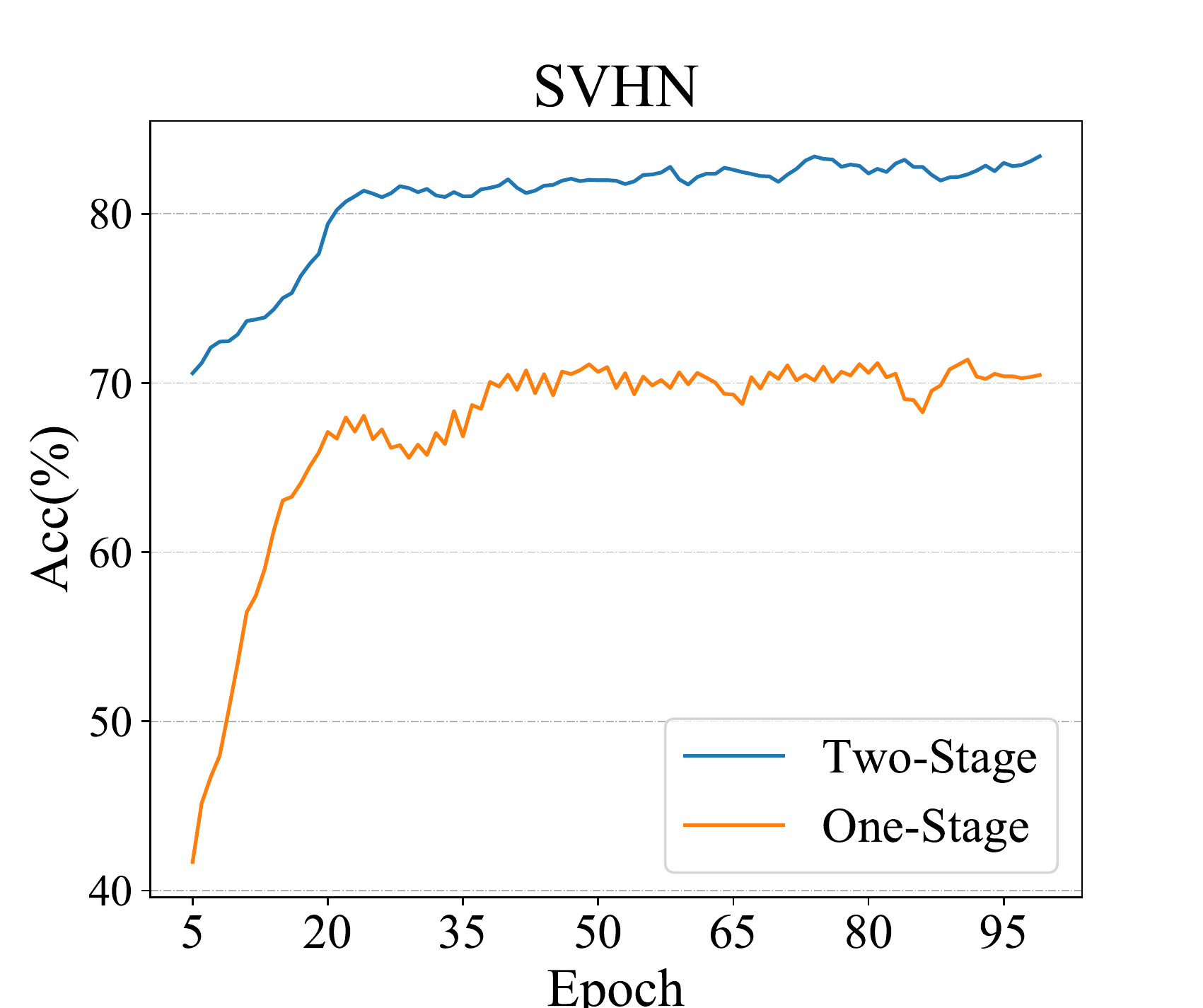}
    	\end{minipage}%
	}
	\subfigure[$\alpha$=0.05]{
    	\begin{minipage}[t]{0.32\linewidth}
    		\centering
    		\includegraphics[width=1.9in]{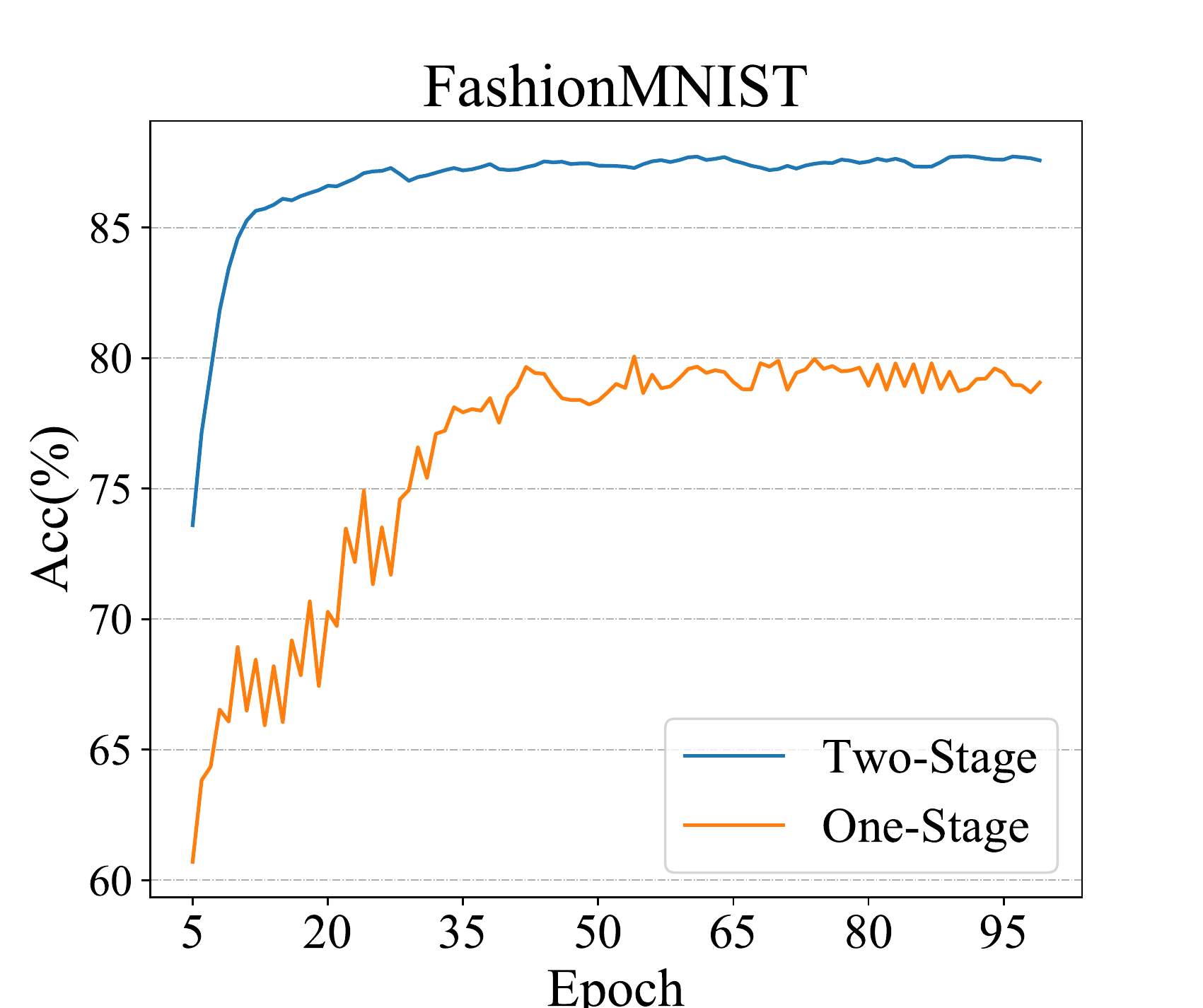}
      \includegraphics[width=1.9in]{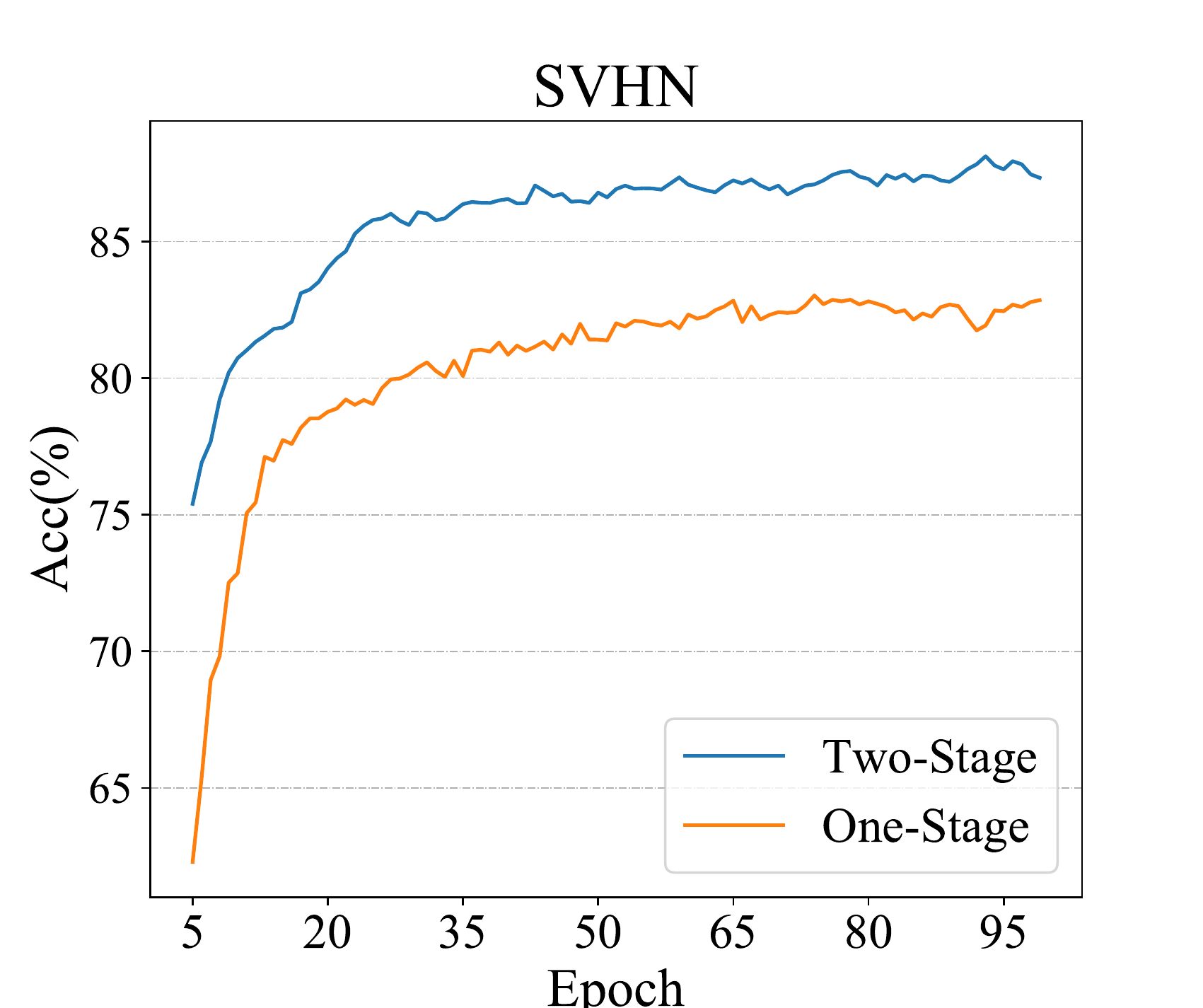}
    	\end{minipage}%
	}
	\subfigure[$\alpha$=0.1]{
    	\begin{minipage}[t]{0.32\linewidth}
    		\centering
    		\includegraphics[width=1.9in]{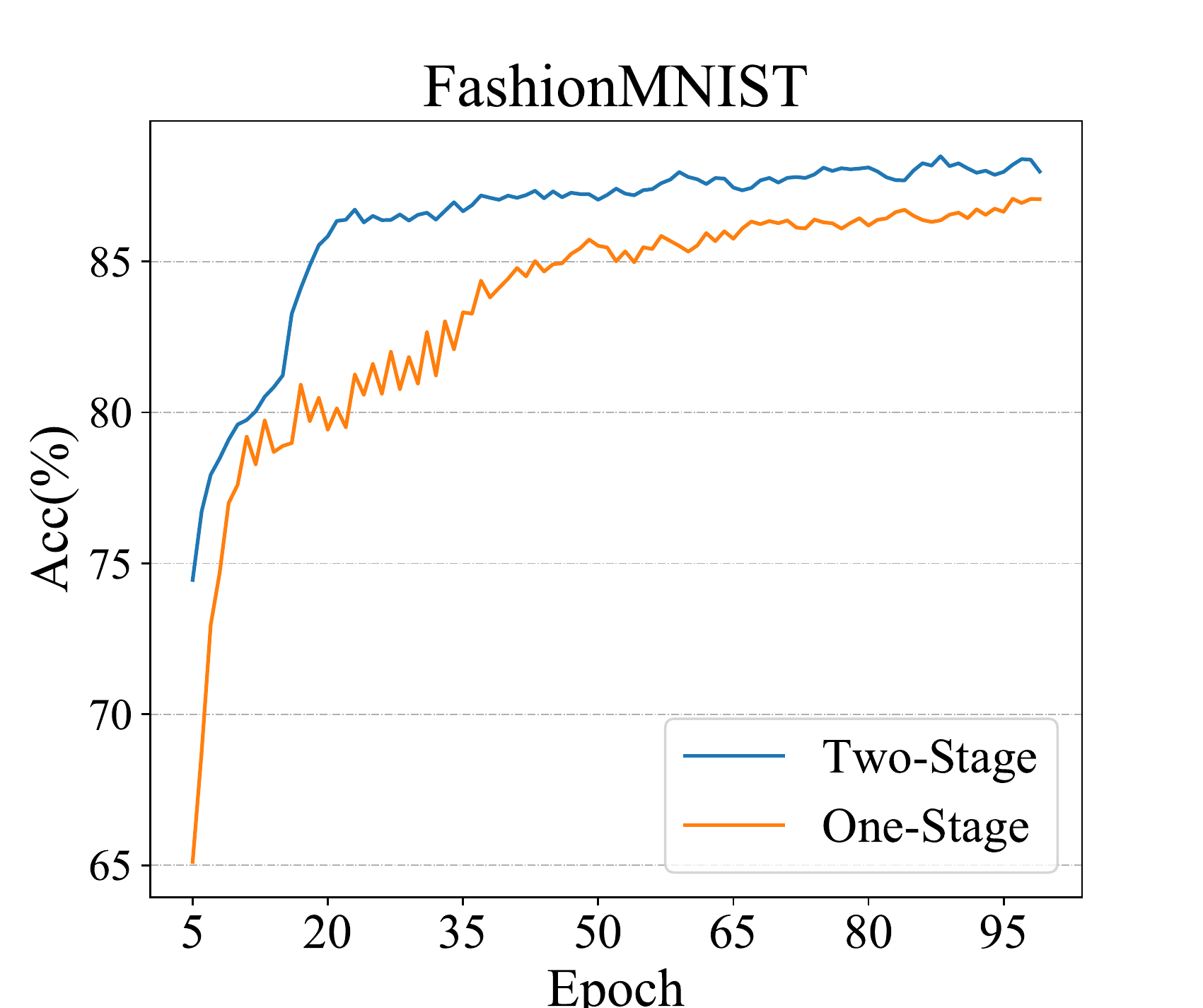}
      \includegraphics[width=1.9in]{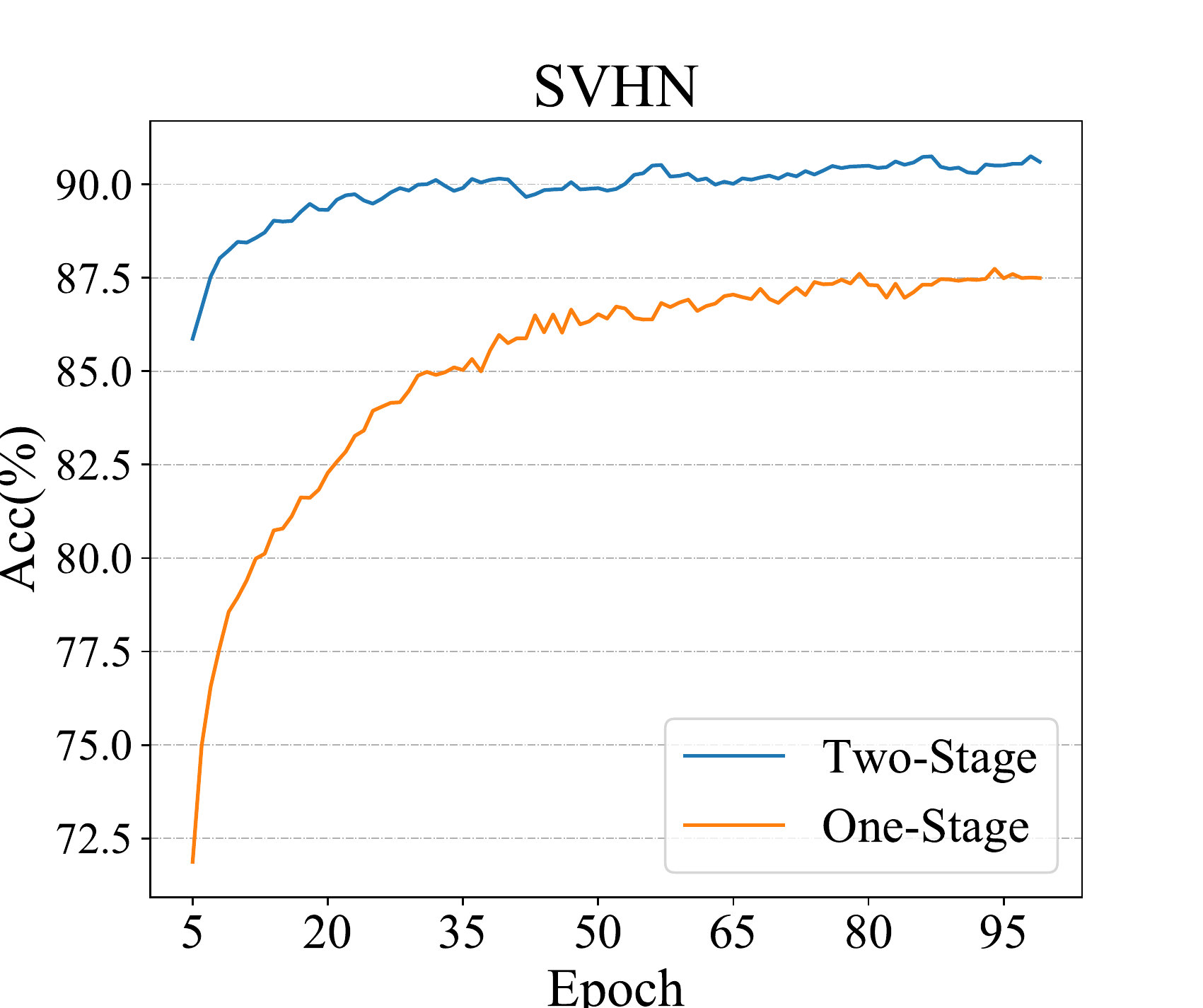}
    	\end{minipage}%
	}
	\caption{Visualization of two-stage vs. single-stage FedMGD training model performance. The results are obtained when number of clients is 5.}
	\label{stage_fig}
\end{figure}

Therefore, in FedMGD we adopt a two-stage update method to improve the performance of the federated learning model on the basis of ensuring the accuracy of the global modeling. We show the variation in accuracy for two different procedures for training the model in Figure~\ref{stage_fig}. 
It is shown from the figure that the accuracy of the model trained using the two-stage procedure is significantly higher than that using the one-stage procedure. Moreover, the difference in the accuracy of the trained models of the two different procedures gradually increases with the increase of the label distribution skew between clients.

\begin{table}[h]
\caption{\textbf{Two-Stage vs. One-Stage FedMGD.} FedMGD trains the model in one and two stages to compare the performance of the model under the two procedures (\%).}
\renewcommand{\arraystretch}{1.2}
 		\small		
% 		\resizebox{0.85\columnwidth}{!}{
 		\setlength{\tabcolsep}{4.5mm}{
\begin{tabular}{c|c|c|c|c}
\hline
\textbf{Dataset}                       & \textbf{Procedure}    & \textbf{$\alpha$=0.01} & \textbf{$\alpha$=0.05} & \textbf{$\alpha$=0.1} \\ \hline
\multirow{2}{*}{\textbf{FashionMNIST}} & \textbf{One-stage} & 51.14±2.26             & 80.10±1.52             & 85.97±2.10            \\ \cline{2-5} 
                                       & \textbf{Two-stage} & \textbf{84.04±0.58} \textcolor{green!70!black}{($\uparrow$32.90)}   & \textbf{87.57±0.40} \textcolor{green!70!black}{($\uparrow$7.47)}   & \textbf{89.29±1.33} \textcolor{green!70!black}{($\uparrow$3.32)}  \\ \hline
\multirow{2}{*}{\textbf{SVHN}}         & \textbf{One-stage} & 70.53±1.41                       & 83.40±1.56                       & 87.56±0.25                      \\ \cline{2-5} 
                                       & \textbf{Two-stage} & \textbf{84.14±0.91} \textcolor{green!70!black}{($\uparrow$13.61)}   & \textbf{88.62±0.39} \textcolor{green!70!black}{($\uparrow$5.22)}   & \textbf{90.47±0.37} \textcolor{green!70!black}{($\uparrow$2.91)}  \\ \hline
\end{tabular}
}

\label{stage}
\end{table}

\subsubsection{Sampling Methods in FedMGD.} \label{4.3.3}
In this section we compare the effects of modeling the global data distribution using different generator preset label sampling methods and validate the rationality of using server-side sampling in FedMGD.
Based on the characteristics of data distribution in a distributed environment, we propose two different sampling methods, Client Sampling and Server Sampling. Client sampling refers to collecting the distribution of each label in the client, so that the preset labels of the generator are sampled according to the label distribution in the client. Server Sampling means that the generator's preset tags are sampled on the server side using a uniform distribution.

\begin{figure}[t]
	\centering
	\subfigure[$\alpha$=0.01]{
    	\begin{minipage}[t]{0.32\linewidth}
    		\centering
    		\includegraphics[width=1.9in]{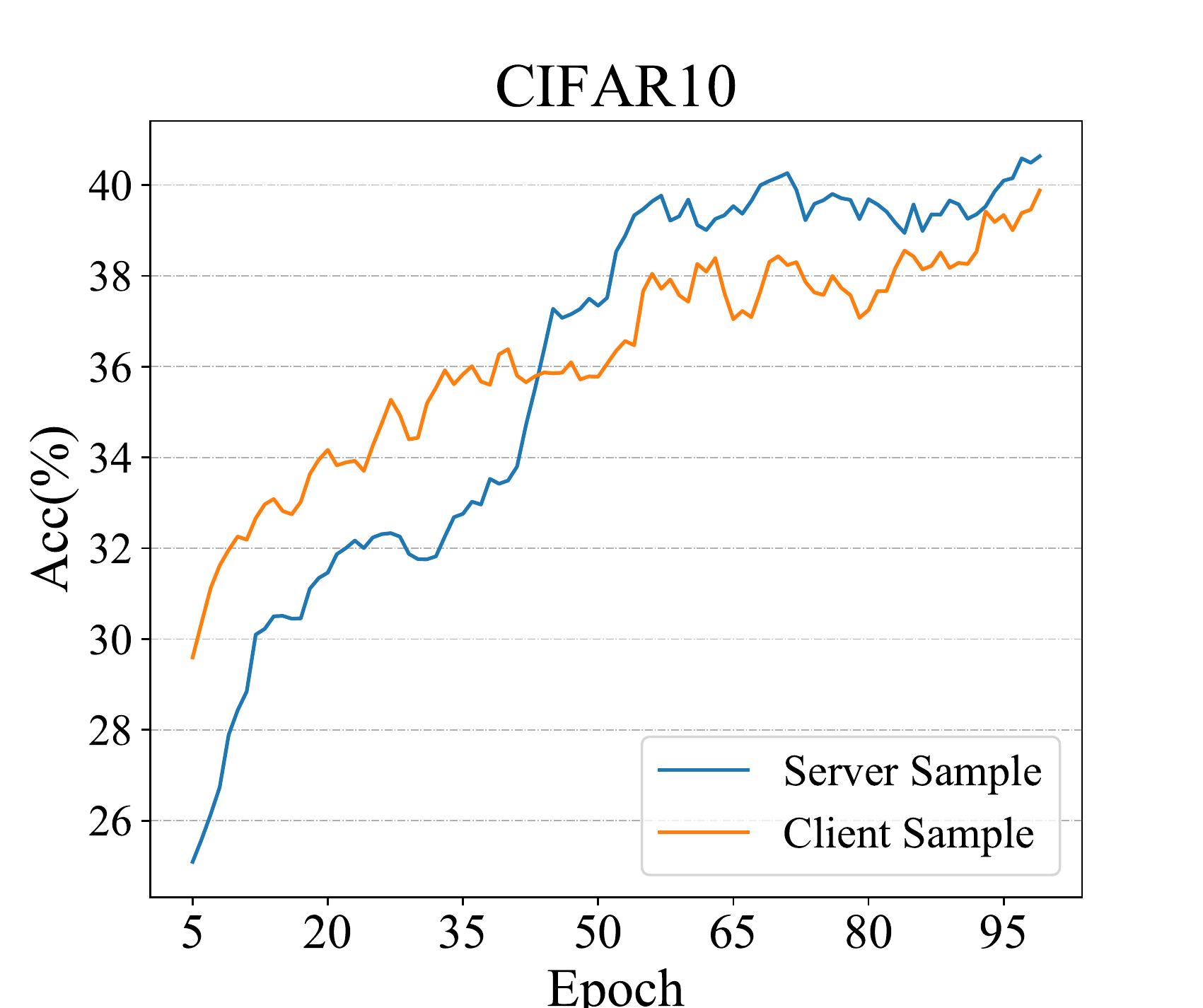}
    	\end{minipage}%
	}
	\subfigure[$\alpha$=0.05]{
    	\begin{minipage}[t]{0.32\linewidth}
    		\centering
    		\includegraphics[width=1.9in]{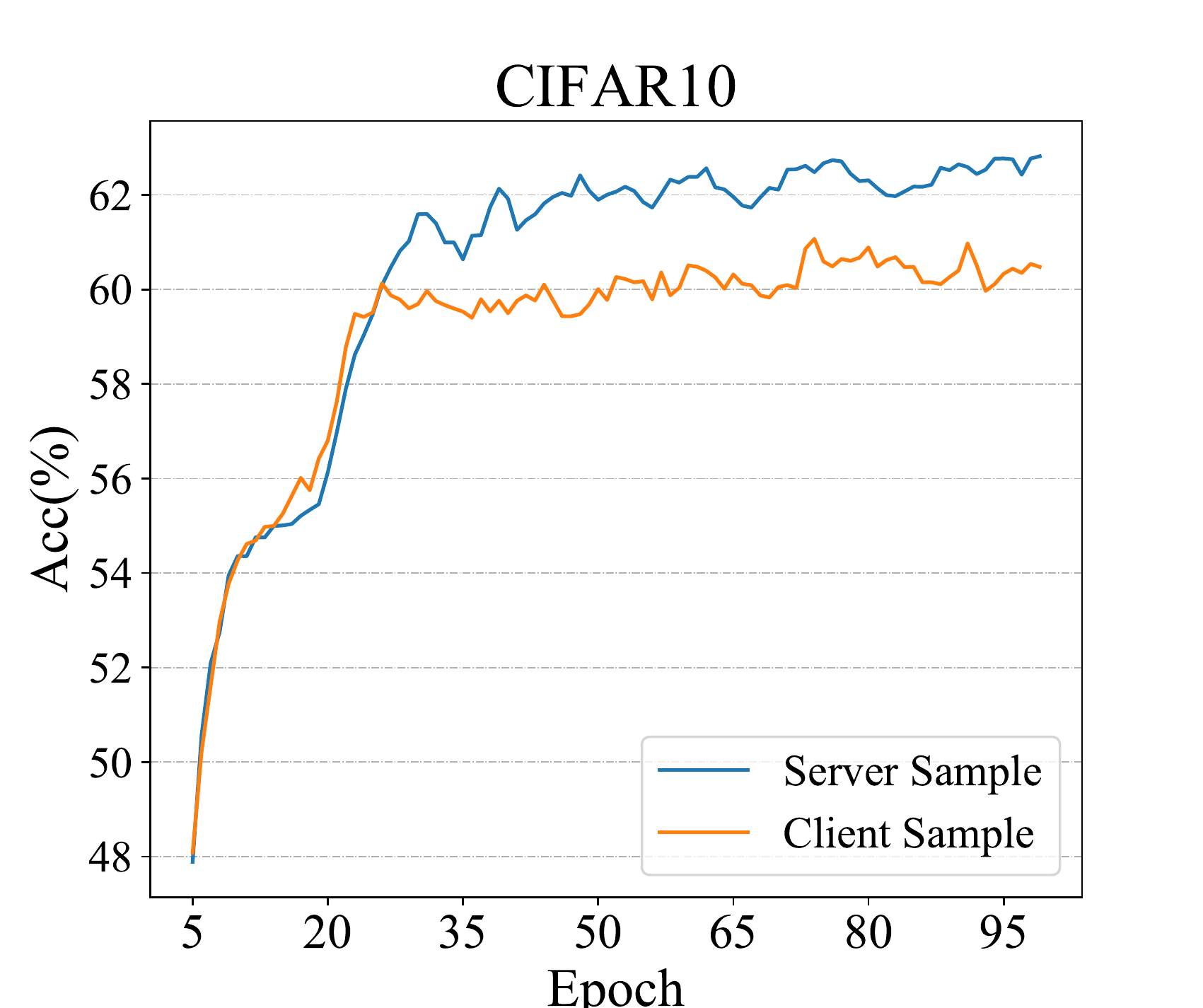}
    	\end{minipage}%
	}
	\subfigure[$\alpha$=0.1]{
    	\begin{minipage}[t]{0.32\linewidth}
    		\centering
    		\includegraphics[width=1.9in]{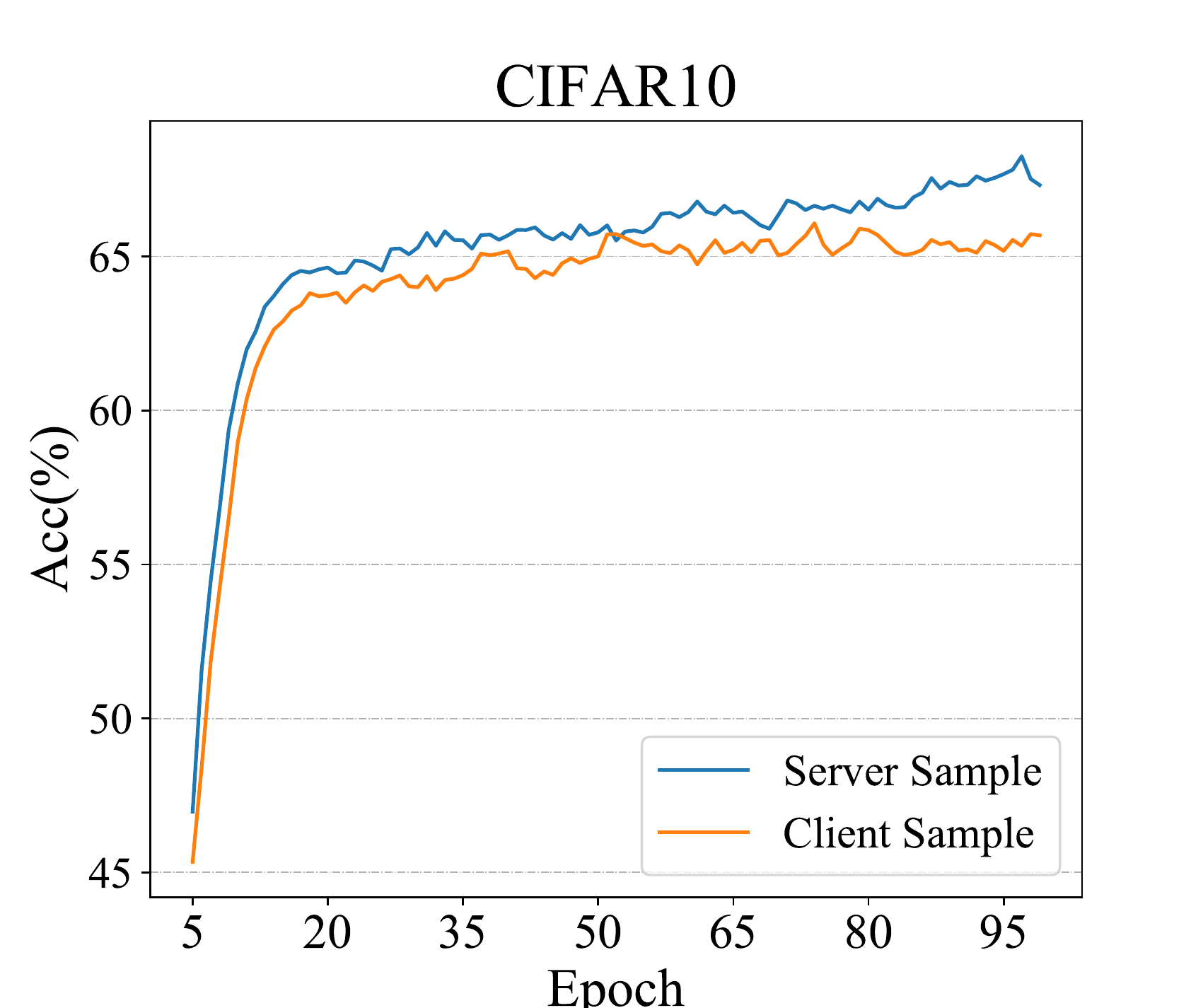}
    	\end{minipage}%
	}
	\caption{Visualization of algorithm performance under different generator sampling methods. The results are obtained when number of clients is 4.}
	\label{sample}
\end{figure}

In this experiment, the CIFAR10 data set is used, and the data is divided into 4 clients according to the Dirichlet distribution to ensure that the label distribution and the total amount of data of each client are different. The experimental results are shown in Table~\ref{tab5}. In different label distribution skew scenarios, the accuracy of the resulting model is higher when the generator preset label uses the server method to sample the result. This is because the use of uniform sampling in the generative adversarial stage in FedMGD can better learn the data information of each category, which is more conducive to refine the overall label distribution skew. We show the accuracy variation of the two methods in Figure~\ref{sample}. The accuracy of the model trained using server sampling in the figure is significantly higher than client sampling, which demonstrates the validity of our method.

\begin{table}[h]
    \footnotesize
	\begin{center}
	\small	
		\caption{\textbf{Compare the sampling method of FedMGD.} Compare the effect of the preset labels obtained by the generator using different sampling methods on the accuracy of the model.}
		% \resizebox{0.9\columnwidth}{!}{
		\renewcommand{\arraystretch}{1.2}
 		\small		
% 		\resizebox{0.85\columnwidth}{!}{
 		\setlength{\tabcolsep}{17.8mm}{

			\begin{tabular}{c|c|c}
                \hline
                \textbf{}              & \textbf{Client Sample} & \textbf{Server Sample} \\ \hline
                \textbf{$\alpha$=0.01} & 39.46±0.80             & \textbf{40.35±0.87} \textcolor{green!70!black}{($\uparrow$0.89)}    \\ \hline
                \textbf{$\alpha$=0.05} & 60.72±1.58             & \textbf{63.23±0.39} \textcolor{green!70!black}{($\uparrow$2.51)}   \\ \hline
                \textbf{$\alpha$=0.1}  & 65.63±0.47             & \textbf{66.91±0.87} \textcolor{green!70!black}{($\uparrow$1.28)}   \\ \hline
            \end{tabular}
		}
	% }
		\label{tab5}
	\end{center}
\end{table}

\section{Conclusion}

In this work, we mainly investigate how to mitigate the impact of label distribution skew on the performance of federated learning models.
We propose a method FedMGD to mitigate the degraded model performance when aggregating heterogeneous data distributions by modeling the global data distribution.
It improves the compatibility of the global model under heterogeneous data distribution by using global information to refine the aggregated model.
In our experiments, FedMGD has better performance in the scenarios of label distribution skew compared to baseline methods. 
In addition, we demonstrate through a series of ablation experiments that FedMGD can indeed better model the global data distribution in a label distribution skew scenario and it provides a novel solution for data source of downstream tasks.
In future research, we will explore a method for global modeling that is based on any federated learning approach to reduce the performance degradation caused by label distribution skew.

\section*{Acknowledgments}
This work was supported in part by the National Key Research and Development Program of China (No.2021YFF12\\01200). This work was carried out in part using computing resources at the High Performance Computing Center of Central South University. 

\bibliographystyle{model1-num-names}
\bibliography{reference}

\begin{thebibliography}{43}
\expandafter\ifx\csname natexlab\endcsname\relax\def\natexlab#1{#1}\fi
\providecommand{\url}[1]{\texttt{#1}}
\providecommand{\href}[2]{#2}
\providecommand{\path}[1]{#1}
\providecommand{\DOIprefix}{doi:}
\providecommand{\ArXivprefix}{arXiv:}
\providecommand{\URLprefix}{URL: }
\providecommand{\Pubmedprefix}{pmid:}
\providecommand{\doi}[1]{\href{http://dx.doi.org/#1}{\path{#1}}}
\providecommand{\Pubmed}[1]{\href{pmid:#1}{\path{#1}}}
\providecommand{\bibinfo}[2]{#2}
\ifx\xfnm\relax \def\xfnm[#1]{\unskip,\space#1}\fi
%Type = Inproceedings
\bibitem[{McMahan et~al.(2017)McMahan, Moore, Ramage, Hampson, and y~Arcas}]{2}
\bibinfo{author}{H.~B. McMahan}, \bibinfo{author}{E.~Moore},
  \bibinfo{author}{D.~Ramage}, \bibinfo{author}{S.~Hampson},
  \bibinfo{author}{B.~A. y~Arcas},
\newblock \bibinfo{title}{Communication-efficient learning of deep networks
  from decentralized data},
\newblock in: \bibinfo{booktitle}{International Conference on Artificial
  Intelligence and Statistics (AISTATS)}, \bibinfo{organization}{PMLR},
  \bibinfo{year}{2017}, pp. \bibinfo{pages}{1273--1282}.
%Type = Article
\bibitem[{Peter et~al.(2021)Peter, H.~Brendan, Brendan, Aurélien, Mehdi, Arjun
  et~al.}]{3}
\bibinfo{author}{K.~Peter}, \bibinfo{author}{M.~H.~Brendan},
  \bibinfo{author}{A.~Brendan}, \bibinfo{author}{B.~Aurélien},
  \bibinfo{author}{B.~Mehdi}, \bibinfo{author}{B.~Arjun, Nitin}, et~al.,
\newblock \bibinfo{title}{Advances and open problems in federated learning},
\newblock \bibinfo{journal}{Foundations and Trends in Machine Learning}
  \bibinfo{volume}{14} (\bibinfo{year}{2021}) \bibinfo{pages}{1--210}.
%Type = Inproceedings
\bibitem[{Li et~al.(2020)Li, Sahu, Zaheer, Sanjabi, Talwalkar, and Smith}]{4}
\bibinfo{author}{T.~Li}, \bibinfo{author}{A.~K. Sahu},
  \bibinfo{author}{M.~Zaheer}, \bibinfo{author}{M.~Sanjabi},
  \bibinfo{author}{A.~Talwalkar}, \bibinfo{author}{V.~Smith},
\newblock \bibinfo{title}{Federated optimization in heterogeneous networks},
\newblock in: \bibinfo{booktitle}{Proceedings of Machine Learning and Systems
  (MLSys)}, \bibinfo{year}{2020}.
%Type = Inproceedings
\bibitem[{Karimireddy et~al.(2020)Karimireddy, Kale, Mohri, Reddi, Stich, and
  Suresh}]{13}
\bibinfo{author}{S.~P. Karimireddy}, \bibinfo{author}{S.~Kale},
  \bibinfo{author}{M.~Mohri}, \bibinfo{author}{S.~Reddi},
  \bibinfo{author}{S.~Stich}, \bibinfo{author}{A.~T. Suresh},
\newblock \bibinfo{title}{Scaffold: Stochastic controlled averaging for
  federated learning},
\newblock in: \bibinfo{booktitle}{International Conference on Machine Learning
  (ICML)}, \bibinfo{organization}{PMLR}, \bibinfo{year}{2020}, pp.
  \bibinfo{pages}{5132--5143}.
%Type = Inproceedings
\bibitem[{Li et~al.(2020)Li, Sanjabi, Beirami, and Smith}]{21}
\bibinfo{author}{T.~Li}, \bibinfo{author}{M.~Sanjabi},
  \bibinfo{author}{A.~Beirami}, \bibinfo{author}{V.~Smith},
\newblock \bibinfo{title}{Fair resource allocation in federated learning},
\newblock in: \bibinfo{booktitle}{International Conference on Learning
  Representations (ICLR)}, \bibinfo{year}{2020}.
%Type = Inproceedings
\bibitem[{Mohri et~al.(2019)Mohri, Sivek, and Suresh}]{22}
\bibinfo{author}{M.~Mohri}, \bibinfo{author}{G.~Sivek}, \bibinfo{author}{A.~T.
  Suresh},
\newblock \bibinfo{title}{Agnostic federated learning},
\newblock in: \bibinfo{booktitle}{International Conference on Machine Learning
  (ICML)}, \bibinfo{organization}{PMLR}, \bibinfo{year}{2019}, pp.
  \bibinfo{pages}{4615--4625}.
%Type = Article
\bibitem[{Zhao et~al.(2018)Zhao, Li, Lai, Suda, Civin, and Chandra}]{5}
\bibinfo{author}{Y.~Zhao}, \bibinfo{author}{M.~Li}, \bibinfo{author}{L.~Lai},
  \bibinfo{author}{N.~Suda}, \bibinfo{author}{D.~Civin},
  \bibinfo{author}{V.~Chandra},
\newblock \bibinfo{title}{Federated learning with non-iid data},
\newblock \bibinfo{journal}{arXiv preprint arXiv:1806.00582}
  (\bibinfo{year}{2018}).
%Type = Inproceedings
\bibitem[{Yoshida et~al.(2020)Yoshida, Nishio, Morikura, Yamamoto, and
  Yonetani}]{23}
\bibinfo{author}{N.~Yoshida}, \bibinfo{author}{T.~Nishio},
  \bibinfo{author}{M.~Morikura}, \bibinfo{author}{K.~Yamamoto},
  \bibinfo{author}{R.~Yonetani},
\newblock \bibinfo{title}{Hybrid-fl for wireless networks: Cooperative learning
  mechanism using non-iid data},
\newblock in: \bibinfo{booktitle}{International Conference on Communications
  (ICC)}, \bibinfo{year}{2020}, pp. \bibinfo{pages}{1--7}.
%Type = Inproceedings
\bibitem[{Lin et~al.(2020)Lin, Kong, Stich, and Jaggi}]{6}
\bibinfo{author}{T.~Lin}, \bibinfo{author}{L.~Kong}, \bibinfo{author}{S.~U.
  Stich}, \bibinfo{author}{M.~Jaggi},
\newblock \bibinfo{title}{Ensemble distillation for robust model fusion in
  federated learning},
\newblock in: \bibinfo{booktitle}{International Conference on Neural
  Information Processing Systems (NeurIPS)}, volume~\bibinfo{volume}{33},
  \bibinfo{year}{2020}, pp. \bibinfo{pages}{2351--2363}.
%Type = Article
\bibitem[{Li and Wang(2019)}]{7}
\bibinfo{author}{D.~Li}, \bibinfo{author}{J.~Wang},
\newblock \bibinfo{title}{Fedmd: Heterogenous federated learning via model
  distillation},
\newblock \bibinfo{journal}{arXiv preprint arXiv:1910.03581}
  (\bibinfo{year}{2019}).
%Type = Article
\bibitem[{Jeong et~al.(2018)Jeong, Oh, Kim, Park, Bennis, and Kim}]{8}
\bibinfo{author}{E.~Jeong}, \bibinfo{author}{S.~Oh}, \bibinfo{author}{H.~Kim},
  \bibinfo{author}{J.~Park}, \bibinfo{author}{M.~Bennis},
  \bibinfo{author}{S.-L. Kim},
\newblock \bibinfo{title}{Communication-efficient on-device machine learning:
  Federated distillation and augmentation under non-iid private data},
\newblock \bibinfo{journal}{arXiv preprint arXiv:1811.11479}
  (\bibinfo{year}{2018}).
%Type = Article
\bibitem[{Yang et~al.(2019)Yang, Liu, Chen, and Tong}]{12}
\bibinfo{author}{Q.~Yang}, \bibinfo{author}{Y.~Liu}, \bibinfo{author}{T.~Chen},
  \bibinfo{author}{Y.~Tong},
\newblock \bibinfo{title}{Federated machine learning: Concept and
  applications},
\newblock \bibinfo{journal}{ACM Transactions on Intelligent Systems and
  Technology (TIST)} \bibinfo{volume}{10} (\bibinfo{year}{2019})
  \bibinfo{pages}{1--19}.
%Type = Incollection
\bibitem[{Long et~al.(2020)Long, Tan, Jiang, and Zhang}]{33}
\bibinfo{author}{G.~Long}, \bibinfo{author}{Y.~Tan},
  \bibinfo{author}{J.~Jiang}, \bibinfo{author}{C.~Zhang},
\newblock \bibinfo{title}{Federated learning for open banking},
\newblock in: \bibinfo{booktitle}{Federated learning},
  \bibinfo{publisher}{Springer}, \bibinfo{year}{2020}, pp.
  \bibinfo{pages}{240--254}.
%Type = Inproceedings
\bibitem[{Liu et~al.(2020)Liu, Huang, Luo, Huang, Liu, Chen, Feng, Chen, Yu,
  and Yang}]{34}
\bibinfo{author}{Y.~Liu}, \bibinfo{author}{A.~Huang}, \bibinfo{author}{Y.~Luo},
  \bibinfo{author}{H.~Huang}, \bibinfo{author}{Y.~Liu},
  \bibinfo{author}{Y.~Chen}, \bibinfo{author}{L.~Feng},
  \bibinfo{author}{T.~Chen}, \bibinfo{author}{H.~Yu},
  \bibinfo{author}{Q.~Yang},
\newblock \bibinfo{title}{Fedvision: An online visual object detection platform
  powered by federated learning},
\newblock in: \bibinfo{booktitle}{AAAI Conference on Artificial Intelligence
  (AAAI)}, volume~\bibinfo{volume}{34}, \bibinfo{year}{2020}, pp.
  \bibinfo{pages}{13172--13179}.
%Type = Inproceedings
\bibitem[{Li et~al.(2019)Li, Milletar{\`\i}, Xu, Rieke, Hancox, Zhu, Baust,
  Cheng, Ourselin, Cardoso et~al.}]{35}
\bibinfo{author}{W.~Li}, \bibinfo{author}{F.~Milletar{\`\i}},
  \bibinfo{author}{D.~Xu}, \bibinfo{author}{N.~Rieke},
  \bibinfo{author}{J.~Hancox}, \bibinfo{author}{W.~Zhu},
  \bibinfo{author}{M.~Baust}, \bibinfo{author}{Y.~Cheng},
  \bibinfo{author}{S.~Ourselin}, \bibinfo{author}{M.~J. Cardoso}, et~al.,
\newblock \bibinfo{title}{Privacy-preserving federated brain tumour
  segmentation},
\newblock in: \bibinfo{booktitle}{International workshop on machine learning in
  medical imaging (MLMI)}, \bibinfo{organization}{Springer},
  \bibinfo{year}{2019}, pp. \bibinfo{pages}{133--141}.
%Type = Article
\bibitem[{Xu et~al.(2021)Xu, Glicksberg, Su, Walker, Bian, and Wang}]{36}
\bibinfo{author}{J.~Xu}, \bibinfo{author}{B.~S. Glicksberg},
  \bibinfo{author}{C.~Su}, \bibinfo{author}{P.~Walker},
  \bibinfo{author}{J.~Bian}, \bibinfo{author}{F.~Wang},
\newblock \bibinfo{title}{Federated learning for healthcare informatics},
\newblock \bibinfo{journal}{Journal of Healthcare Informatics Research}
  \bibinfo{volume}{5} (\bibinfo{year}{2021}) \bibinfo{pages}{1--19}.
%Type = Article
\bibitem[{Brisimi et~al.(2018)Brisimi, Chen, Mela, Olshevsky, Paschalidis, and
  Shi}]{37}
\bibinfo{author}{T.~S. Brisimi}, \bibinfo{author}{R.~Chen},
  \bibinfo{author}{T.~Mela}, \bibinfo{author}{A.~Olshevsky},
  \bibinfo{author}{I.~C. Paschalidis}, \bibinfo{author}{W.~Shi},
\newblock \bibinfo{title}{Federated learning of predictive models from
  federated electronic health records},
\newblock \bibinfo{journal}{International journal of medical informatics}
  \bibinfo{volume}{112} (\bibinfo{year}{2018}) \bibinfo{pages}{59--67}.
%Type = Article
\bibitem[{Rieke et~al.(2020)Rieke, Hancox, Li, Milletari, Roth, Albarqouni,
  Bakas, Galtier, Landman, Maier-Hein et~al.}]{38}
\bibinfo{author}{N.~Rieke}, \bibinfo{author}{J.~Hancox},
  \bibinfo{author}{W.~Li}, \bibinfo{author}{F.~Milletari},
  \bibinfo{author}{H.~R. Roth}, \bibinfo{author}{S.~Albarqouni},
  \bibinfo{author}{S.~Bakas}, \bibinfo{author}{M.~N. Galtier},
  \bibinfo{author}{B.~A. Landman}, \bibinfo{author}{K.~Maier-Hein}, et~al.,
\newblock \bibinfo{title}{The future of digital health with federated
  learning},
\newblock \bibinfo{journal}{NPJ digital medicine} \bibinfo{volume}{3}
  (\bibinfo{year}{2020}) \bibinfo{pages}{1--7}.
%Type = Inproceedings
\bibitem[{Tan et~al.(2020)Tan, Liu, Zheng, and Yang}]{39}
\bibinfo{author}{B.~Tan}, \bibinfo{author}{B.~Liu}, \bibinfo{author}{V.~Zheng},
  \bibinfo{author}{Q.~Yang},
\newblock \bibinfo{title}{A federated recommender system for online services},
\newblock in: \bibinfo{booktitle}{ACM Conference on Recommender Systems
  (RecSys)}, \bibinfo{year}{2020}, pp. \bibinfo{pages}{579--581}.
%Type = Inproceedings
\bibitem[{Minto et~al.(2021)Minto, Haller, Livshits, and Haddadi}]{40}
\bibinfo{author}{L.~Minto}, \bibinfo{author}{M.~Haller},
  \bibinfo{author}{B.~Livshits}, \bibinfo{author}{H.~Haddadi},
\newblock \bibinfo{title}{Stronger privacy for federated collaborative
  filtering with implicit feedback},
\newblock in: \bibinfo{booktitle}{ACM Conference on Recommender Systems
  (RecSys)}, \bibinfo{year}{2021}, pp. \bibinfo{pages}{342--350}.
%Type = Inproceedings
\bibitem[{Wu et~al.(2021)Wu, Liu, Huang, Ning, Wang, Chen, Yi, and Zhou}]{41}
\bibinfo{author}{J.~Wu}, \bibinfo{author}{Q.~Liu}, \bibinfo{author}{Z.~Huang},
  \bibinfo{author}{Y.~Ning}, \bibinfo{author}{H.~Wang},
  \bibinfo{author}{E.~Chen}, \bibinfo{author}{J.~Yi},
  \bibinfo{author}{B.~Zhou},
\newblock \bibinfo{title}{Hierarchical personalized federated learning for user
  modeling},
\newblock in: \bibinfo{booktitle}{The Web Conference (WWW)},
  \bibinfo{year}{2021}, pp. \bibinfo{pages}{957--968}.
%Type = Inproceedings
\bibitem[{Hitaj et~al.(2017)Hitaj, Ateniese, and Perez-Cruz}]{25}
\bibinfo{author}{B.~Hitaj}, \bibinfo{author}{G.~Ateniese},
  \bibinfo{author}{F.~Perez-Cruz},
\newblock \bibinfo{title}{Deep models under the gan: Information leakage from
  collaborative deep learning},
\newblock in: \bibinfo{booktitle}{ACM SIGSAC Conference on Computer and
  Communications Security (CCS)}, \bibinfo{publisher}{ACM},
  \bibinfo{year}{2017}, p. \bibinfo{pages}{603–618}.
%Type = Inproceedings
\bibitem[{Truex et~al.(2019)Truex, Baracaldo, Anwar, Steinke, Ludwig, Zhang,
  and Zhou}]{42}
\bibinfo{author}{S.~Truex}, \bibinfo{author}{N.~Baracaldo},
  \bibinfo{author}{A.~Anwar}, \bibinfo{author}{T.~Steinke},
  \bibinfo{author}{H.~Ludwig}, \bibinfo{author}{R.~Zhang},
  \bibinfo{author}{Y.~Zhou},
\newblock \bibinfo{title}{A hybrid approach to privacy-preserving federated
  learning},
\newblock in: \bibinfo{booktitle}{ACM workshop on artificial intelligence and
  security (AISec)}, \bibinfo{year}{2019}, pp. \bibinfo{pages}{1--11}.
%Type = Article
\bibitem[{Mothukuri et~al.(2021)Mothukuri, Parizi, Pouriyeh, Huang,
  Dehghantanha, and Srivastava}]{43}
\bibinfo{author}{V.~Mothukuri}, \bibinfo{author}{R.~M. Parizi},
  \bibinfo{author}{S.~Pouriyeh}, \bibinfo{author}{Y.~Huang},
  \bibinfo{author}{A.~Dehghantanha}, \bibinfo{author}{G.~Srivastava},
\newblock \bibinfo{title}{A survey on security and privacy of federated
  learning},
\newblock \bibinfo{journal}{Future Generation Computer Systems}
  \bibinfo{volume}{115} (\bibinfo{year}{2021}) \bibinfo{pages}{619--640}.
%Type = Article
\bibitem[{Ryffel et~al.(2018)Ryffel, Trask, Dahl, Wagner, Mancuso, Rueckert,
  and Passerat-Palmbach}]{44}
\bibinfo{author}{T.~Ryffel}, \bibinfo{author}{A.~Trask},
  \bibinfo{author}{M.~Dahl}, \bibinfo{author}{B.~Wagner},
  \bibinfo{author}{J.~Mancuso}, \bibinfo{author}{D.~Rueckert},
  \bibinfo{author}{J.~Passerat-Palmbach},
\newblock \bibinfo{title}{A generic framework for privacy preserving deep
  learning},
\newblock \bibinfo{journal}{arXiv preprint arXiv:1811.04017}
  (\bibinfo{year}{2018}).
%Type = Inproceedings
\bibitem[{Li et~al.(2021)Li, Jiang, Zhang, Kamp, and Dou}]{45}
\bibinfo{author}{X.~Li}, \bibinfo{author}{M.~Jiang},
  \bibinfo{author}{X.~Zhang}, \bibinfo{author}{M.~Kamp},
  \bibinfo{author}{Q.~Dou},
\newblock \bibinfo{title}{Fed{\{}bn{\}}: Federated learning on non-{\{}iid{\}}
  features via local batch normalization},
\newblock in: \bibinfo{booktitle}{International Conference on Learning
  Representations (ICLR)}, \bibinfo{year}{2021}.
%Type = Inproceedings
\bibitem[{Duan et~al.(2021)Duan, Li, and Lu}]{46}
\bibinfo{author}{J.-H. Duan}, \bibinfo{author}{W.~Li}, \bibinfo{author}{S.~Lu},
\newblock \bibinfo{title}{Feddna: Federated learning with decoupled
  normalization-layer aggregation for non-iid data},
\newblock in: \bibinfo{booktitle}{Joint European Conference on Machine Learning
  and Knowledge Discovery in Databases}, \bibinfo{organization}{Springer},
  \bibinfo{year}{2021}, pp. \bibinfo{pages}{722--737}.
%Type = Inproceedings
\bibitem[{Han and Zhang(2020)}]{24}
\bibinfo{author}{Y.~Han}, \bibinfo{author}{X.~Zhang},
\newblock \bibinfo{title}{Robust federated learning via collaborative machine
  teaching},
\newblock in: \bibinfo{booktitle}{AAAI Conference on Artificial Intelligence
  (AAAI)}, volume~\bibinfo{volume}{34}, \bibinfo{year}{2020}, pp.
  \bibinfo{pages}{4075--4082}.
%Type = Inproceedings
\bibitem[{Zhu et~al.(2021)Zhu, Hong, and Zhou}]{27}
\bibinfo{author}{Z.~Zhu}, \bibinfo{author}{J.~Hong}, \bibinfo{author}{J.~Zhou},
\newblock \bibinfo{title}{Data-free knowledge distillation for heterogeneous
  federated learning},
\newblock in: \bibinfo{booktitle}{International Conference on Machine Learning
  (ICML)}, \bibinfo{organization}{PMLR}, \bibinfo{year}{2021}, pp.
  \bibinfo{pages}{12878--12889}.
%Type = Inproceedings
\bibitem[{Goodfellow et~al.(2014)Goodfellow, Pouget-Abadie, Mirza, Xu,
  Warde-Farley, Ozair, Courville et~al.}]{14}
\bibinfo{author}{I.~J. Goodfellow}, \bibinfo{author}{J.~Pouget-Abadie},
  \bibinfo{author}{M.~Mirza}, \bibinfo{author}{B.~Xu},
  \bibinfo{author}{D.~Warde-Farley}, \bibinfo{author}{S.~Ozair},
  \bibinfo{author}{B.~Courville}, et~al.,
\newblock \bibinfo{title}{Generative adversarial nets},
\newblock in: \bibinfo{booktitle}{International Conference on Neural
  Information Processing Systems (NeurIPS)}, volume~\bibinfo{volume}{27},
  \bibinfo{publisher}{MIT Press}, \bibinfo{year}{2014}, pp.
  \bibinfo{pages}{2672--2680}.
%Type = Inproceedings
\bibitem[{Wang et~al.(2019)Wang, Song, Zhang, Song, Wang, and Qi}]{26}
\bibinfo{author}{Z.~Wang}, \bibinfo{author}{M.~Song},
  \bibinfo{author}{Z.~Zhang}, \bibinfo{author}{Y.~Song},
  \bibinfo{author}{Q.~Wang}, \bibinfo{author}{H.~Qi},
\newblock \bibinfo{title}{Beyond inferring class representatives: User-level
  privacy leakage from federated learning},
\newblock in: \bibinfo{booktitle}{International Conference on Computer
  Communications (INFOCOM)}, \bibinfo{organization}{IEEE},
  \bibinfo{year}{2019}, pp. \bibinfo{pages}{2512--2520}.
%Type = Inproceedings
\bibitem[{Augenstein et~al.(2020)Augenstein, McMahan, Ramage, Ramaswamy,
  Kairouz, Chen, Mathews et~al.}]{15}
\bibinfo{author}{S.~Augenstein}, \bibinfo{author}{H.~B. McMahan},
  \bibinfo{author}{D.~Ramage}, \bibinfo{author}{S.~Ramaswamy},
  \bibinfo{author}{P.~Kairouz}, \bibinfo{author}{M.~Chen},
  \bibinfo{author}{R.~Mathews}, et~al.,
\newblock \bibinfo{title}{Generative models for effective ml on private,
  decentralized datasets},
\newblock in: \bibinfo{booktitle}{International Conference on Learning
  Representations (ICLR)}, \bibinfo{year}{2020}.
%Type = Article
\bibitem[{Rasouli et~al.(2020)Rasouli, Sun, and Rajagopal}]{16}
\bibinfo{author}{M.~Rasouli}, \bibinfo{author}{T.~Sun},
  \bibinfo{author}{R.~Rajagopal},
\newblock \bibinfo{title}{Fedgan: Federated generative adversarial networks for
  distributed data},
\newblock \bibinfo{journal}{arXiv preprint arXiv:2006.07228}
  (\bibinfo{year}{2020}).
%Type = Inproceedings
\bibitem[{Hardy et~al.(2019)Hardy, Le~Merrer, and Sericola}]{9}
\bibinfo{author}{C.~Hardy}, \bibinfo{author}{E.~Le~Merrer},
  \bibinfo{author}{B.~Sericola},
\newblock \bibinfo{title}{Md-gan: Multi-discriminator generative adversarial
  networks for distributed datasets},
\newblock in: \bibinfo{booktitle}{International Parallel and Distributed
  Processing Symposium (IPDPS)}, \bibinfo{publisher}{IEEE},
  \bibinfo{year}{2019}, pp. \bibinfo{pages}{866--877}.
%Type = Inproceedings
\bibitem[{Chang et~al.(2020)Chang, Qu, Zhang, Sabuncu, Chen, Zhang, and
  Metaxas}]{11}
\bibinfo{author}{Q.~Chang}, \bibinfo{author}{H.~Qu},
  \bibinfo{author}{Y.~Zhang}, \bibinfo{author}{M.~Sabuncu},
  \bibinfo{author}{C.~Chen}, \bibinfo{author}{T.~Zhang}, \bibinfo{author}{D.~N.
  Metaxas},
\newblock \bibinfo{title}{Synthetic learning: Learn from distributed
  asynchronized discriminator gan without sharing medical image data},
\newblock in: \bibinfo{booktitle}{The IEEE Conference on Computer Vision and
  Pattern Recognition (CVPR)}, \bibinfo{publisher}{IEEE}, \bibinfo{year}{2020},
  pp. \bibinfo{pages}{13856--13866}.
%Type = Article
\bibitem[{Yonetani et~al.(2019)Yonetani, Takahashi, Hashimoto, and Ushiku}]{10}
\bibinfo{author}{R.~Yonetani}, \bibinfo{author}{T.~Takahashi},
  \bibinfo{author}{A.~Hashimoto}, \bibinfo{author}{Y.~Ushiku},
\newblock \bibinfo{title}{Decentralized learning of generative adversarial
  networks from non-iid data},
\newblock \bibinfo{journal}{arXiv preprint arXiv:1905.09684}
  (\bibinfo{year}{2019}).
%Type = Inproceedings
\bibitem[{Cohen et~al.(2017)Cohen, Afshar, Tapson, and van Schaik}]{18}
\bibinfo{author}{G.~Cohen}, \bibinfo{author}{S.~Afshar},
  \bibinfo{author}{J.~Tapson}, \bibinfo{author}{A.~van Schaik},
\newblock \bibinfo{title}{Emnist: Extending mnist to handwritten letters},
\newblock in: \bibinfo{booktitle}{International Joint Conference on Neural
  Networks (IJCNN)}, \bibinfo{publisher}{IEEE}, \bibinfo{year}{2017}, pp.
  \bibinfo{pages}{2921--2926}.
%Type = Article
\bibitem[{Xiao et~al.(2017)Xiao, Rasul, and Vollgraf}]{19}
\bibinfo{author}{H.~Xiao}, \bibinfo{author}{K.~Rasul},
  \bibinfo{author}{R.~Vollgraf},
\newblock \bibinfo{title}{Fashion-mnist: a novel image dataset for benchmarking
  machine learning algorithms},
\newblock \bibinfo{journal}{arXiv preprint arXiv:1708.07747}
  (\bibinfo{year}{2017}).
%Type = Inproceedings
\bibitem[{Netzer et~al.(2011)Netzer, Wang, Coates, Bissacco, Wu, and Ng}]{28}
\bibinfo{author}{Y.~Netzer}, \bibinfo{author}{T.~Wang},
  \bibinfo{author}{A.~Coates}, \bibinfo{author}{A.~Bissacco},
  \bibinfo{author}{B.~Wu}, \bibinfo{author}{A.~Y. Ng},
\newblock \bibinfo{title}{Reading digits in natural images with unsupervised
  feature learning},
\newblock in: \bibinfo{booktitle}{NeurIPS Workshop on Deep Learning and
  Unsupervised Feature Learning}, \bibinfo{year}{2011}.
%Type = Article
\bibitem[{Krizhevsky et~al.(2009)Krizhevsky, Hinton et~al.}]{32}
\bibinfo{author}{A.~Krizhevsky}, \bibinfo{author}{G.~Hinton}, et~al.,
\newblock \bibinfo{title}{Learning multiple layers of features from tiny
  images}  (\bibinfo{year}{2009}).
%Type = Article
\bibitem[{Hsu et~al.(2019)Hsu, Qi, and Brown}]{29}
\bibinfo{author}{T.-M.~H. Hsu}, \bibinfo{author}{H.~Qi},
  \bibinfo{author}{M.~Brown},
\newblock \bibinfo{title}{Measuring the effects of non-identical data
  distribution for federated visual classification},
\newblock \bibinfo{journal}{arXiv preprint arXiv:1909.06335}
  (\bibinfo{year}{2019}).
%Type = Inproceedings
\bibitem[{Isola et~al.(2017)Isola, Zhu, Zhou, and Efros}]{31}
\bibinfo{author}{P.~Isola}, \bibinfo{author}{J.-Y. Zhu},
  \bibinfo{author}{T.~Zhou}, \bibinfo{author}{A.~A. Efros},
\newblock \bibinfo{title}{Image-to-image translation with conditional
  adversarial networks},
\newblock in: \bibinfo{booktitle}{The IEEE Conference on Computer Vision and
  Pattern Recognition (CVPR)}, \bibinfo{year}{2017}, pp.
  \bibinfo{pages}{1125--1134}.
%Type = Inproceedings
\bibitem[{He et~al.(2016)He, Zhang, Ren, and Sun}]{20}
\bibinfo{author}{K.~He}, \bibinfo{author}{X.~Zhang}, \bibinfo{author}{S.~Ren},
  \bibinfo{author}{J.~Sun},
\newblock \bibinfo{title}{Deep residual learning for image recognition},
\newblock in: \bibinfo{booktitle}{The IEEE Conference on Computer Vision and
  Pattern Recognition (CVPR)}, \bibinfo{publisher}{IEEE}, \bibinfo{year}{2016},
  pp. \bibinfo{pages}{770--778}.

\end{thebibliography}

\hspace*{\fill} \\
\textbf{Tao Sheng} received the B.S. degree in Software Engineering from the Xinjiang University, Urumqi, Xinjiang, China, in 2020. He is currently pursuing the M.S. degree in Software Engineering with the Central South University, Changsha, China. His current research interests include federated learning and deep learning.

\hspace*{\fill} \\
\textbf{Chenchao Shen} received the PhD degree from Zhejiang University, China, in 2021. He is currently an associate professor at the School of Computer, Central South University, Changsha, Hunan, China. His research interests include computer vision, deep learning, and self-supervised learning.

\hspace*{\fill} \\
\textbf{Yuan Liu} obtained his Bachelor's degree in Software Engineering and is currently a Ph.D. student in the School of Computer Science and Engineering at Central South University where he works on federated learning for different distributions and generalization.

\hspace*{\fill} \\
\textbf{Yeyu Ou} received the B.S. degree in Engineering Mechanics from the Hehai University, Nanjing, China, in 2020. She is currently pursuing the M.S. degree in Computer Technology with the Central South University, Changsha, China. Her current research interests include graph neural networks and big data analysis.

\hspace*{\fill} \\
\textbf{Zhe Qu} received the M.S. degree from the University of Delaware, Newark, DE, USA, in 2017. He is currently a Ph.D. candidate in the University of South Florida, Tampa, FL, USA. His research interests include network and mobile system security, and machine learning for networks.

\hspace*{\fill} \\
\textbf{Jianxin Wang} (Senior Member, IEEE) received the Ph.D. degree from Central South University, Changsha, Hunan, China. He is currently the Chair and a Professor with the School of Computer Science and Engineering, Central South University. His research interests include algorithm analysis and optimization, parameterized algorithm, bioinformatics, and computer networks.

\end{document}

\shorttitle{Modeling Global Distribution for Federated Learning with Label Distribution Skew}
	\shortauthors{Tao Sheng et~al.}
	%	\clearpage
	\onecolumn 
	\section*{Appendix} \label{Appendix}
	
	\begin{figure}[h]
		\centering
		% \subfigure[EMNIST]{
		% 	\begin{minipage}[t]{0.34\linewidth}
		% 		\centering
		% 		\includegraphics[width=1.8in]{fig/data/c5/emnist/NonIID_class_dirichlet_0.01/emnist_5_0.01.pdf}
		% 		%\caption{fig1}
		% 	\end{minipage}%
		% 	\begin{minipage}[t]{0.34\linewidth}
		% 		\centering
		% 		\includegraphics[width=1.8in]{fig/data/c5/emnist/NonIID_class_dirichlet_0.05/emnist_5_0.05.pdf}
		% 		%\caption{fig1}
		% 	\end{minipage}%
		% 	\begin{minipage}[t]{0.34\linewidth}
		% 		\centering
		% 		\includegraphics[width=1.8in]{fig/data/c5/emnist/NonIID_class_dirichlet_0.1/emnist_5_0.1.pdf}
		% 		%\caption{fig1}
		% 	\end{minipage}%
		% }
		\subfigure[FashionMNIST]{
			\begin{minipage}[t]{0.34\linewidth}
				\centering
				\includegraphics[width=1.9in]{fig/data/c5/fashionmnist/NonIID_class_dirichlet_0.01/fashionmnist_5_0.01.pdf}
				%\caption{fig1}
			\end{minipage}%
			\begin{minipage}[t]{0.34\linewidth}
				\centering
				\includegraphics[width=1.9in]{fig/data/c5/fashionmnist/NonIID_class_dirichlet_0.05/fashionmnist_5_0.05.pdf}
				%\caption{fig1}
			\end{minipage}%
			\begin{minipage}[t]{0.34\linewidth}
				\centering
				\includegraphics[width=1.9in]{fig/data/c5/fashionmnist/NonIID_class_dirichlet_0.1/fashionmnist_5_0.1.pdf}
				%\caption{fig1}
			\end{minipage}%
		}
		\subfigure[SVHN]{
			\begin{minipage}[t]{0.34\linewidth}
				\centering
				\includegraphics[width=1.9in]{fig/data/c5/svhn/NonIID_class_dirichlet_0.01/svhn_5_0.01.pdf}
				%\caption{fig1}
			\end{minipage}%
			\begin{minipage}[t]{0.34\linewidth}
				\centering
				\includegraphics[width=1.9in]{fig/data/c5/svhn/NonIID_class_dirichlet_0.05/svhn_5_0.05.pdf}
				%\caption{fig1}
			\end{minipage}%
			\begin{minipage}[t]{0.34\linewidth}
				\centering
				\includegraphics[width=1.9in]{fig/data/c5/svhn/NonIID_class_dirichlet_0.1/svhn_5_0.1.pdf}
				%\caption{fig1}
			\end{minipage}%
		}
		\subfigure[CIFAR10]{
			\begin{minipage}[t]{0.34\linewidth}
				\centering
				\includegraphics[width=1.9in]{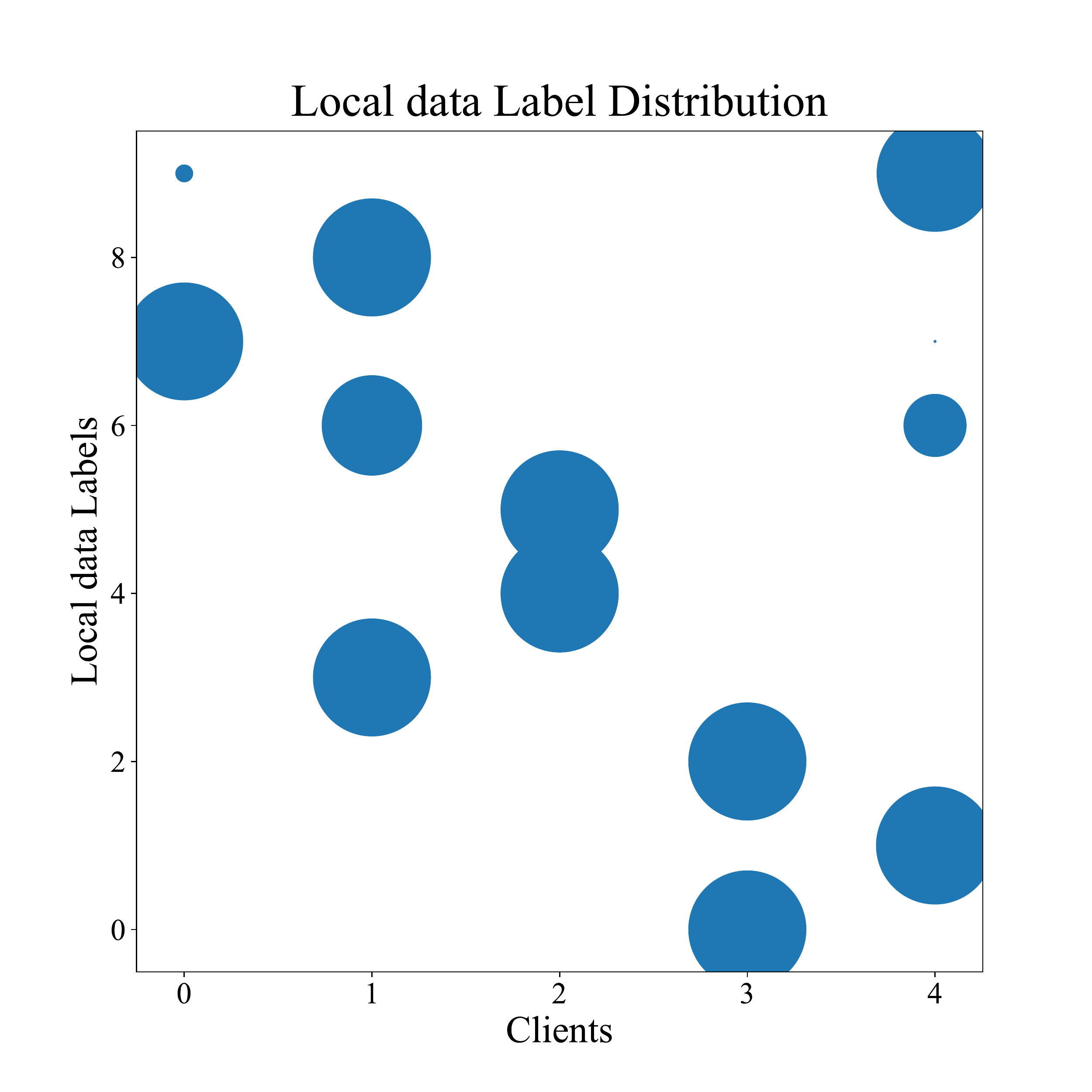}
				%\caption{fig1}
			\end{minipage}%
			\begin{minipage}[t]{0.34\linewidth}
				\centering
				\includegraphics[width=1.9in]{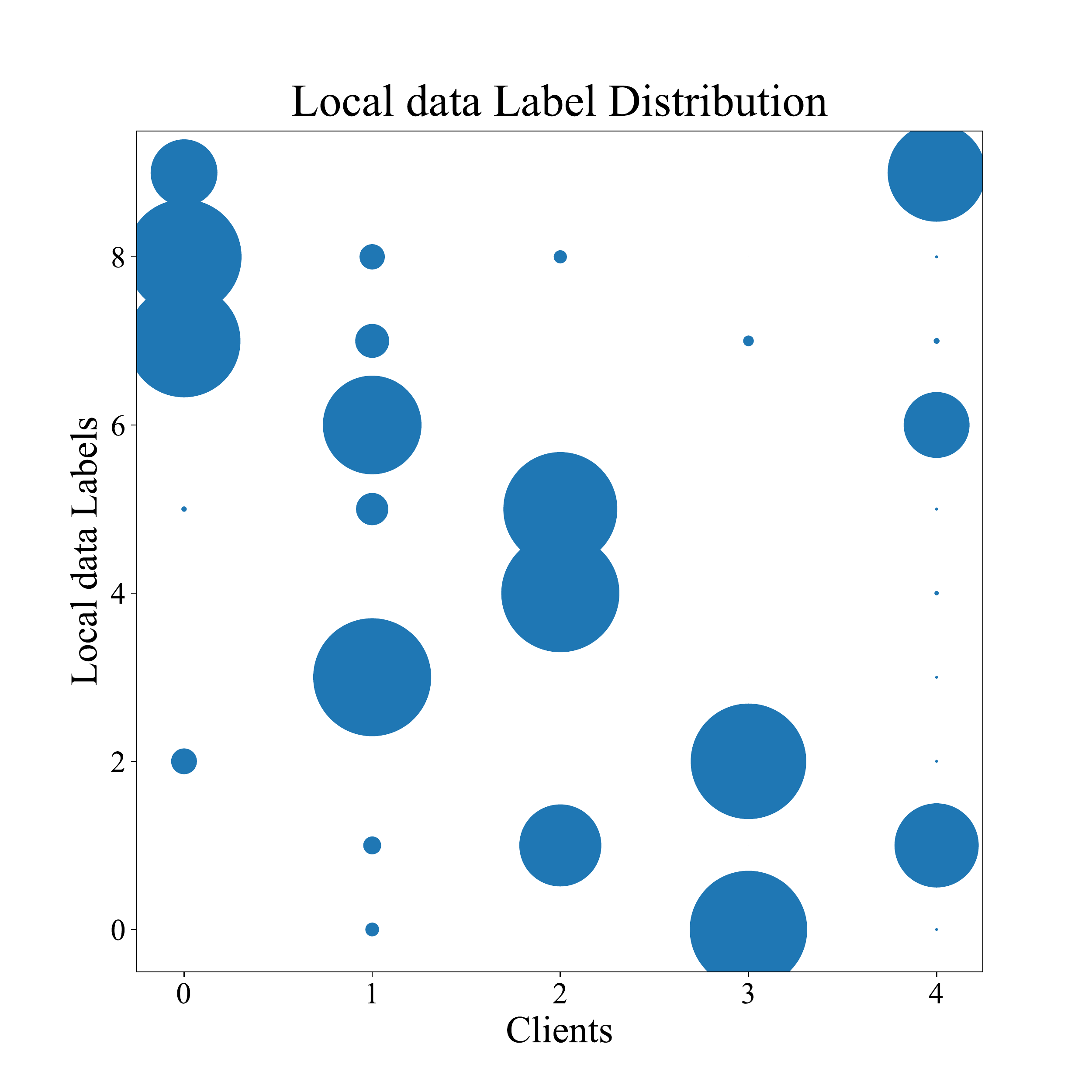}
				%\caption{fig1}
			\end{minipage}%
			\begin{minipage}[t]{0.34\linewidth}
				\centering
				\includegraphics[width=1.9in]{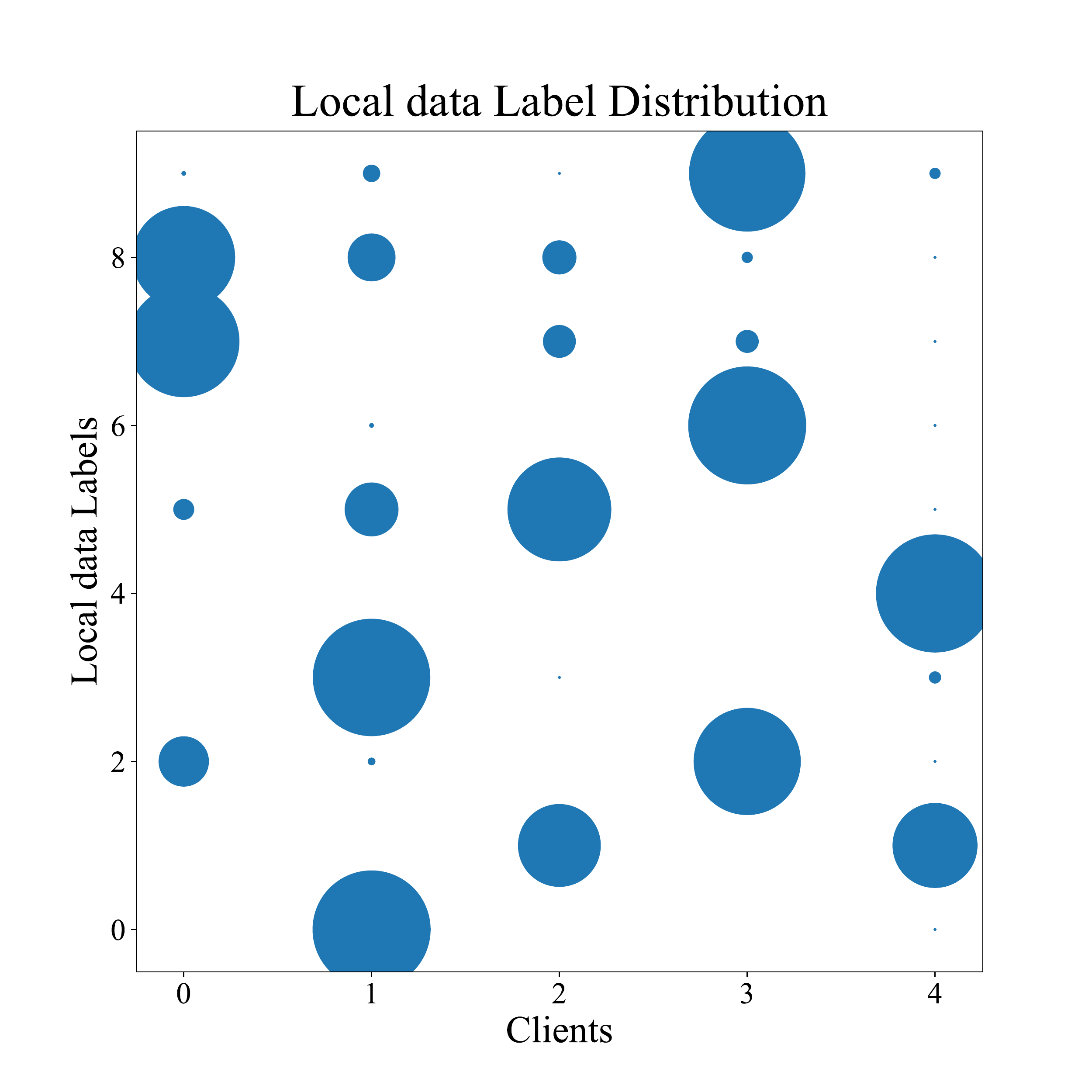}
				%\caption{fig1}
			\end{minipage}%
		}
		\centering
		\caption{Visualization of statistical heterogeneity between users on different datasets when the number of clients is 5. Where the x-axis represents the different clients, the y-axis indicates the class labels, and the size of the scattered points indicates the number of training samples with available labels for that user. From left to right $\alpha$ is 0.01, 0.05, and 0.1.}
		\label{fig4}
	\end{figure}
	
	\begin{figure}[h]
		\centering
		% \subfigure[EMNIST]{
		% 	\begin{minipage}[t]{0.34\linewidth}
		% 		\centering
		% 		\includegraphics[width=1.8in]{fig/data/c10/emnist/NonIID_class_dirichlet_0.01/emnist_10_0.01.pdf}
		% 		%\caption{fig1}
		% 	\end{minipage}%
		% 	\begin{minipage}[t]{0.34\linewidth}
		% 		\centering
		% 		\includegraphics[width=1.8in]{fig/data/c10/emnist/NonIID_class_dirichlet_0.05/emnist_10_0.05.pdf}
		% 		%\caption{fig1}
		% 	\end{minipage}%
		% 	\begin{minipage}[t]{0.34\linewidth}
		% 		\centering
		% 		\includegraphics[width=1.8in]{fig/data/c10/emnist/NonIID_class_dirichlet_0.1/emnist_10_0.1.pdf}
		% 		%\caption{fig1}
		% 	\end{minipage}%
		% }
		\subfigure[FashionMNIST]{
			\begin{minipage}[t]{0.34\linewidth}
				\centering
				\includegraphics[width=1.9in]{fig/data/c10/fashionmnist/NonIID_class_dirichlet_0.01/fashionmnist_10_0.01.pdf}
				%\caption{fig1}
			\end{minipage}%
			\begin{minipage}[t]{0.34\linewidth}
				\centering
				\includegraphics[width=1.9in]{fig/data/c10/fashionmnist/NonIID_class_dirichlet_0.05/fashionmnist_10_0.05.pdf}
				%\caption{fig1}
			\end{minipage}%
			\begin{minipage}[t]{0.34\linewidth}
				\centering
				\includegraphics[width=1.9in]{fig/data/c10/fashionmnist/NonIID_class_dirichlet_0.1/fashionmnist_10_0.1.pdf}
				%\caption{fig1}
			\end{minipage}%
		}
		\subfigure[SVHN]{
			\begin{minipage}[t]{0.34\linewidth}
				\centering
				\includegraphics[width=1.9in]{fig/data/c10/svhn/NonIID_class_dirichlet_0.01/svhn_10_0.01.pdf}
				%\caption{fig1}
			\end{minipage}%
			\begin{minipage}[t]{0.34\linewidth}
				\centering
				\includegraphics[width=1.8in]{fig/data/c10/svhn/NonIID_class_dirichlet_0.05/svhn_10_0.05.pdf}
				%\caption{fig1}
			\end{minipage}%
			\begin{minipage}[t]{0.34\linewidth}
				\centering
				\includegraphics[width=1.9in]{fig/data/c10/svhn/NonIID_class_dirichlet_0.1/svhn_10_0.1.pdf}
				%\caption{fig1}
			\end{minipage}%
		}
		\subfigure[CIFAR10]{
			\begin{minipage}[t]{0.34\linewidth}
				\centering
				\includegraphics[width=1.9in]{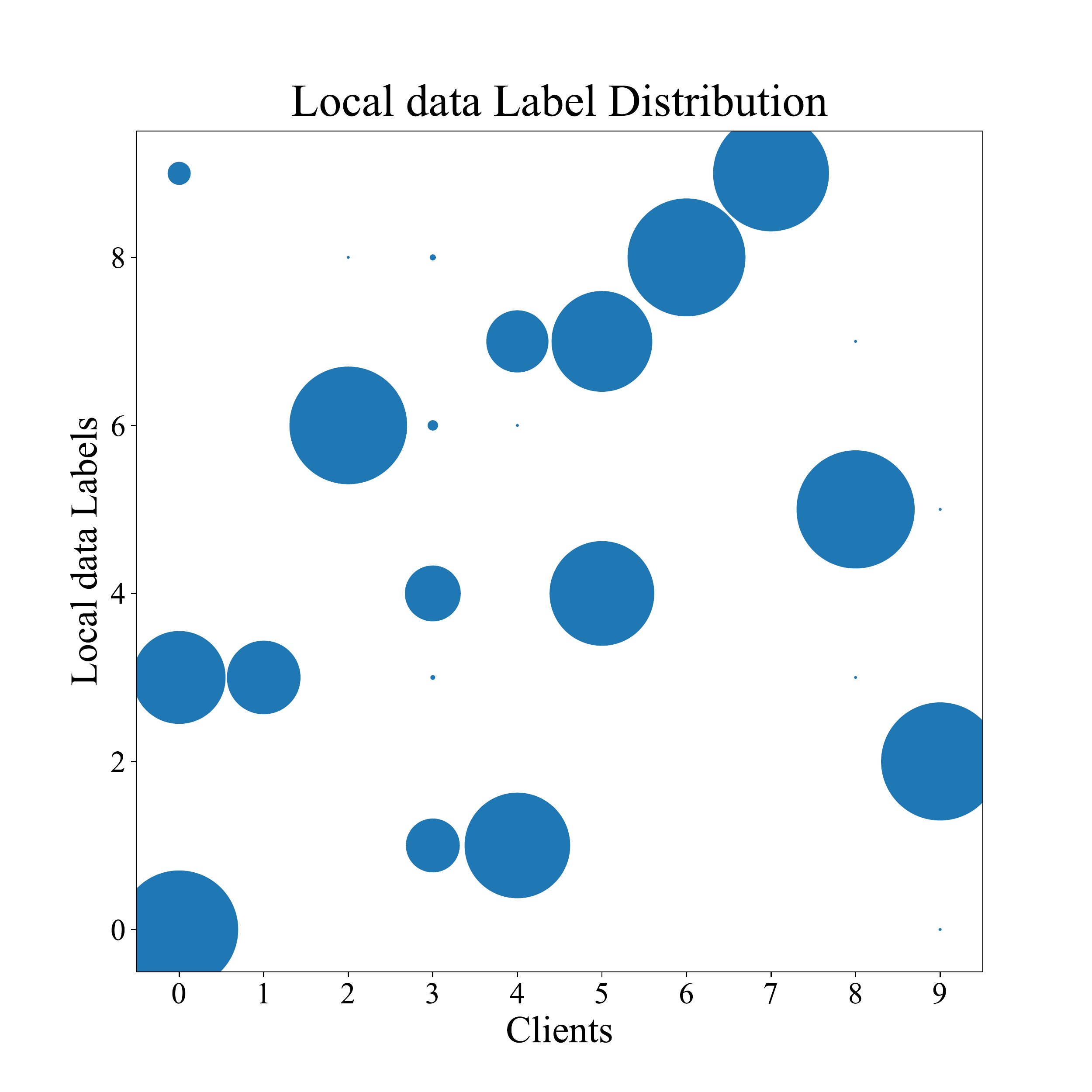}
				%\caption{fig1}
			\end{minipage}%
			\begin{minipage}[t]{0.34\linewidth}
				\centering
				\includegraphics[width=1.9in]{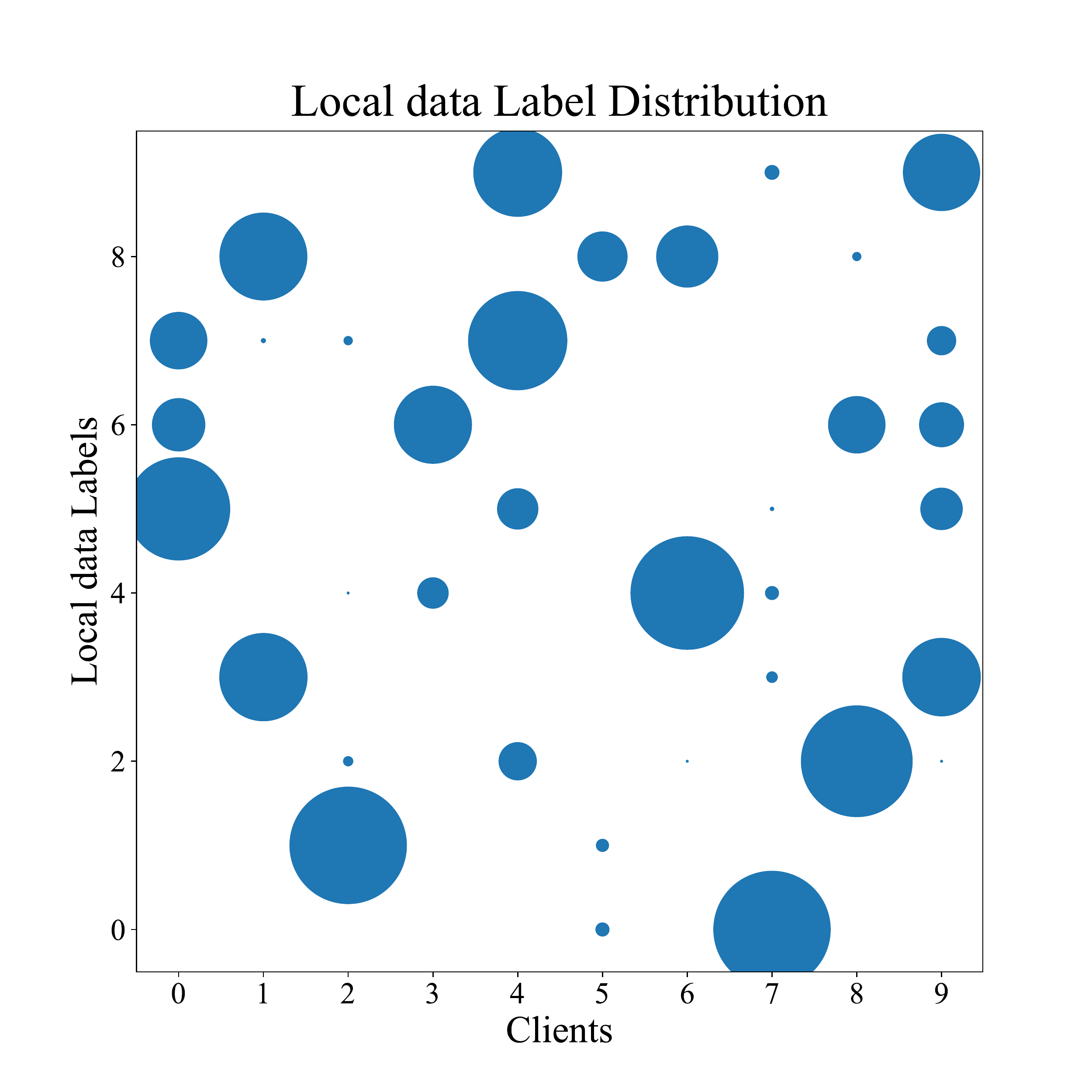}
				%\caption{fig1}
			\end{minipage}%
			\begin{minipage}[t]{0.34\linewidth}
				\centering
				\includegraphics[width=1.9in]{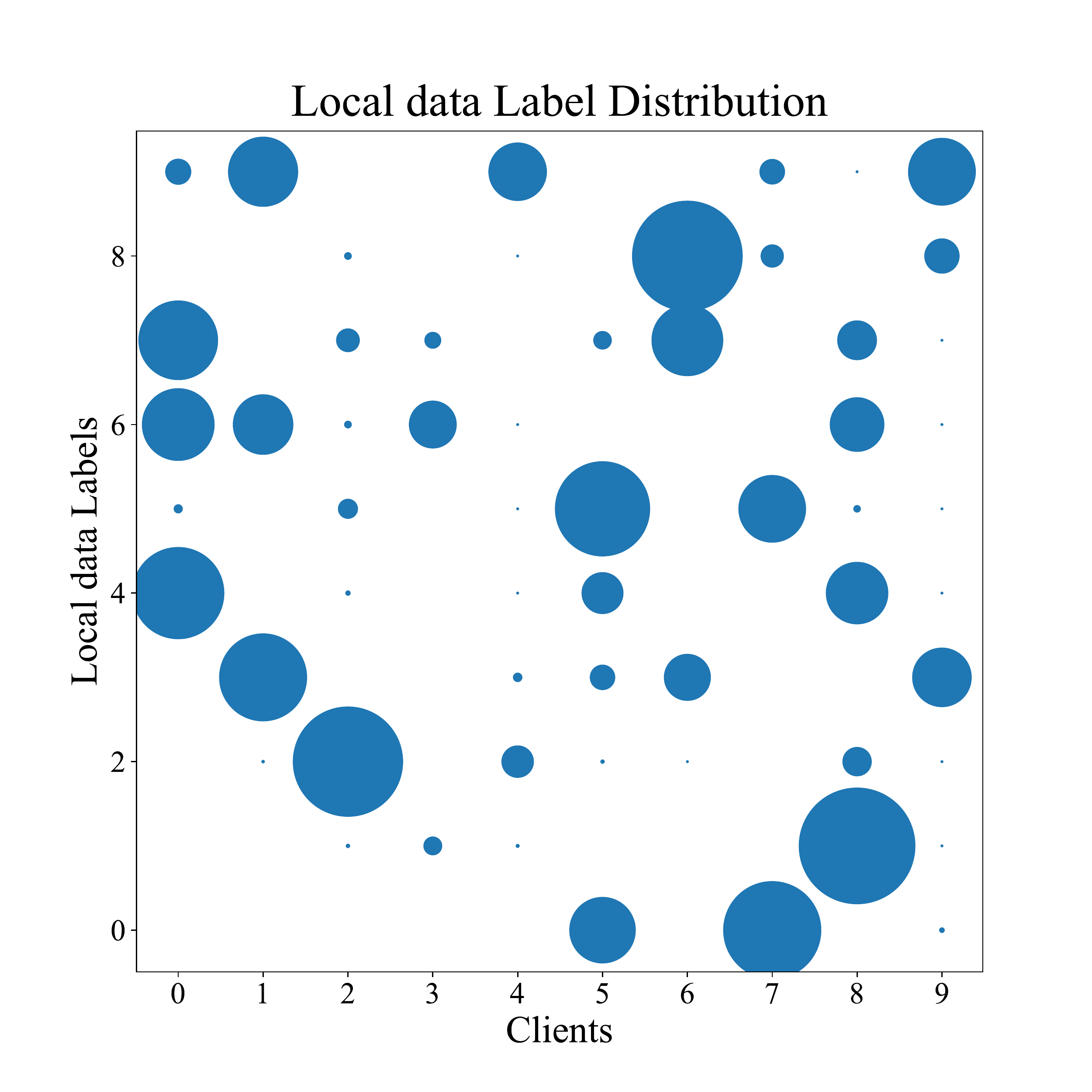}
				%\caption{fig1}
			\end{minipage}%
		}
		\centering
		\caption{Visualization of statistical heterogeneity between users on different datasets when the number of clients is 10. Where the x-axis represents the different clients, the y-axis indicates the class labels, and the size of the scattered points indicates the number of training samples with available labels for that user. From left to right $\alpha$ is 0.01, 0.05, and 0.1.}
		\label{fig44}
	\end{figure}

	\begin{figure}[h]
		\centering
		\begin{minipage}[t]{0.33\linewidth}
			\centering
			\includegraphics[width=1.7in]{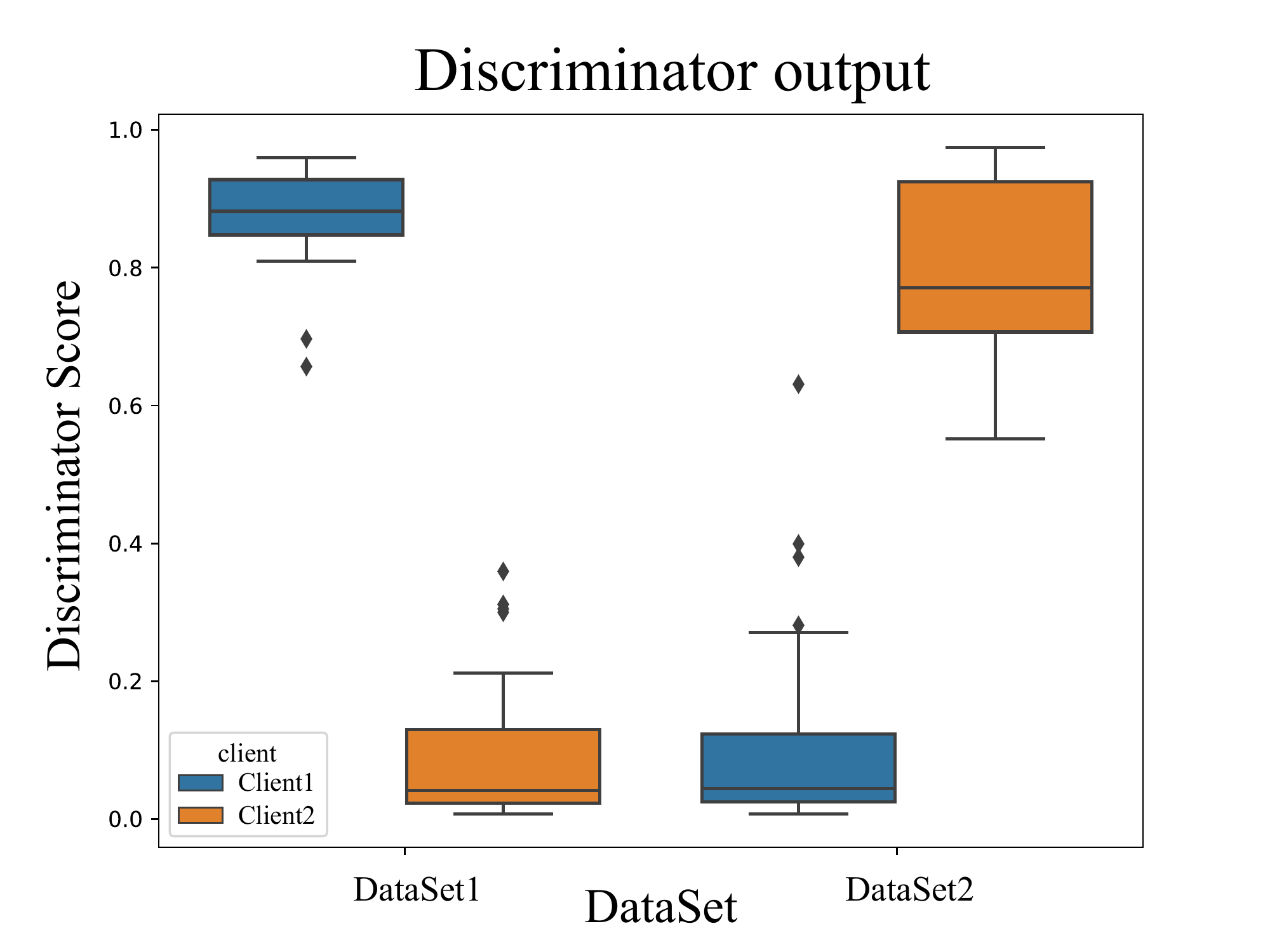}
		\end{minipage}%
		\begin{minipage}[t]{0.33\linewidth}
			\centering
			\includegraphics[width=1.7in]{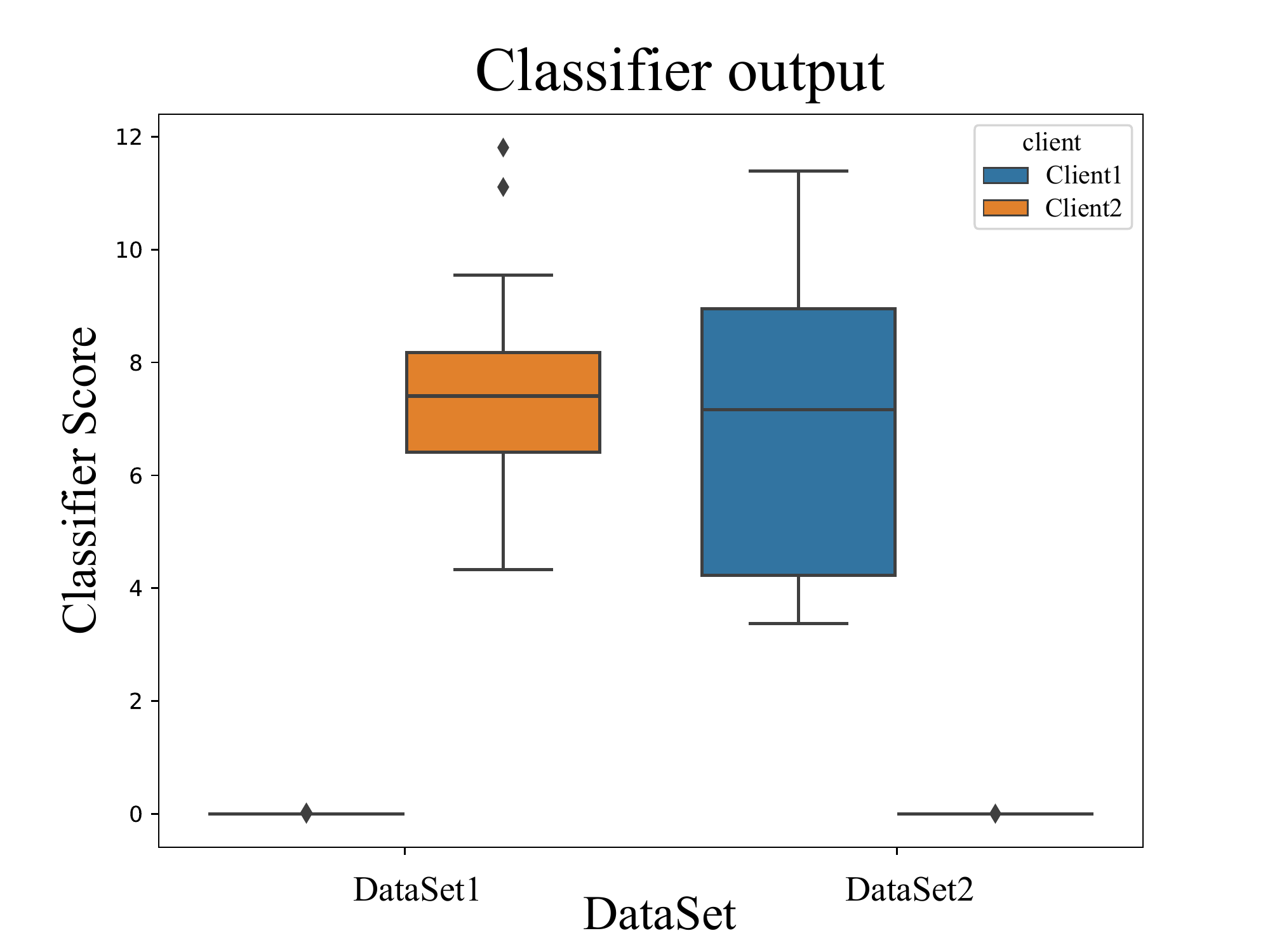}
		\end{minipage}%
		\begin{minipage}[t]{0.33\linewidth}
			\centering
			\includegraphics[width=1.7in]{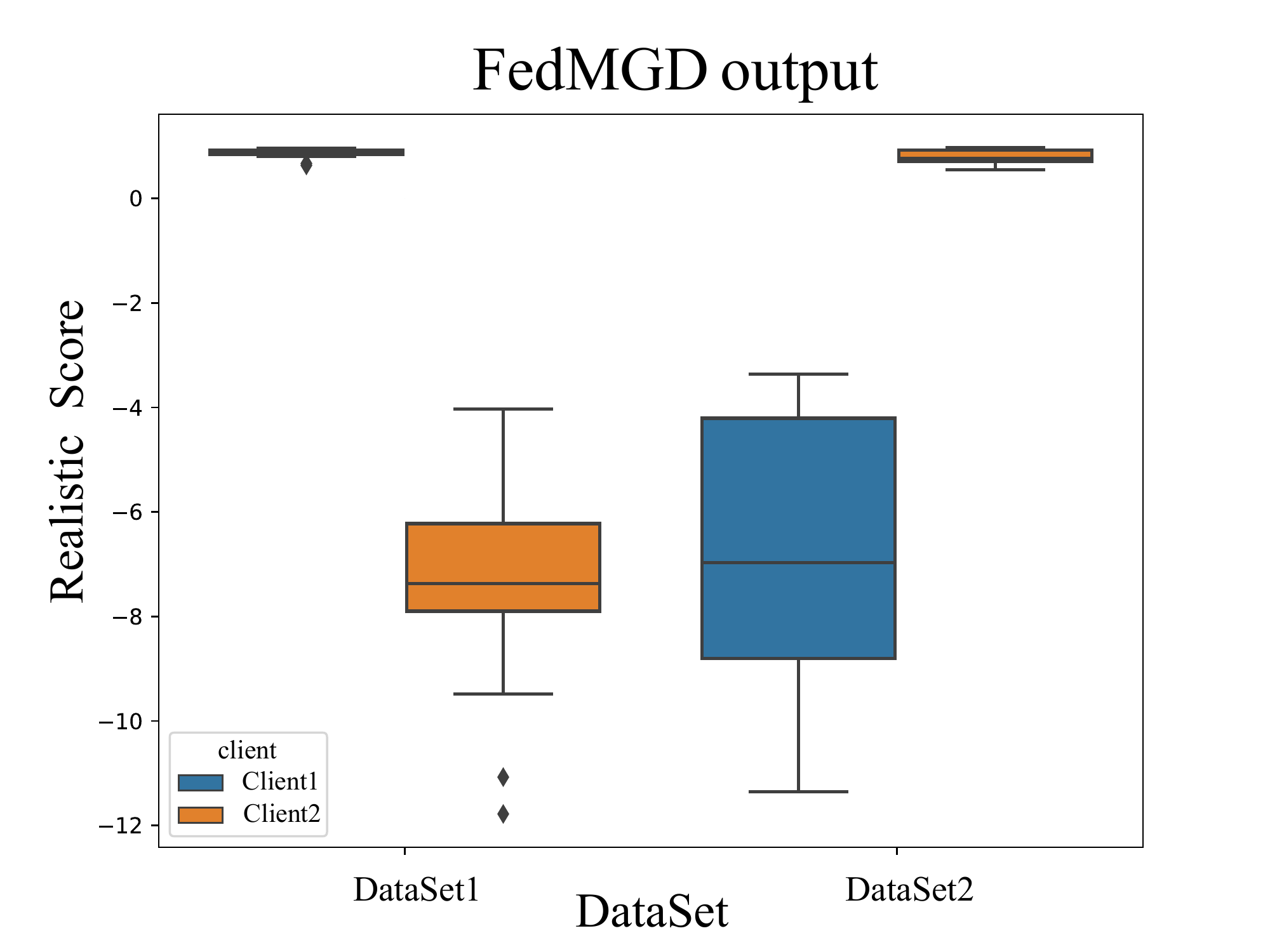}
		\end{minipage}%
		\caption{The classifiers and discriminators are trained in two clients with different data classes. To verify that the classifiers and discriminators perform worse on data categories that have never been seen before, we take each other's data as input. From left to right, the discriminator score, the classifier score (distance from the true label), and the FedMGD (merged) realistic score are shown.}
		\label{cf2u}
	\end{figure}